\documentclass[11pt]{article}

\usepackage[final]{acl}

\usepackage{morefloats}
\extrafloats{500}

\usepackage{times}
\usepackage{latexsym}
\usepackage{comment}

\usepackage[T1]{fontenc}

\usepackage[utf8]{inputenc}

\usepackage{microtype}

\usepackage{inconsolata}

\usepackage{graphicx}

\usepackage{url}
\usepackage{booktabs}
\usepackage{xcolor}
\usepackage{amsmath}
\usepackage{amssymb}
\usepackage{bm}
\usepackage{makecell}
\usepackage{multirow}
\usepackage{placeins}

\usepackage{cleveref}
\Crefname{section}{\S}{\S\S}

\DeclareMathOperator*{\argmax}{arg\,max}

\usepackage{accents}

\newcommand{\todo}[1]{}
\renewcommand{\todo}[1]{{\color{red} TODO: {#1}}}

\title{LLMs Faithfully and Iteratively Compute Answers During CoT: \\ A Systematic Analysis With Multi-step Arithmetics}

\author{
    Keito Kudo${}^{1,2}$,
    Yoichi Aoki${}^{1,2}$,  
    Tatsuki Kuribayashi${}^{3,1}$, Shusaku Sone${}^{1}$ \\ 
    {\bf Masaya Taniguchi${}^{2,1}$, Ana Brassard${}^{2}$, Keisuke Sakaguchi${}^{1,2}$, Kentaro Inui${}^{3,1,2}$} \\
    ${}^{1}$Tohoku University,
    ${}^{2}$RIKEN, 
    ${}^{3}$MBZUAI \\ 
    \texttt{\{keito.kudo.q4, youichi.aoki.p2, sone.shusaku.r8\}@dc.tohoku.ac.jp, } \\
    \texttt{\{tatsuki.kuribayashi, kentaro.inui\}@mbzuai.ac.ae, } \\
    \texttt{keisuke.sakaguchi@tohoku.ac.jp,}
    \texttt{\{masaya.taniguchi, ana.brassard\}@riken.jp} \\
}

\begin{document}
\maketitle

\begin{abstract}
This study investigates the internal information flow of large language models (LLMs) while performing chain-of-thought (CoT) style reasoning.
Specifically, with a particular interest in the faithfulness of the CoT explanation to LLMs' final answer, we explore (i) when the LLMs' answer is (pre)determined, especially before the CoT begins or after, and (ii) how strongly the information from CoT specifically has a causal effect on the final answer.
Our experiments with controlled arithmetic tasks reveal a systematic internal reasoning mechanism of LLMs.
They have not derived an answer at the moment when input was fed into the model.
Instead, they compute (sub-)answers while generating the reasoning chain on the fly.
Therefore, the generated reasoning chains can be regarded as faithful reflections of the model's internal computation.
\end{abstract}

\section{Introduction}
\label{sec:introduction}
Modern large language models (LLMs), given a query from users, typically produce intermediate reasoning steps as well as a final answer.
Such reasoning steps to achieve the answer, i.e., explanation, are helpful to grasp the model's underlying reasoning process and the plausible justification for the produced answer. 
One critical concern is how \textit{faithful} the explanation, typically with a chain-of-thought (CoT)~\cite{wei2022chain} style, is to their final answer, i.e., the causal relationship between the CoT process and the final answer.
In particular, how models \textit{internally} associate the CoT part and its following answer is an intriguing question in the interpretability field.
In this study, we analyze the faithfulness between the CoT explanation and the model's final answer through the lens of (mechanistic) interpretability.
This question involves two key subquestions---when, in the first place, models come up with (sub-)answers during CoT-style reasoning, and how these are referred to internally.
One possible strategy, for example, would be to reach the final answer even during reading problem statements before CoT generation (the first pass), and then the model produces a fluent explanation just to follow an expected format, and ultimately, the model simply refers to their predetermined answers.
In this case, the final answer is no longer strictly faithful to the CoT explanation, and users should be aware of this limitation in interpreting the model's outputs.

\begin{figure*}[t]
    \centering
    \includegraphics[width=1.0\linewidth]{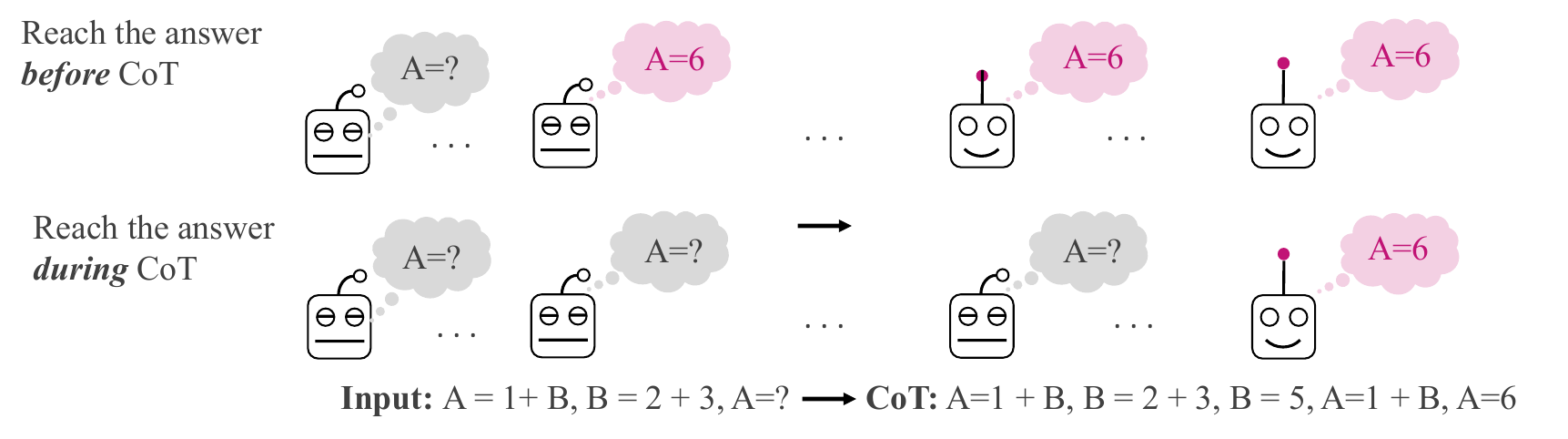}
    \caption{
    Using linear probes, we investigated at which time during the LLM's problem-solving process it is possible to determine the values of each variable, illustrating the model's problem-solving process.
    Our analysis indicates that LLMs come up with (sub-)answers during CoT.
    This conclusion is also consistent with the findings from the causal experiments in \cref{sec:intervention}.
    }
    \label{fig:overview}
\end{figure*}

In our first experiments, we exploratorily analyze \textit{when} (at each layer at each timestep) models represent answers internally, using linear probes.
To prevent situations where CoT is not required for deriving the answer, we set up controlled testbeds of symbolic arithmetic reasoning and observed when/where probes could accurately extract and control the answer (Figure~\ref{fig:overview}). 
By comparing accuracies across each timestep, one can observe at which point models internally start being informative (the values of the variables are encoded in the hidden states as linearly separable representations) to the probes, illustrating the model's internal reasoning pattern. 
We found that models have not derived an answer at the moment they first read the problem; instead, they obtain (sub-)answers while generating the reasoning chain, suggesting the faithfulness of CoT to the final answer.

Based on the probing results, we further conducted causal intervention analyses to clarify the causal relationship between the model's internal representations and answers (\cref{sec:intervention}).
We found that, after CoT generation, the model's final answer can be changed by causal intervention to internal representations for the CoT explanation part, while not for the first pass part.
The discovered causal graph follows recency bias, i.e., each reasoning step in CoT causally depends on its previous step, and the final answer heavily relies on the latter part of CoT.
Thus, we tentatively conclude that models derive an answer during CoT on the fly, and the generated reasoning chains can be regarded as faithful reflections of the model's final answer.
\footnote{Code and data are available at \url{https://github.com/keitokudo/faithful-cot-multistep-arithmetic}}

\begin{table*}[t]
  \scriptsize
  \centering
  \tabcolsep  2pt
  \begin{tabular}{cp{28em}p{23em}rrr}
  \toprule
  Level & \textsc{Input} & \textsc{Output} & \#Step & \#Stack & \#Dist. \\
  \midrule
  1 & \({\underline{\mathrm{A}=1+\mathrm{B}}}_{\hspace{0.05cm}\mathbf{-3}}\), \({\underline{\mathrm{B}=2}}_{\hspace{0.05cm}\mathbf{-2}}\); \({\underline{\mathrm{A}=?}}_{\hspace{0.05cm}\mathbf{-1}}\) & \({\underline{\mathrm{A}=1+\mathrm{B}}}_{\hspace{0.05cm}\mathbf{0}}\), \({\underline{\mathrm{B}=2}}_{\hspace{0.05cm}\mathbf{1}}\), \({\underline{\mathrm{A}=1+\mathrm{B}}}_{\hspace{0.05cm}\mathbf{2}}\), \({\underline{\mathrm{A}=1+2}}_{\hspace{0.05cm}\mathbf{3}}\), \({\underline{\mathrm{A}=3}}_{\hspace{0.05cm}\mathbf{4}}\) & 1 & 1 & 0 \\[0.4cm]
  2 & \({\underline{\mathrm{A}=2+3}}_{\hspace{0.05cm}\mathbf{-3}}\), \({\underline{\mathrm{B}=1+\mathrm{A}}}_{\hspace{0.05cm}\mathbf{-2}}\); \({\underline{\mathrm{B}=?}}_{\hspace{0.05cm}\mathbf{-1}}\) & \({\underline{\mathrm{B}=1+\mathrm{A}}}_{\hspace{0.05cm}\mathbf{0}}\), \({\underline{\mathrm{A}=2+3}}_{\hspace{0.05cm}\mathbf{1}}\), \({\underline{\mathrm{A}=5}}_{\hspace{0.05cm}\mathbf{2}}\), \({\underline{\mathrm{B}=1+\mathrm{A}}}_{\hspace{0.05cm}\mathbf{3}}\), \({\underline{\mathrm{B}=1+5}}_{\hspace{0.05cm}\mathbf{4}}\), \({\underline{\mathrm{B}=6}}_{\hspace{0.05cm}\mathbf{5}}\) & 2 & 0 & 0 \\[0.4cm]
  3 & \({\underline{\mathrm{A}=1+\mathrm{B}}}_{\hspace{0.05cm}\mathbf{-3}}\), \({\underline{\mathrm{B}=2+3}}_{\hspace{0.05cm}\mathbf{-2}}\); \({\underline{\mathrm{A}=?}}_{\hspace{0.05cm}\mathbf{-1}}\) & \({\underline{\mathrm{A}=1+\mathrm{B}}}_{\hspace{0.05cm}\mathbf{0}}\), \({\underline{\mathrm{B}=2+3}}_{\hspace{0.05cm}\mathbf{1}}\), \({\underline{\mathrm{B}=5}}_{\hspace{0.05cm}\mathbf{2}}\), \({\underline{\mathrm{A}=1+\mathrm{B}}}_{\hspace{0.05cm}\mathbf{3}}\), \({\underline{\mathrm{A}=1+5}}_{\hspace{0.05cm}\mathbf{4}}\), \({\underline{\mathrm{A}=6}}_{\hspace{0.05cm}\mathbf{5}}\) & 2 & 1 & 0 \\[0.4cm]
  4 & \({\underline{\mathrm{A}=1+\mathrm{B}}}_{\hspace{0.05cm}\mathbf{-4}}\), \({\underline{\mathrm{B}=2+3}}_{\hspace{0.05cm}\mathbf{-3}}\), \({\underline{\mathrm{C}=4+5}}_{\hspace{0.05cm}\mathbf{-2}}\);   \({\underline{\mathrm{A}=?}}_{\hspace{0.05cm}\mathbf{-1}}\) & \({\underline{\mathrm{A}=1+\mathrm{B}}}_{\hspace{0.05cm}\mathbf{0}}\), \({\underline{\mathrm{B}=2+3}}_{\hspace{0.05cm}\mathbf{1}}\), \({\underline{\mathrm{B}=5}}_{\hspace{0.05cm}\mathbf{2}}\), \({\underline{\mathrm{A}=1+\mathrm{B}}}_{\hspace{0.05cm}\mathbf{3}}\), \({\underline{\mathrm{A}=1+5}}_{\hspace{0.05cm}\mathbf{4}}\), \({\underline{\mathrm{A}=6}}_{\hspace{0.05cm}\mathbf{5}}\) & 2 & 1 & 1 \\[0.4cm]
  5 & \({\underline{\mathrm{A}=1+\mathrm{B}}}_{\hspace{0.05cm}\mathbf{-4}}\), \({\underline{\mathrm{B}=2+\mathrm{C}}}_{\hspace{0.05cm}\mathbf{-3}}\), \({\underline{\mathrm{C}=1+2}}_{\hspace{0.05cm}\mathbf{-2}}\); \({\underline{\mathrm{A}=?}}_{\hspace{0.05cm}\mathbf{-1}}\) & \({\underline{\mathrm{A}=1+\mathrm{B}}}_{\hspace{0.05cm}\mathbf{0}}\), \({\underline{\mathrm{B}=2+\mathrm{C}}}_{\hspace{0.05cm}\mathbf{1}}\), \({\underline{\mathrm{C}=1+2}}_{\hspace{0.05cm}\mathbf{2}}\), \({\underline{\mathrm{C}=3}}_{\hspace{0.05cm}\mathbf{3}}\), \({\underline{\mathrm{B}=2+\mathrm{C}}}_{\hspace{0.05cm}\mathbf{4}}\), \({\underline{\mathrm{B}=2+3}}_{\hspace{0.05cm}\mathbf{5}}\), \({\underline{\mathrm{B}=5}}_{\hspace{0.05cm}\mathbf{6}}\), \({\underline{\mathrm{A}=1+\mathrm{B}}}_{\hspace{0.05cm}\mathbf{7}}\), \({\underline{\mathrm{A}=1+5}}_{\hspace{0.05cm}\mathbf{8}}\), \({\underline{\mathrm{A}=6}}_{\hspace{0.05cm}\mathbf{9}}\) &3 &2 &0 \\
  \bottomrule
\end{tabular}

  \caption{
  Examples of arithmetic reasoning tasks used in our experiments at each complexity level. 
  \#Step indicates the number of required operations to reach the final answer.
  \#Stack indicates how many variables' values are not immediately determined in their first appearing equation.
  \#Dist. is the number of unnecessary distractor equations. The number (e.g., $\mathbf{-3}$) indicated in the lower right corner of each equation represents the equation's position.
  This position is used as a reference point for calculating $t^*_{\mathrm{eq}}$ in \cref{subsec:evaluation_metrics}.
  }
  \label{table:tasks}
\end{table*}

\section{General settings}
\label{sec:general_setting}

\subsection{Arithmetic problems}
\label{subsec:input}
We prepared a synthetic dataset of multi-hop arithmetic problems similarly to~\citet{kudo-etal-2023-deep} and~\citet{Yu2024DoLR}.
Each problem instance consists of strings of assignments (e.g., \texttt{A=1}) and operations (e.g., \texttt{B=1+3} or \texttt{B=1+A}) ending with a query for a variable's value (e.g., \texttt{B=?}).
By using this synthetic dataset, we can conduct controlled experiments across different levels of problem complexity and confirm the generality of the obtained results.
Moreover, by training and evaluating probes for each common problem format (level), we can conduct fine-grained, token-level analyses.
Such controlled analyses are difficult with datasets written in natural language.

We specifically defined five complexity levels  (Table~\ref{table:tasks}), depending on (i) how many equations need to be resolved to reach the answer (\#Step in Table~\ref{table:tasks}), (ii) how many variables' values cannot be immediately resolved (and thus pended to a stack) in their first appearance when incrementally reading the problem from left to right (\#Stack), (iii) and the number of unnecessary distractor equations (\#Dist.). 
For example, in the level 5 example in Table~\ref{table:tasks}, where \#Step is three and \#Stack is two, calculating \texttt{A} requires at least three steps of reasoning: \texttt{C(=1+2)=3}, \texttt{B(=2+3)=5}, and then \texttt{A(=1+5)=6}, and two variables need to be resolved before reaching \texttt{A}: \texttt{B} and \texttt{C}.

\paragraph{Notation.}
Formally, let $v$ denote a variable name sampled from the 26 letters of the English alphabet $\Sigma = \{\text{a}, \text{b}, \text{c}, \dots, \text{z}\}$, and $d$ denote a number sampled from the set of decimal digits $D = \{0, 1, 2, \ldots, 9\}$.
Each instance consists of multiple equations $[e_1, e_2,\cdots, e_n]$ followed by a final query $q$. Each equation follows the format $v=d$, $v=d \pm d$, or $v=d \pm v$.
We denote $i$-th variable in an instance from the left as $v_i$; e.g., in the Level 5 example in Table~\ref{table:tasks}, $v_1=\texttt{A}$, $v_2=\texttt{B}$, and  $v_3=\texttt{C}$.\footnote{For brevity, all examples in this paper use the uppercased variables \texttt{A}, \texttt{B}, \texttt{C}, and only the operator \texttt{+}, but the actual instances have a variety in the variable names and operator types.}
The value assigned to a variable $v_i$ is denoted as $\$\{v_i\} \in D$.

\paragraph{Data constraints.}
We ensure that $\$\{v_i\}$ for any $v_i$ is also a single-digit number, and $\$\{v_i\}$ is constant within the same instance (i.e., we exclude cases such as \texttt{A=1+2,A=B+2,B=6}).
All samples of the same complexity level follow the exact same format except for the actual numbers, variable names, and operators.
In other words, an answer $\$\{v_i\}$ in each level can be obtained with exactly the same procedure. For example, first calculate $\$\{v_3\}$, and then calculate $\$\{v_2\}$ with $\$\{v_3\}$, and then finally calculate $\$\{v_1\}$ with $\$\{v_2\}$. Non-duplicated instances are created for each level by varying the variable names, numbers, and operators appearing in the equations.
We use 10,000 instances to train probing classifiers and 2,000 to test their accuracy.
To prevent the probe from simply memorizing (sub-)answers, the training and test sets were constructed with no overlap in arithmetic expressions.\footnote{The overlap is computed at the equation-level. For example, if \texttt{1+2} appears in the training set, then \texttt{1+2} does not appear in any test instances.}

\subsection{CoT-style inference}
\label{subsec:output}

Given the input described in~\cref{subsec:input}, the model generates CoT $z$ and a final answer $y$.
Henceforth, we refer to the part before CoT as the \textsc{Input} $x$ and the CoT-reasoning part ($y, z$) as the \textsc{Output} of an instance, as shown in the right part of Table~\ref{table:tasks}.
For example, for the Level 1 instance in Table~\ref{table:tasks} (topmost), $x=$ ``\texttt{A=1+B, B=2},'' $z=$ ``\texttt{A=1+B, B=2, A=1+2, A=}'', and $y=$  ``\texttt{3}.''
We demonstrate three examples in the same problem level for the model to inform the problem setting and expected CoT-style reasoning style (Table~\ref{table:tasks}).
That is, the task is, given a demonstration of three examples, and an \textsc{Input} $x$, to generate intermediate steps $z$ and derive a final answer $y$.
As a sanity check, we confirmed that the target models could follow the expected \textsc{Output} format and solve the tasks with nearly 100\% accuracy in this setting (\cref{appendix:subsec:performance_of_llms_on_tasks}).

\subsection{Research scope}
\label{subsec:scope}
We are particularly concerned about the faithfulness between the CoT part $z$ and its following final answer $y$, since these two parts are typically shown to users, and thus are of primary concern regarding the model's reliability from the user's perspective.
In contrast, some recent studies~\cite{afzal-etal-2025-knowing,cox2025-posthoc-cot,ye2025physics} have investigated whether the CoT $z$ depends on the internal state of the model $x$ before CoT begins (whether the CoT $z$ is \textit{post hoc} or not).
This causal relationship $x\rightarrow z$ is out of scope in this study, while we are still interested in whether the model $x$ has a predetermined answer $y$ in order to distinguish the possibilities of $x\rightarrow y$ and $z\rightarrow y$, i.e., which part is causally relied on to the final answer $y$.

\section{Linear probing}
\label{sec:probing}
To begin with the analysis, it will be good to know where the final answer, or the necessary sub-answer for it, can possibly come from.
We first analyze where final-/sub- answers can be extracted from the model's internal representations, using linear probes.

\paragraph{Position $t$.}
We denote a token position within the entire concatenated sequence $x\oplus z\oplus y$ with $t \in \mathbb{Z}$.
The position $t$ is relative to CoT; that is, $t$ is zero at the moment CoT begins, negative within the \textsc{Input}, and positive within the \textsc{Output}.
Similarly, we assign an equation position $t_\mathrm{eq} \in \mathbb{Z}$ to each equation in the \textsc{Input} and \textsc{Output} (subscripts on the underlines in Table~\ref{table:tasks}).

\subsection{Training linear probes}
\label{subsec:probe_setting}
Linear probes~\cite{DBLP:conf/iclr/AlainB17} are trained to identify when the model derives the answer internally. 
We separately train a probe for each combination of token position $t \in \mathbb{Z}$, layer depth $l \in \mathbb{N}$, and the variable of interest $v_i \in \Sigma$ in each level of the problem.
That is, each probe is solely responsible for the position $(t, l, v_i)$ to identify the answer $\hat{\$\{v_i\}}$ represented. 
Formally, given a model's $d$-dimensional hidden state $\bm h_{t,l} \in \mathbb{R}^{d}$, the probing classifier $f_{t,l,v_i}(\cdot): \mathbb{R}^{d} \to D$ is trained to predict the correct answer $\$\{v_i\}$.
If a probe $f_{t,l,v_i}$ achieves high accuracy, this suggests that the answer for a particular variable $\$\{v_i\}$ is already computed at the corresponding position $(t, l)$.
The probe $f_{t,l,v_i}$ is a single linear transformation:
\begin{equation}
    \begin{aligned}
    \hat{\$\{v_{i}\}}_{t,l} & = f_{t,l,v_i}(\bm h_{t,l}) \\
    & = \argmax_D \bm W_{t,l,v_{i}}\bm h_{t,l} + \bm b_{t,l,v_{i}} \;\;\mathrm{,}
    \end{aligned}
\end{equation}

\noindent
where, $\bm W_{t,l,v_{i}} \in \mathbb{R}^{|D| \times d}$ and $\bm b_{t,l,v_{i}} \in \mathbb{R}^{|D|}$ are the trainable weight and bias parameters of the probe, respectively.
The symbol $\hat{\cdot}$ refers to the model's predicted answer.
We train the probes using stochastic gradient descent~\cite{Robbins1951ASA} to minimize the cross entropy loss.
The hyperparameters are listed in Table~\ref{table:probing_hyperparameter} in the appendix.

 Each probe classifier is trained with 10,000 of $\bm h_{t,l}$ from training instances and then computes the accuracy with 2,000 hidden states from test instances.

\begin{figure*}[t]
    \centering
    \includegraphics[width=\linewidth]{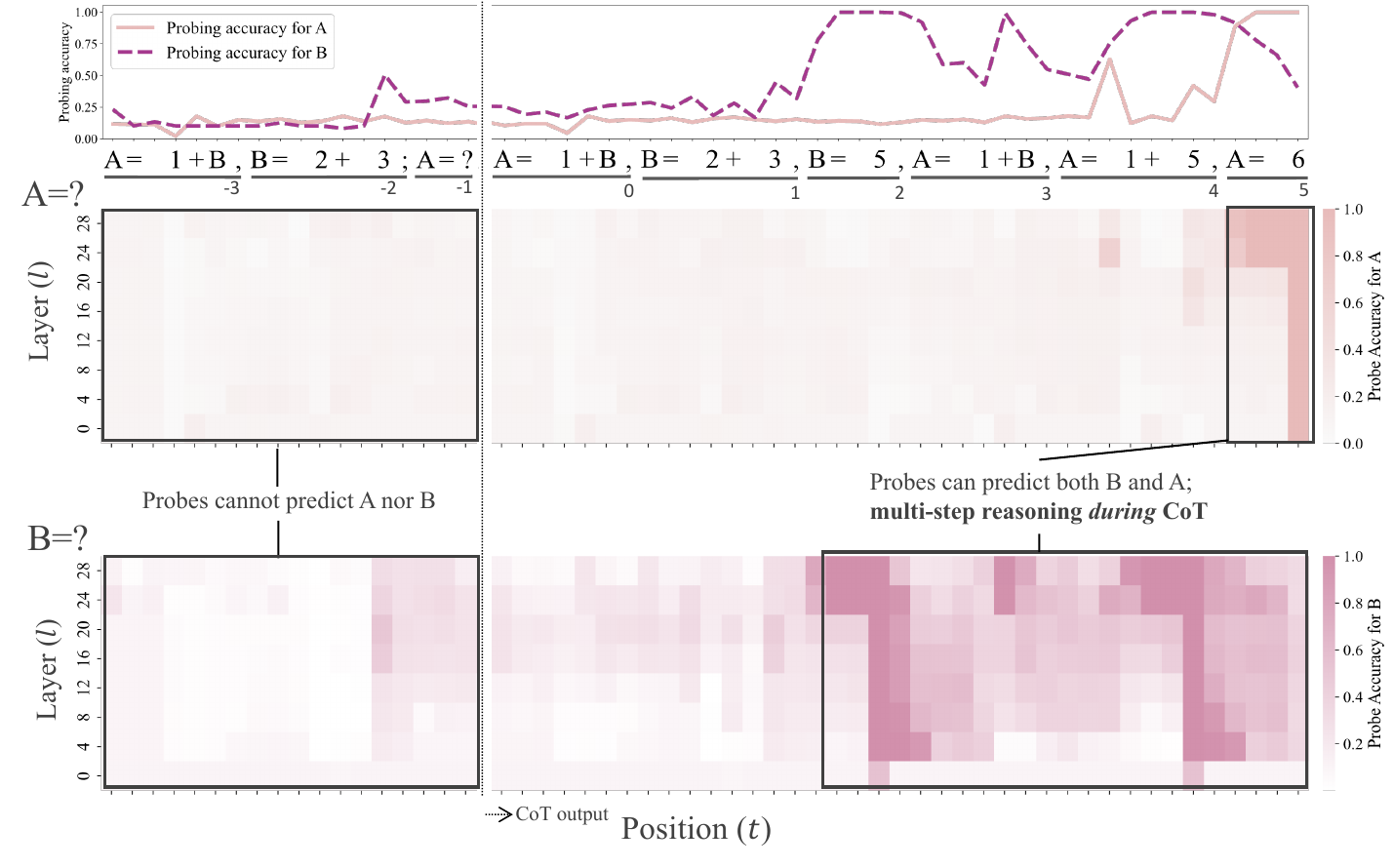}
    \caption{
    Probing results for Qwen2.5-7B at the Level 3 task.
    The heatmaps in the lower section represent the accuracy of probes computed on the evaluation set.
    Each cell shows the probing accuracies in each token $t$, layer $l$.
    The upper part indicates the maximum probing accuracy achieved at each token position $t$.
    The input sequence below the line graphs is just an example; in the actual evaluation set, each variable name, number, and operator are randomly sampled from $(D, \Sigma, \{+, -\})$.
    }
    \label{fig:probing_results_step2}
\end{figure*}

\subsection{Evaluation metrics}
\label{subsec:evaluation_metrics}
The probing results from all the token positions $t$ and layers $l$ are aggregated as follows:

\begin{align}
        t^*(v_i) &=\min \{ t \mid \max_{l} \mathrm{acc}(t,l,v_i) > \tau\} \text{,} \label{eq:rsp_tok}
\end{align}

\noindent
where $\mathrm{acc}(t,l,v_i) \in [0,1]$ denotes the probing accuracy of $f_{t,l,v_i}$.
The position $t^*(v_i) \in \mathbb{Z}$ indicates when the first time the probing classifier achieved a reasonable accuracy above $\tau$.
We set $\tau=0.9$ in our main experiment, and we show additional results with different thresholds in Appendix~\ref{appendix:subsec:all_probing_results}.
As a more coarse but comprehensive value, we also report $t_\mathrm{eq}^*(v_i)$, indicating which equation $t_\mathrm{eq}$ in the input $x$ the position $t^*(v_i)$ falls into.
Given that the $t$ (and $t_\mathrm{eq}$) is relative to the CoT-beginning position, if $t_\mathrm{(eq)}^*(v_i)$ is negative, the value $\$\{v_i\}$ is computed before CoT begins.
In Figure~\ref{fig:probing_results_step2}, for example, both $t^*(\mathrm{\texttt{A}})$ and  $t^*(\mathrm{\texttt{B}})$ are positive.

As complementary information of the position, we also report two types of accuracy:
\begin{align}
        \mathrm{Acc}_{\mathrm{\prec CoT}}(v_{i}) &= \max_{t < 0, l} \mathrm{acc}(t,l,v_i) \text{,}\\
        \mathrm{Acc}_{\mathrm{\succ CoT}}(v_{i}) &= \max_{t \geq 0, l} \mathrm{acc}(t,l,v_i) \text{.}
\end{align}

\noindent
If $\mathrm{Acc}_{\mathrm{\prec CoT}}(v_{i})$ is sufficiently high, $\$\{v_{i}\}$ is resolved internally before CoT begins. 
Conversely, if $\mathrm{Acc}_{\mathrm{\prec CoT}}(v_{i})$ is low and $\mathrm{Acc}_{\mathrm{\succ CoT}}(v_{i})$ is high, the answer is derived while performing CoT reasoning.

\subsection{Results}
\label{subsec:experiments}

\paragraph{Across task levels.}
We first analyze Qwen2.5-7B~\cite{qwen2.5} across the five task levels.
Table~\ref{table:rsp_task_dimmension} reports the position $t_\mathrm{eq}^*$ for each variable and its lower bound $t_\mathrm{eq}^{\mathrm{inf}}$ where the answer can be first identifiable.
In most cases, (sub)answers are represented in the model’s hidden states corresponding to the \textsc{Output} part ($t^*_\mathrm{eq}>0$) in a linearly separable form, even when deriving relatively simple sub-answers, e.g., $\$\{\texttt{B}\}$ in Level 2. %
The exceptions are: (i) $v_2$ in Level 1, and (ii) $v_3$ in Level 4.
Here, $v_2$ in Level 1 requires no computation (\#Steps$=0$), and $v_3$ in Level 4 is a distractor that is not needed to derive the final answer.
To sum up, all calculations (at least 1 step) required to obtain the final answer are resolved \textbf{during CoT}, suggesting that it's unlikely to refer to a predetermined answer as model's final answer.

\begin{table}[t!]
  \tabcolsep 1mm
  \small
  \centering

\begin{tabular}{cllrrrr}
\toprule
&\multicolumn{2}{c}{Variable} &\multicolumn{2}{c}{When ($\downarrow$)} & \multicolumn{2}{c}{$\mathrm{Acc.}$ ($\uparrow$)} \\
\cmidrule(lr){2-3} \cmidrule(lr){4-5} \cmidrule(lr){6-7}
Level & variable & \#Step & $t_\mathrm{eq}^*$ & $t_\mathrm{eq}^{\mathrm{inf}}$ & $\mathrm{\prec CoT}$ & $\mathrm{\succ CoT}$ \\ 
\midrule
1 & $v_1$ ($\mathrm{A}$) & 1 & 4 & $-$2 & 35.8 & 100 \\
  & $v_2$ ($\mathrm{B}$) & 0 & $-$2 & $-$2 & 100 & 100 \\
\cmidrule(l){1-7}
2 & $v_1$ ($\mathrm{A}$) & 1 & 2 & $-$3 & 49.2 & 100 \\
  & $v_2$ ($\mathrm{B}$) & 2 & 5 & $-$2 & 21.6 & 94.7 \\
\cmidrule(l){1-7}
3 & $v_1$ ($\mathrm{A}$) & 2 & 5 & $-$2 & 17.9 & 97.4 \\
  & $v_2$ ($\mathrm{B}$) & 1 & 2 & $-$2 & 50.5 & 100 \\
\cmidrule(l){1-7}
4 & $v_1$ ($\mathrm{A}$) & 2 & 5 & $-$3 & 17.2 & 100 \\
  & $v_2$ ($\mathrm{B}$) & 1 & 2 & $-$3 & 47.7 & 100 \\
  & $v_3$ ($\mathrm{C}$) & 1 & N/A & $-$2 & 43.7 & 23.7 \\
\cmidrule(l){1-7}
5 & $v_1$ ($\mathrm{A}$) & 3 & 9 & $-$2 & 18.1 & 100 \\
  & $v_2$ ($\mathrm{B}$) & 2 & 6 & $-$2 & 22.6 & 100 \\
  & $v_3$ ($\mathrm{C}$) & 1 & 3 & $-$2 & 50.6 & 100 \\
\bottomrule
\end{tabular}

  \caption{
  The results of Qwen2.5-7B on the five levels. 
  The $t^*_\mathrm{eq}$ is the time when the model comes up with the correct answer (see \cref{subsec:evaluation_metrics}).
  The $t_\mathrm{eq}^{\mathrm{inf}}$ column indicates the lower bound of $t^*_\mathrm{eq}$ score. 
  The $\mathrm{\prec CoT}$ and $\mathrm{\succ CoT}$ scores correspond to the accuracies introduced in \cref{subsec:evaluation_metrics}.
  N/A indicates that the threshold $\tau$ was not exceeded at any position $t$.
  }
  \label{table:rsp_task_dimmension}
\end{table}

\paragraph{Across models.}
We analyzed nine models (Table~\ref{table:rsp_model_dimmension}) on the Level 3 task.
Similarly to Table~\ref{table:rsp_task_dimmension}, we observed $t^*_\mathrm{eq}>0$. %
This pattern holds for other tasks and for different thresholds $\tau$ (Tables~\ref{table:rsp_model_dimmension_tau_0.85_level_1}--\ref{table:rsp_model_dimmension_tau_0.95_level_5} in \cref{appendix:subsec:all_probing_results}).
On the other hand, in the \textsc{Input} part---particularly the latter half of equation~$-2$ ($\underline{\mathrm{B}=2+3,}_{\hspace{0.05cm}\mathbf{-2}}$)---the pre-CoT accuracy $\mathrm{Acc}_{\mathrm{\prec CoT}}(v_{i})$ increases modestly, peaking around $60\%$.%

\begin{table}[t]
  \tabcolsep 0.3mm
  \footnotesize
  \centering

\begin{tabular}{lcccccc}
\toprule
&&\multicolumn{2}{c}{When ($\downarrow$)} & \multicolumn{2}{c}{$\mathrm{Acc}$ ($\uparrow$)} \\
\cmidrule(r){3-4} \cmidrule(r){5-6}
 & Var. & $t^*_\mathrm{eq}$ & $t^*$ & $\mathrm{\prec CoT}$ & $\mathrm{\succ CoT}$ \\ 
\midrule
Qwen2.5 (7B)  & $v_1$ ($\mathrm{A}$) & 5  & 36  & 17.9  & 100 \\  
\cite{qwen2.5}                    & $v_2$ ($\mathrm{B}$) & 2 & 16  & 50.5  & 100 \\ 
\cmidrule(l){2-6}
Qwen2.5 (14B)  & $v_1$ ($\mathrm{A}$) & 5  & 35  & 17.8  & 100 \\  
\cite{qwen2.5}                    & $v_2$ ($\mathrm{B}$) & 2 & 16  & 50.5  & 100 \\ 
\cmidrule(l){2-6}
Qwen2.5 (32B)  & $v_1$ ($\mathrm{A}$) & 5  & 36  & 17.8  & 100 \\  
\cite{qwen2.5}                    & $v_2$ ($\mathrm{B}$) & 2 & 15  & 67.4  & 100 \\ 
\cmidrule(l){2-6}
Qwen2.5-Math (7B)  & $v_1$ ($\mathrm{A}$) & 5  & 35  & 18.6  & 100 \\  
\cite{yang2024qwen25mathtechnicalreportmathematical} & $v_2$ ($\mathrm{B}$) & 2 & 15  & 56.1  & 100 \\  
\cmidrule(l){2-6}
Yi1.5 (9B)  & $v_1$ ($\mathrm{A}$) & 5  & 41  & 17.8  & 100 \\  
\cite{DBLP:journals/corr/abs-2403-04652}      & $v_2$ ($\mathrm{B}$) & 2 & 18  & 36.9  & 100 \\  
\cmidrule(l){2-6}
Yi1.5 (34B)  & $v_1$ ($\mathrm{A}$) & 5  & 41  & 22.4  & 100 \\  
\cite{DBLP:journals/corr/abs-2403-04652}      & $v_2$ ($\mathrm{B}$) & 2 & 18  & 37.4  & 100 \\ 
\cmidrule(l){2-6}
Llama3.1 (8B)  & $v_1$ ($\mathrm{A}$) & 5  & 35  & 26.0  & 100 \\  
\cite{DBLP:journals/corr/abs-2407-21783}  & $v_2$ ($\mathrm{B}$) & 2 & 16  & 29.6  & 100 \\ 
\cmidrule(l){2-6}
Llama3.2 (3B)  & $v_1$ ($\mathrm{A}$) & 5  & 36  & 17.8  & 93.2 \\  
\cite{DBLP:journals/corr/abs-2407-21783}        & $v_2$ ($\mathrm{B}$) & 2 & 17  & 33.2  & 95.4 \\ 
\cmidrule(l){2-6}
Mistral-Nemo (12B)  & $v_1$ ($\mathrm{A}$) & 5  & 36  & 17.8  & 100 \\  
\cite{mistral-nemo}   & $v_2$ ($\mathrm{B}$) & 2  & 16  & 28.9  & 100 \\ 
\bottomrule
\end{tabular}

  \caption{
  Results for various models on the Level 3 task. 
  The $t^*$ column shows the token-wise time (described in \cref{subsec:evaluation_metrics}), and the other columns are the same as Table~\ref{table:rsp_task_dimmension}.
  The $t^*$ and $t^*_\mathrm{eq}$ scores that are the same as their lower bounds are bolded.
  }
  \label{table:rsp_model_dimmension}
\end{table}

\begin{figure*}[t]
    \centering
    \includegraphics[width=\linewidth]{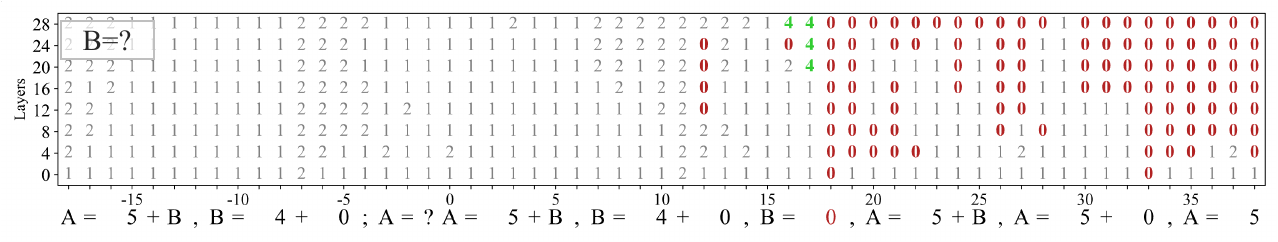}
    \caption{
    Error analysis of cases where Llama3.2-3B generated \emph{incorrect} answers.
    The vertical axis represents the index of the transformer layer.
    The horizontal axis represents the tokens input to the model over time.
    The numbers highlighted in green represent the gold labels for the predictions, while those highlighted in red denote the values incorrectly generated by the model.}
    \label{fig:error_analysis}
\end{figure*}

\begin{figure}[t]
  \centering
  \includegraphics[width=\linewidth]{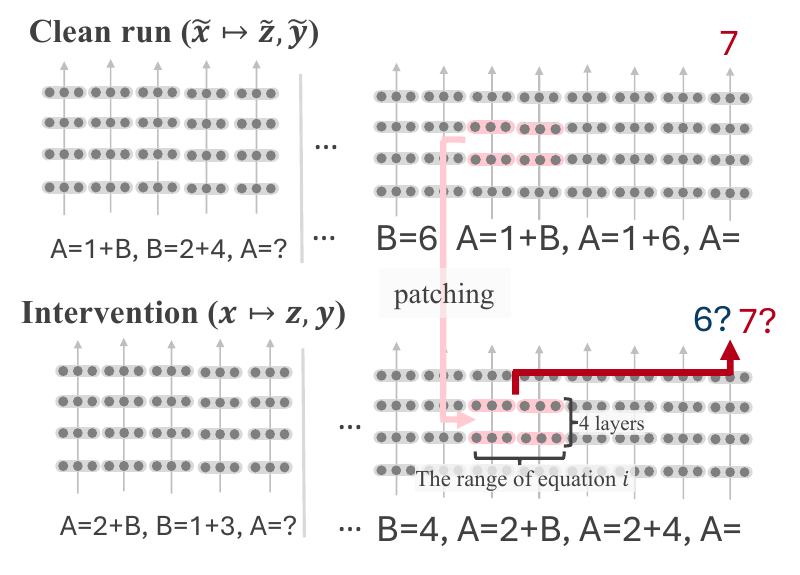}
  \caption{
  Overview of the causal intervention experiment.
  First, we perform normal inference (Clean run) and cache its hidden states.
  Subsequently, we evaluate whether the output changes by replacing some of the hidden states of a model solving a different problem with the cached hidden states.
  }
  \label{fig:intervention_overview}
\end{figure}

\begin{figure}[t]
  \centering
  \includegraphics[width=\linewidth]{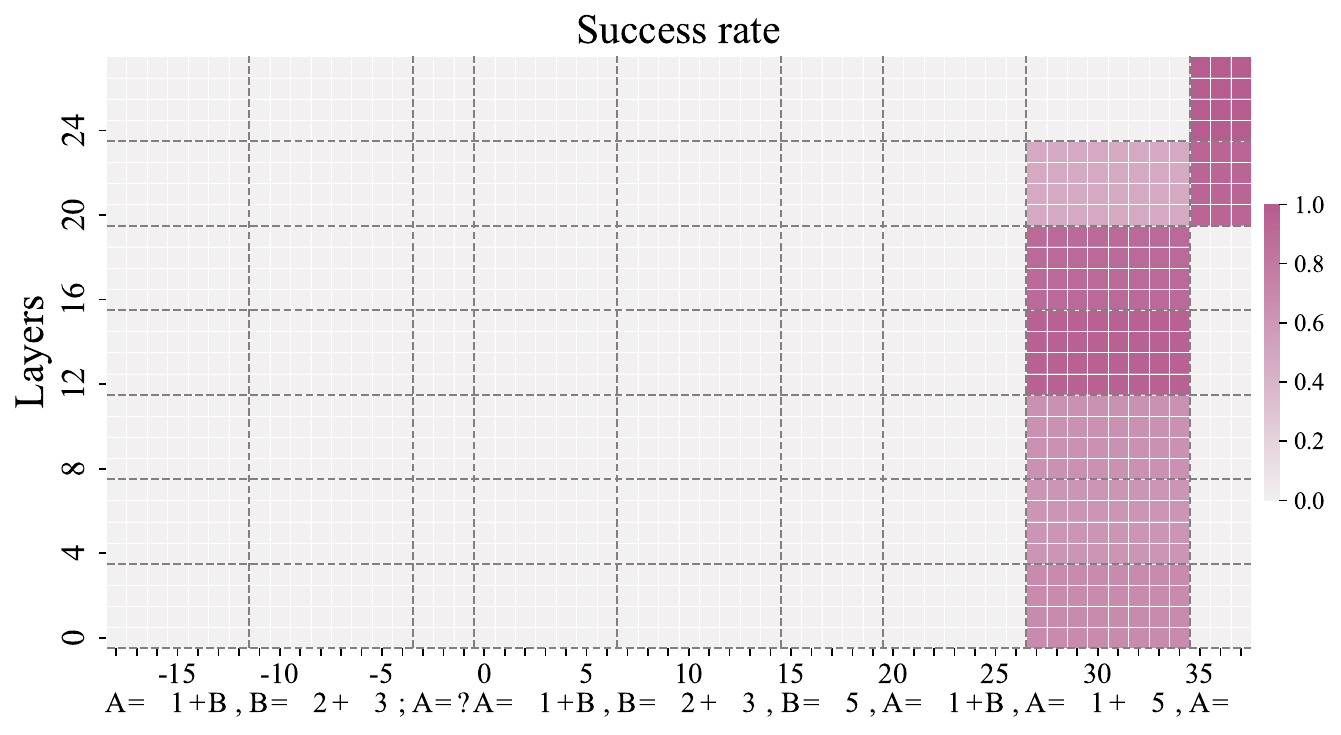}
  \caption{
  Results of the causal intervention on Qwen2.5-7B. Each grid cell shows the success rate when the final answer $y$ \(({\underline{\textcolor{gray}{\mathrm{A}=}\textbf{6}}}_{\hspace{0.05cm}\mathbf{5}})\) is the target token.
  }
  \label{fig:sliding_windowed_intervention_results}
\end{figure}

\begin{figure*}[t]
  \centering
  \includegraphics[width=\linewidth]{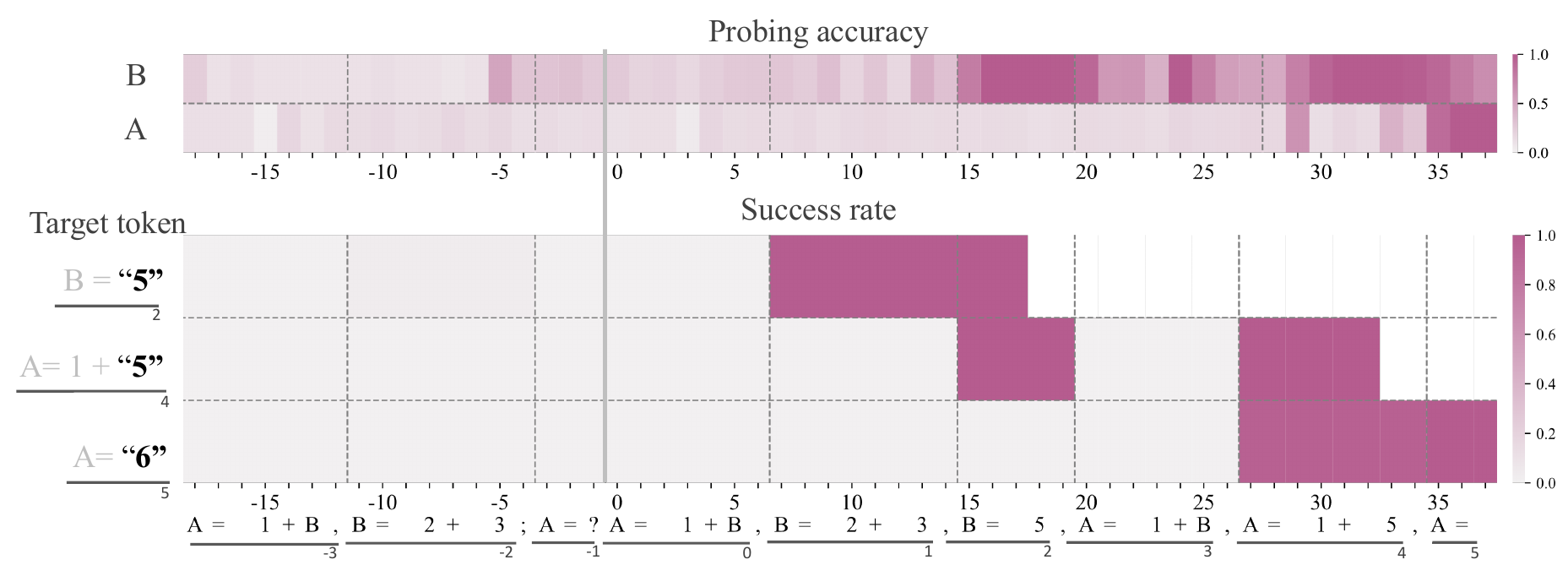}
  \caption{
  The upper part is the accuracy of probes, as shown in Figure~\ref{fig:probing_results_step2}.
  The lower part is the result of max pooling the Success rates from Figure~\ref{fig:sliding_windowed_intervention_results} in the layer direction.  
  }
  \label{fig:pooled_intervention_results}
\end{figure*}

\subsection{Analysis}
\label{subsec:probing_analysis}
In this section, we present the results of a prompt-sensitivity analysis and an error analysis.
We also analyze the effects of factors such as reasoning-chain format (including implicit reasoning).
For details, see~\cref{appendix:subsec:alternative_cot_formats}.

\paragraph{Equation order.}
The difference between these two tasks lies in the order of the equations.
Level 3 tasks involve forward references, whereas Level 2 tasks do not.
An autoregressive language model can attend only to past inputs. 
Therefore, the model may be able to reach the final answer more quickly on Level 2 tasks than on Level 3 tasks.
In practice, Table~\ref{table:rsp_task_dimmension} indicates that the probe predicts the answer during CoT at both levels.
Therefore, the impact of equation order (dependency structure) on the model’s internal mechanism is thought to be limited.

\paragraph{Distractors.}
In the Level 4 task (see Table~\ref{table:tasks}), the variable $v_3$ (\texttt{C}) is a distractor, that is, $\$\{v_3\}$ is not necessary to derive the final answer.
The models can infer this fact from the few-shot examples.
According to Table~\ref{table:rsp_task_dimmension}, the $\mathrm{Acc}(v_3)$ in Level 4 was at most 44\%, a relatively low accuracy.
From this result, we can see that, unlike the variables required to derive the final answer, $v_3$ is not encoded in a simple form that can be extracted by a linear transformation alone.
This suggests the possibility that the model employs an efficient internal mechanism that does not derive variables unnecessary for obtaining the final answer.
It is also consistent with the finding that the CoT is faithful to the final answer.

\paragraph{Effect of depth of reasoning.}
The difference between Level 3 and Level 5 is whether two or three calculation steps are required to reach the final answer.
Qualitatively, the probe’s prediction accuracy for each variable increases a few tokens before the answer appears in the output text, indicating alignment between the internal state and the output.
This finding supports the conclusion that the generated explanations are faithful to the internal state and that this conclusion generalizes across levels.

\begin{table}[t!]
  \tabcolsep 0.5mm
  \small
  \centering

\begin{tabular}{lccrrrr}
\toprule
&\multicolumn{2}{c}{Variable} &\multicolumn{2}{c}{When ($\downarrow$)} & \multicolumn{2}{c}{$\mathrm{Acc}$ ($\uparrow$)} \\
\cmidrule(lr){2-3} \cmidrule(lr){4-5} \cmidrule(lr){6-7}
Setting & Variable  & \#Step & $t_\mathrm{eq}^*$ & $t_\mathrm{eq}^\dagger$ &  $\mathrm{\prec CoT}$ & $\mathrm{\succ CoT}$ \\ 
\midrule
Same level & $v_1$ & 2 & 5 & $-$3  & 17.2 & 100 \\
  & $v_2$ & 1 & 2 & $-$3  & 47.7 & 100 \\
  & $v_3$ & 1 & N/A & $-$2 & 43.7 & 23.7 \\
\midrule
General & $v_1$ & 2 &  5  & $-$3 & 18.1 & 94.7 \\
   & $v_2$ & 1 & 2 & $-$3 & 49.4 & 100 \\
   & $v_3$ & 1 & N/A & $-$2 & 43.3 & 28.1 \\
\bottomrule
\end{tabular}

  \caption{
  Evaluation results of same level prompting and general prompting.
  Each column is the same as Table~\ref{table:rsp_task_dimmension}.
  }
  \label{table:rsp_prompt_comparision}
\end{table}

\paragraph{Prompt differences.}
We investigated whether the format of few-shot prompting influences the model's internal reasoning patterns.
We analyzed the differences between two scenarios: \emph{1. Same level prompting} and \emph{2. General prompting}.
In same level prompting, we provide three equations that match the level of the evaluation task, following the general setting (\cref{sec:general_setting}).
In contrast, under general prompting the model receives 50 randomly generated demonstrations (three equations each).\footnote{We increased the number of demonstrations in the general prompts because task performance was otherwise insufficient.}
Table~\ref{table:rsp_prompt_comparision} reports $t_\mathrm{eq}^*$ for both strategies for Level 4 task.
We observe no difference in $t_\mathrm{eq}^*$, suggesting that the few-shot prompt format has minimal effect on the model's internal reasoning patterns.

\paragraph{Behavior in incorrect instances.}
Figure~\ref{fig:error_analysis} shows the top-1 predictions of the probe for $\mathrm{B}$ in instances where Llama3.2-3B produced \textit{incorrect} answers for level 3 (where \texttt{B} happened to be 4).
When the model produced an incorrect answer, we often found that the correct answer had appeared at an earlier decoding step (see~\cref{appendix:subsec:error_analysis} for the probe’s predictions on additional samples).
The underlying reason for these mistakes remains unclear, but through our probing framework, one can track how such an incorrect answer is propagated to the subsequent generation, which might be practically helpful in debugging the model internals.

\section{Causal interventions}
\label{sec:intervention}

From our probing experiments in \cref{sec:probing}, we tentatively concluded that the final answer is determined after CoT-style generation begins. 
That is, CoT is faithful to the final answer.
In this section, we corroborate this conclusion using causal intervention analysis and further clarify how input information flows internally while generating CoT reasoning and the final answer.

\subsection{Settings}
\paragraph{Activation patching.}
We employ activation patching~\cite{NEURIPS2020_92650b2e,NEURIPS2022_6f1d43d5,zhang2024towards}, which is a widely adopted technique for causal intervention analysis.
To inspect the causal relationship between specific hidden states $\bm h_{t,l}$ and a final answer $y$, we compare two generation scenarios: (i) the ordinary inference and (ii) the intervened inference.
In the latter scenario, we replace the specific hidden states $\bm h_{t,l}$ with other variants $\Tilde{\bm h}_{t,l}$ obtained from the same model but with a different input $\Tilde{x}$ (Intervention in Figure~\ref{fig:intervention_overview}).

The input $x$ and $\Tilde{x}$ have different correct answer $y$ and $\Tilde{y}$ as well as different chains $z$ and $\Tilde{z}$, respectively.
For example, for the triple ($\Tilde{x} =``$\texttt{A=1+B,B=2+4;A=?},'' $\Tilde{z} = ``$\texttt{A=1+B,B=2+4,B=6,A=1+B,A=1+6,A=7},'' $\Tilde{y}=$\texttt{7}), one may use ($x =``$\texttt{A=2+B,B=1+3;A=?},'' $z = ``$\texttt{A=2+B,B=1+3,B=4,A=2+B,A=2+4,A=6},'' $y=$\texttt{6}).
If the model's output turns from $y$ into $\Tilde{y}$ or $z_{t}$ into $\Tilde{z_{t}}$ due to the intervention to $\bm h_{t,l}$ with $\Tilde{\bm h}_{t,l}$, we can confirm the causal relationship between $\bm h_{t,l}$ and the original answer $y$.
Henceforth, to precisely explain the setting, we denote the model's final output with intervention as $\hat{y}^{\mathrm{patch}}$ and that without intervention as $\hat{y}$, respectively ($y$ and $\Tilde{y}$ denote respective gold answers).
In the same way, we denote the generated reasoning chain with intervention as $\hat{z_{t}}^{\mathrm{patch}}$ and that without intervention as $\hat{z_{t}}$, respectively.

\paragraph{Evaluation metrics.}
We report \emph{Success Rate} as a metric for this experiment.
The Success rate indicates how frequently (\%) the intervened output $ \hat{y}^{\mathrm{patch}}$ aligns with the correct answer $\Tilde{y}$.
For reasoning chains, we report the Success rate for $\hat{z_{t}}^{\mathrm{patch}}$ as well.

\paragraph{Patching targets.}
We specifically focus on Level 3 tasks and Qwen2.5-7B.
The experimental results for the other models are reported in~\cref{appendix:subsec:causal_intervention_results}.
Inspired by sliding window patching~\cite{zhang2024towards}, we partition the hidden states into coarse grids, corresponding to each equation and every four layers, and perform activation patching on each grid separately (illustrated in Figure~\ref{fig:pooled_intervention_results}).\footnote{All the hidden states in each grid are intervened at once.}
For every grid, we compute the \emph{Success rate} by applying activation patching.  
We also examine multiple target tokens, specifically, at (i) the end of the equation 2 ($z_{17}$ in ${\underline{\textcolor{gray}{\mathrm{B}=}\textbf{5}}}_{\hspace{0.05cm}\mathbf{2}}$ in Figure~\ref{fig:probing_results_step2}), (ii) the end of the equation 4 ($z_{32}$ in ${\underline{\textcolor{gray}{\mathrm{A}=1+}\textbf{5}}}_{\hspace{0.05cm}\mathbf{4}}$ in Figure~\ref{fig:probing_results_step2}), and (iii) the final answer ($y$).
When we apply activation patching, we generate only the target token with greedy decoding while forced-decoding the context.
Note that the above equations are examples.
We create a test set of 2,000 instances for evaluation.

\subsection{Results.}
\paragraph{CoT is faithful to the final answer.}
Figure~\ref{fig:sliding_windowed_intervention_results} presents the success rate at each grid position when the final answer $y$ (${\underline{\textcolor{gray}{\mathrm{A}=}\textbf{6}}}_{\hspace{0.05cm}\mathbf{2}}$) is the target token.
The results show a strong causal dependence of the final answer on the \textsc{Output} part, whereas the \textsc{Input} part has only a limited effect.
We observe the same pattern for other target tokens (e.g., $z_{17}$ and $z_{32}$).
This trend is broadly consistent across models (\cref{appendix:subsec:causal_intervention_results}).
Furthermore, we focus on the generation of ${\underline{\textcolor{gray}{\mathrm{B}=}\textbf{5}}}_{\hspace{0.05cm}\mathbf{2}}$, which is a necessary subanswer to reach the final answer (see Figure~\ref{fig:pooled_intervention_results}).
Intervention to the immediately preceding equation ${\underline{\textcolor{gray}{\mathrm{B}=2+3}}}_{\hspace{0.05cm}\mathbf{1}}$ affects the generation of ${\underline{\textcolor{gray}{\mathrm{B}=}\textbf{5}}}_{\hspace{0.05cm}\mathbf{2}}$, but patching the equation ${\underline{\textcolor{gray}{\mathrm{B}=2+3}}}_{\hspace{0.05cm}\mathbf{-2}}$ in the problem statement $x$ (the region where a slight improvement in probing accuracy before the CoT was observed; $t=-11$ to $t=-4$) had almost no effect on the output of ${\underline{\textcolor{gray}{\mathrm{B}=}\textbf{5}}}_{\hspace{0.05cm}\mathbf{2}}$.
Similarly, when we focus on the case where the target token is ${\underline{\textcolor{gray}{\mathrm{A}=1+}\textbf{5}}}_{\hspace{0.05cm}\mathbf{4}}$, patching the equation ${\underline{\textcolor{gray}{\mathrm{B}=2+3}}}_{\hspace{0.05cm}\mathbf{-2}}$ again resulted in almost no change in the output.
Therefore, the influence of the information processed in the \textsc{Input} part on the sub-answers ($\textcolor{gray}{\mathrm{B}=}\textbf{5}$) is limited. %
This tendency is especially strong in the Qwen family (except for Qwen2.5-Math-7B).
By contrast, other model families show a weak causal relationship between ${\underline{\textcolor{gray}{\mathrm{B}=}\textbf{5}}}_{\hspace{0.05cm}\mathbf{2}}$ and the \textsc{Input} part (\cref{appendix:subsec:causal_intervention_results}).
This aligns with the results observed in probing experiments~\cref{subsec:experiments}.

\paragraph{Recency bias.}
We further investigate which parts of CoT output causally depend on which input.
The bottom part of Figure~\ref{fig:pooled_intervention_results} suggests strong \textit{recency bias} in the causal relationship between hidden states and output tokens for Qwen2.5-7B.
That is, intervention succeeded only when the target hidden state is (i) in the same equation as the target token, (ii) in the last equation where necessary information is stated (e.g., \texttt{B=2+3}$\rightarrow$\texttt{B=5}), or (iii) in the last equation where a value of a relevant variable is explicitly stated (e.g., \texttt{B=5}$\rightarrow$\texttt{A=1+5}).
This finding suggests strong recency bias in the internal process of LLMs' multi-hop reasoning.
This aligns with our finding that models derive an answer during CoT on the fly, and the generated reasoning chains can be regarded as faithful reflections of the model's final answer.

\section{Related Work}
\label{sec:related_work}

\paragraph{Post hoc nature of CoT.}
Several prior studies investigate similar questions~\cite{afzal-etal-2025-knowing,ye2025physics,cox2025-posthoc-cot,DBLP:journals/corr/abs-2508-19827}.
For instance, \citet{cox2025-posthoc-cot} examined whether CoT explanations are post-hoc or not, specifically, confirmed that, in a binary classification task, the correct label can be predicted from the hidden state at the end of the input (prompt).
In their setting, the CoT $z$ is (freely) produced from the intervened hidden state $x$, while we force-decode the CoT chain. 
In this sense, their scope is rather on the causality of $x \rightarrow z$, and thus, as stated in~\cref{subsec:scope}, the scope is slightly different.
They also showed that, in some tasks (e.g., sports understanding), the final answer can be predicted with the model's final state in the prompt, apparently contradicting our probing experiments, but their tasks would be answerable with factual or commonsense knowledge without CoT, and in their specific logical deduction task---where on-the-fly reasoning is required, similarly to our arithmetic tasks---the probe indeed struggles to predict the final answer, similarly to our results.
\citet{afzal-etal-2025-knowing} also suggests the model reaches the answer internally before CoT begins, but their experimental design is somewhat different from ours.
Specifically, they try to predict whether the model will succeed/fail the task before CoT begins as a binary problem, rather than tracing how models reach the correct answer during CoT or identifying where the concrete answer comes up with during CoT reasoning.
\citet{ye2025physics} also find that the model, in advance, anticipates which components (variables) will be required to reach the final answer by tracking whether a variable has been computed, and how models retain computed values during reasoning.
Although the experiments are similar, they do not investigate causal relationships and therefore do not fully answer the question of whether the reasoning chain is faithful.

\paragraph{Faithfulness of CoT.}
\citet{paul-etal-2024-making, bentham2024chainofthought} similarly evaluate the faithfulness of CoT to the final answer.
They test the consistency between the final answer and the CoT by intervening on the CoT text.
Specifically, they replace it with counterfactual content and then check whether the final answer changes.
They report that CoT is not sufficiently faithful to the final answer.
In addition, \citet{NEURIPS2023_ed3fea90,DBLP:journals/corr/abs-2505-05410} define CoT as faithful when the following two conditions hold:
(i) The CoT explicitly refers to the hint provided in the prompt.
(ii) The model relies on the hint to reach the answer (that is, without the hint, the model would not have produced the hinted answer).
Using these conditions, they evaluate faithfulness indirectly by testing whether the model's internal computation depends on the hinted information.

By contrast, in our work, we trace the model's internal reasoning more directly by probing the values of intermediate answers, and thus track the faithfulness of CoT to the final answer using a more direct method and a controlled testbed than these prior studies.
Through token-level probing, we analyze when and at which token position the reasoning is actually carried out.
In this respect, our work differs in that it also allows us to observe the faithfulness of the internal states to the CoT text along the temporal axis.

\paragraph{Interpreting multi-hop reasoning in language models.}
Interpreting internal mechanisms of LLMs has been actively investigated~\cite{conneau-etal-2018-cram,tenney2018what,niven-kao-2019-probing,nostalgebraist2020logitlens,geva2023dissecting,DBLP:journals/corr/abs-2408-05147,DBLP:conf/icml/GhandehariounCP24,ferrando-voita-2024-information}. 
They revealed, for example, specialized attention heads responsible for specific operations \citep{cabannes2024iteration} or decision-making, copying, and induction \citep{dutta2024how}. 
With a more concrete example, \citet{yang-etal-2024-large-language-models} showed that, even during the first pass of the problem statements such as \textit{The mother of the singer of Thriller is \_\_\_}, language models first resolve a \emph{bridge entity}, Stevie Wonder in this case, then identify the final answer.
This study is more focused on the difference between the first pass of the problem statements (before CoT generation) and their second pass involving explicit problem solving (while CoT generation).

\paragraph{Arithmetic representations in LLMs.}
How models handle numerical information has also been closely studied. For instance, 
\citet{heinzerling-inui-2024-monotonic} used partial least squares regression~\cite{WOLD2001109} to demonstrate that numeric attributes, such as birth years and population numbers, are encoded as monotonic representations in the activation space of LLMs and can be manipulated with interventions. 
In turn, \citet{stolfo-etal-2023-mechanistic} showed that, in autoregressive LLMs, the operations and numerical information necessary for solving quantitative reasoning are processed in the lower layers of the model, and these results are used by the attention layers to predict the final calculation outcomes. 
\citet{zhu-etal-2025-language} studied the representation of numbers in language models' hidden states during single-hop arithmetic tasks (e.g., \texttt{What is the sum of 12 and 34?}). Their analysis revealed that numerical information is encoded linearly within the hidden states and demonstrated that individual digits could be manipulated independently. In this study, we add to this literature by introducing incremental arithmetic problem solving, i.e., what numerical information is contained in the model's hidden states at each time step of multi-hop arithmetic reasoning.

\paragraph{Model interpretability methods.} 
Linear probing~\cite{alain2017understanding} is one of the representative methods for analyzing the internal representations of neural models---a small model predicts a specific feature from them, thereby determining whether the input contains information about that feature. 
In this study, we use them to derive the models' intermediate answers.
The causality with the model's output can be further verified by examining if a model's predictions change when the hidden states are intervened~\cite{li2023emergent,wu2023interpretability}.
One representative intervention method is activation patching~\cite{NEURIPS2020_92650b2e,NEURIPS2022_6f1d43d5,zhang2024towards}, where hidden states obtained from one model instance are transplanted onto another during inference to change its predictions. 
Such techniques can be applied as a way to control model behavior in practical scenarios such as mitigating inherent biases~\cite{zhao-etal-2019-gender,ganguli2023capacity,yang-etal-2024-mitigating}. Here, we employ activation patching to validate the plausibility of the probing results.

\section{Conclusions}
\label{sec:conclusion}
We conducted probing experiments to determine when (sub)answers are derived during the CoT-style reasoning, using synthetic arithmetic problems as a controlled testbed.
Across multiple models and task difficulties, we found that models tend to resolve the necessary (sub)answer after the CoT begins, i.e., computing the answer on the fly during CoT generation.
Moreover, causal experiments support that the intervention to the CoT part specifically and causally impacts the final answer; that is, we conclude that the CoT is faithful to the final answer at least in our controlled experimental settings.

\clearpage

\section*{Limitations}
\label{sec:limitations}

\paragraph{Variety of tasks}
We analyzed the internal reasoning patterns of language models using synthetic arithmetic reasoning tasks.
The use of synthetic data allows for more detailed control compared to experiments on natural language tasks.
However, vocabulary and expression diversity, for example, are limited compared to natural language tasks.
Therefore, conducting similar analyses on more realistic reasoning tasks in natural language will verify whether the results of this study apply to other broader, realistic contexts as well.
In addition, we focused on the tasks that presumably require step-by-step reasoning as the CoT process, but as suggested in existing studies (\cref{sec:related_work}), the situation may be different if the task is, in the first place, so simple that CoT is no longer needed.

\paragraph{Probing methods}
Interpreting internal mechanisms of LLMs using probing has been actively conducted in our field \cite{conneau-etal-2018-cram,tenney2018what,campbell2023localizing,li2023emergent}; however, there are criticisms regarding the validity of some probing approaches~\cite{liu-etal-2023-cognitive,burns2023discovering}.
One way to overcome such concerns will be to analyze the generality of obtained results through more diverse methodologies~\cite{gurnee2023finding,Bricken2023Monosemanticity}.

\paragraph{Causal intervention purity}
Hidden states carry mixed information (e.g., features of the input text itself).
Thus, even though activation patching is standard in prior work, we cannot fully rule out noise introduced by this intervention.

\section*{Ethics statement}
This paper will not raise particular ethical concerns, considering that (i) no human experiments were conducted, and (ii) our tasks do not involve ethically sensitive topics.

\section*{Acknowledgments}
This work was supported by JST CREST Grant Number JPMJCR20D2, JSPS KAKENHI Grant Numbers JP25KJ0615 and JP25K03175, JST SPRING Grant Number JPMJSP2114, Google Research grant, and the Nakajima Foundation.
Ana Brassard's contribution was supported by a RIKEN Incentive Research Project (FY2024). 
Part of this work was carried out using the computer resource offered under the category of ``General Projects'' by Research Institute for Information Technology, Kyushu University.
We thank the member of the Tohoku NLP Group for their cooperation in this research.

\bibliography{custom}

\newpage
\appendix

\begin{table}[t]
  \scriptsize
  \centering
  \tabcolsep  2pt
  \begin{tabular}{cll}
  \toprule
  & \textsc{Input} & \textsc{Output} \\
  \midrule
  Simple & \({\underline{\mathrm{A}=1+\mathrm{B}}}_{\hspace{0.05cm}\mathbf{-3}}\), \({\underline{\mathrm{B}=2+3}}_{\hspace{0.05cm}\mathbf{-2}}\); \({\underline{\mathrm{A}=?}}_{\hspace{0.05cm}\mathbf{-1}}\) & \({\underline{\mathrm{B}=5}}_{\hspace{0.05cm}\mathbf{0}}\), \({\underline{\mathrm{A}=6}}_{\hspace{0.05cm}\mathbf{1}}\) \\[0.4cm]
  Implicit & \({\underline{\mathrm{A}=1+\mathrm{B}}}_{\hspace{0.05cm}\mathbf{-3}}\), \({\underline{\mathrm{B}=2+3}}_{\hspace{0.05cm}\mathbf{-2}}\); \({\underline{\mathrm{A}=?}}_{\hspace{0.05cm}\mathbf{-1}}\) & \({\underline{\mathrm{A}=6}}_{\hspace{0.05cm}\mathbf{0}}\) \\
  \bottomrule
\end{tabular}

  \caption{
  Examples of arithmetic reasoning tasks used for reasoning chain format comparison. 
  This position is used as a reference point for calculating $t^*_{\mathrm{eq}}$ in \cref{subsec:evaluation_metrics}.
  }
  \label{table:alternative_cot_format_tasks}
\end{table}

\begin{table}[t]
  \tabcolsep 0.5mm
  \small
  \centering \begin{tabular}{lccrrrr}
\toprule
&\multicolumn{2}{c}{Variable} &\multicolumn{2}{c}{When ($\downarrow$)} & \multicolumn{2}{c}{$\mathrm{Acc}$ ($\uparrow$)} \\
\cmidrule(lr){2-3} \cmidrule(lr){4-5} \cmidrule(lr){6-7}
Setting & Variable  & \#Step & $t_\mathrm{eq}^*$ & $t_\mathrm{eq}^\dagger$ &  $\mathrm{\prec CoT}$ & $\mathrm{\succ CoT}$ \\ 
\midrule
Simple CoT & $v_1$ & 2 & 1 & 5 & 34.8 & 99.7 \\
  & $v_2$ & 1 & 0 & 1 & 69.7 & 1.0 \\
\midrule
Implicit & $v_1$ & 2 & N/A & N/A & 40.8 & 80.7 \\
   & $v_2$ & 1 & N/A & N/A & 69.1 & 79.2 \\
\bottomrule
\end{tabular}

  \caption{
  Probing evaluation results with different reasoning chains.
  Each column is the same as Table~\ref{table:rsp_task_dimmension}.
  }
  \label{table:rsp_reasoning_chain_format_comparision}
\end{table}

\section{Alternative CoT formats.}
\label{appendix:subsec:alternative_cot_formats}
For comparison, we also ran probing experiments on Qwen2.5-7B under two reasoning chain formats.
Specifically, we defined two formats.
The first is the Simple CoT setting, in which the reasoning chain outputs only the intermediate sub-results.
The second is the Implicit (reasoning) setting, which omits intermediate computation steps in the output text.
Table~\ref{table:alternative_cot_format_tasks} lists example equations.

\begin{figure*}[p]
    \centering
    \includegraphics[width=0.95\linewidth]{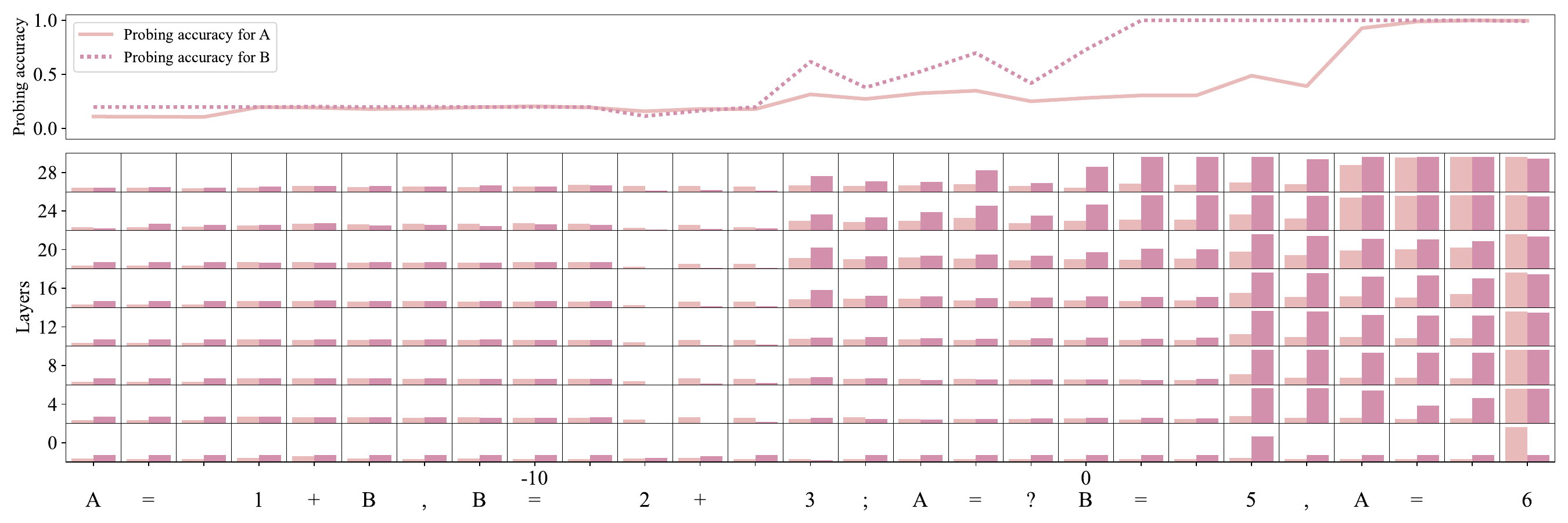}
    \caption{Probing results when Qwen2.5-7B solves Level \texttt{2} with \textbf{Simple CoT}.}
    \label{fig:probing_qwen2.5_7B_task2_only_answer_cot}
\end{figure*}

\begin{figure*}[p]
    \centering
    \includegraphics[width=0.95\linewidth]{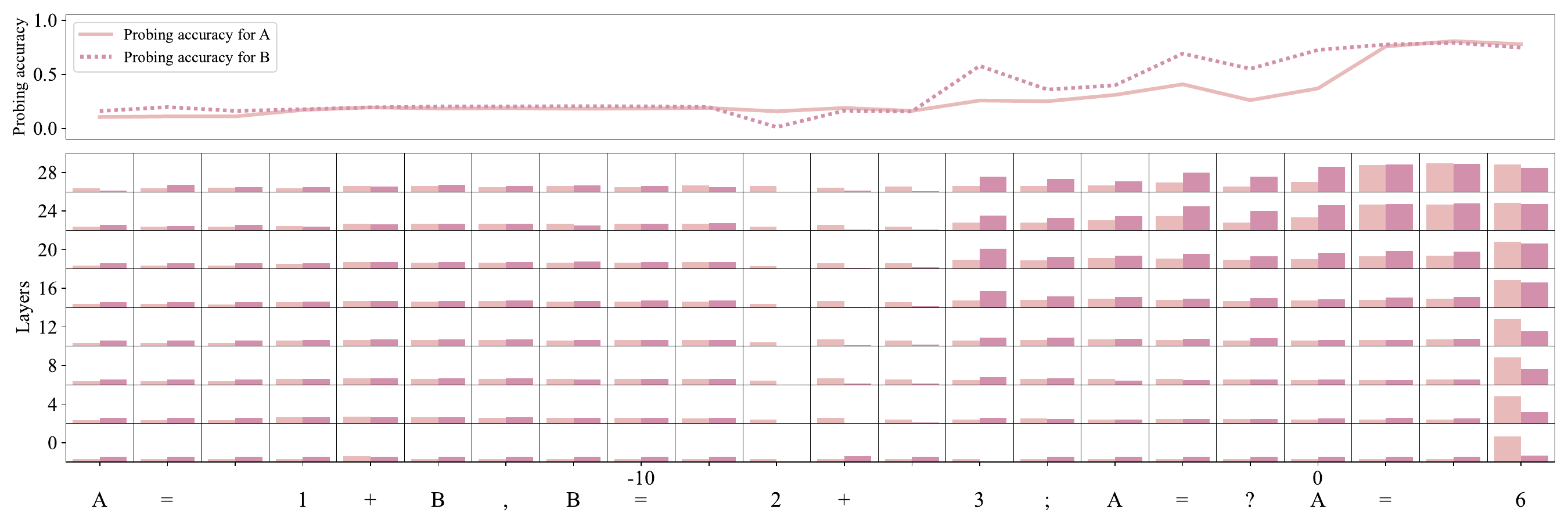}
    \caption{Probing results when Qwen2.5-7B solves Level \texttt{2} with \textbf{Implicit reasoning}.}
    \label{fig:probing_qwen2.5_7B_task2_implicit_reasoning}
\end{figure*}

The experimental results are shown in Table~\ref{table:rsp_reasoning_chain_format_comparision} and Figures~\ref{fig:probing_qwen2.5_7B_task2_only_answer_cot} and~\ref{fig:probing_qwen2.5_7B_task2_implicit_reasoning}.
Under the Simple CoT setting, the model’s task accuracy was 99.5\%, whereas under the Implicit reasoning setting it was 77.8\%\footnote{Please note that, unlike the other experimental configurations, under the implicit reasoning setting the model is unable to answer the problems with sufficiently high accuracy.}.
As in the Simple CoT setting, intermediate results are linearly separable in the \textsc{Output} segment ($t^*_\mathrm{eq}>0$).
This trend is consistent with the results obtained using the default CoT format in \cref{subsec:experiments}.
In the implicit reasoning setting, the accuracy did not reach the threshold at any position (both $\mathrm{Acc}_{\mathrm{\prec CoT}}$ and $\mathrm{Acc}_{\mathrm{\succ CoT}}$ are N/A).
These results suggest that when the model fails to solve a problem, the corresponding (sub-)answers are also unlikely to be represented in its internal states. This, in turn, indicates that the answers decoded from internal states are faithful, in the sense that when the model cannot solve the problem, there may be no linearly decodable representation either.

\section{Supplemental results}
\label{appendix:sec:supplemental_results}

\begin{table}[t]
  \tabcolsep 0.5mm
  \footnotesize
  \centering

\begin{tabular}{lccccc}
\toprule
&  &  & Level &  & \\
& 1 & 2 & 3 & 4 & 5 \\
\midrule
Qwen2.5 (7B) & 100.00 & 100.00 & 100.00 & 100.00 & 100.00 \\
Qwen2.5 (14B) & 100.00 & 100.00 & 100.00 & 100.00 & 98.25 \\ 
Qwen2.5 (32B)  & 100.00 & 100.00 & 100.00 & 100.00 & 100.00 \\
Qwen2.5-Math (7B) & 100.00 & 100.00 & 100.00 & 100.00 & 100.00 \\
Yi1.5 (9B) & 100.00 & 100.00 & 100.00 & 100.00 & 100.00 \\ 
Yi1.5 (34B) & 100.00 & 100.00 & 100.00 & 100.00 & 100.00 \\
Llama3.1 (8B) & 100.00 & 100.00 & 100.00 & 100.00 & 99.40 \\
Llama3.2 (3B) & 99.95 & 99.75 & 93.15 & 90.90 & 38.45 \\
Mistral-Nemo (12B) & 100.00 & 100.00 & 99.95 & 100.00 & 99.85 \\
 \bottomrule
\end{tabular}

  \caption{
    The performance of language models on the arithmetic reasoning tasks.
    The $\mathrm{Task}$ column shows the accuracy for the evaluation set (exact match).
  }\label{table:task_accuracy}
\end{table}

\begin{figure}[t]
  \centering
  \includegraphics[width=1.0\linewidth]{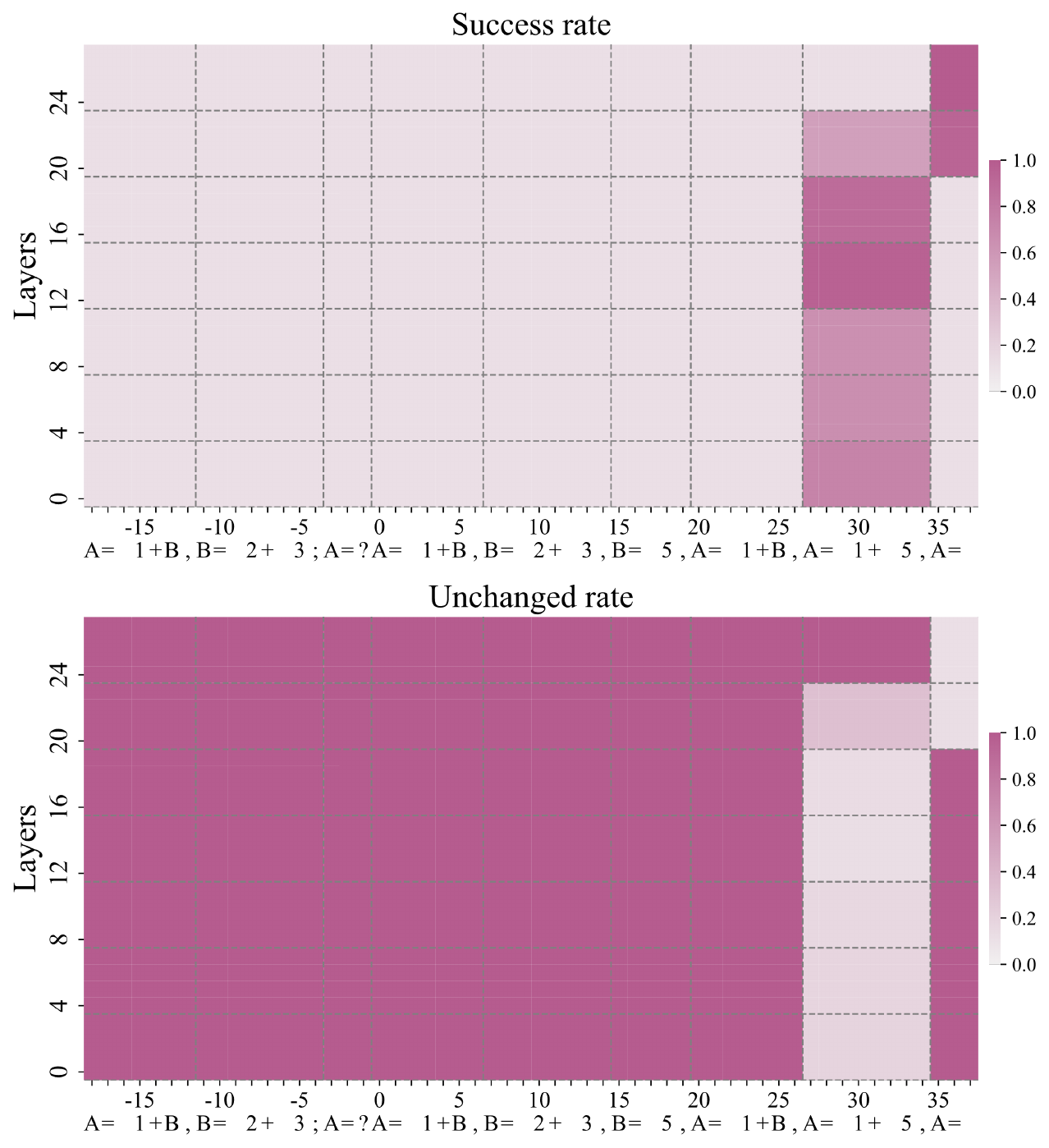}
  \caption{
  Success and Unchanged rates for each grid when the final answer $y$ \(({\underline{\textcolor{gray}{\mathrm{A}=}\textbf{6}}}_{\hspace{0.05cm}\mathbf{5}}\)) is the target token.
  The Success rate heatmap at the top is the same as Figure 5.
  }
  \label{fig:chain_intervention_results_for_final_answer}
\end{figure}

\begin{figure}[t!]
     \centering
     \includegraphics[width=1.0\linewidth]{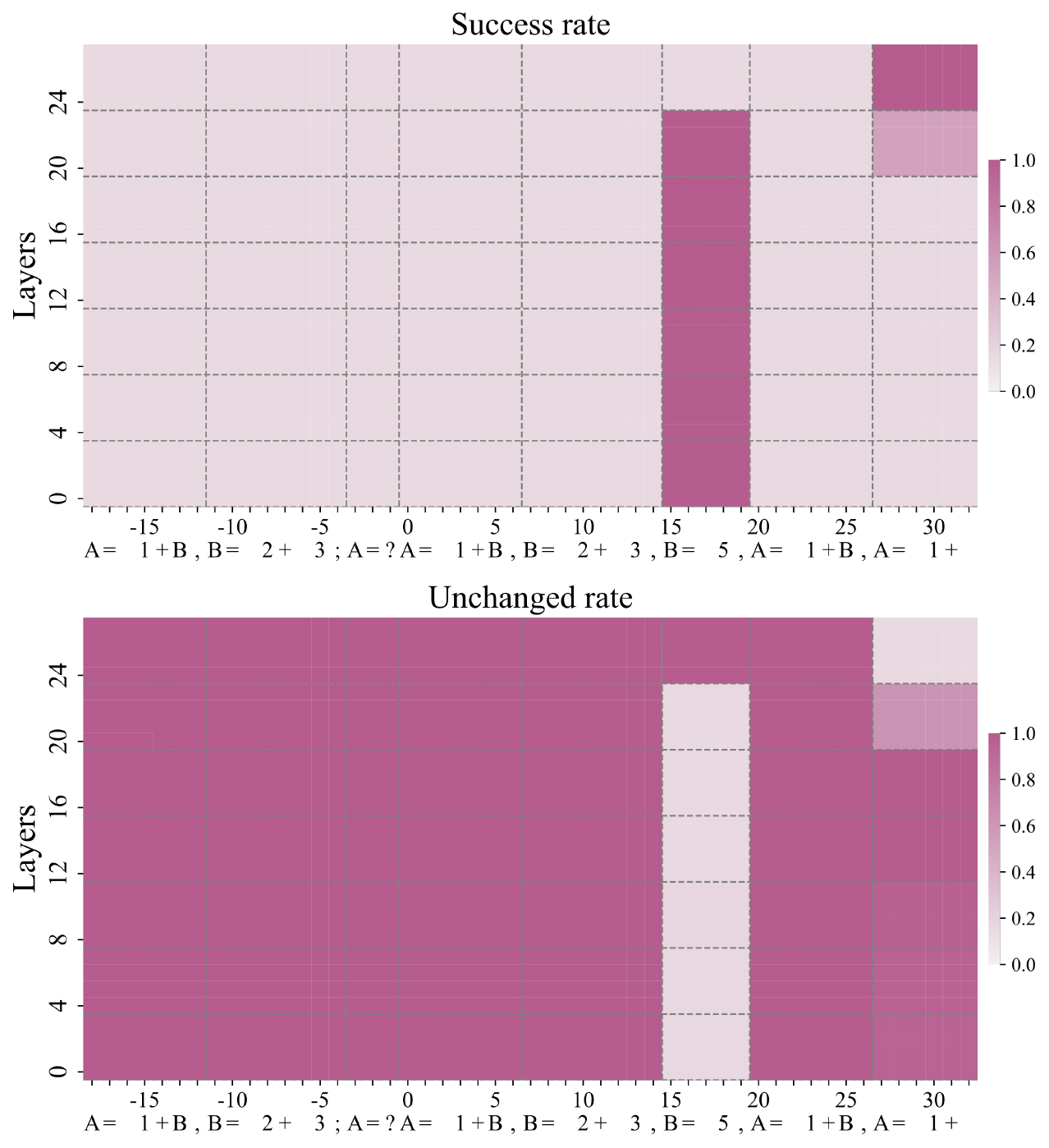}
     \caption{
     Success rate and Unchanged rate for each grid when intervention was performed with $z_{17}$
     (\({\underline{\textcolor{gray}{\mathrm{A}=1+}5}}_{\hspace{0.05cm}\mathbf{4}}\)) as the target token.
     }\label{fig:chain_intervention_results_for_last_5}
\end{figure}

\begin{figure}[t!]
     \centering
     \includegraphics[width=1.0\linewidth]{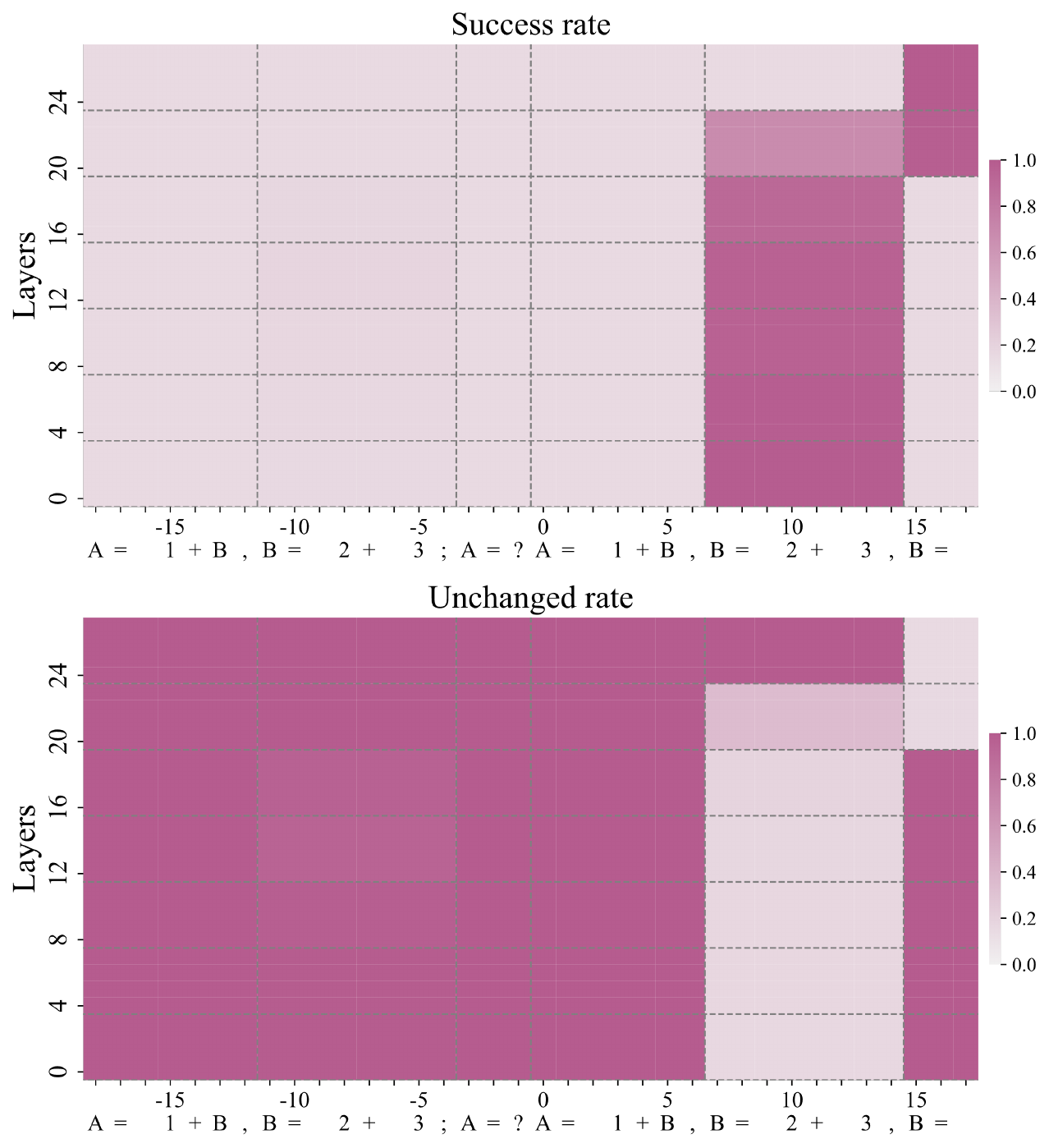}
     \caption{
     Success rate and Unchanged rate for each grid when intervention was performed with
     $z_{32}$
     (\({\underline{\textcolor{gray}{\mathrm{B}=}5}}_{\hspace{0.05cm}\mathbf{2}}\)) as the target token.
     }\label{fig:chain_intervention_results_for_mid_5}
\end{figure}

\subsection{Model performance on arithmetic tasks}
\label{appendix:subsec:performance_of_llms_on_tasks}
Table~\ref{table:task_accuracy} shows the accuracy of language models on arithmetic reasoning tasks for each experimental setting.
We computed the accuracy based on exact matches between the output, including the chain ($\hat{z} \oplus \hat{y}$), and the gold labels ($z \oplus y$). 
The accuracy for all models is nearly 100\%, indicating that they are capable of solving the arithmetic reasoning tasks used in this experiment.

\subsection{Additional probing predictions for correct/incorrect instances.}
\label{appendix:subsec:error_analysis}
\begin{figure*}[t]
    \centering
    \includegraphics[width=0.95\linewidth]{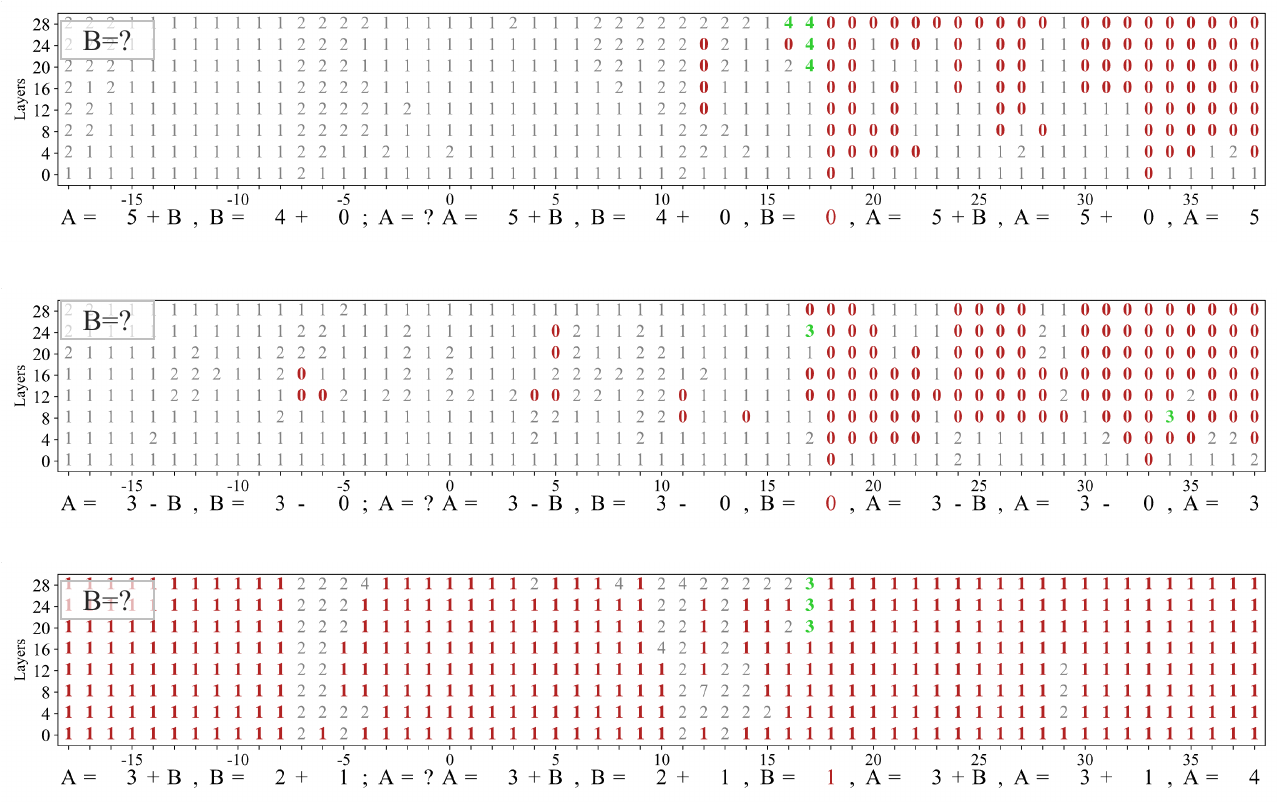}
    \caption{
    Error analysis of cases where Llama3.2-3B generated \emph{incorrect} answers.
    The vertical axis represents the index of the transformer layer.
    The horizontal axis represents the tokens input to the model over time.
    The numbers in the figure indicate the labels predicted by each probe (top-1) at each layer and time step.
    The numbers highlighted in green represent the gold labels for the predictions, while those highlighted in red denote the values incorrectly generated by the model.}
    \label{fig:error_incorrect_analysis}
\end{figure*}

\begin{figure*}[t]
    \centering
    \includegraphics[width=0.95\linewidth]{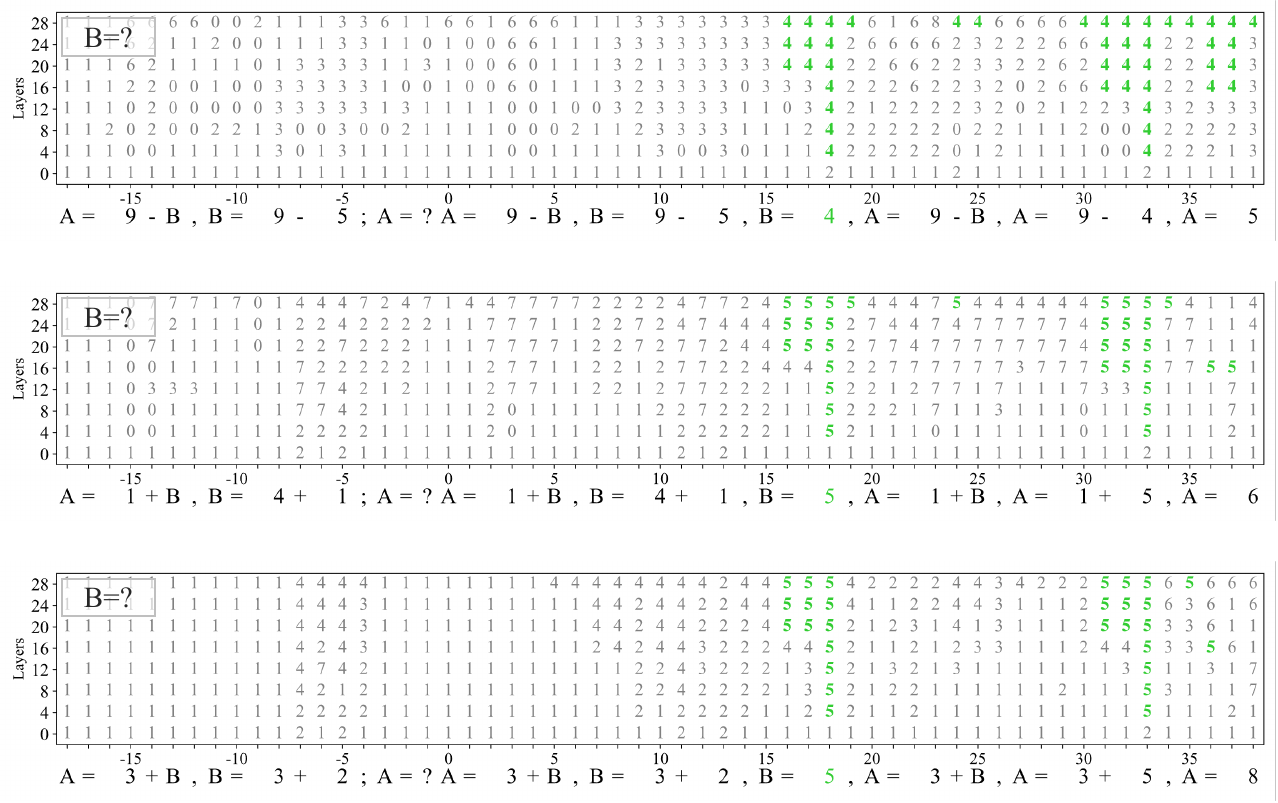}
    \caption{
    Top 1 predictions of probes where Llama3.2-3B generated \emph{correct} answers.
    The vertical axis represents the index of the transformer layer.
    The horizontal axis represents the tokens input to the model over time.
    The numbers in the figure indicate the labels predicted by each probe (top-1) at each layer and time step.
    The numbers highlighted in green represent the gold labels for the predictions.
    }
    \label{fig:error_analysis_correct}
\end{figure*}

Figure~\ref{fig:error_incorrect_analysis} shows the top-1 predictions of the probe for $\mathrm{B}$ in instances where Llama3.2-3B produced \textit{incorrect} answers for Task 3.
In contrast, Figure \ref{fig:error_analysis_correct} illustrates the probe predictions for instances where Llama3.2-3B generates correct responses.
In Figure~\ref{fig:error_analysis} the model should output (\texttt{B=4}) but instead produces \texttt{0}.
We observe the same pattern in other examples in Figure~\ref{fig:error_incorrect_analysis}.
One possible cause is that digits for computed results (e.g., \texttt{4}) and digits from the input (e.g., \texttt{0}) are represented in neighboring subspaces, which may cause confusion; we leave a detailed analysis for future work.

\begin{table}[t]
  \tabcolsep 1mm
  \small
  \centering
  \begin{tabular}{lp{42mm}}
\toprule
 Train instances            &  10,000 \\
 Optimizer      &   SGD~\cite{Robbins1951ASA}  \\
 Learning rate      & $1.0\times10^{-3}$ (constant) \\
 Batch size             & 10,000 \\
 Epochs      &   10,000 \\
\bottomrule
\end{tabular}

  \caption{Hyperparameters for training the probe}
  \label{table:probing_hyperparameter}
\end{table}

\subsection{All probing results}
\label{appendix:subsec:all_probing_results}
\begin{figure*}[p]
    \centering
    \includegraphics[width=0.95\linewidth]{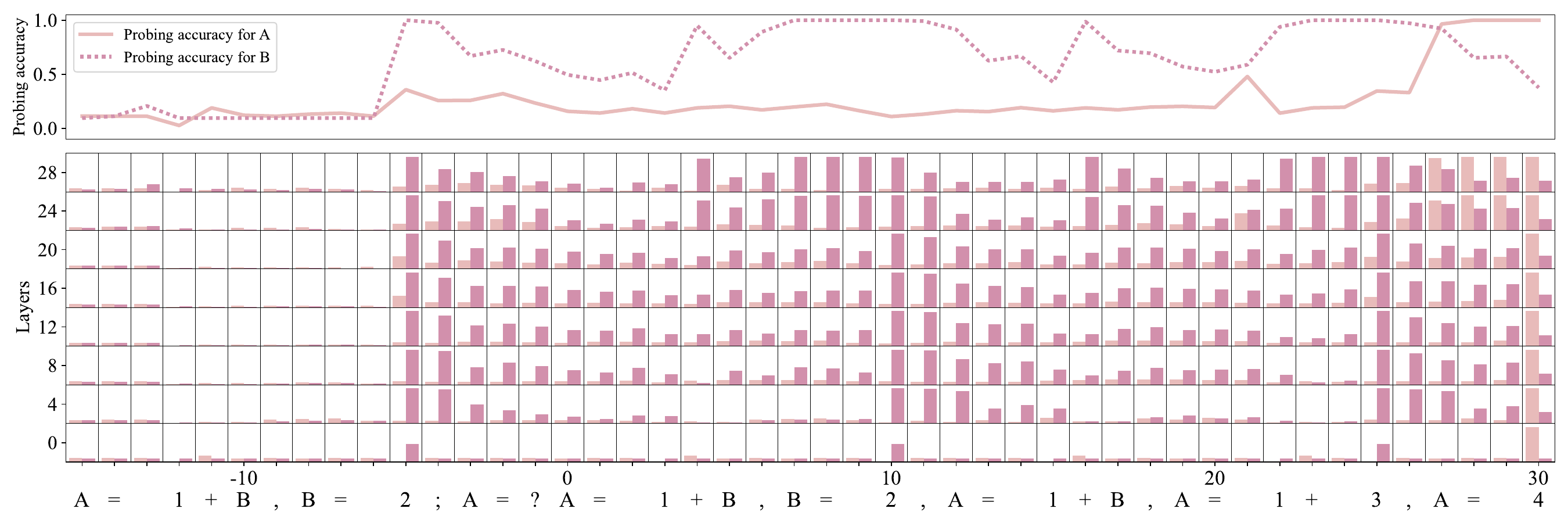}
    \caption{Probing results when Qwen2.5-7B solves Level \texttt{1}.}
    \label{fig:probing_qwen2.5_7B_task1}
\end{figure*}

\begin{figure*}[p]
    \centering
    \includegraphics[width=0.95\linewidth]{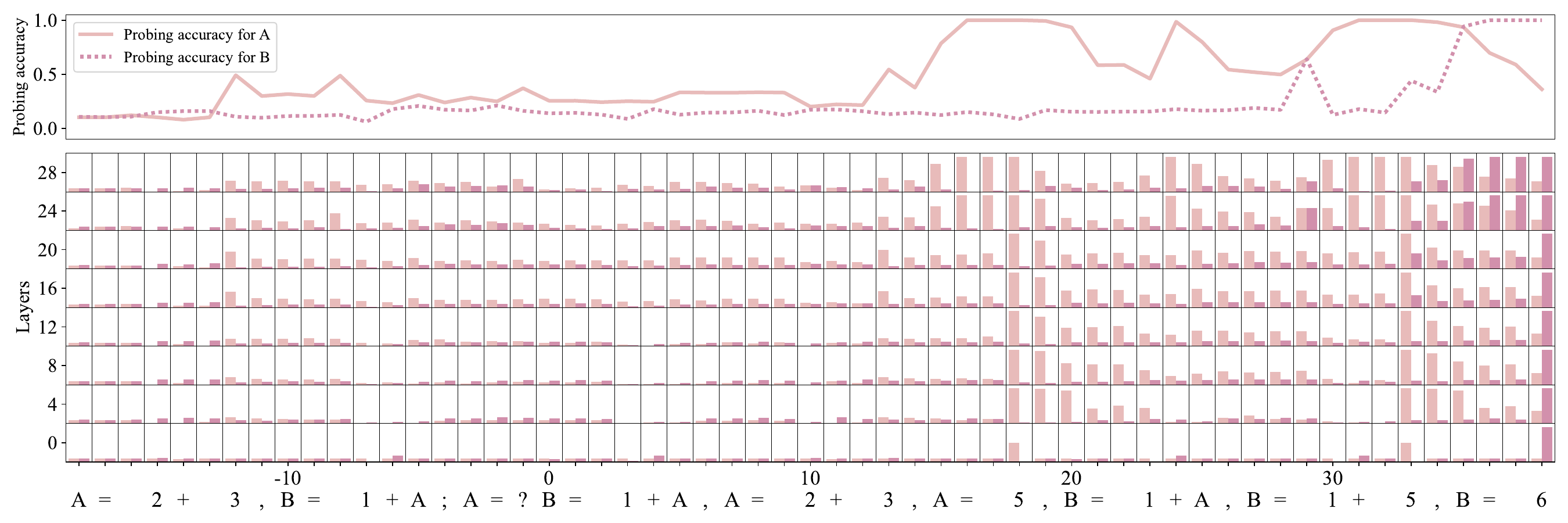}
    \caption{Probing results when Qwen2.5-7B solves Level \texttt{2}.}
    \label{fig:probing_qwen2.5_7B_task2}
\end{figure*}

\begin{figure*}[p]
    \centering
    \includegraphics[width=0.95\linewidth]{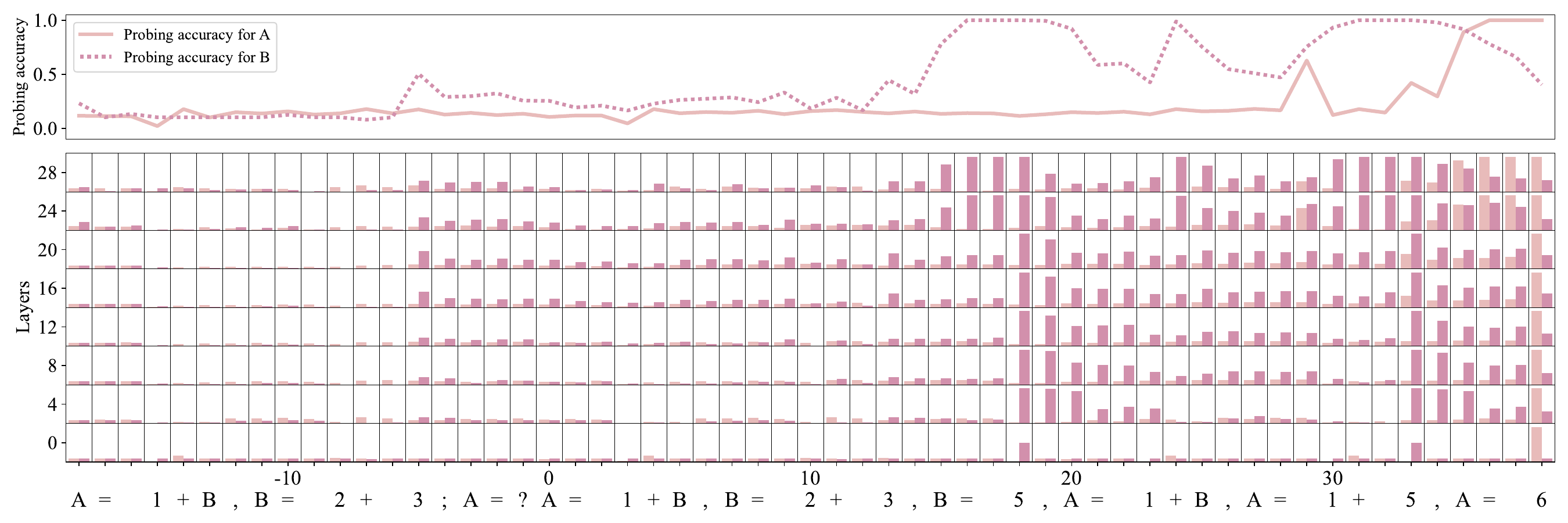}
    \caption{Probing results when Qwen2.5-7B solves Level \texttt{3}.}
    \label{fig:probing_qwen2.5_7B_task3}
\end{figure*}

\begin{figure*}[p]
    \centering
    \includegraphics[width=0.95\linewidth]{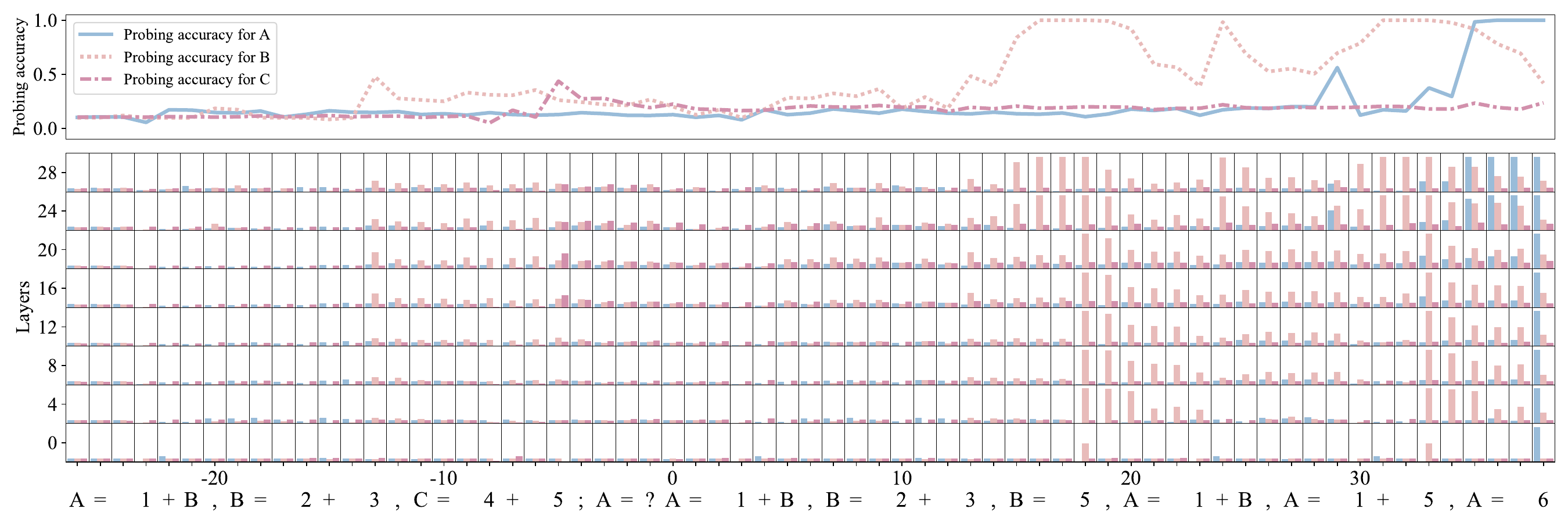}
    \caption{Probing results when Qwen2.5-7B solves Level \texttt{4}.}
    \label{fig:probing_qwen2.5_7B_task4}
\end{figure*}

\begin{figure*}[p]
    \centering
    \includegraphics[width=0.95\linewidth]{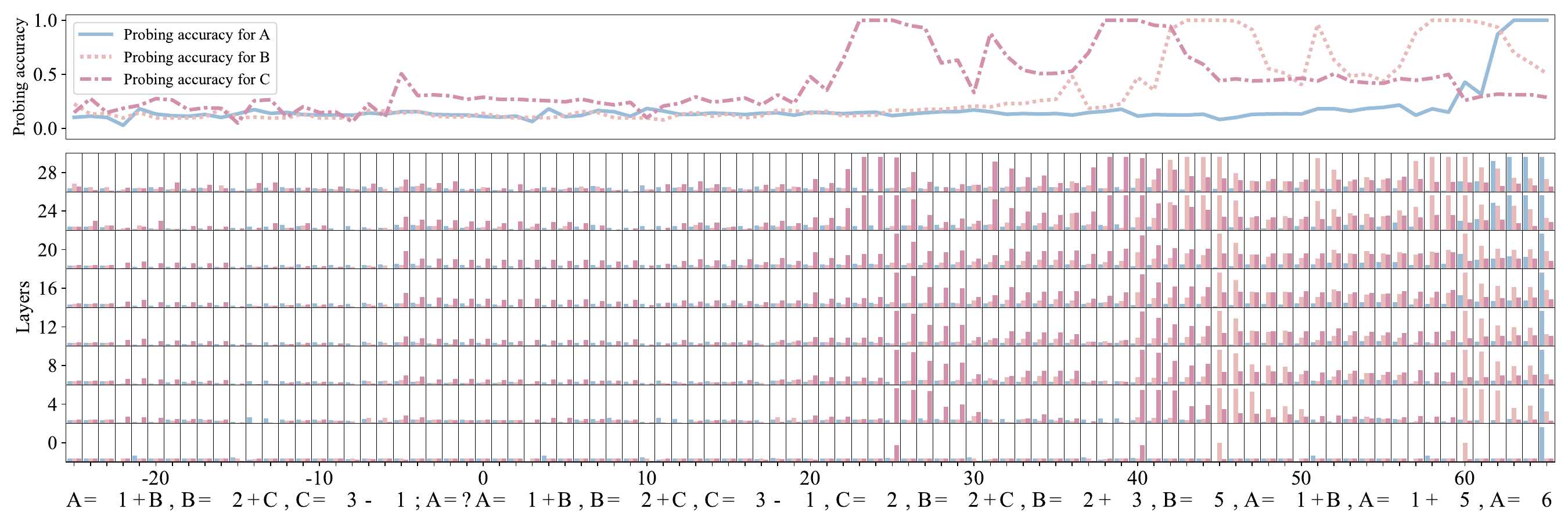}
    \caption{Probing results when Qwen2.5-7B solves Level \texttt{5}.}
    \label{fig:probing_qwen2.5_7B_task5}
\end{figure*}

\begin{figure*}[p]
    \centering
    \includegraphics[width=0.95\linewidth]{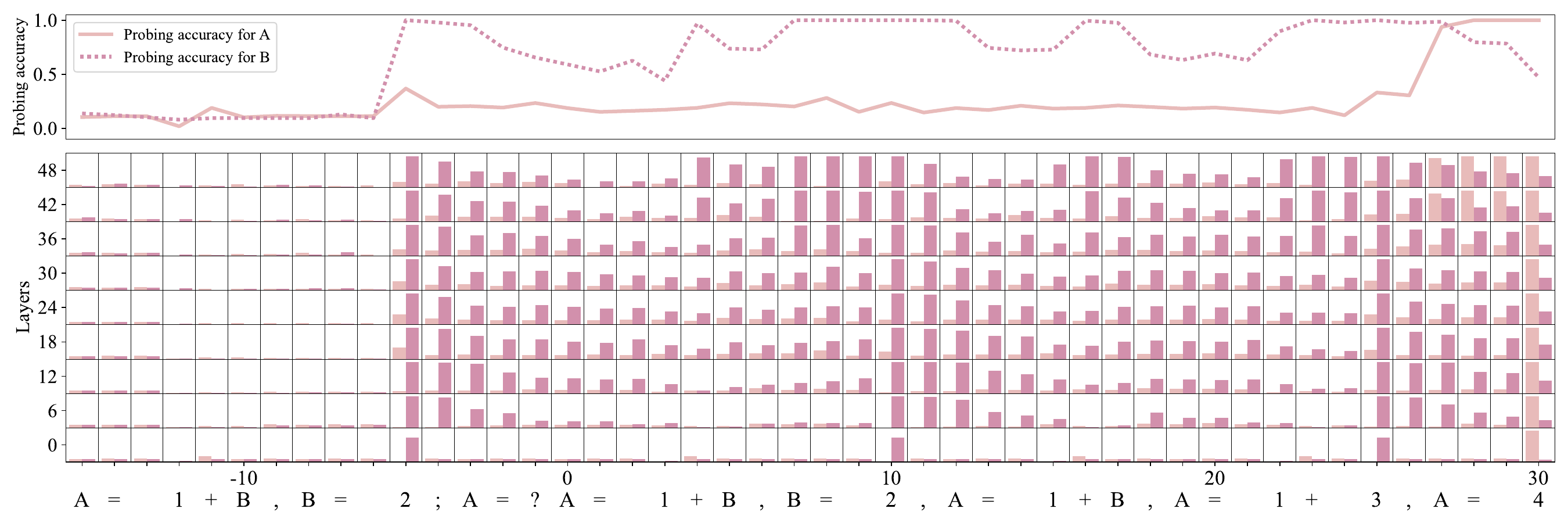}
    \caption{Probing results when Qwen2.5-14B solves Level \texttt{1}.}
    \label{fig:probing_qwen2.5_14B_task1}
\end{figure*}

\begin{figure*}[p]
    \centering
    \includegraphics[width=0.95\linewidth]{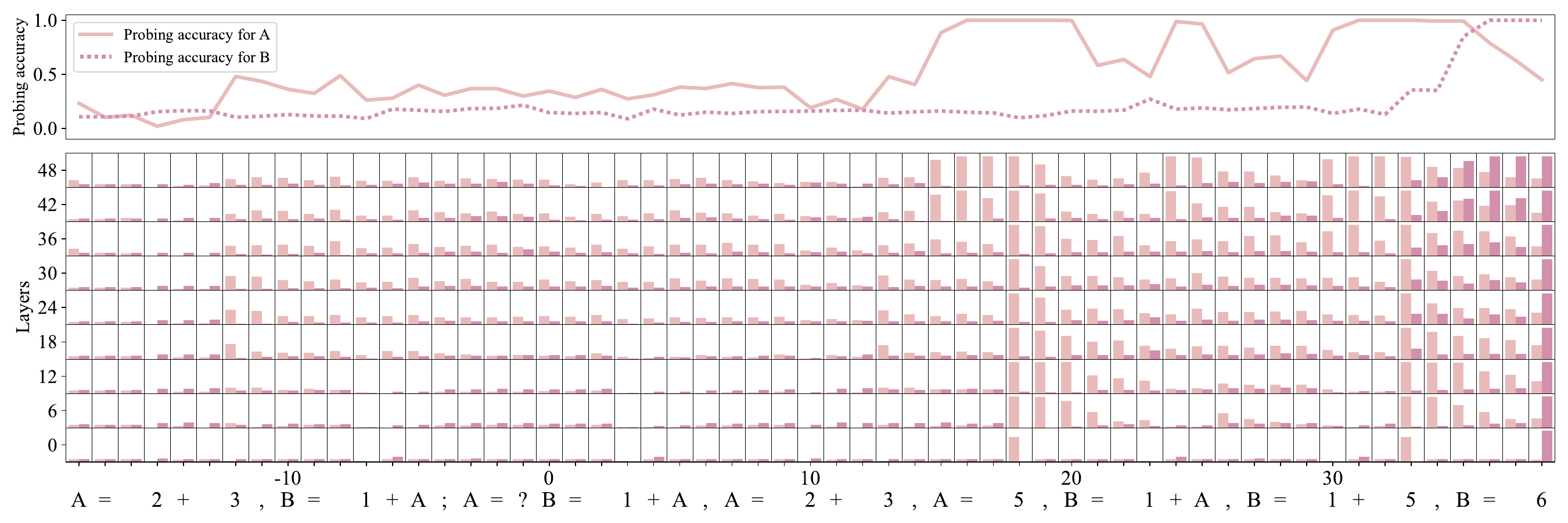}
    \caption{Probing results when Qwen2.5-14B solves Level \texttt{2}.}
    \label{fig:probing_qwen2.5_14B_task2}
\end{figure*}

\begin{figure*}[p]
    \centering
    \includegraphics[width=0.95\linewidth]{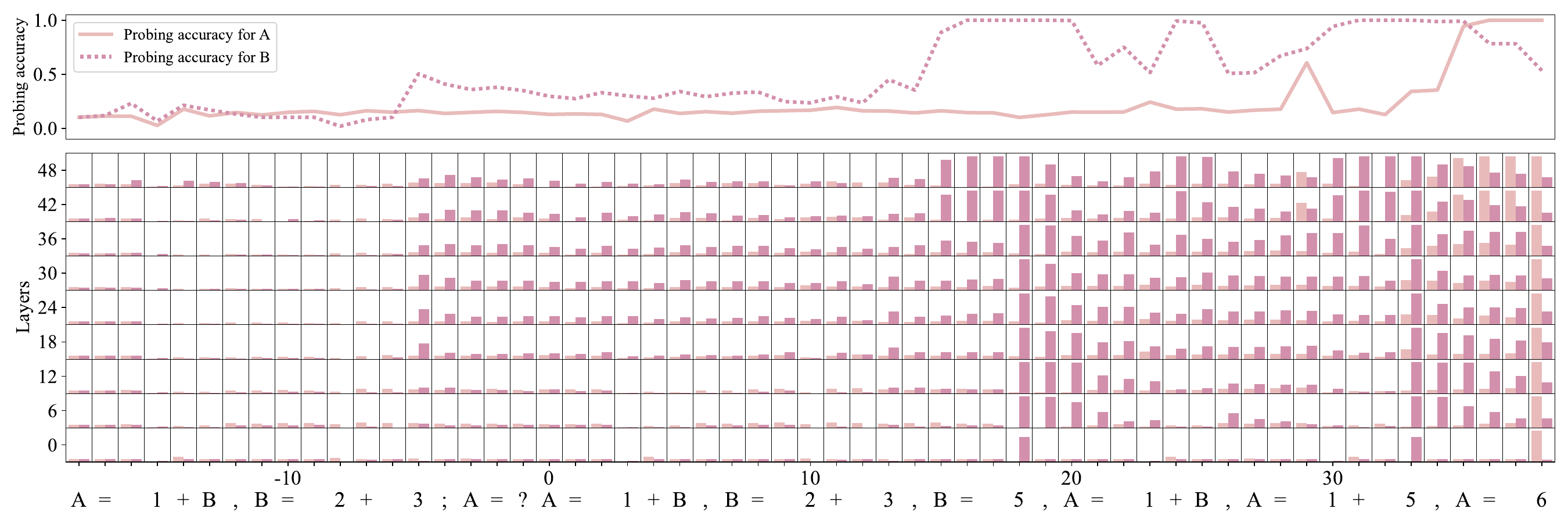}
    \caption{Probing results when Qwen2.5-14B solves Level \texttt{3}.}
    \label{fig:probing_qwen2.5_14B_task3}
\end{figure*}

\begin{figure*}[p]
    \centering
    \includegraphics[width=0.95\linewidth]{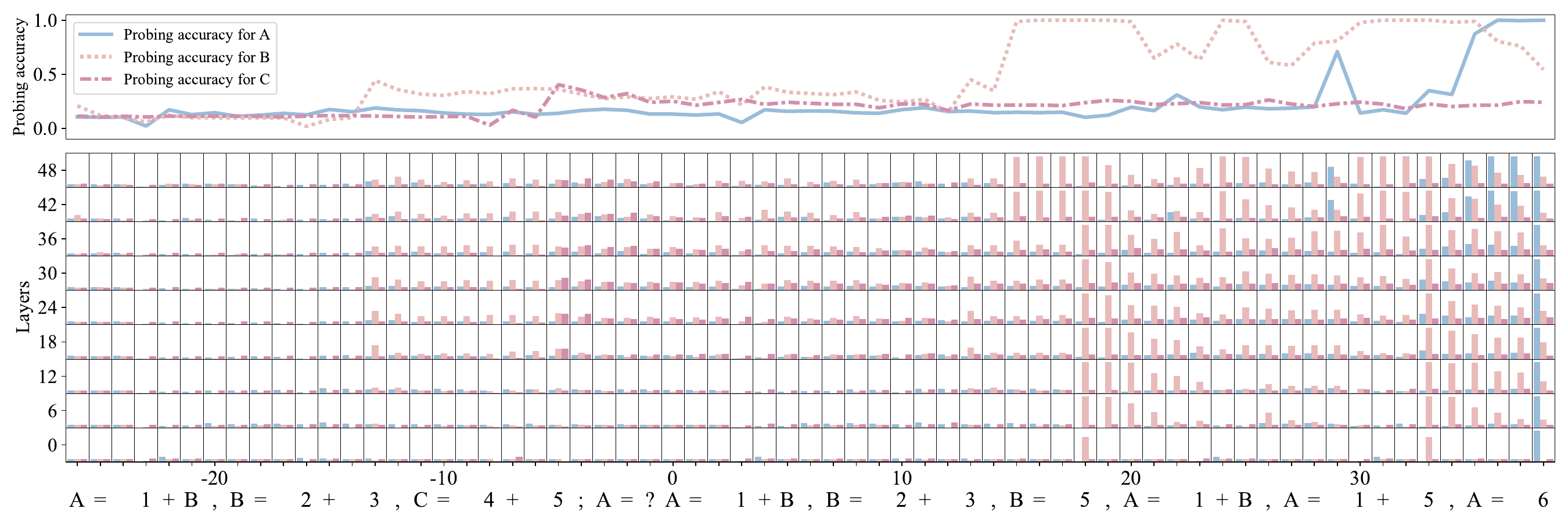}
    \caption{Probing results when Qwen2.5-14B solves Level \texttt{4}.}
    \label{fig:probing_qwen2.5_14B_task4}
\end{figure*}

\begin{figure*}[p]
    \centering
    \includegraphics[width=0.95\linewidth]{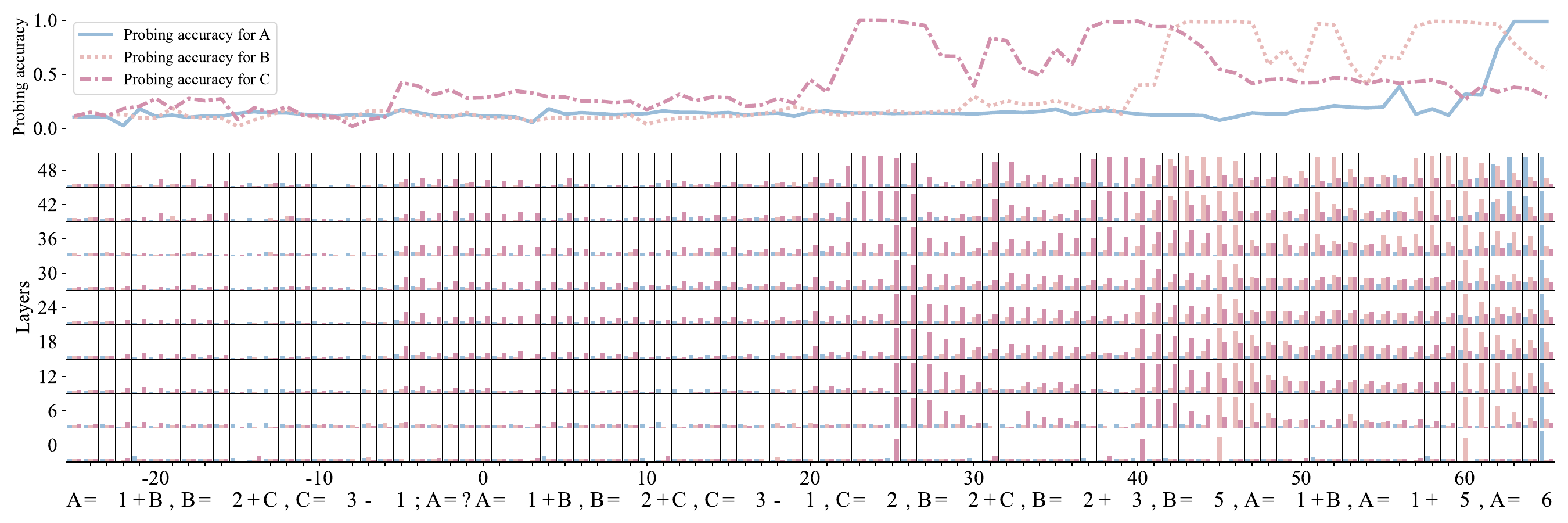}
    \caption{Probing results when Qwen2.5-14B solves Level \texttt{5}.}
    \label{fig:probing_qwen2.5_14B_task5}
\end{figure*}

\begin{figure*}[p]
    \centering
    \includegraphics[width=0.95\linewidth]{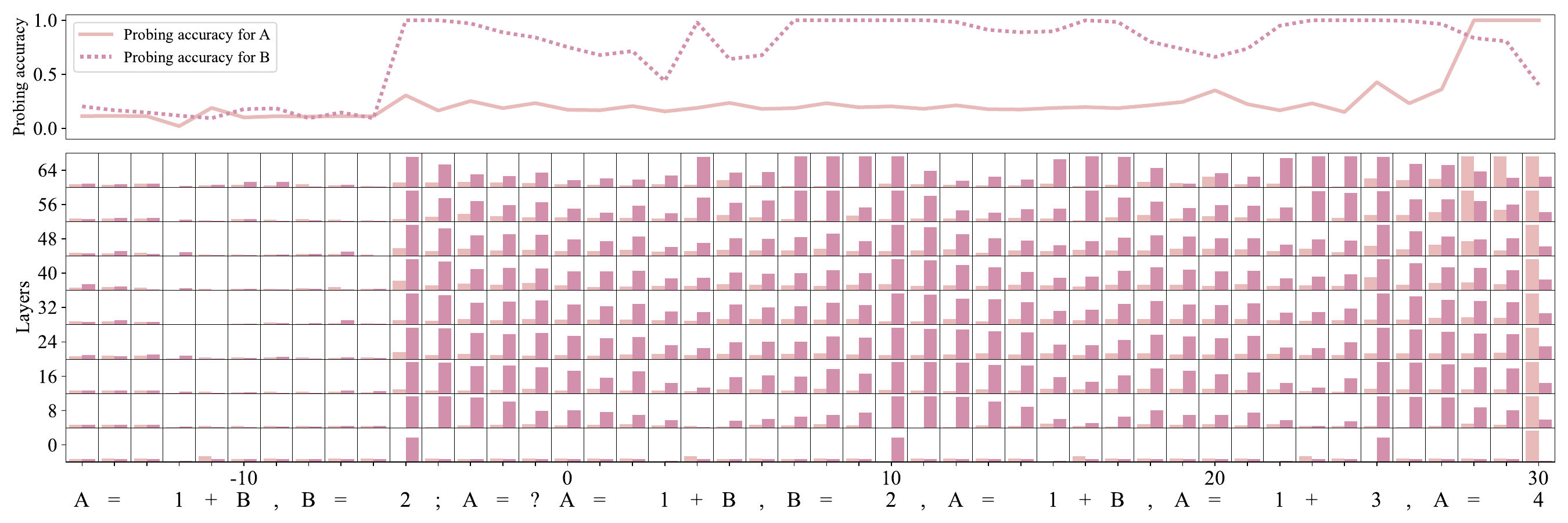}
    \caption{Probing results when Qwen2.5-32B solves Level \texttt{1}.}
    \label{fig:probing_qwen2.5_32B_task1}
\end{figure*}

\begin{figure*}[p]
    \centering
    \includegraphics[width=0.95\linewidth]{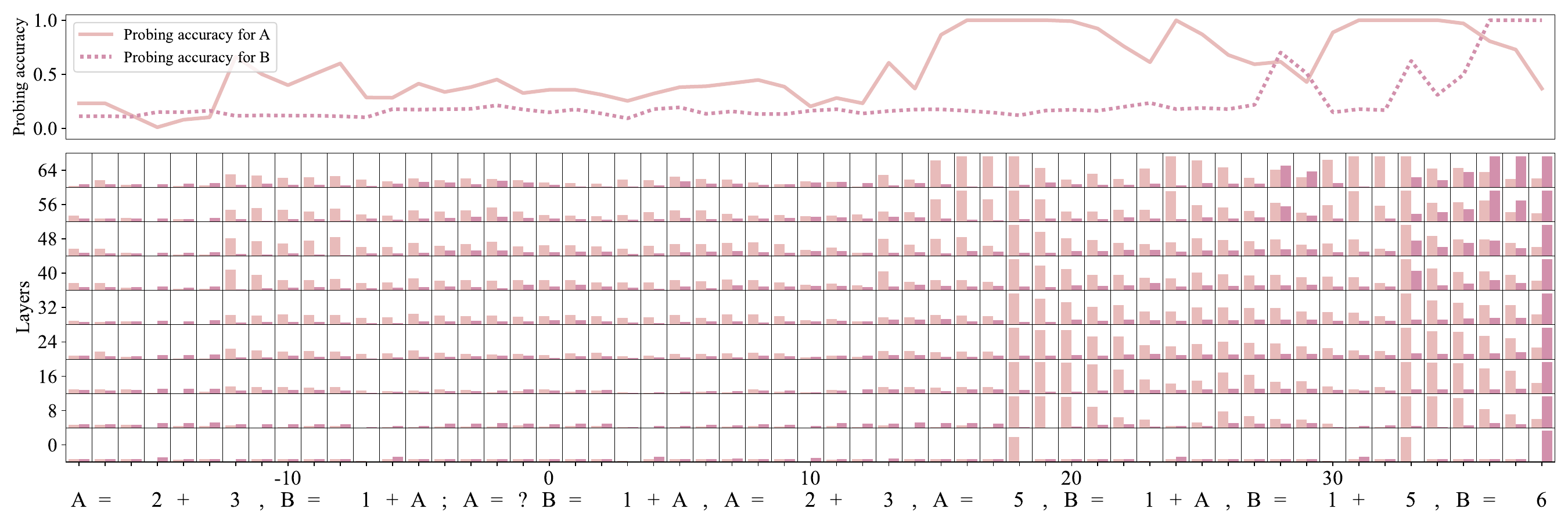}
    \caption{Probing results when Qwen2.5-32B solves Level \texttt{2}.}
    \label{fig:probing_qwen2.5_32B_task2}
\end{figure*}

\begin{figure*}[p]
    \centering
    \includegraphics[width=0.95\linewidth]{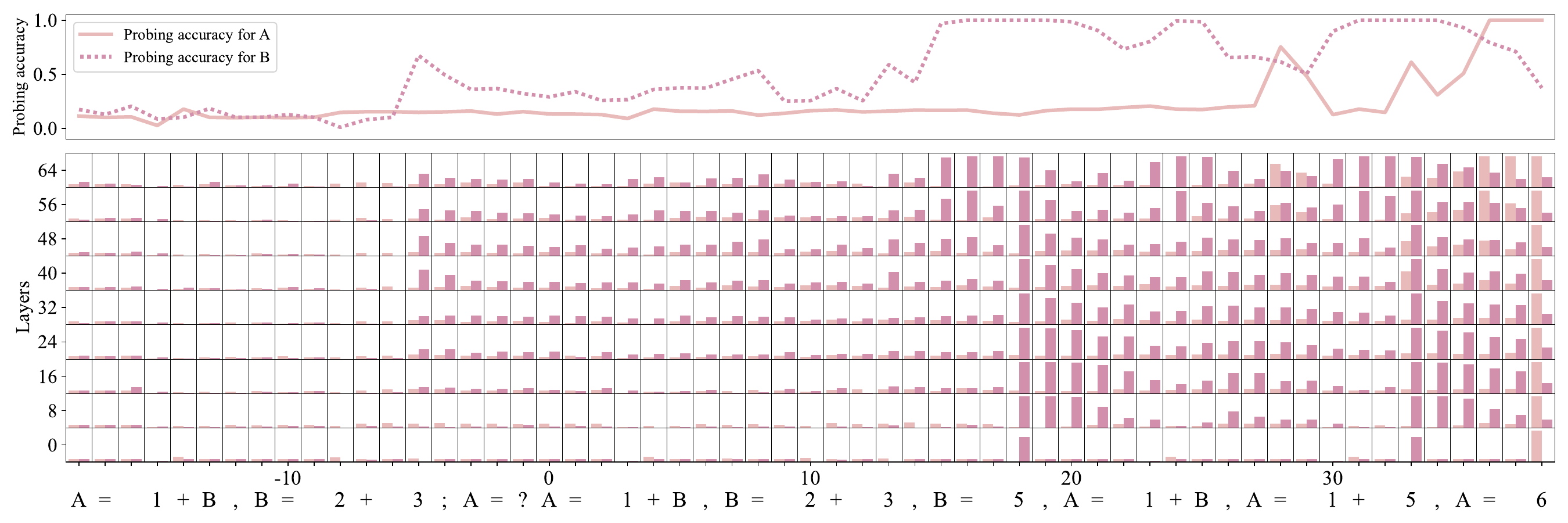}
    \caption{Probing results when Qwen2.5-32B solves Level \texttt{3}.}
    \label{fig:probing_qwen2.5_32B_task3}
\end{figure*}

\begin{figure*}[p]
    \centering
    \includegraphics[width=0.95\linewidth]{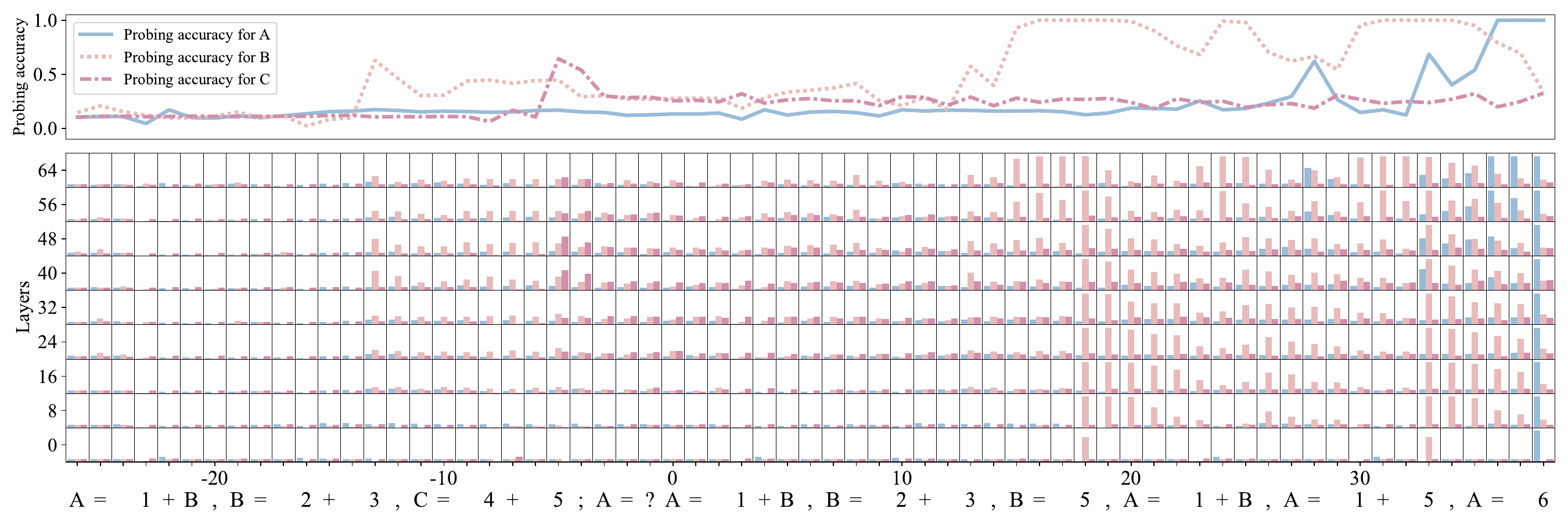}
    \caption{Probing results when Qwen2.5-32B solves Level \texttt{4}.}
    \label{fig:probing_qwen2.5_32B_task4}
\end{figure*}

\begin{figure*}[p]
    \centering
    \includegraphics[width=0.95\linewidth]{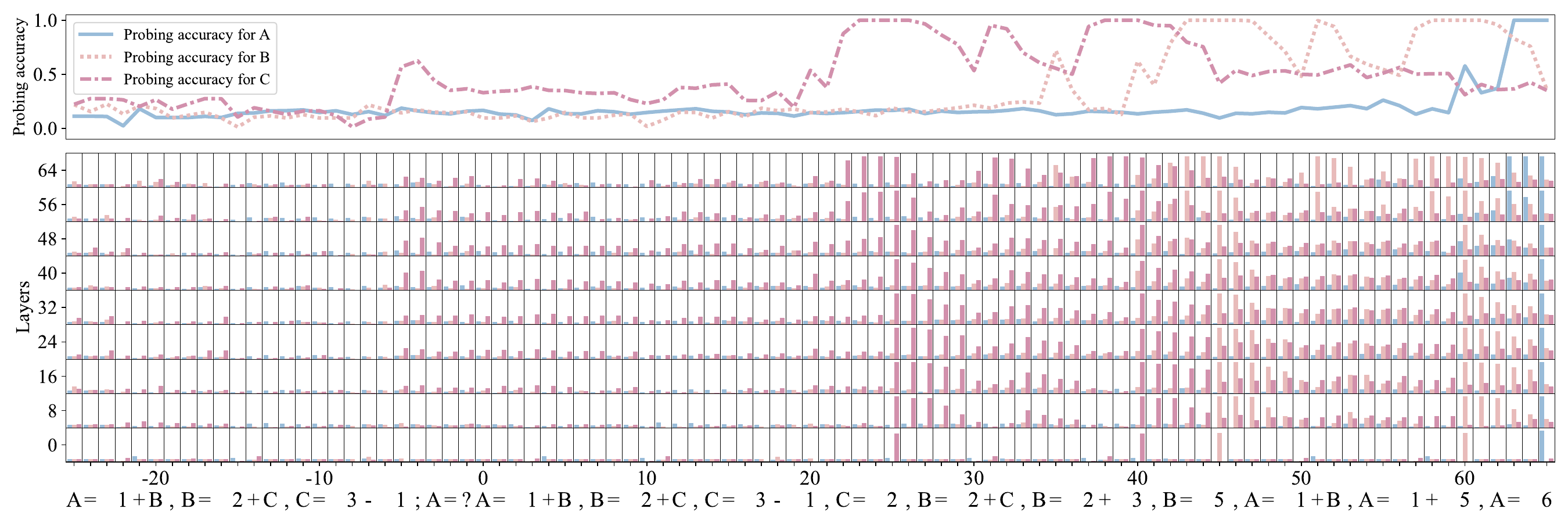}
    \caption{Probing results when Qwen2.5-32B solves Level \texttt{5}.}
    \label{fig:probing_qwen2.5_32B_task5}
\end{figure*}

\begin{figure*}[p]
    \centering
    \includegraphics[width=0.95\linewidth]{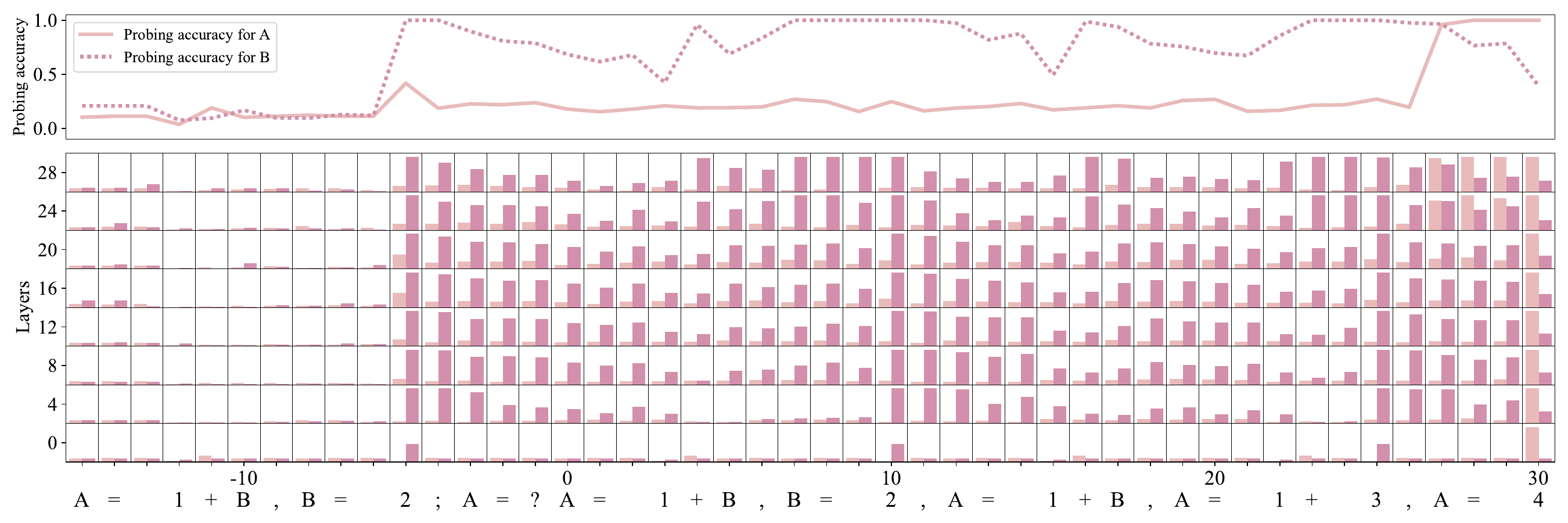}
    \caption{Probing results when Qwen2.5-Math-7B solves Level \texttt{1}.}
    \label{fig:probing_qwen2.5_math_7B_task1}
\end{figure*}

\begin{figure*}[p]
    \centering
    \includegraphics[width=0.95\linewidth]{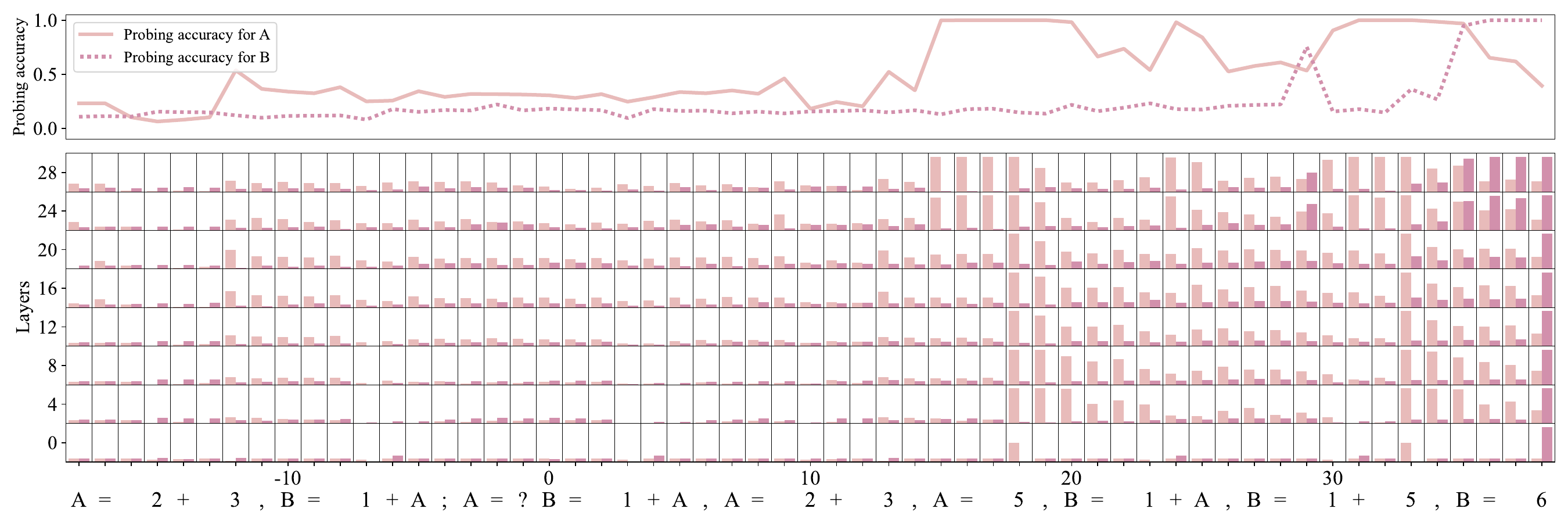}
    \caption{Probing results when Qwen2.5-Math-7B solves Level \texttt{2}.}
    \label{fig:probing_qwen2.5_math_7B_task2}
\end{figure*}

\begin{figure*}[p]
    \centering
    \includegraphics[width=0.95\linewidth]{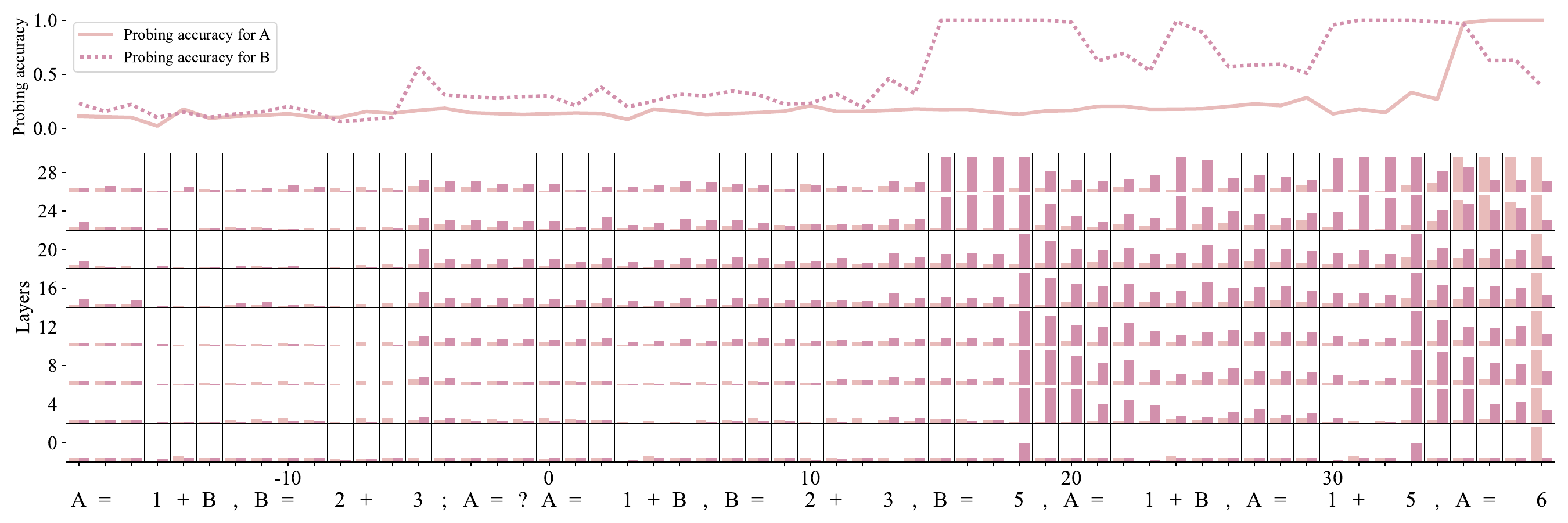}
    \caption{Probing results when Qwen2.5-Math-7B solves Level \texttt{3}.}
    \label{fig:probing_qwen2.5_math_7B_task3}
\end{figure*}

\begin{figure*}[p]
    \centering
    \includegraphics[width=0.95\linewidth]{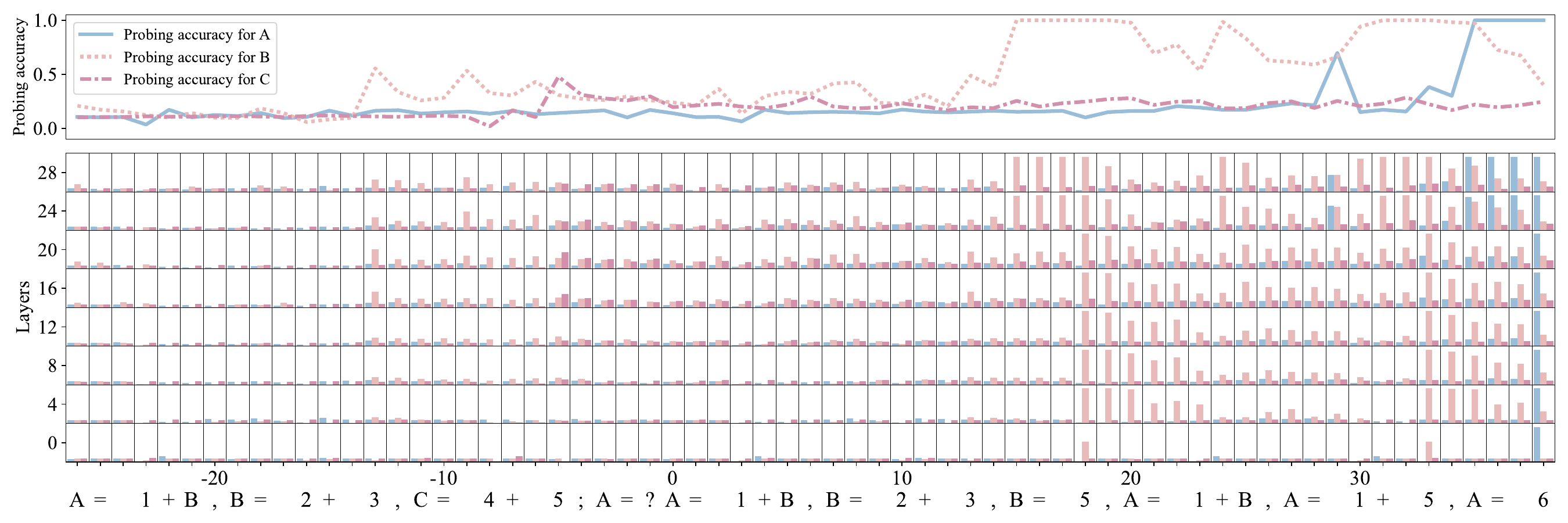}
    \caption{Probing results when Qwen2.5-Math-7B solves Level \texttt{4}.}
    \label{fig:probing_qwen2.5_math_7B_task4}
\end{figure*}

\begin{figure*}[p]
    \centering
    \includegraphics[width=0.95\linewidth]{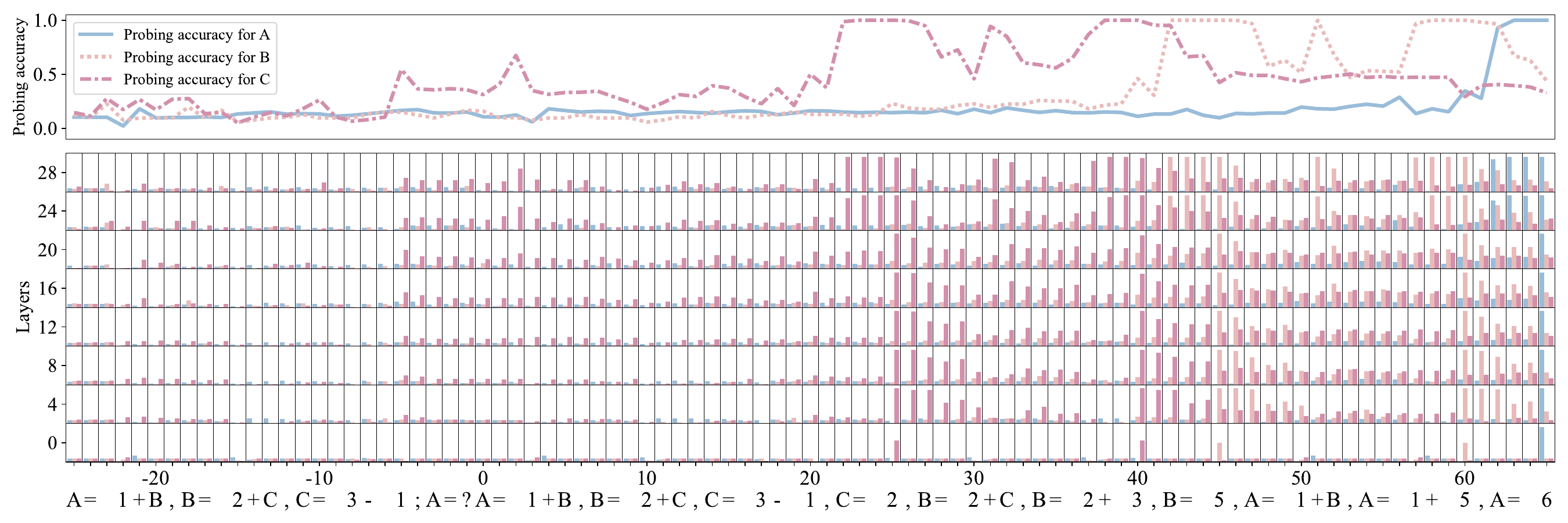}
    \caption{Probing results when Qwen2.5-Math-7B solves Level \texttt{5}.}
    \label{fig:probing_qwen2.5_math_7B_task5}
\end{figure*}

\begin{figure*}[p]
    \centering
    \includegraphics[width=0.95\linewidth]{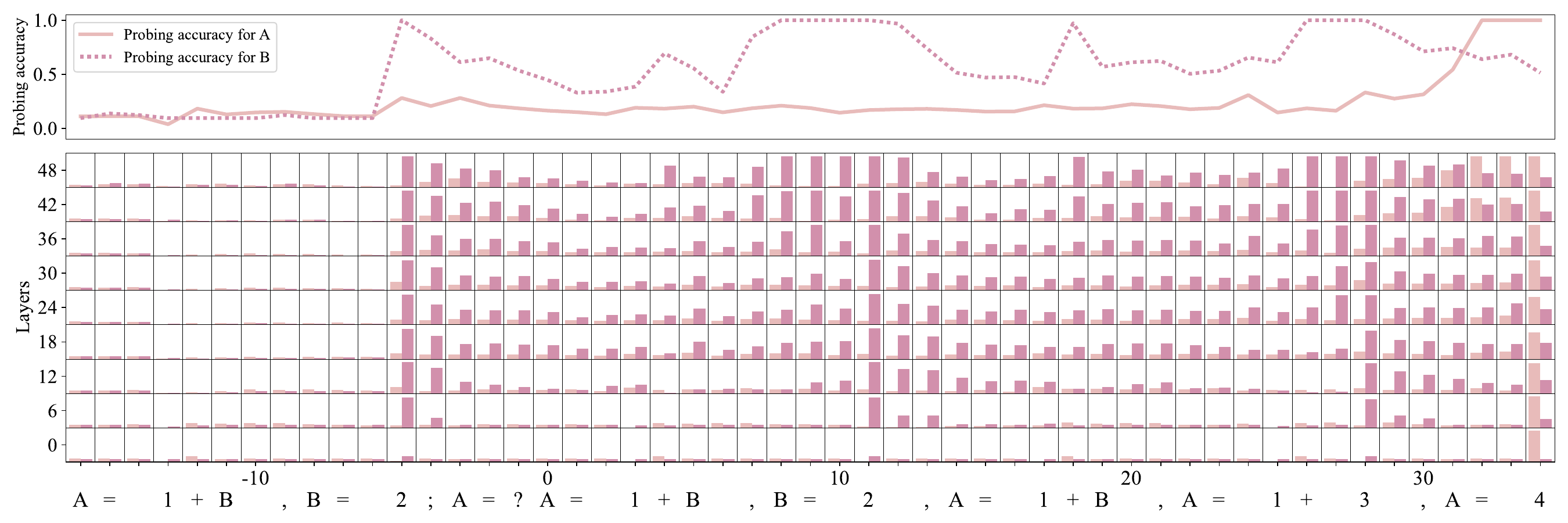}
    \caption{Probing results when Yi-1.5-9B solves Level \texttt{1}.}
    \label{fig:probing_yi_1.5_9B_task1}
\end{figure*}

\begin{figure*}[p]
    \centering
    \includegraphics[width=0.95\linewidth]{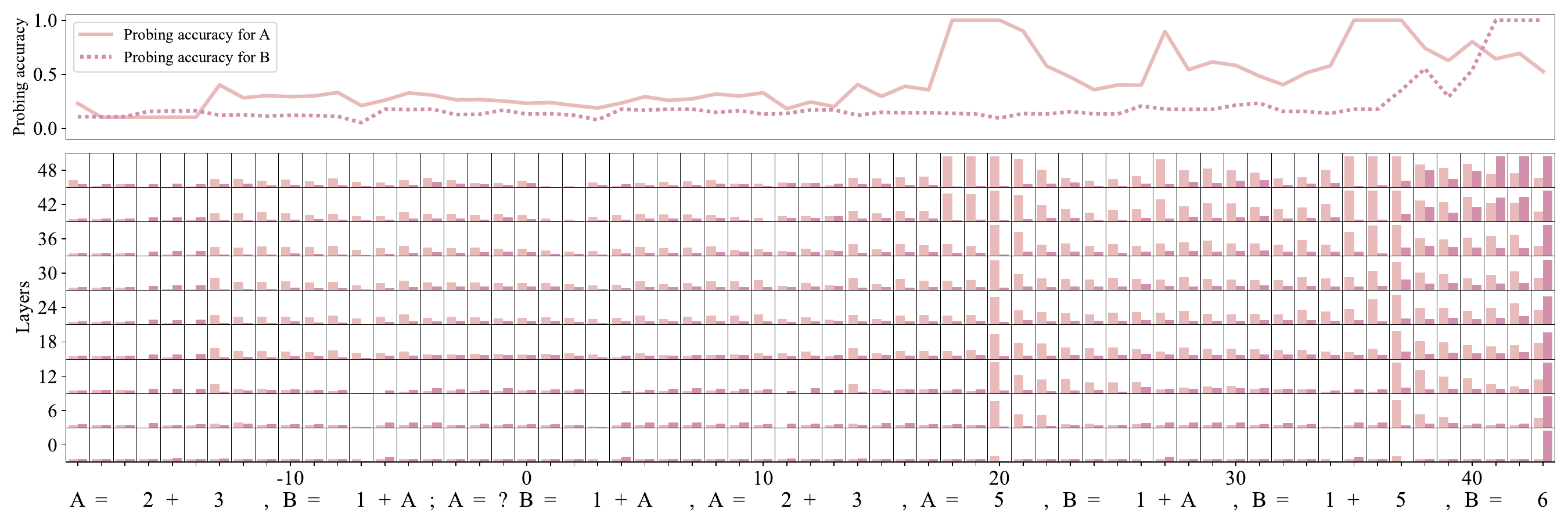}
    \caption{Probing results when Yi-1.5-9B solves Level \texttt{2}.}
    \label{fig:probing_yi_1.5_9B_task2}
\end{figure*}

\begin{figure*}[p]
    \centering
    \includegraphics[width=0.95\linewidth]{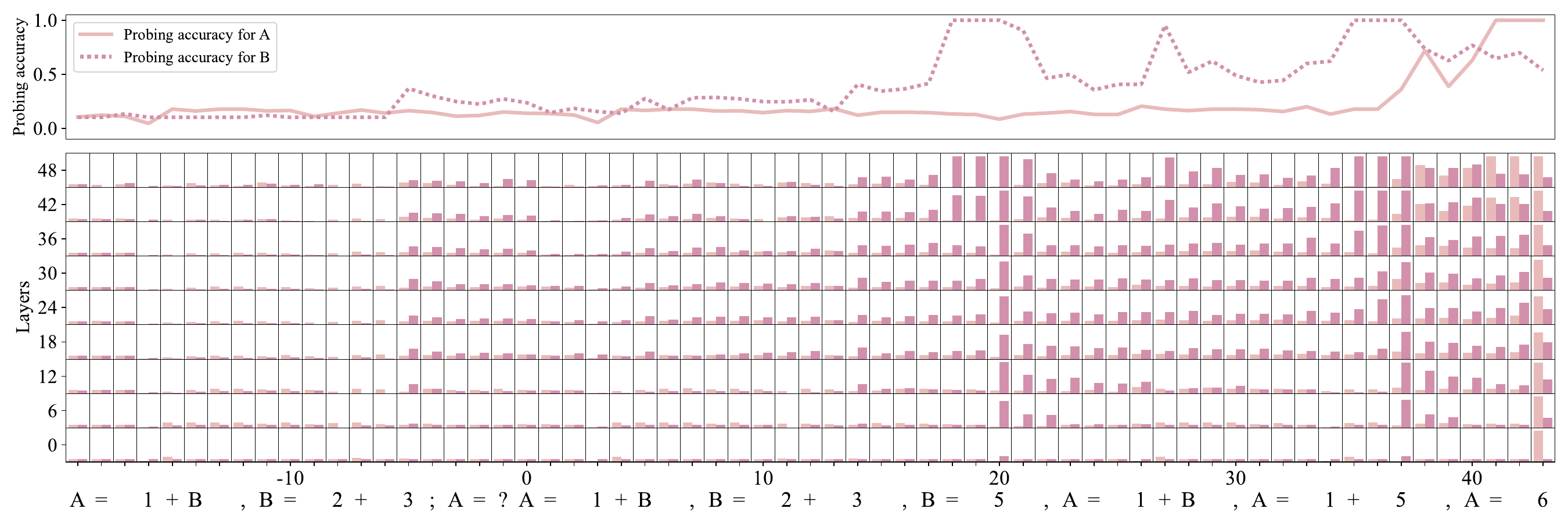}
    \caption{Probing results when Yi-1.5-9B solves Level \texttt{3}.}
    \label{fig:probing_yi_1.5_9B_task3}
\end{figure*}

\begin{figure*}[p]
    \centering
    \includegraphics[width=0.95\linewidth]{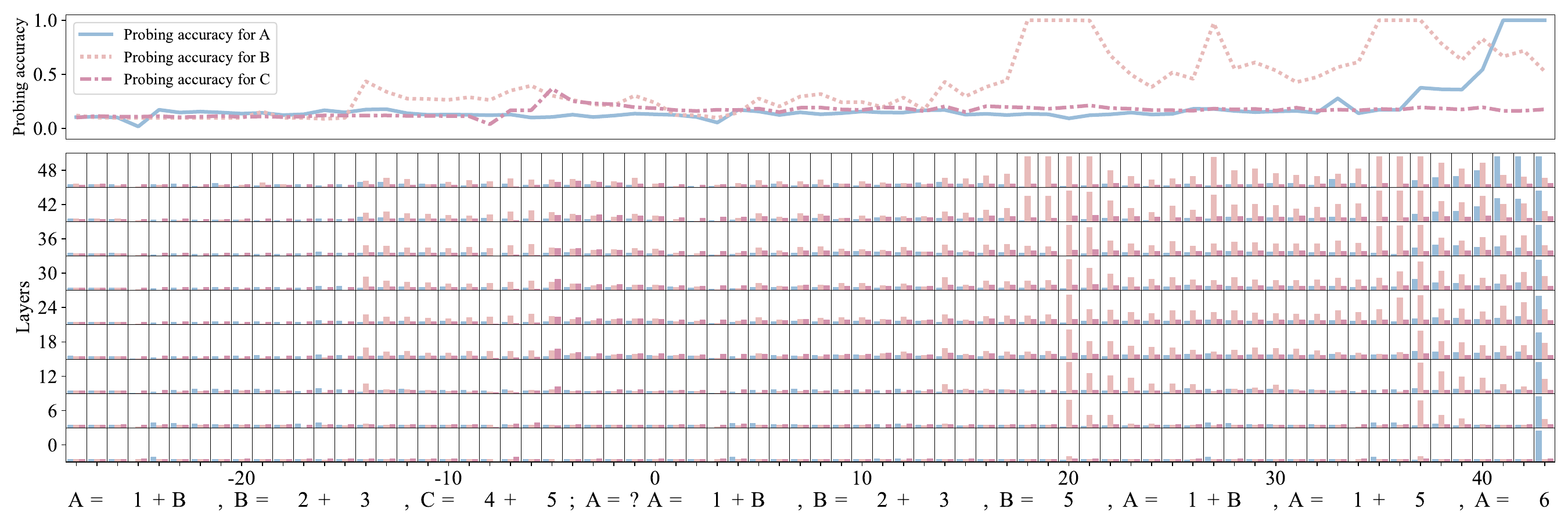}
    \caption{Probing results when Yi-1.5-9B solves Level \texttt{4}.}
    \label{fig:probing_yi_1.5_9B_task4}
\end{figure*}

\begin{figure*}[p]
    \centering
    \includegraphics[width=0.95\linewidth]{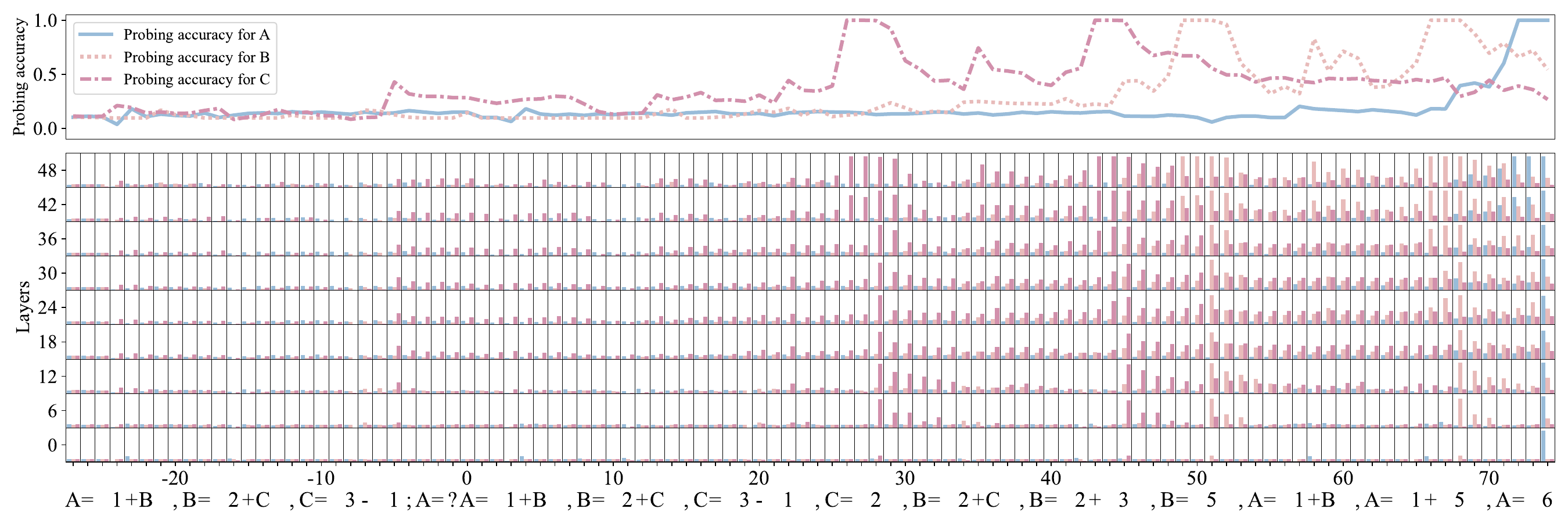}
    \caption{Probing results when Yi-1.5-9B solves Level \texttt{5}.}
    \label{fig:probing_yi_1.5_9B_task5}
\end{figure*}

\begin{figure*}[p]
    \centering
    \includegraphics[width=0.95\linewidth]{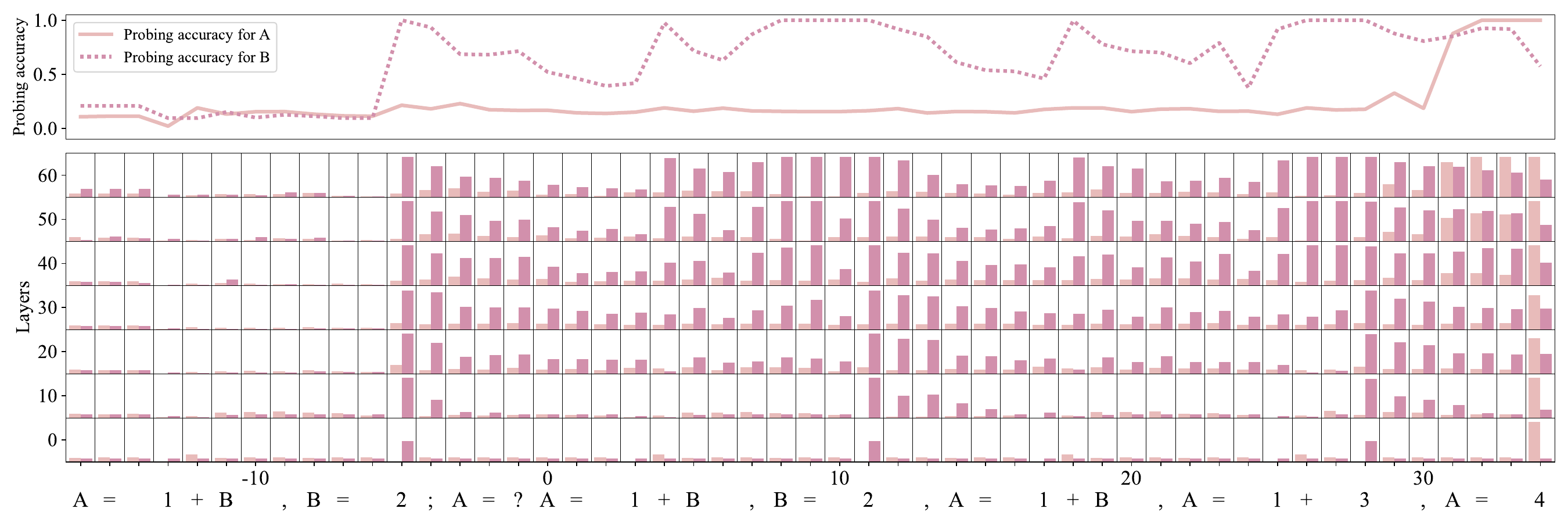}
    \caption{Probing results when Yi-1.5-34B solves Level \texttt{1}.}
    \label{fig:probing_yi_1.5_34B_task1}
\end{figure*}

\begin{figure*}[p]
    \centering
    \includegraphics[width=0.95\linewidth]{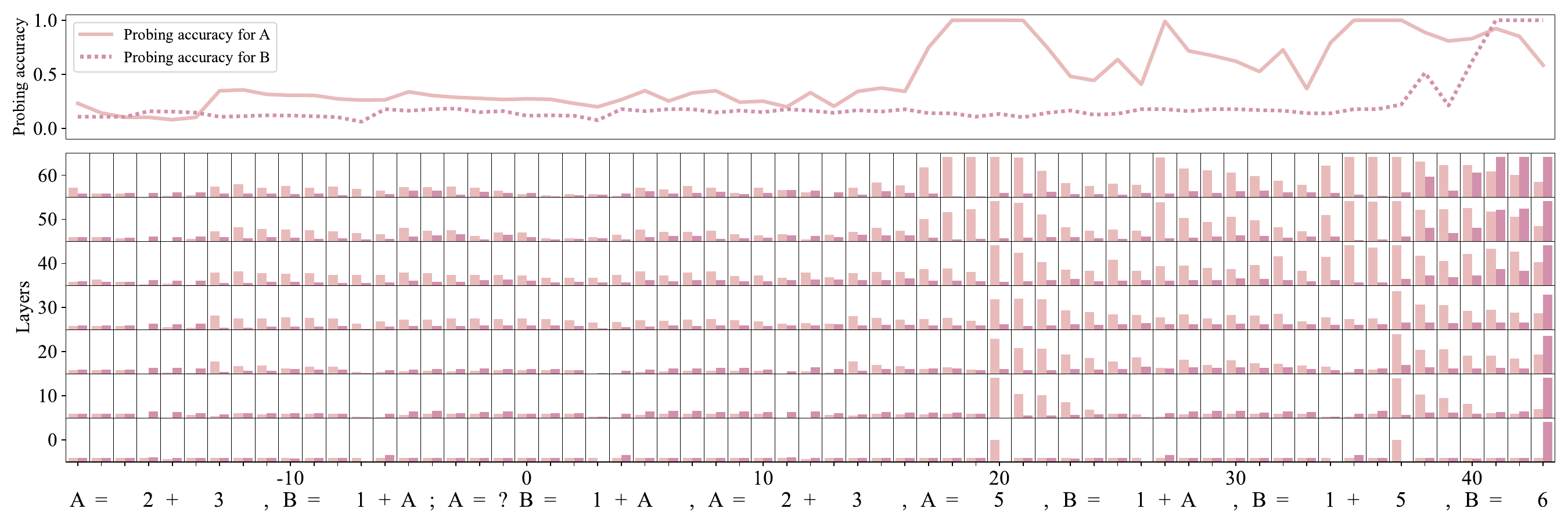}
    \caption{Probing results when Yi-1.5-34B solves Level \texttt{2}.}
    \label{fig:probing_yi_1.5_34B_task2}
\end{figure*}

\begin{figure*}[p]
    \centering
    \includegraphics[width=0.95\linewidth]{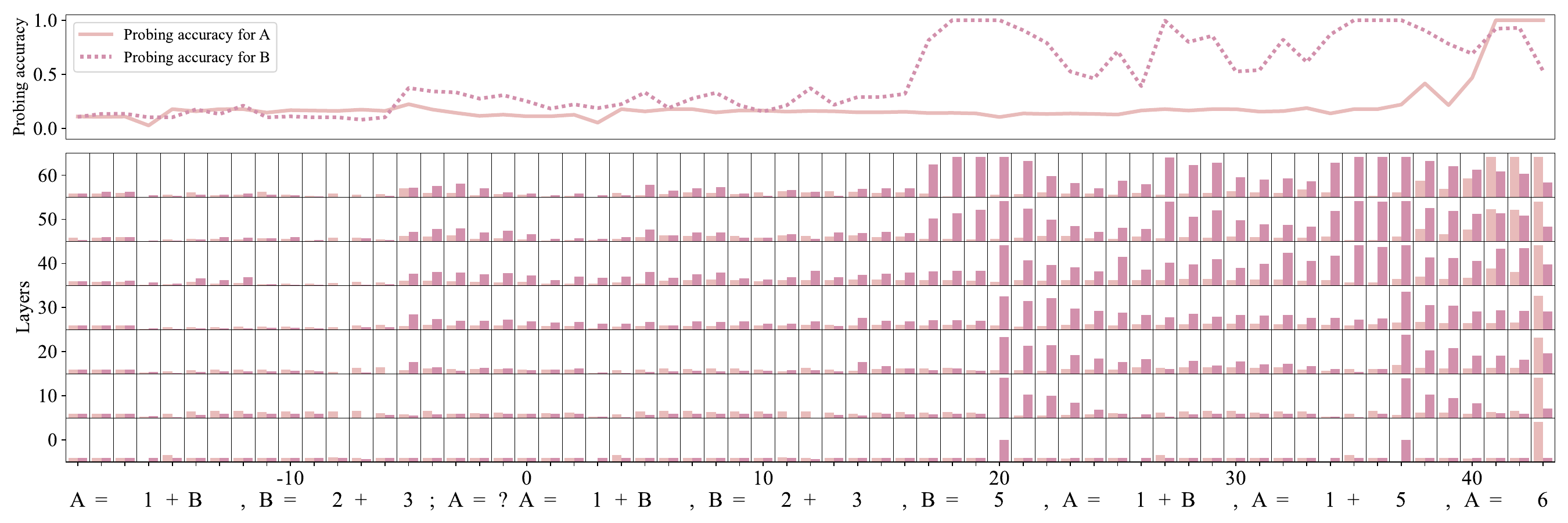}
    \caption{Probing results when Yi-1.5-34B solves Level \texttt{3}.}
    \label{fig:probing_yi_1.5_34B_task3}
\end{figure*}

\begin{figure*}[p]
    \centering
    \includegraphics[width=0.95\linewidth]{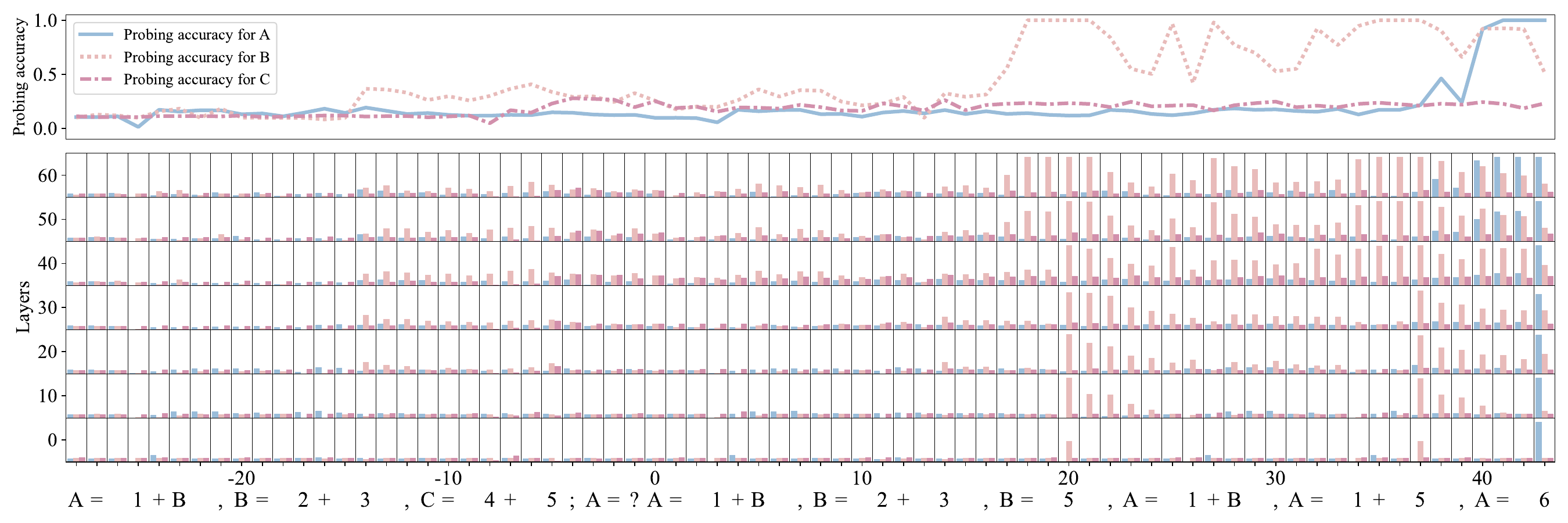}
    \caption{Probing results when Yi-1.5-34B solves Level \texttt{4}.}
    \label{fig:probing_yi_1.5_34B_task4}
\end{figure*}

\begin{figure*}[p]
    \centering
    \includegraphics[width=0.95\linewidth]{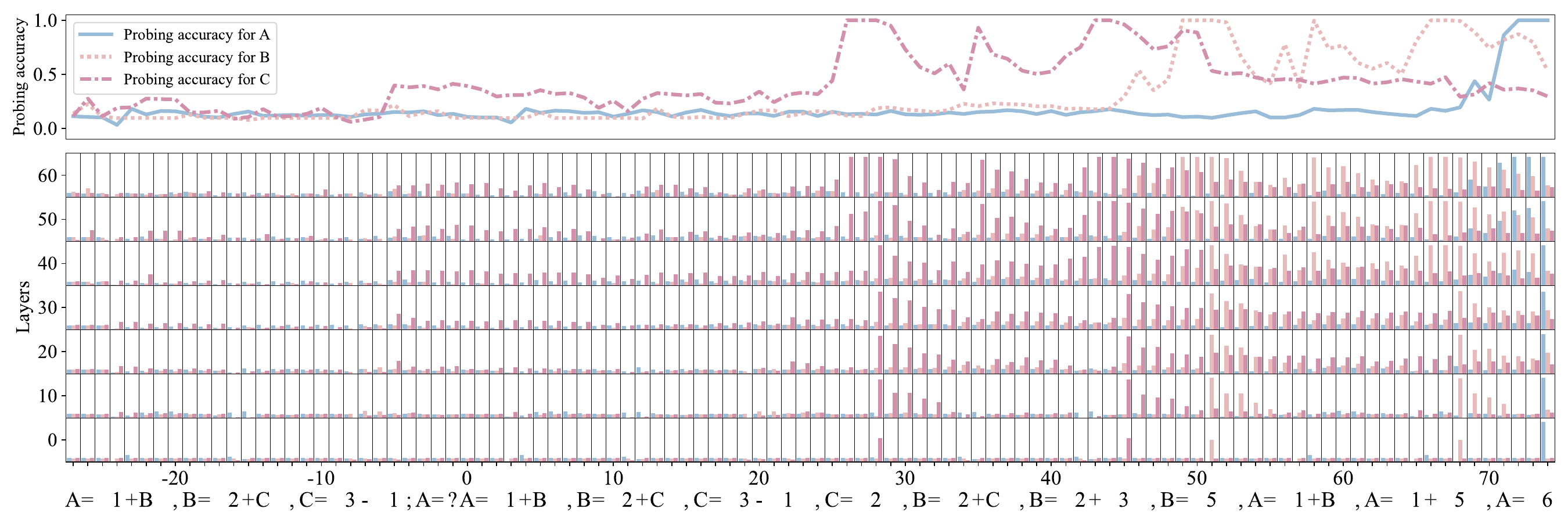}
    \caption{Probing results when Yi-1.5-34B solves Level \texttt{5}.}
    \label{fig:probing_yi_1.5_34B_task5}
\end{figure*}

\begin{figure*}[p]
    \centering
    \includegraphics[width=0.95\linewidth]{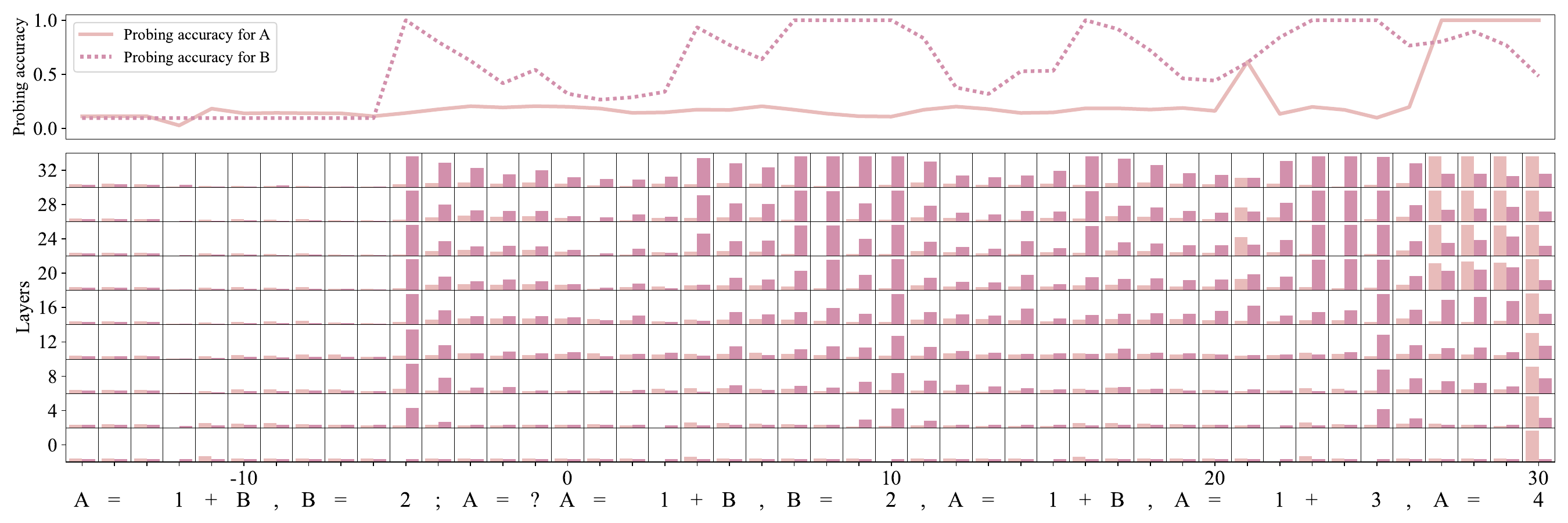}
    \caption{Probing results when Llama-3.1-8B solves Level \texttt{1}.}
    \label{fig:probing_llama3.1_8B_task1}
\end{figure*}

\begin{figure*}[p]
    \centering
    \includegraphics[width=0.95\linewidth]{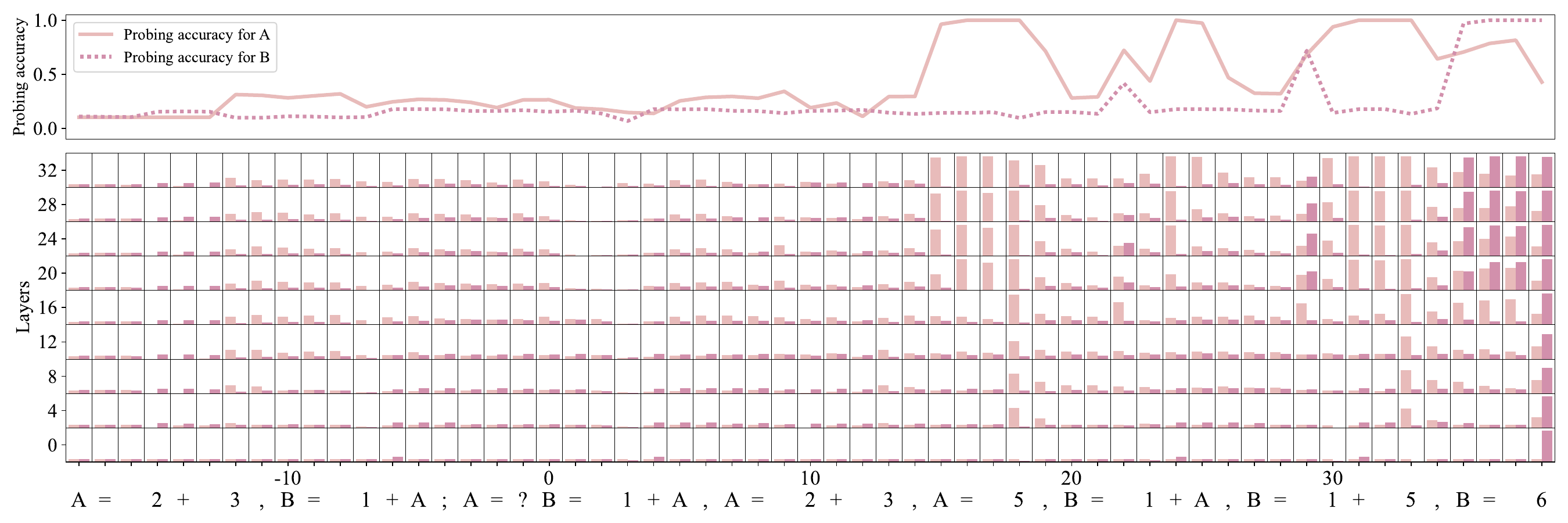}
    \caption{Probing results when Llama-3.1-8B solves Level \texttt{2}.}
    \label{fig:probing_llama3.1_8B_task2}
\end{figure*}

\begin{figure*}[p]
    \centering
    \includegraphics[width=0.95\linewidth]{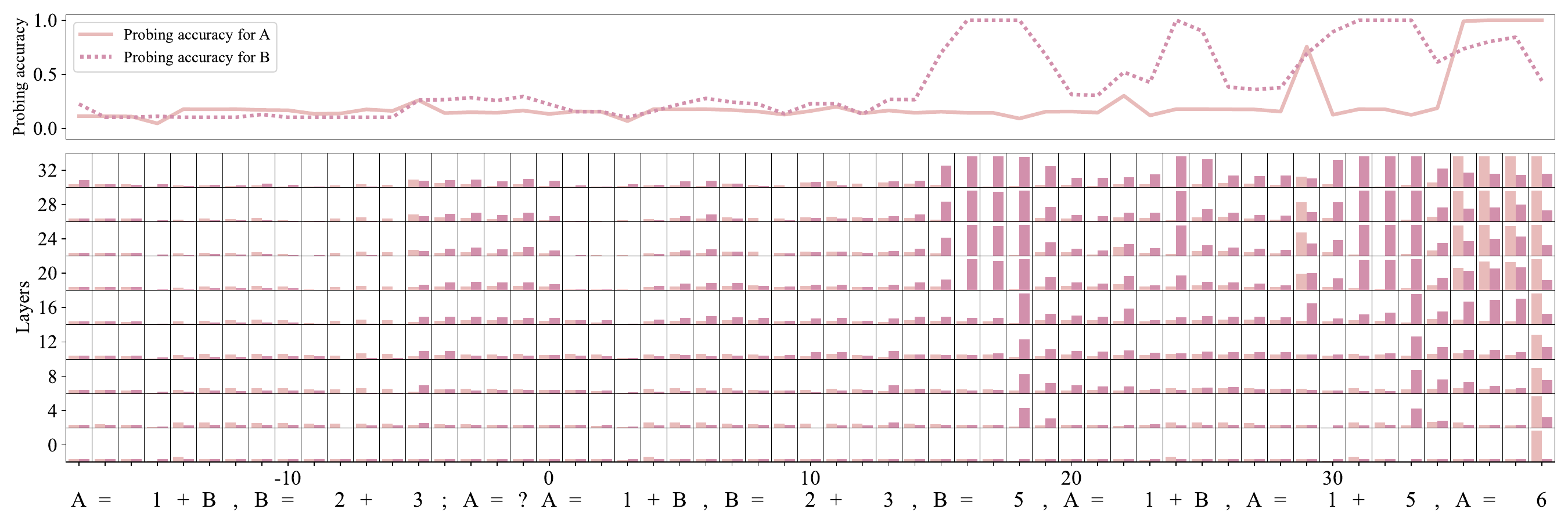}
    \caption{Probing results when Llama-3.1-8B solves Level \texttt{3}.}
    \label{fig:probing_llama3.1_8B_task3}
\end{figure*}

\begin{figure*}[p]
    \centering
    \includegraphics[width=0.95\linewidth]{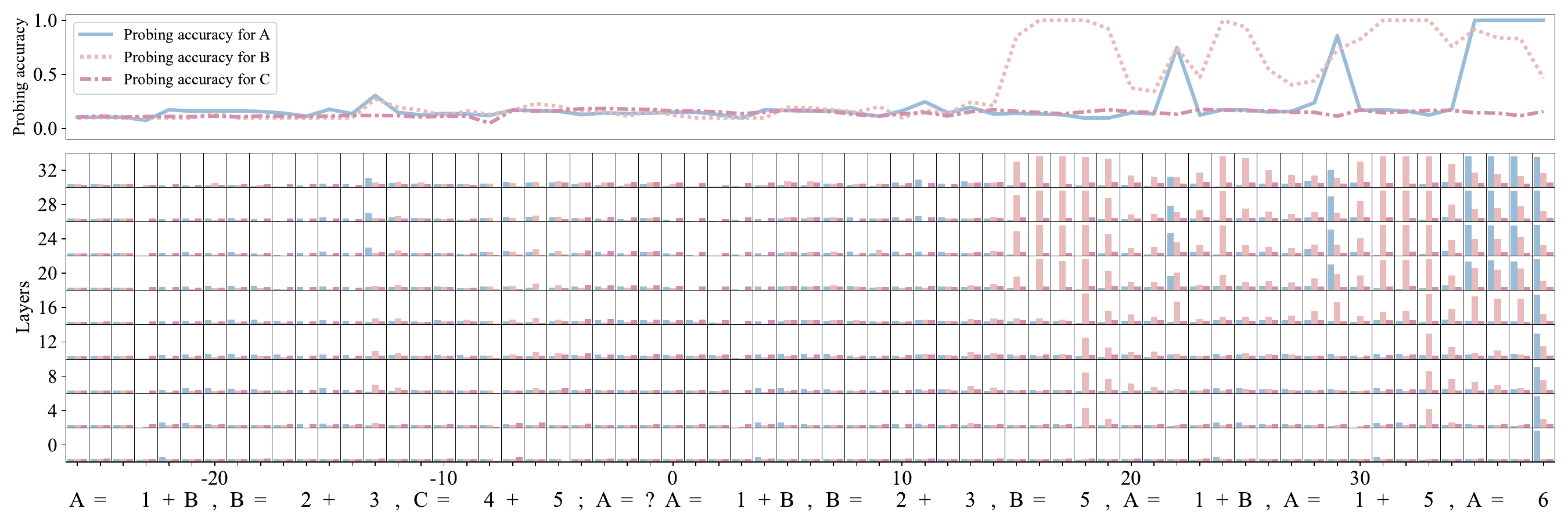}
    \caption{Probing results when Llama-3.1-8B solves Level \texttt{4}.}
    \label{fig:probing_llama3.1_8B_task4}
\end{figure*}

\begin{figure*}[p]
    \centering
    \includegraphics[width=0.95\linewidth]{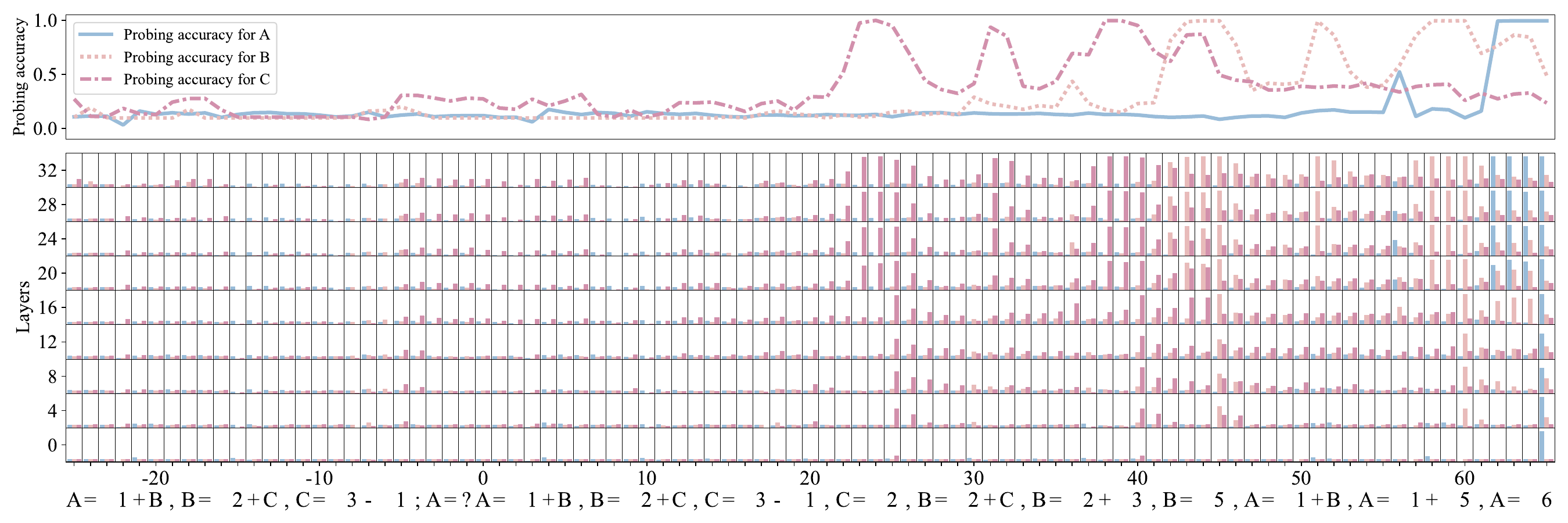}
    \caption{Probing results when Llama-3.1-8B solves Level \texttt{5}.}
    \label{fig:probing_llama3.1_8B_task5}
\end{figure*}

\begin{figure*}[p]
    \centering
    \includegraphics[width=0.95\linewidth]{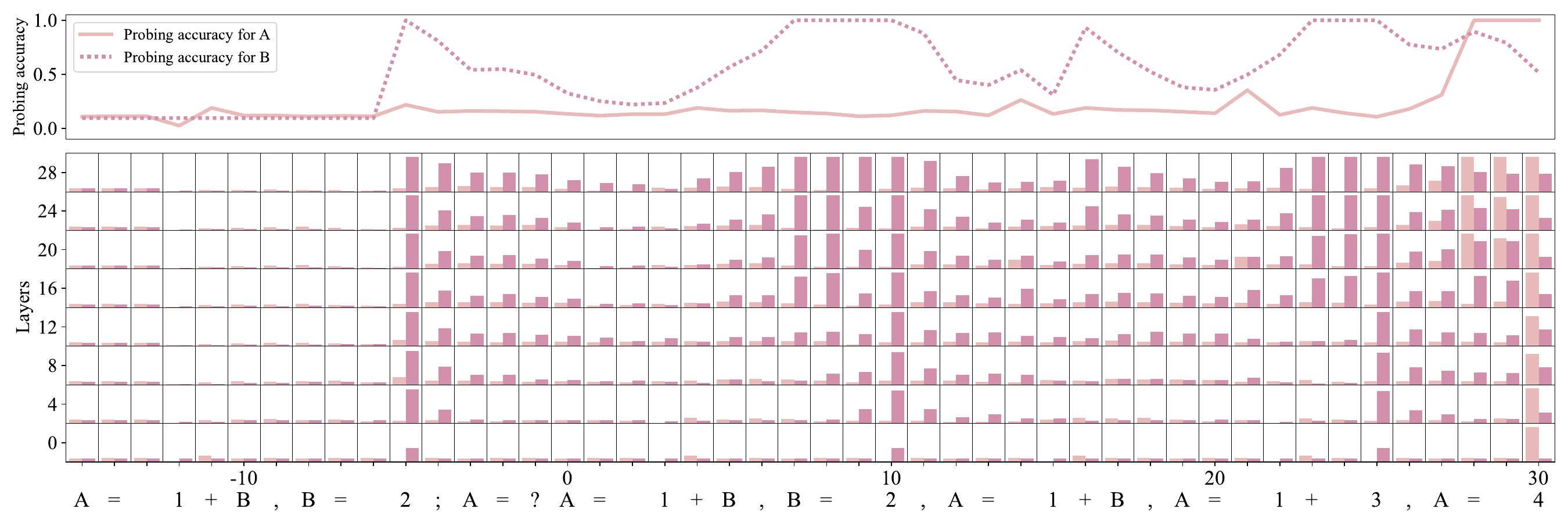}
    \caption{Probing results when Llama-3.2-3B solves Level \texttt{1}.}
    \label{fig:probing_llama3.2_3B_task1}
\end{figure*}

\begin{figure*}[p]
    \centering
    \includegraphics[width=0.95\linewidth]{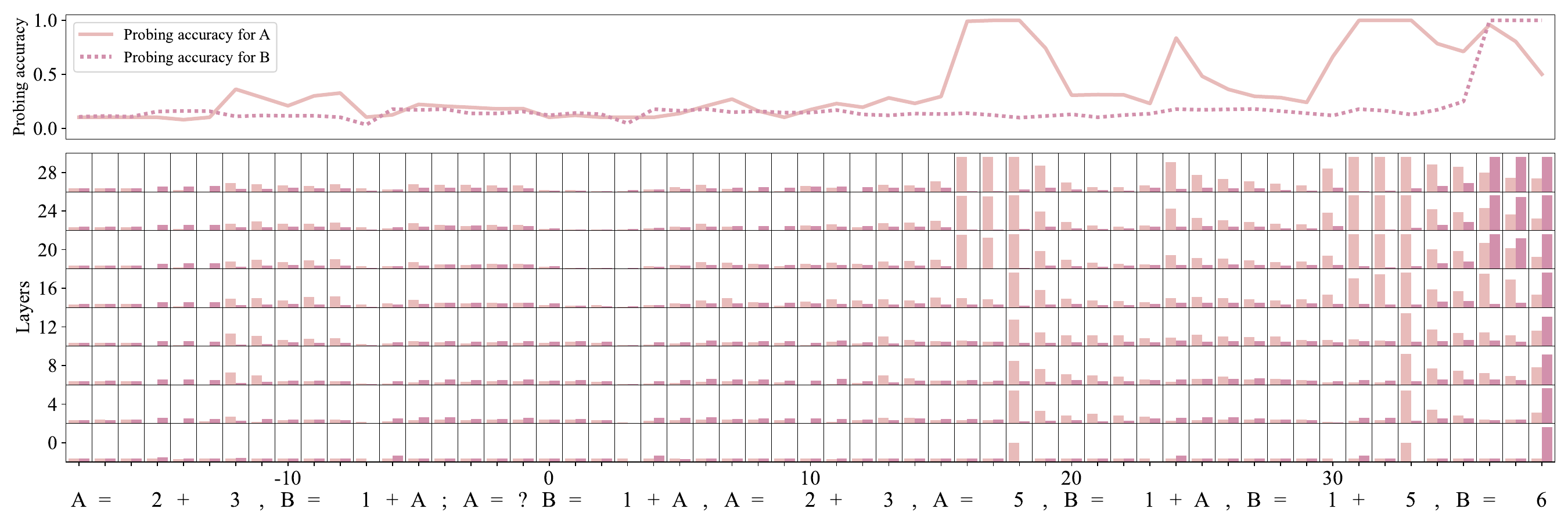}
    \caption{Probing results when Llama-3.2-3B solves Level \texttt{2}.}
    \label{fig:probing_llama3.2_3B_task2}
\end{figure*}

\begin{figure*}[p]
    \centering
    \includegraphics[width=0.95\linewidth]{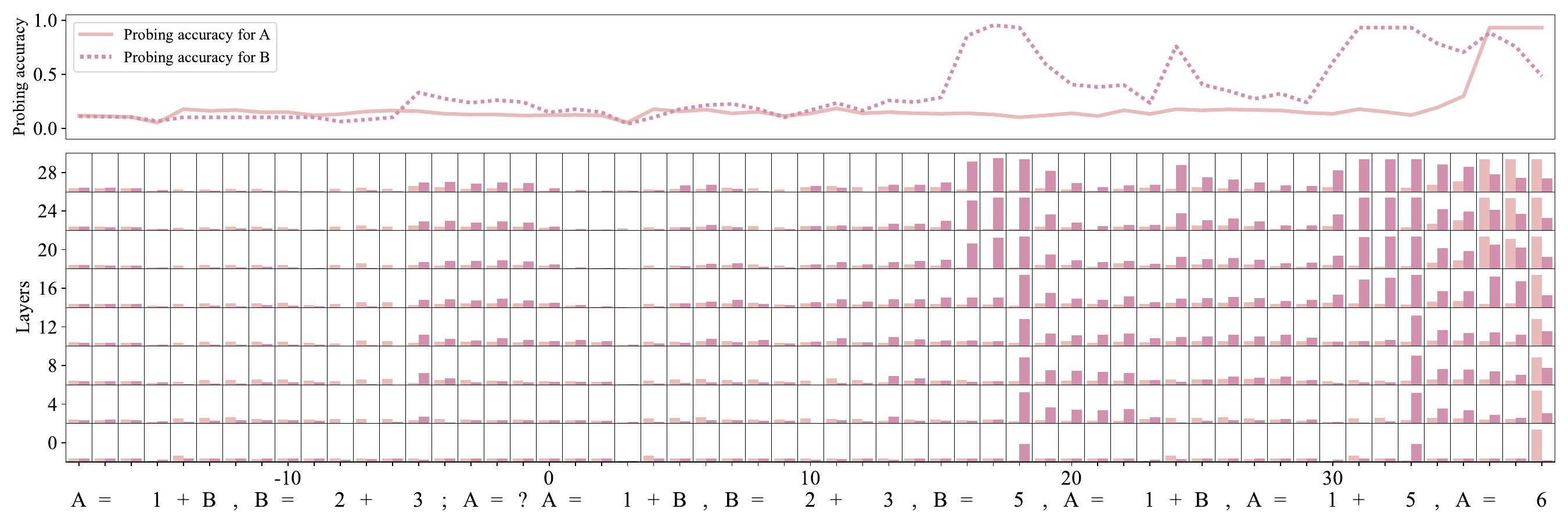}
    \caption{Probing results when Llama-3.2-3B solves Level \texttt{3}.}
    \label{fig:probing_llama3.2_3B_task3}
\end{figure*}

\begin{figure*}[p]
    \centering
    \includegraphics[width=0.95\linewidth]{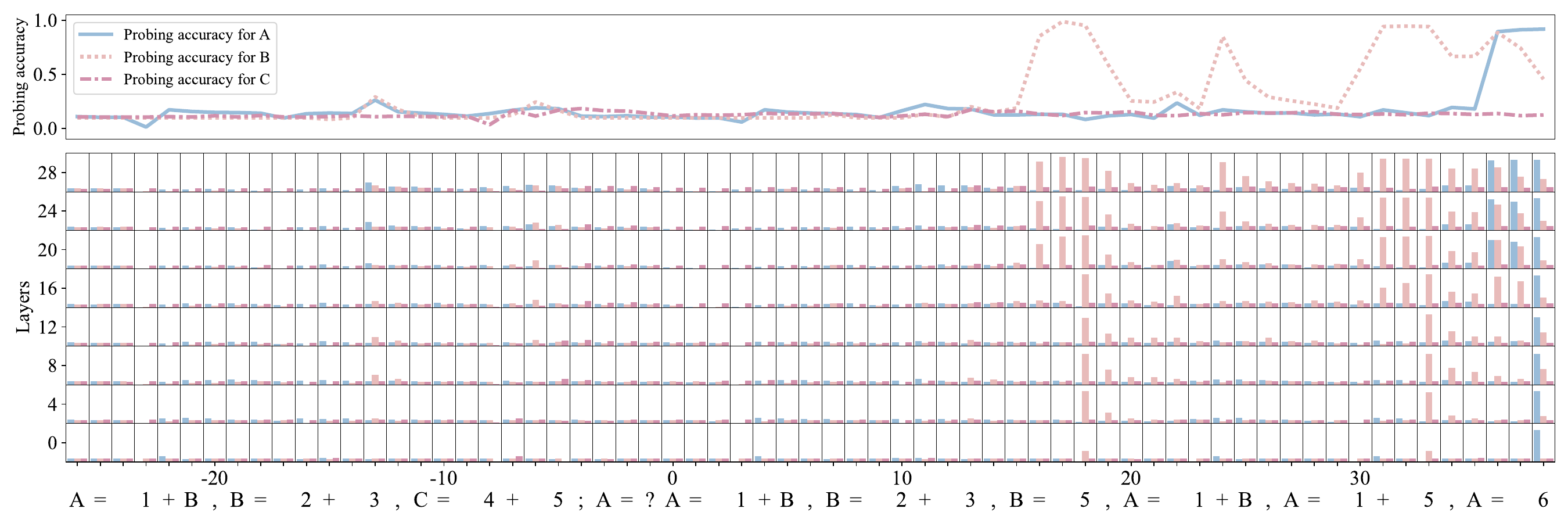}
    \caption{Probing results when Llama-3.2-3B solves Level \texttt{4}.}
    \label{fig:probing_llama3.2_3B_task4}
\end{figure*}

\begin{figure*}[p]
    \centering
    \includegraphics[width=0.95\linewidth]{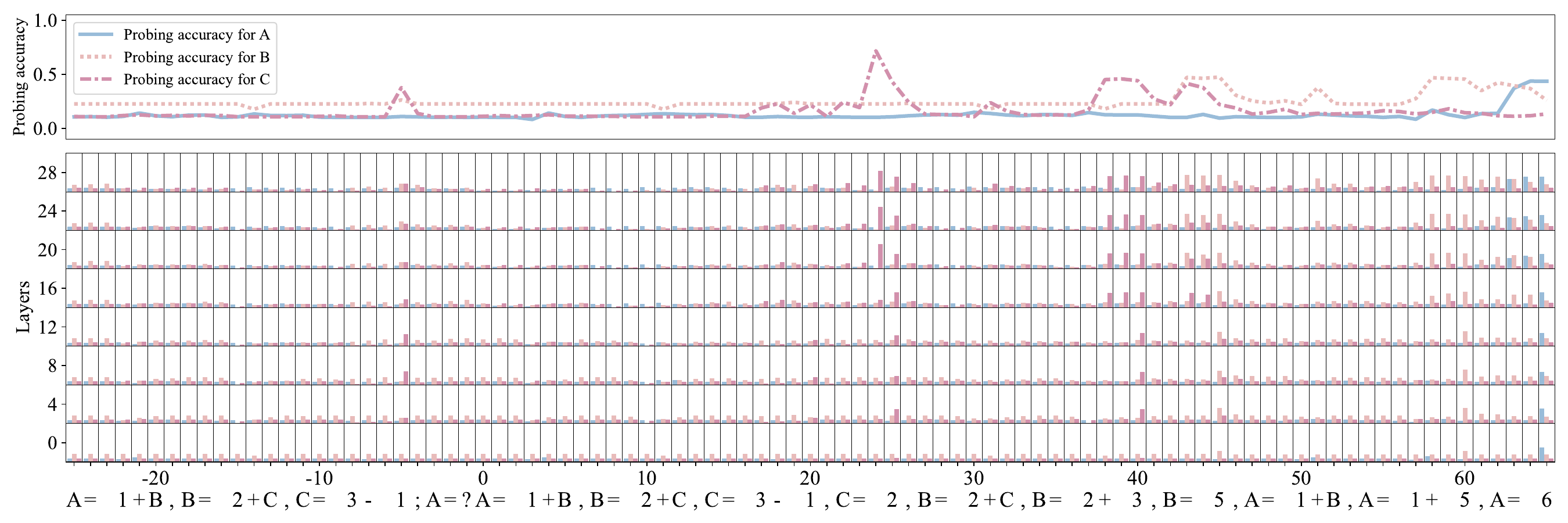}
    \caption{Probing results when Llama-3.2-3B solves Level \texttt{5}.}
    \label{fig:probing_llama3.2_3B_task5}
\end{figure*}

\begin{figure*}[p]
    \centering
    \includegraphics[width=0.95\linewidth]{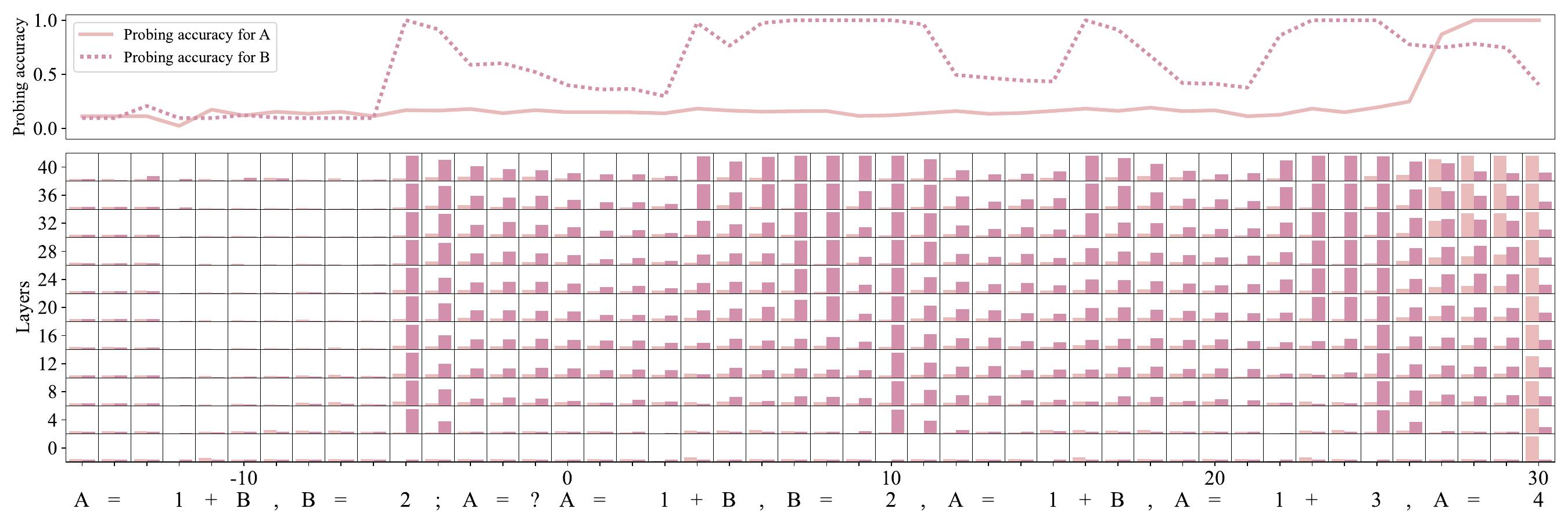}
    \caption{Probing results when Mistral-Nemo-Base-2407 solves Level \texttt{1}.}
    \label{fig:probing_mistral_nemo_base_2407_task1}
\end{figure*}

\begin{figure*}[p]
    \centering
    \includegraphics[width=0.95\linewidth]{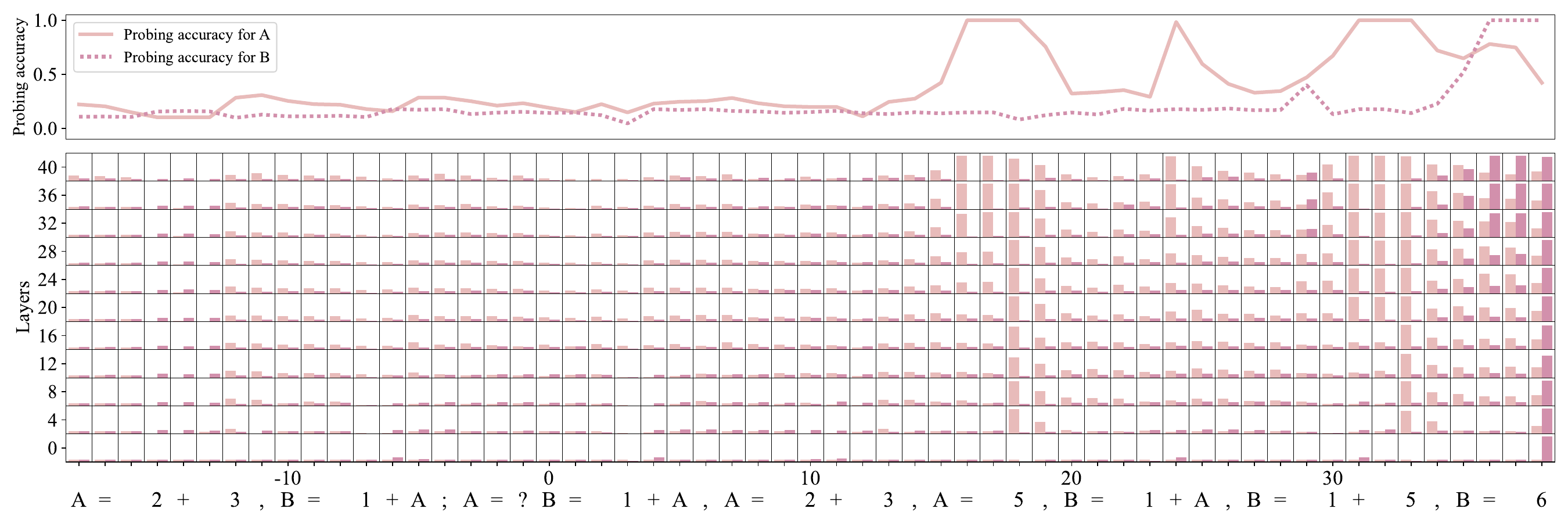}
    \caption{Probing results when Mistral-Nemo-Base-2407 solves Level \texttt{2}.}
    \label{fig:probing_mistral_nemo_base_2407_task2}
\end{figure*}

\begin{figure*}[p]
    \centering
    \includegraphics[width=0.95\linewidth]{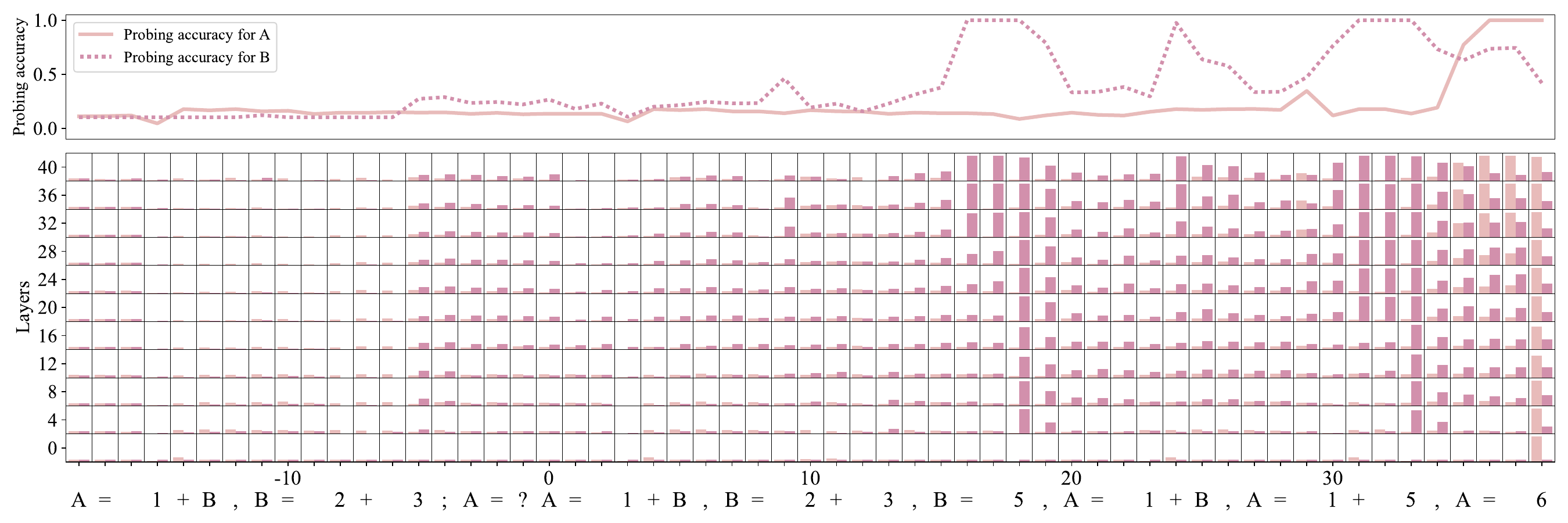}
    \caption{Probing results when Mistral-Nemo-Base-2407 solves Level \texttt{3}.}
    \label{fig:probing_mistral_nemo_base_2407_task3}
\end{figure*}

\begin{figure*}[p]
    \centering
    \includegraphics[width=0.95\linewidth]{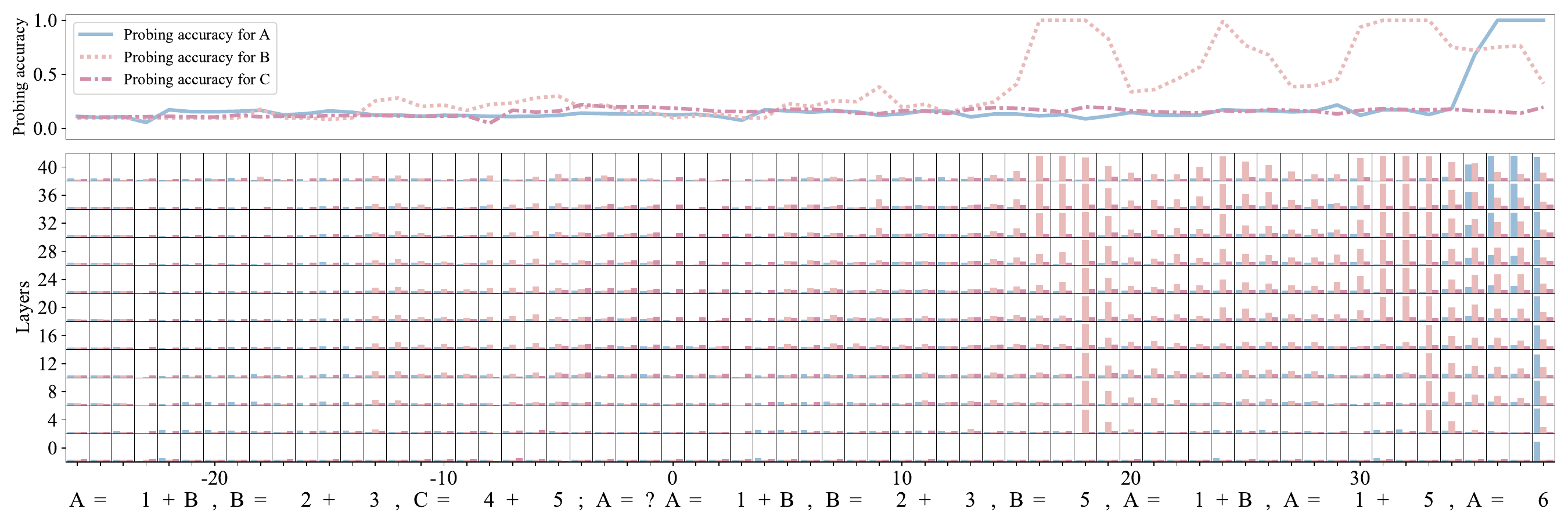}
    \caption{Probing results when Mistral-Nemo-Base-2407 solves Level \texttt{4}.}
    \label{fig:probing_mistral_nemo_base_2407_task4}
\end{figure*}

\begin{figure*}[p]
    \centering
    \includegraphics[width=0.95\linewidth]{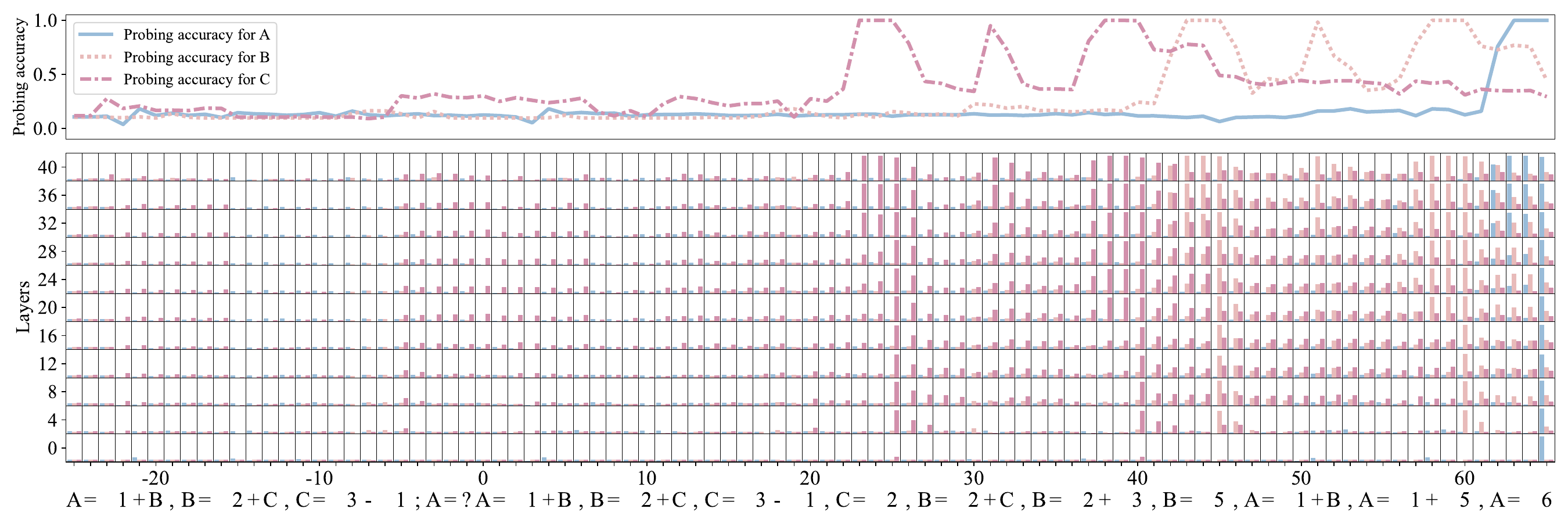}
    \caption{Probing results when Mistral-Nemo-Base-2407 solves Level \texttt{5}.}
    \label{fig:probing_mistral_nemo_base_2407_task5}
\end{figure*}

Figures~\ref{fig:probing_qwen2.5_7B_task1} through~\ref{fig:probing_mistral_nemo_base_2407_task5} present the probing results for all models and tasks discussed in this paper.
For each figure, the input sequence below the graphs is one example (the results are averaged over the test set).
The upper part indicates the maximum probing accuracy achieved at each token position $t$.
The bottom part shows the probing accuracies in each token $t$, layer $l$, and variable $v_i$.
Tables~\ref{table:rsp_model_dimmension_tau_0.85_level_1} through~\ref{table:rsp_model_dimmension_tau_0.95_level_5} summarize these results for thresholds ($\tau$) ranging from 0.85 to 0.95.
From these results, we observe trends similar to those described in \cref{subsec:experiments} across many settings.
However, for the smaller model Llama3.2 (3B), increasing the threshold $\tau$ often leads to cases where the accuracy does not reach the threshold (N/A).
Nonetheless, a consistent pattern remains: $\mathrm{Acc}_{\mathrm{\prec CoT}}(v_{i})$ is low whereas $\mathrm{Acc}_{\mathrm{\succ CoT}}(v_{i})$ is high.%

\subsection{Additional causal intervention results}
\label{appendix:subsec:causal_intervention_results}
\begin{figure}[t]
  \centering
  \includegraphics[width=0.97\linewidth]{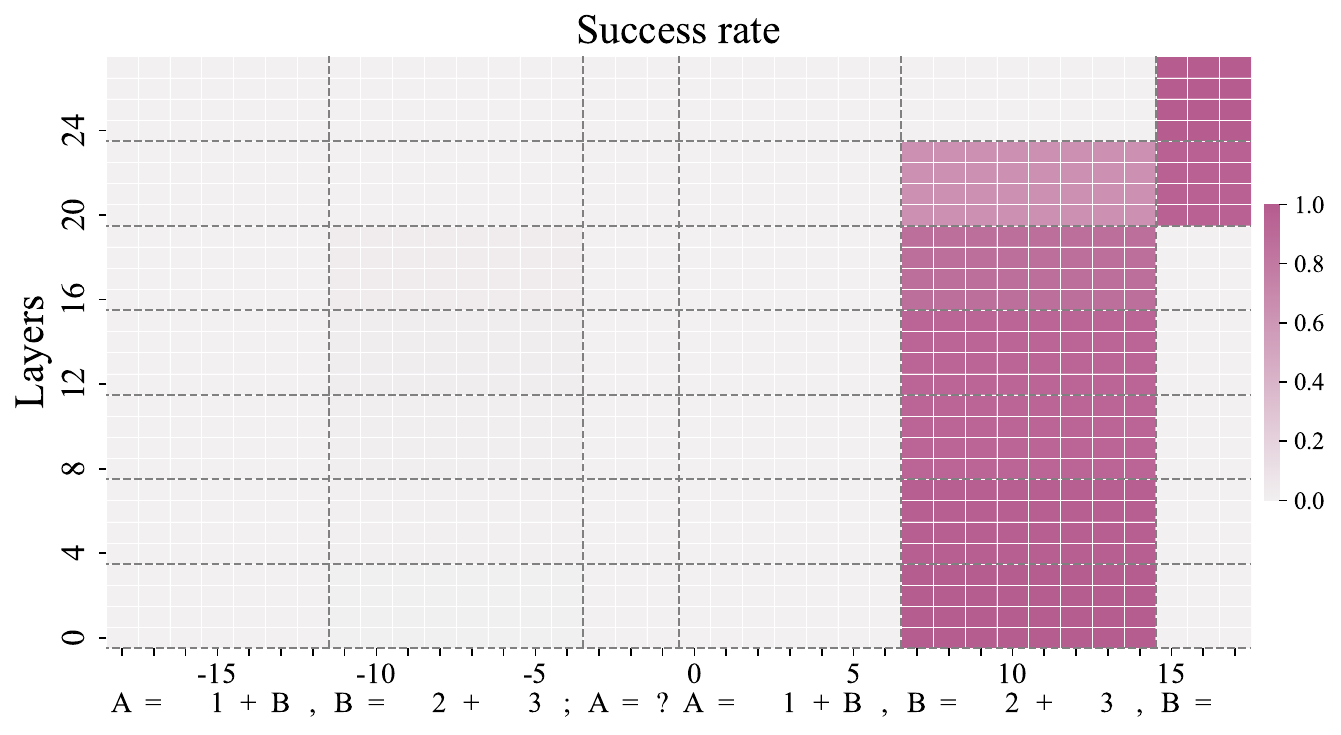}
  \caption{Results of the causal intervention on Qwen2.5-7B. Each grid cell shows the success rate when the intermediate token $z_{17}$ \(({\underline{\textcolor{gray}{\mathrm{B}=}5}}_{\hspace{0.05cm}\mathbf{2}})\) is the target token.}
  \label{fig:intervention_qwen2.5_7B_mid_5}
\end{figure}

\begin{figure}[t]
  \centering
  \includegraphics[width=0.97\linewidth]{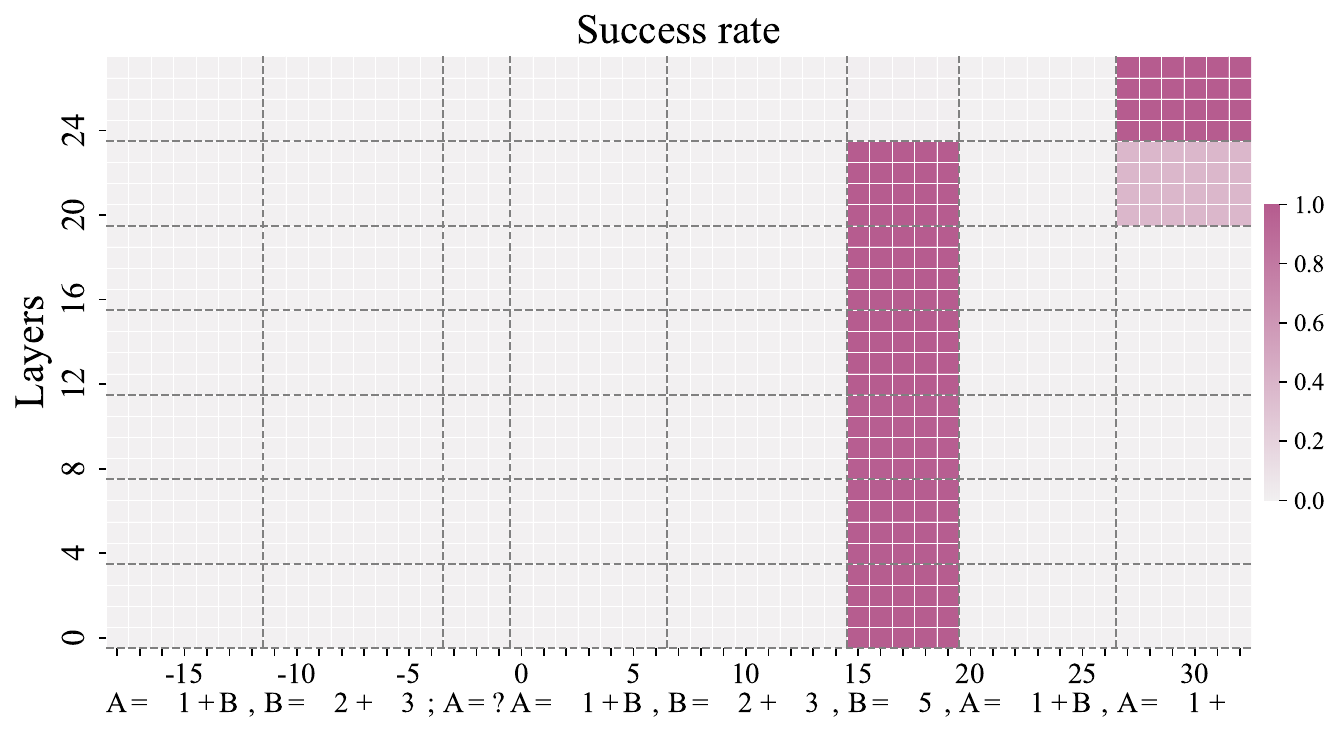}
  \caption{Results of the causal intervention on Qwen2.5-7B. Each grid cell shows the success rate when $z_{32}$ \(({\underline{\textcolor{gray}{\mathrm{A}=1+}5}}_{\hspace{0.05cm}\mathbf{4}})\) is the target token.}
  \label{fig:intervention_qwen2.5_7B_last_5}
\end{figure}

\begin{figure}[t]
  \centering
  \includegraphics[width=0.97\linewidth]{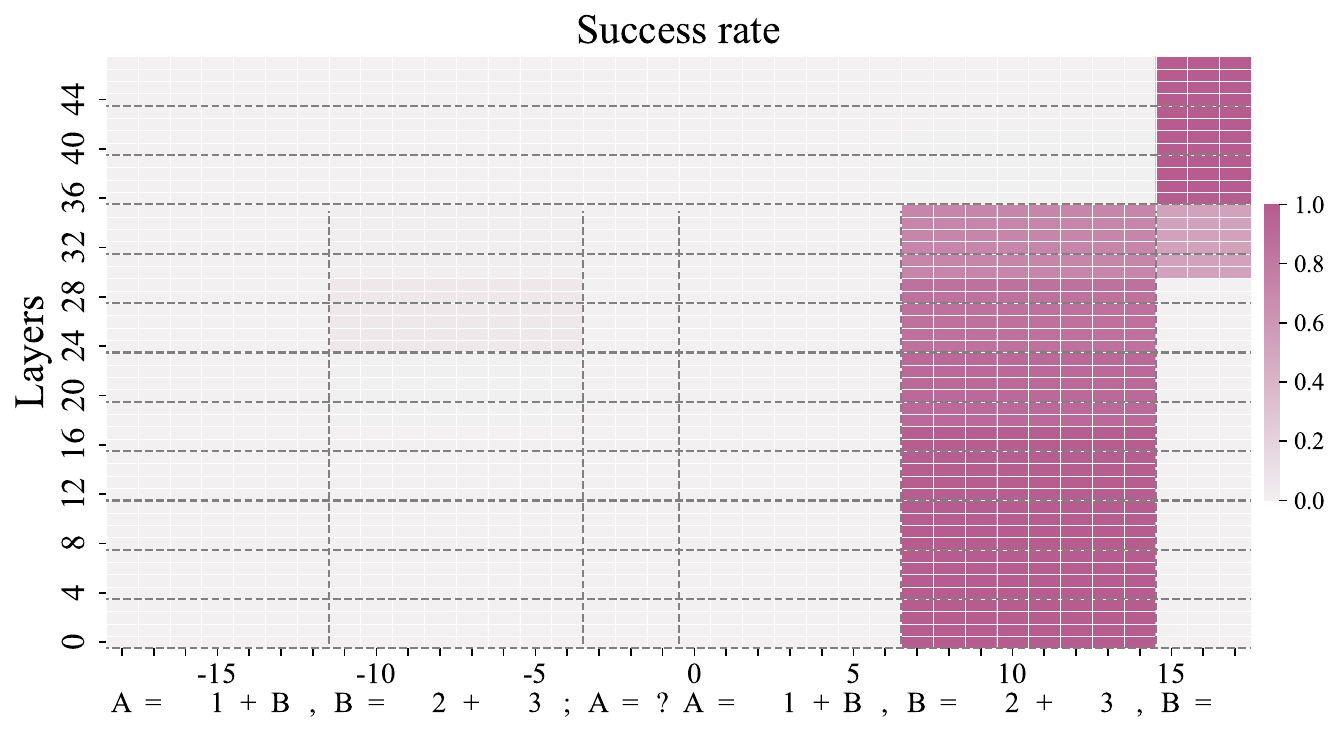}
  \caption{Results of the causal intervention on Qwen2.5-14B. Each grid cell shows the success rate when the intermediate token $z_{17}$ \(({\underline{\textcolor{gray}{\mathrm{B}=}5}}_{\hspace{0.05cm}\mathbf{2}})\) is the target token.}
  \label{fig:intervention_qwen2.5_14B_mid_5}
\end{figure}

\begin{figure}[t]
  \centering
  \includegraphics[width=0.97\linewidth]{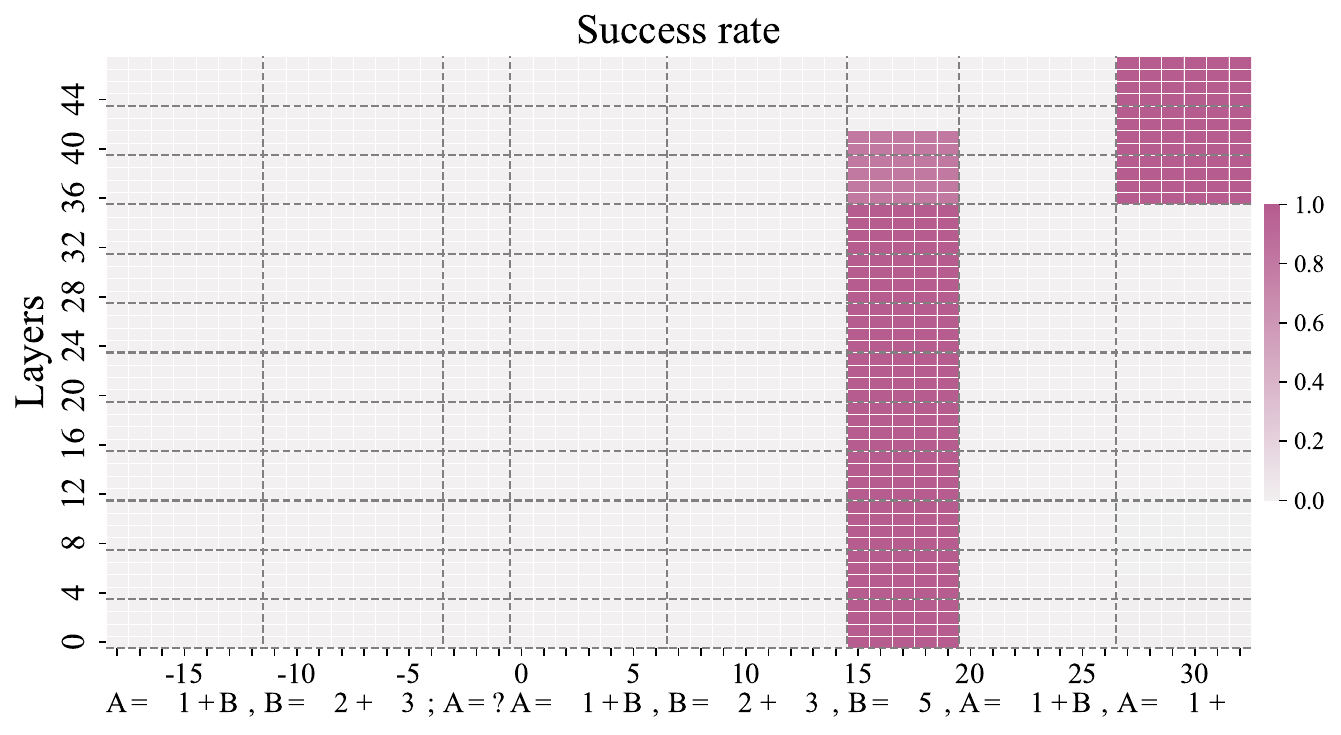}
  \caption{Results of the causal intervention on Qwen2.5-14B. Each grid cell shows the success rate when $z_{32}$ \(({\underline{\textcolor{gray}{\mathrm{A}=1+}5}}_{\hspace{0.05cm}\mathbf{4}})\) is the target token.}
  \label{fig:intervention_qwen2.5_14B_last_5}
\end{figure}

\begin{figure}[t]
  \centering
  \includegraphics[width=0.97\linewidth]{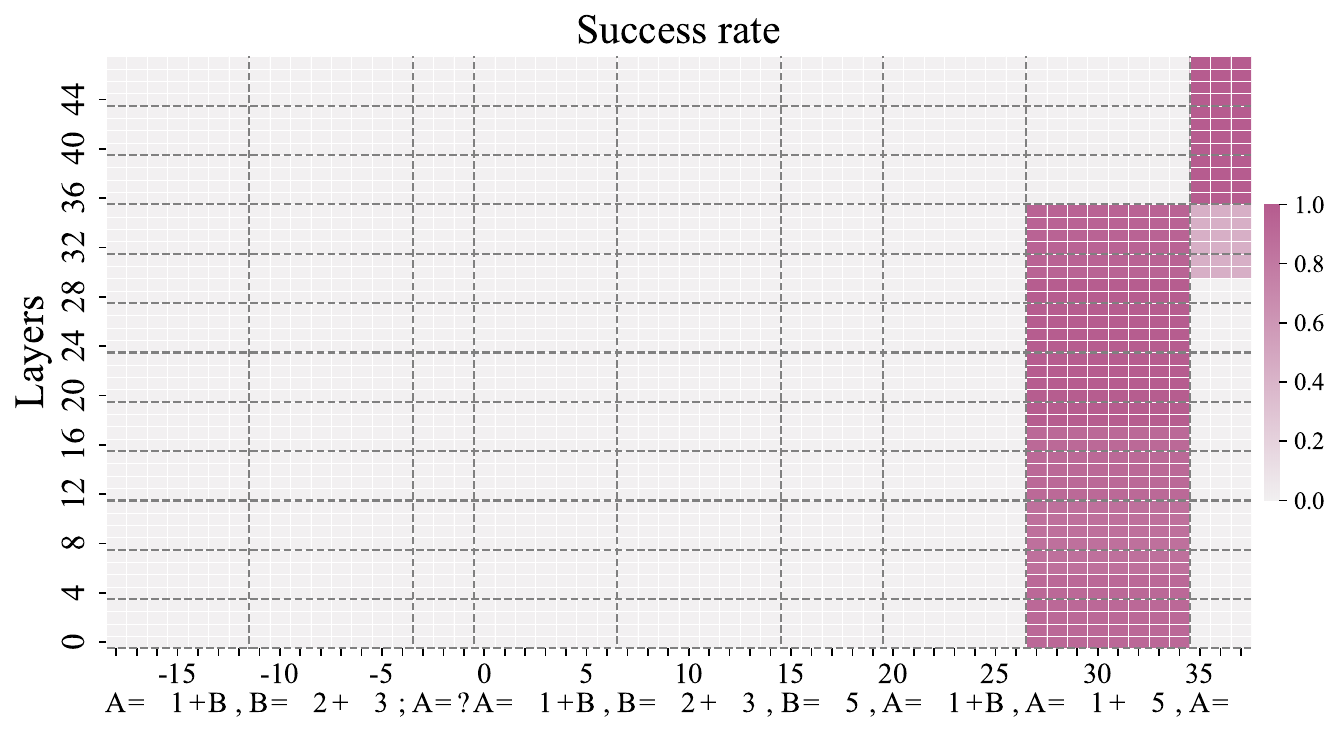}
  \caption{Results of the causal intervention on Qwen2.5-14B. Each grid cell shows the success rate when the final answer $y$ \(({\underline{\textcolor{gray}{\mathrm{A}=}\textbf{6}}}_{\hspace{0.05cm}\mathbf{5}})\) is the target token.}
  \label{fig:intervention_qwen2.5_14B_last_answer}
\end{figure}

\begin{figure}[t]
  \centering
  \includegraphics[width=0.97\linewidth]{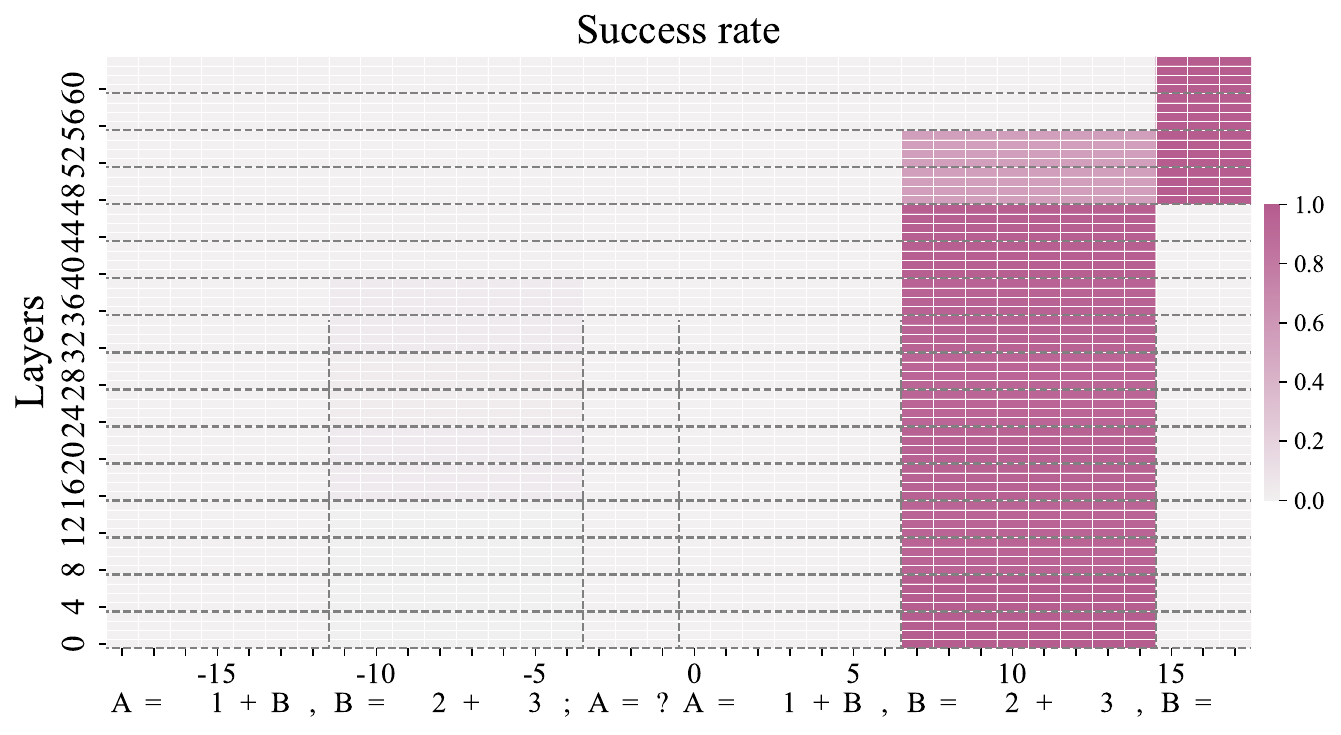}
  \caption{Results of the causal intervention on Qwen2.5-32B. Each grid cell shows the success rate when the intermediate token $z_{17}$ \(({\underline{\textcolor{gray}{\mathrm{B}=}5}}_{\hspace{0.05cm}\mathbf{2}})\) is the target token.}
  \label{fig:intervention_qwen2.5_32B_mid_5}
\end{figure}

\begin{figure}[t]
  \centering
  \includegraphics[width=0.97\linewidth]{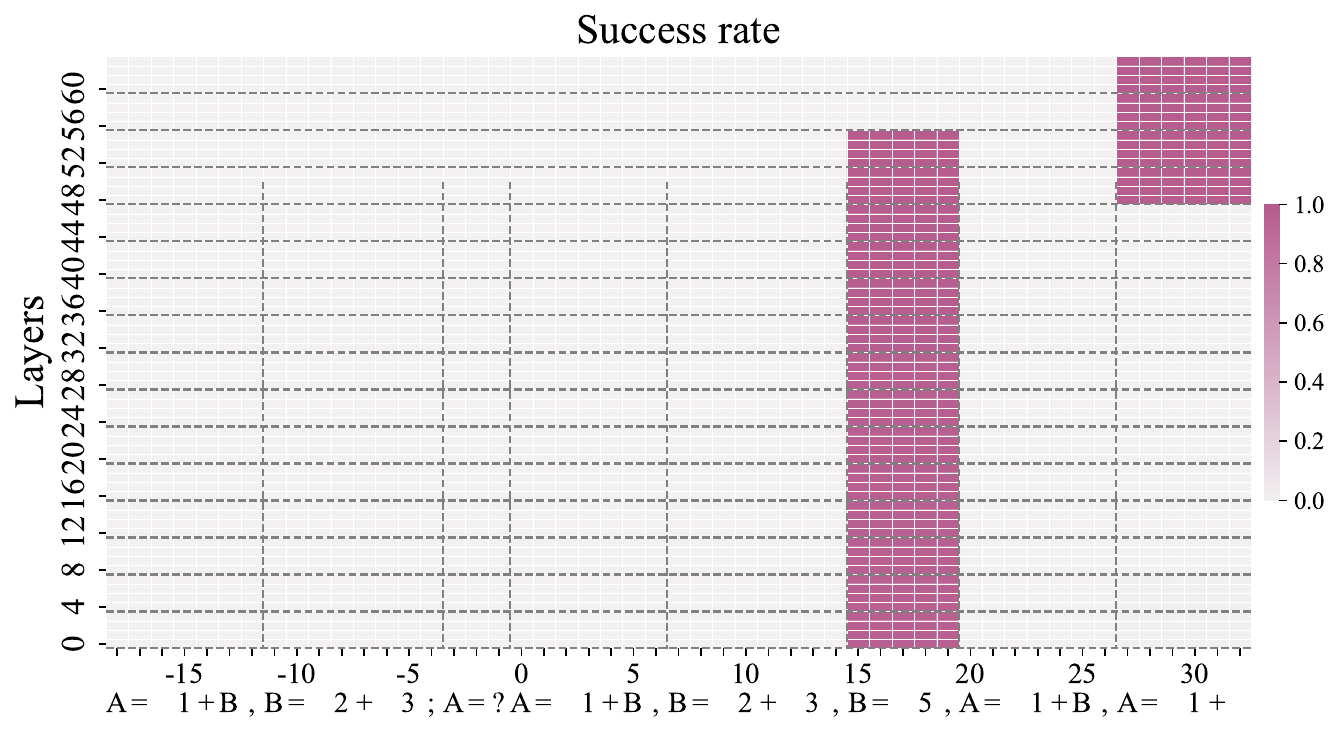}
  \caption{Results of the causal intervention on Qwen2.5-32B. Each grid cell shows the success rate when $z_{32}$ \(({\underline{\textcolor{gray}{\mathrm{A}=1+}5}}_{\hspace{0.05cm}\mathbf{4}})\) is the target token.}
  \label{fig:intervention_qwen2.5_32B_last_5}
\end{figure}

\begin{figure}[t]
  \centering
  \includegraphics[width=0.97\linewidth]{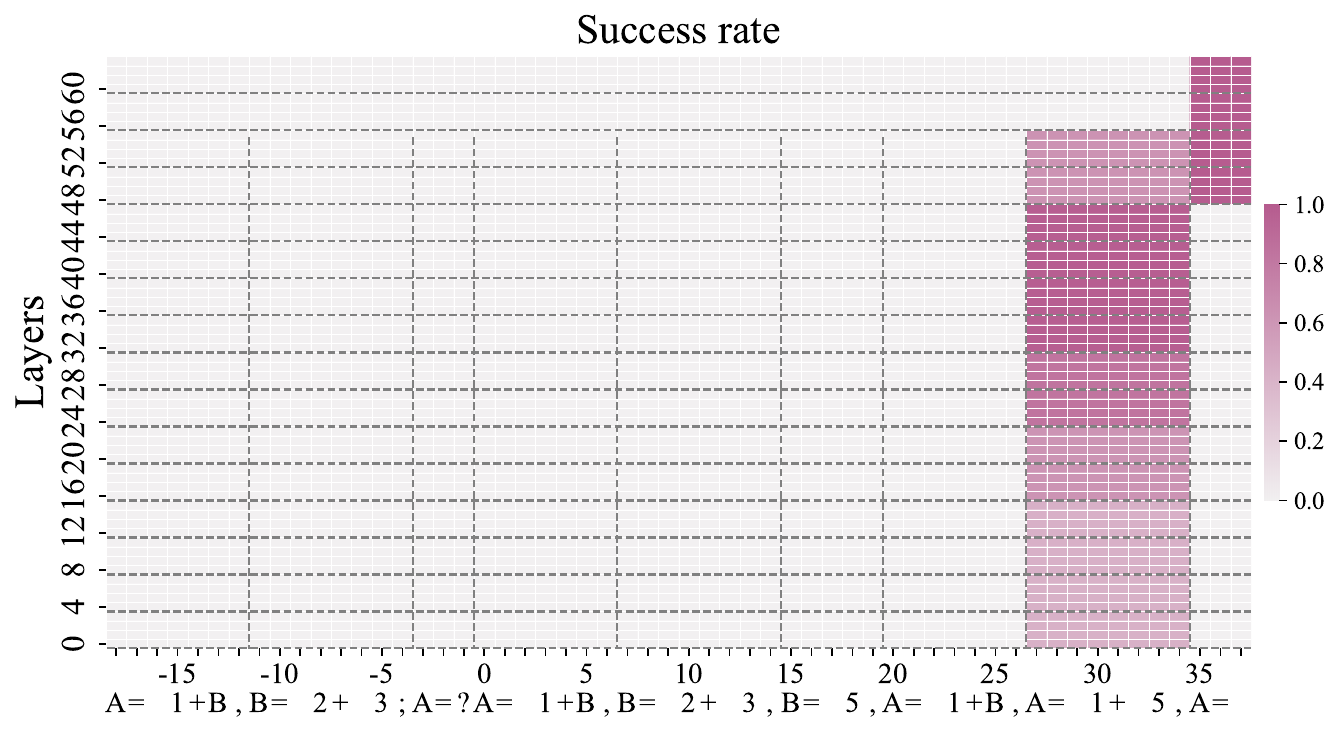}
  \caption{Results of the causal intervention on Qwen2.5-32B. Each grid cell shows the success rate when the final answer $y$ \(({\underline{\textcolor{gray}{\mathrm{A}=}\textbf{6}}}_{\hspace{0.05cm}\mathbf{5}})\) is the target token.}
  \label{fig:intervention_qwen2.5_32B_last_answer}
\end{figure}

\begin{figure}[t]
  \centering
  \includegraphics[width=0.97\linewidth]{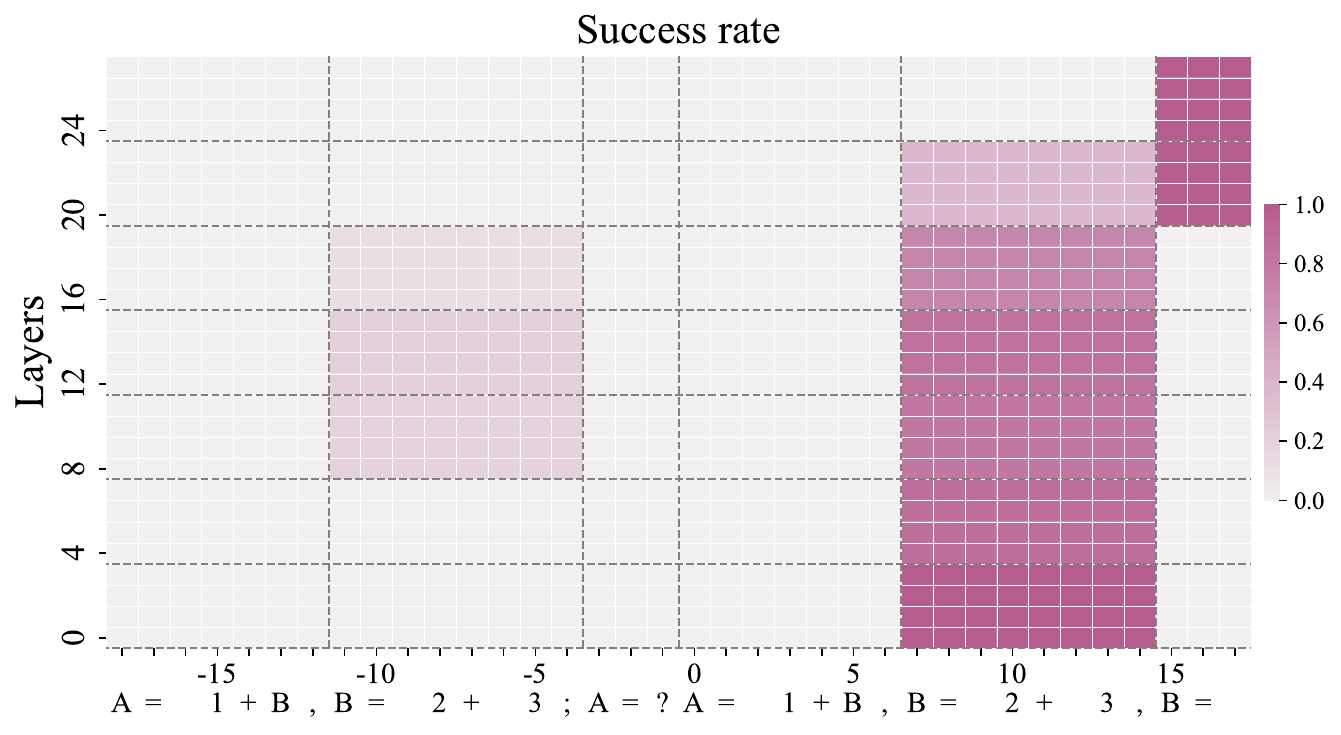}
  \caption{Results of the causal intervention on Qwen2.5-Math-7B. Each grid cell shows the success rate when the intermediate token $z_{17}$ \(({\underline{\textcolor{gray}{\mathrm{B}=}5}}_{\hspace{0.05cm}\mathbf{2}})\) is the target token.}
  \label{fig:intervention_qwen2.5_math_7B_mid_5}
\end{figure}

\begin{figure}[t]
  \centering
  \includegraphics[width=0.97\linewidth]{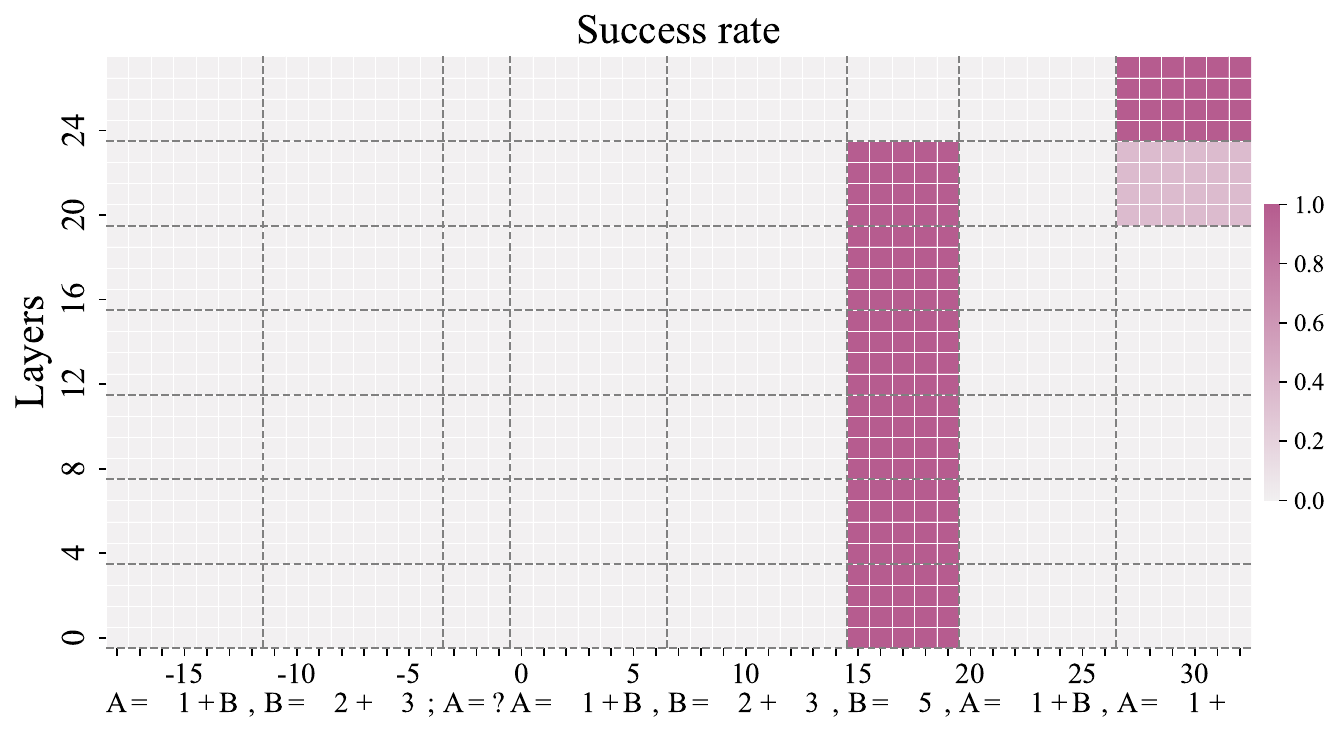}
  \caption{Results of the causal intervention on Qwen2.5-Math-7B. Each grid cell shows the success rate when $z_{32}$ \(({\underline{\textcolor{gray}{\mathrm{A}=1+}5}}_{\hspace{0.05cm}\mathbf{4}})\) is the target token.}
  \label{fig:intervention_qwen2.5_math_7B_last_5}
\end{figure}

\begin{figure}[t]
  \centering
  \includegraphics[width=0.97\linewidth]{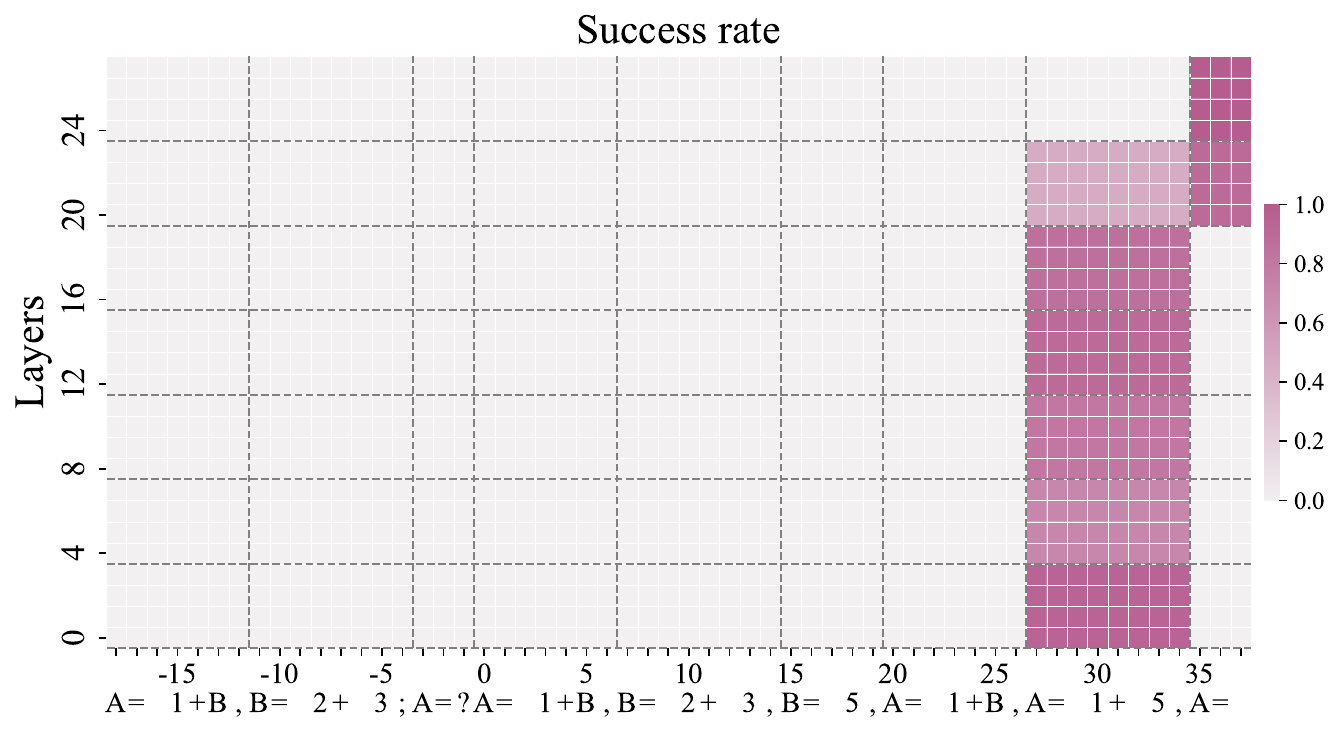}
  \caption{Results of the causal intervention on Qwen2.5-Math-7B. Each grid cell shows the success rate when the final answer $y$ \(({\underline{\textcolor{gray}{\mathrm{A}=}\textbf{6}}}_{\hspace{0.05cm}\mathbf{5}})\) is the target token.}
  \label{fig:intervention_qwen2.5_math_7B_last_answer}
\end{figure}

\begin{figure}[t]
  \centering
  \includegraphics[width=0.97\linewidth]{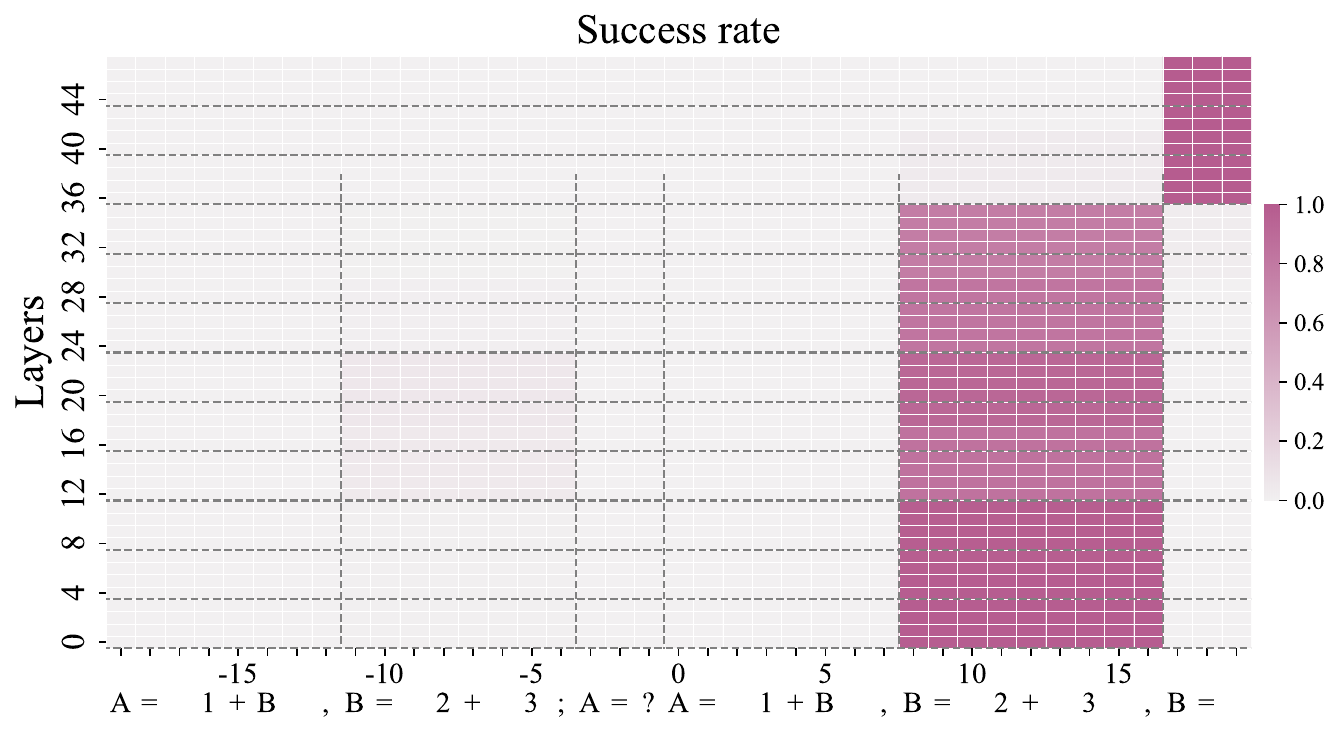}
  \caption{Results of the causal intervention on Yi-1.5-9B. Each grid cell shows the success rate when the intermediate token $z_{17}$ \(({\underline{\textcolor{gray}{\mathrm{B}=}5}}_{\hspace{0.05cm}\mathbf{2}})\) is the target token.}
  \label{fig:intervention_yi_1.5_9B_mid_5}
\end{figure}

\begin{figure}[t]
  \centering
  \includegraphics[width=0.97\linewidth]{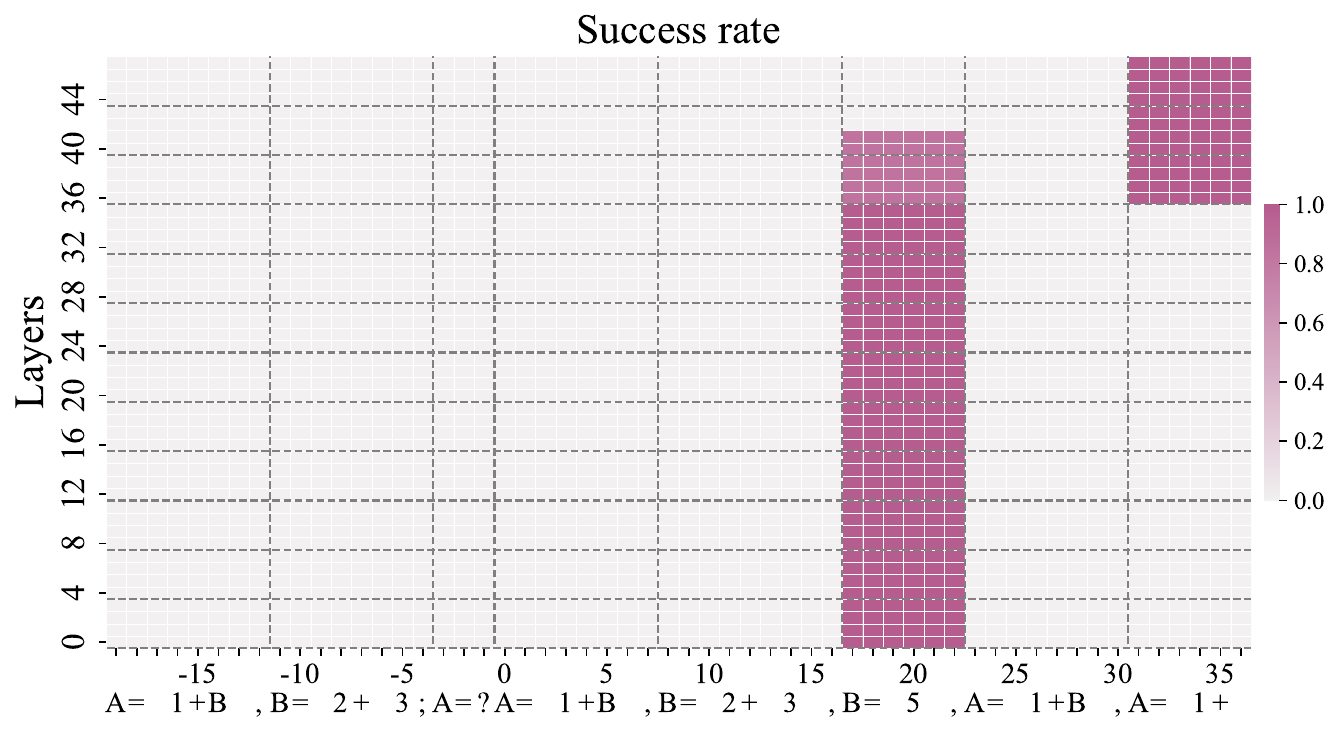}
  \caption{Results of the causal intervention on Yi-1.5-9B. Each grid cell shows the success rate when $z_{32}$ \(({\underline{\textcolor{gray}{\mathrm{A}=1+}5}}_{\hspace{0.05cm}\mathbf{4}})\) is the target token.}
  \label{fig:intervention_yi_1.5_9B_last_5}
\end{figure}

\begin{figure}[t]
  \centering
  \includegraphics[width=0.97\linewidth]{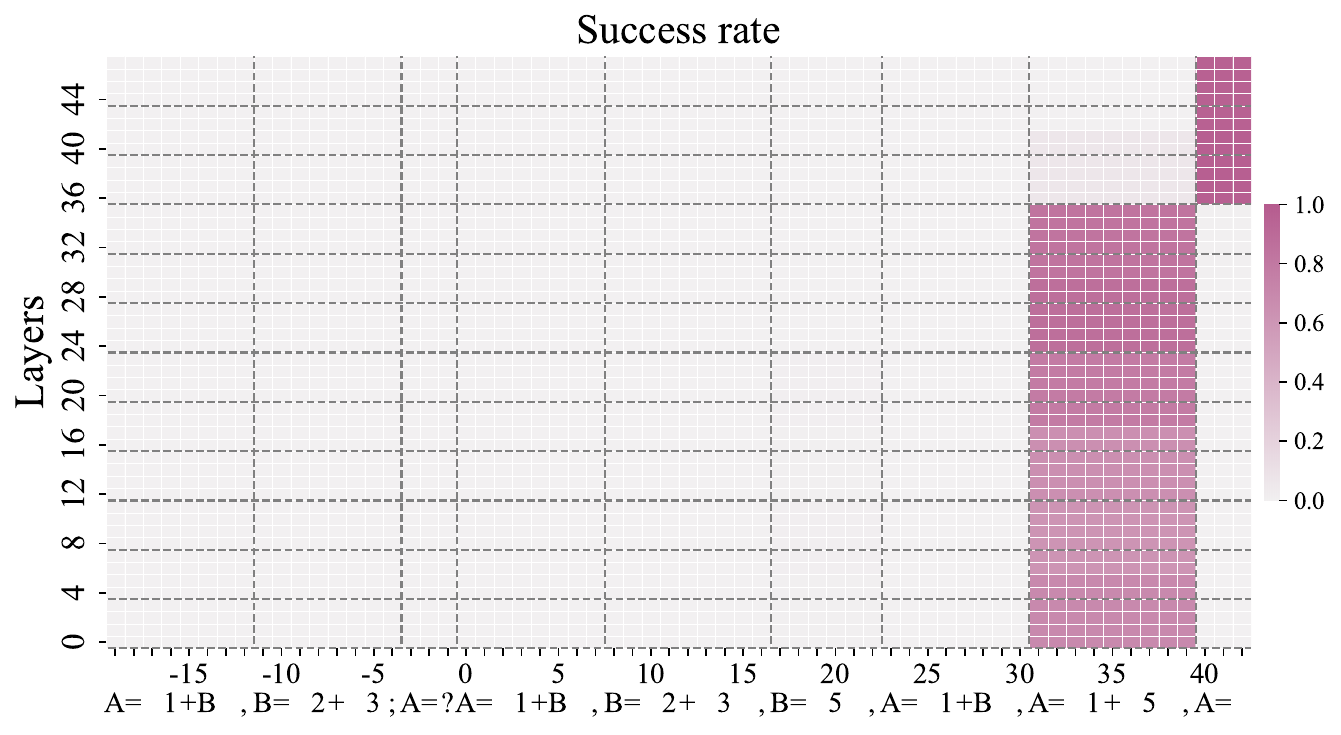}
  \caption{Results of the causal intervention on Yi-1.5-9B. Each grid cell shows the success rate when the final answer $y$ \(({\underline{\textcolor{gray}{\mathrm{A}=}\textbf{6}}}_{\hspace{0.05cm}\mathbf{5}})\) is the target token.}
  \label{fig:intervention_yi_1.5_9B_last_answer}
\end{figure}

\begin{figure}[t]
  \centering
  \includegraphics[width=0.97\linewidth]{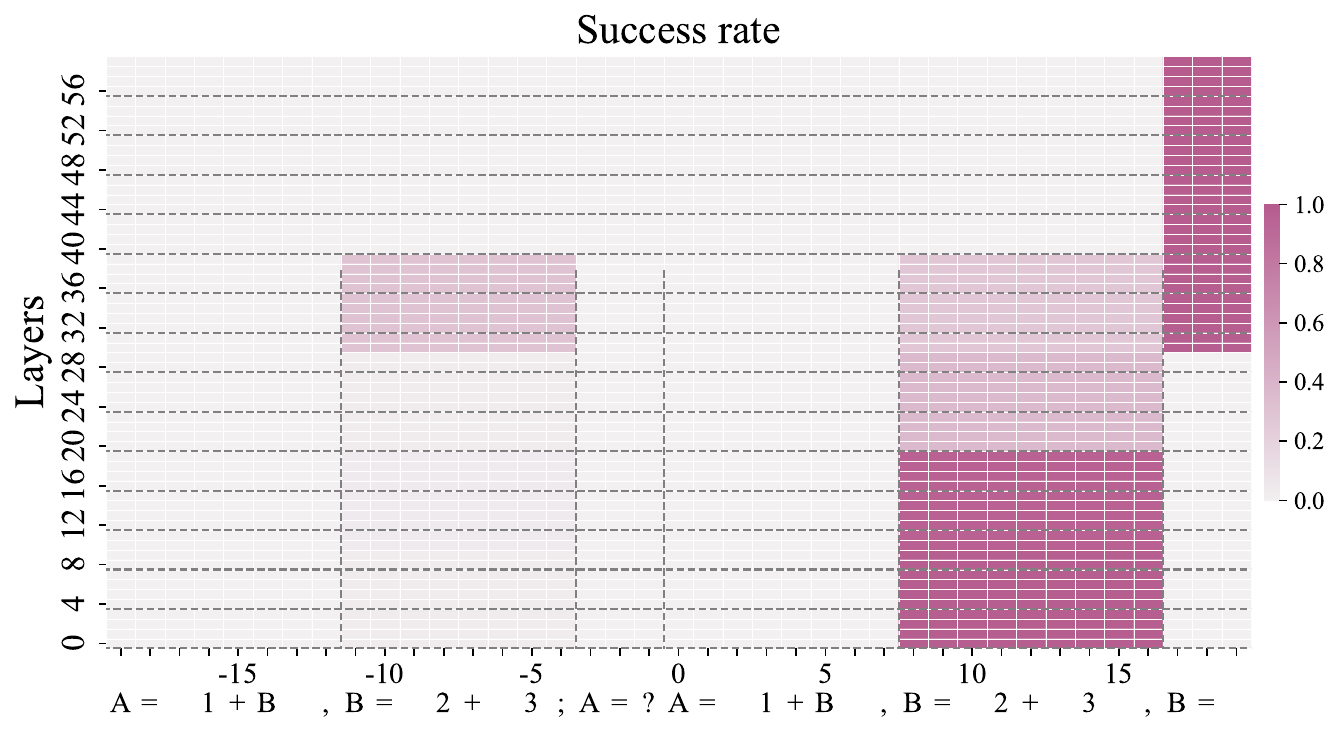}
  \caption{Results of the causal intervention on Yi-1.5-34B. Each grid cell shows the success rate when the intermediate token $z_{17}$ \(({\underline{\textcolor{gray}{\mathrm{B}=}5}}_{\hspace{0.05cm}\mathbf{2}})\) is the target token.}
  \label{fig:intervention_yi_1.5_34B_mid_5}
\end{figure}

\begin{figure}[t]
  \centering
  \includegraphics[width=0.97\linewidth]{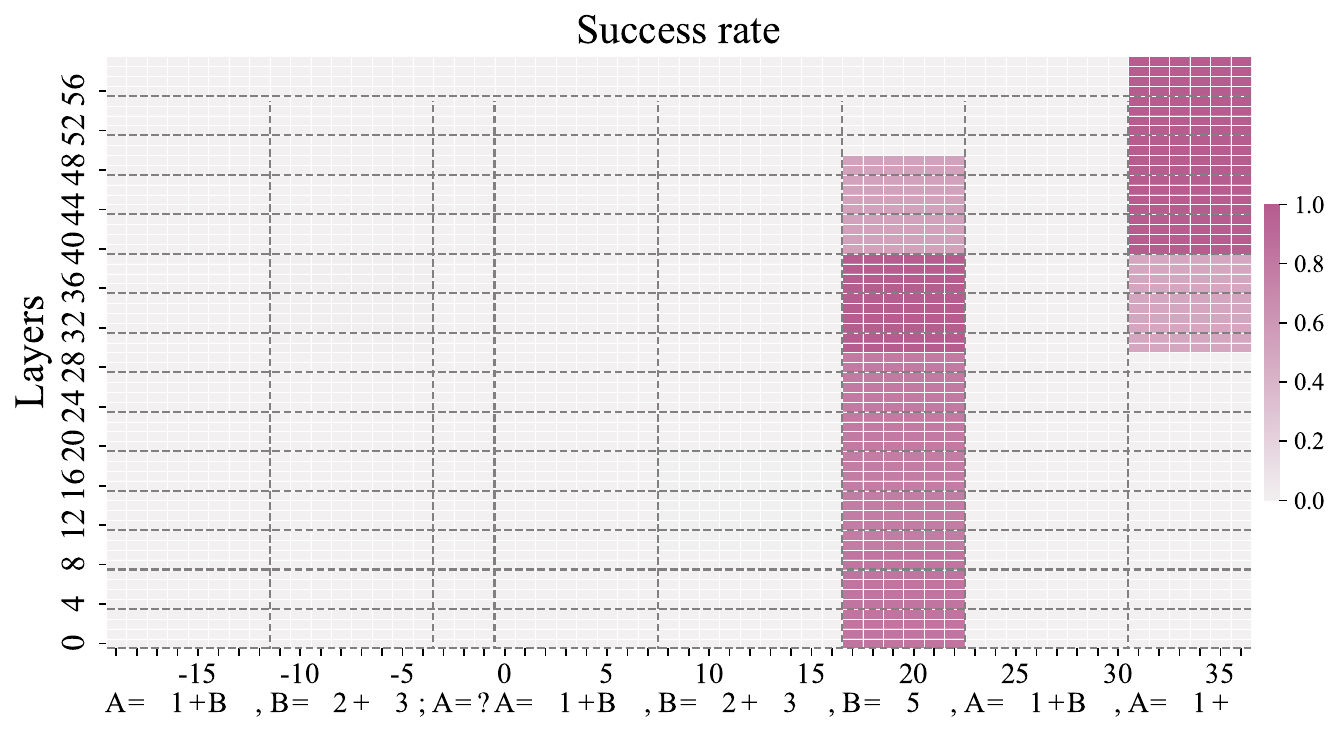}
  \caption{Results of the causal intervention on Yi-1.5-34B. Each grid cell shows the success rate when $z_{32}$ \(({\underline{\textcolor{gray}{\mathrm{A}=1+}5}}_{\hspace{0.05cm}\mathbf{4}})\) is the target token.}
  \label{fig:intervention_yi_1.5_34B_last_5}
\end{figure}

\begin{figure}[t]
  \centering
  \includegraphics[width=0.97\linewidth]{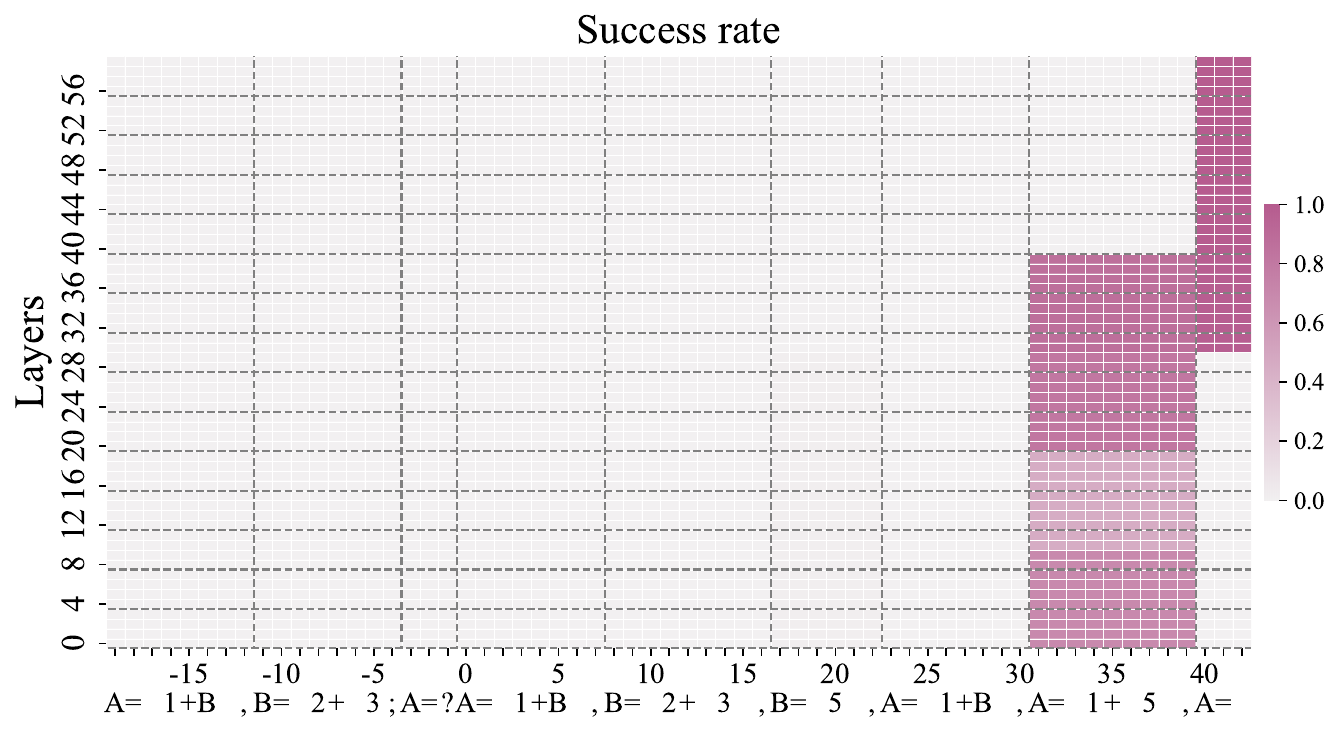}
  \caption{Results of the causal intervention on Yi-1.5-34B. Each grid cell shows the success rate when the final answer $y$ \(({\underline{\textcolor{gray}{\mathrm{A}=}\textbf{6}}}_{\hspace{0.05cm}\mathbf{5}})\) is the target token.}
  \label{fig:intervention_yi_1.5_34B_last_answer}
\end{figure}

\begin{figure}[t]
  \centering
  \includegraphics[width=0.97\linewidth]{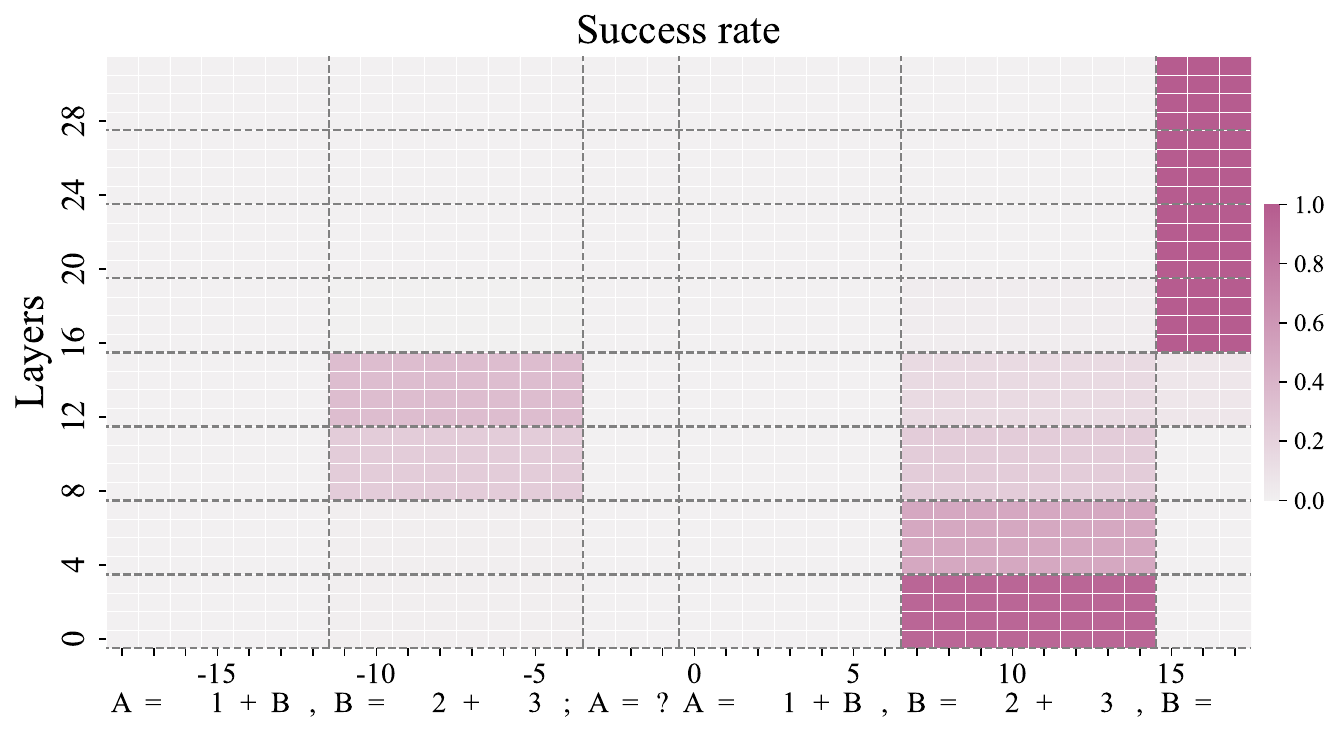}
  \caption{Results of the causal intervention on Llama-3.1-8B. Each grid cell shows the success rate when the intermediate token $z_{17}$ \(({\underline{\textcolor{gray}{\mathrm{B}=}5}}_{\hspace{0.05cm}\mathbf{2}})\) is the target token.}
  \label{fig:intervention_llama3.1_8B_mid_5}
\end{figure}

\begin{figure}[t]
  \centering
  \includegraphics[width=0.97\linewidth]{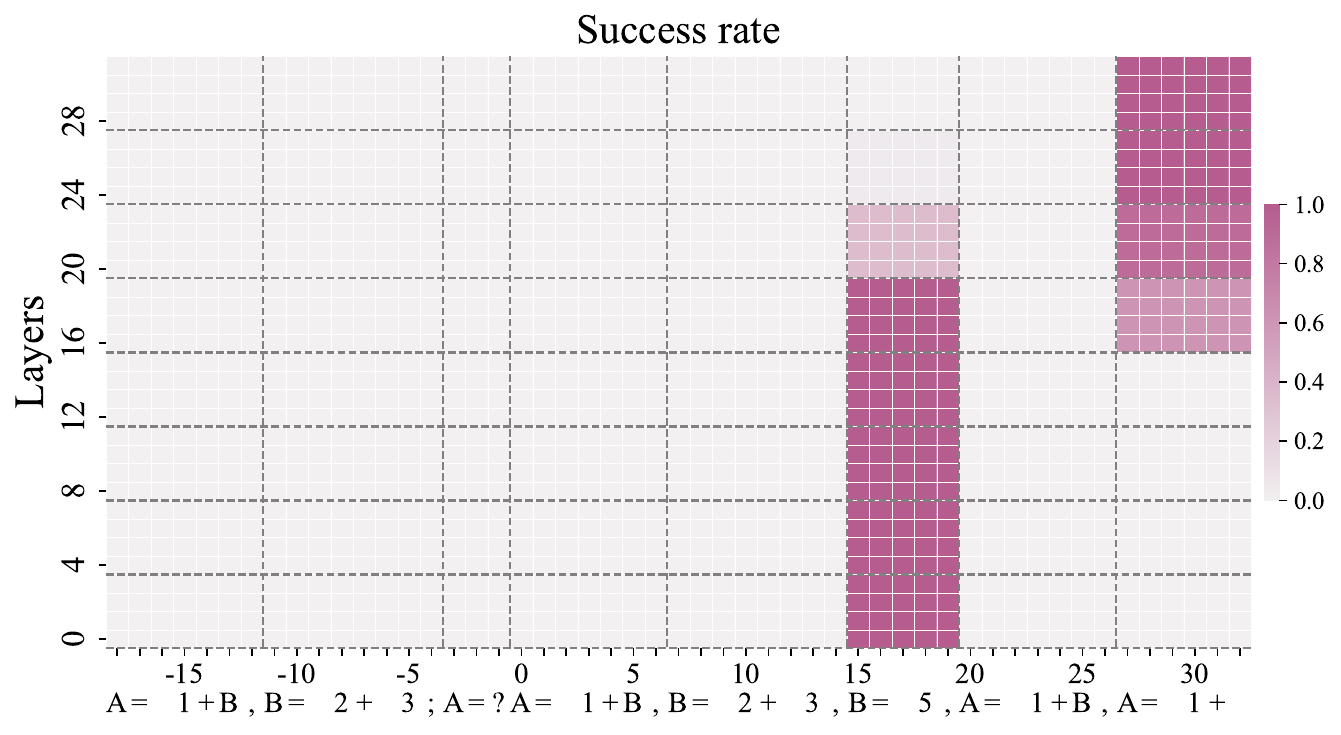}
  \caption{Results of the causal intervention on Llama-3.1-8B. Each grid cell shows the success rate when $z_{32}$ \(({\underline{\textcolor{gray}{\mathrm{A}=1+}5}}_{\hspace{0.05cm}\mathbf{4}})\) is the target token.}
  \label{fig:intervention_llama3.1_8B_last_5}
\end{figure}

\begin{figure}[t]
  \centering
  \includegraphics[width=0.97\linewidth]{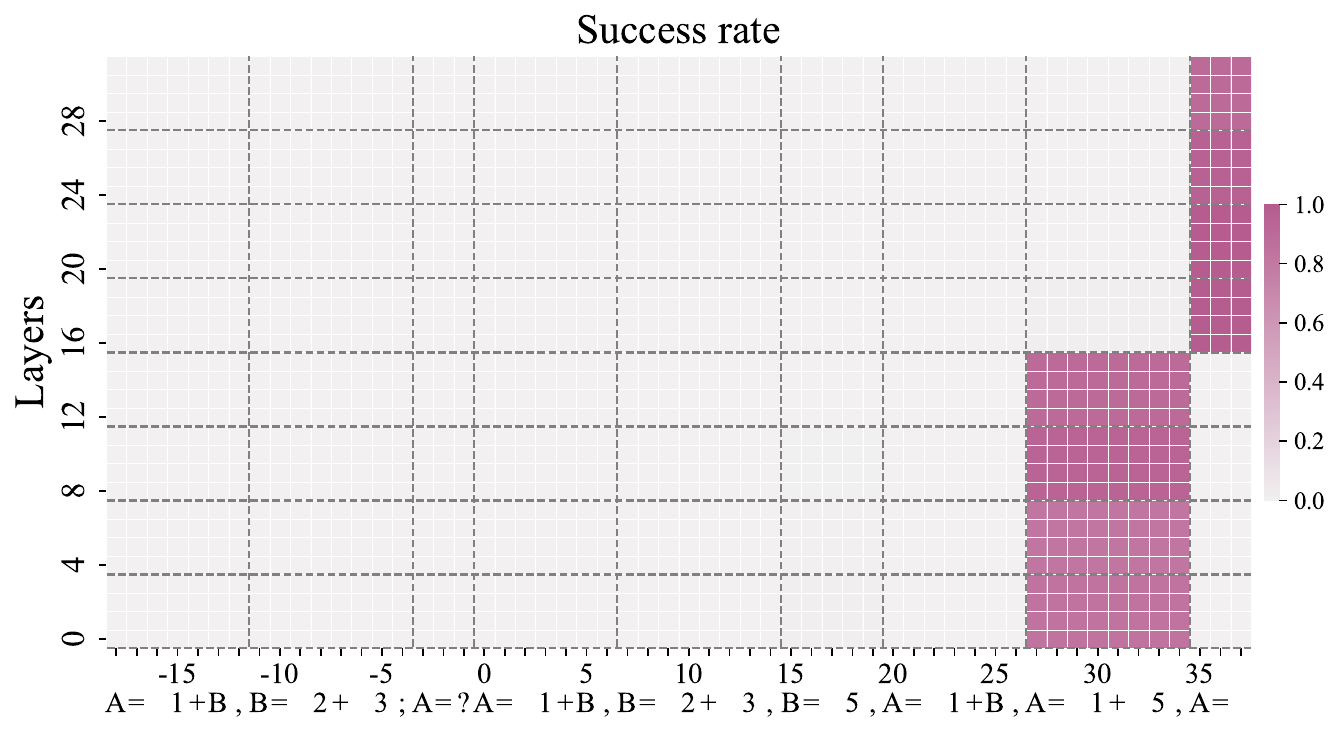}
  \caption{Results of the causal intervention on Llama-3.1-8B. Each grid cell shows the success rate when the final answer $y$ \(({\underline{\textcolor{gray}{\mathrm{A}=}\textbf{6}}}_{\hspace{0.05cm}\mathbf{5}})\) is the target token.}
  \label{fig:intervention_llama3.1_8B_last_answer}
\end{figure}

\begin{figure}[t]
  \centering
  \includegraphics[width=0.97\linewidth]{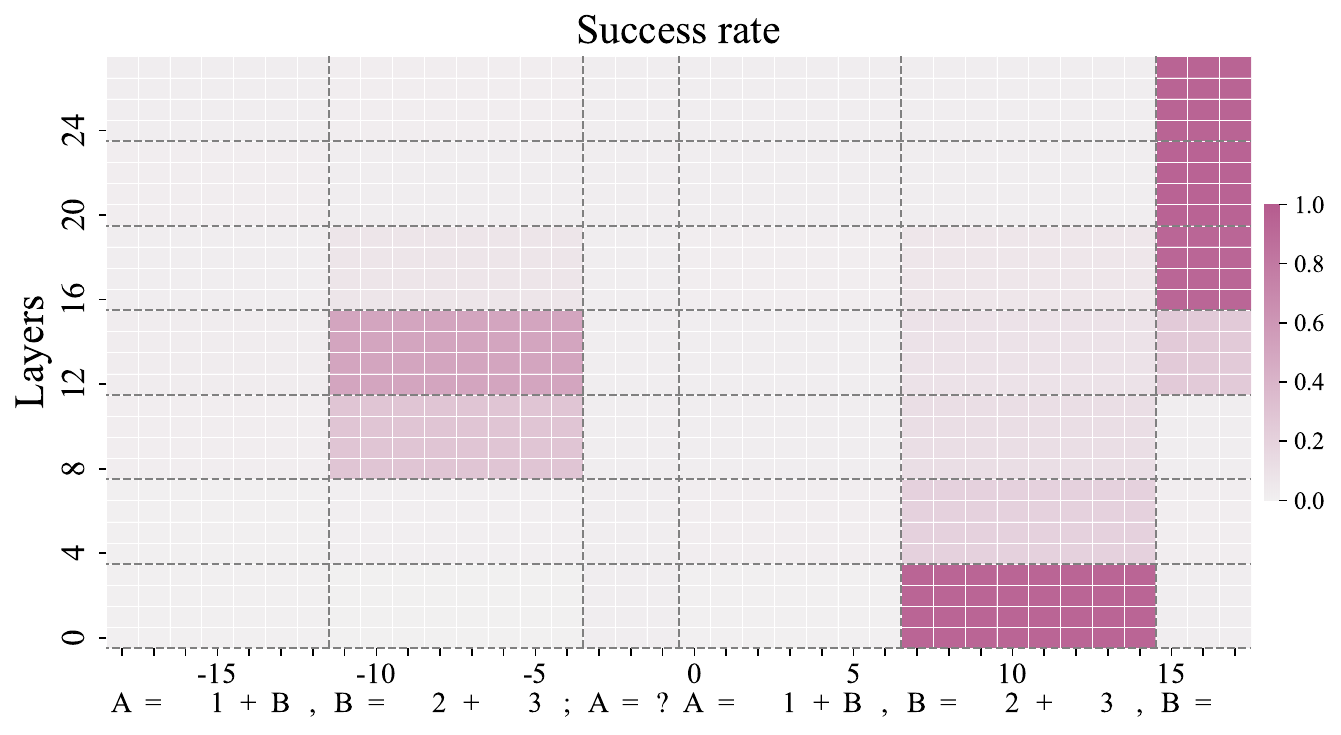}
  \caption{Results of the causal intervention on Llama-3.2-3B. Each grid cell shows the success rate when the intermediate token $z_{17}$ \(({\underline{\textcolor{gray}{\mathrm{B}=}5}}_{\hspace{0.05cm}\mathbf{2}})\) is the target token.}
  \label{fig:intervention_llama3.2_3B_mid_5}
\end{figure}

\begin{figure}[t]
  \centering
  \includegraphics[width=0.97\linewidth]{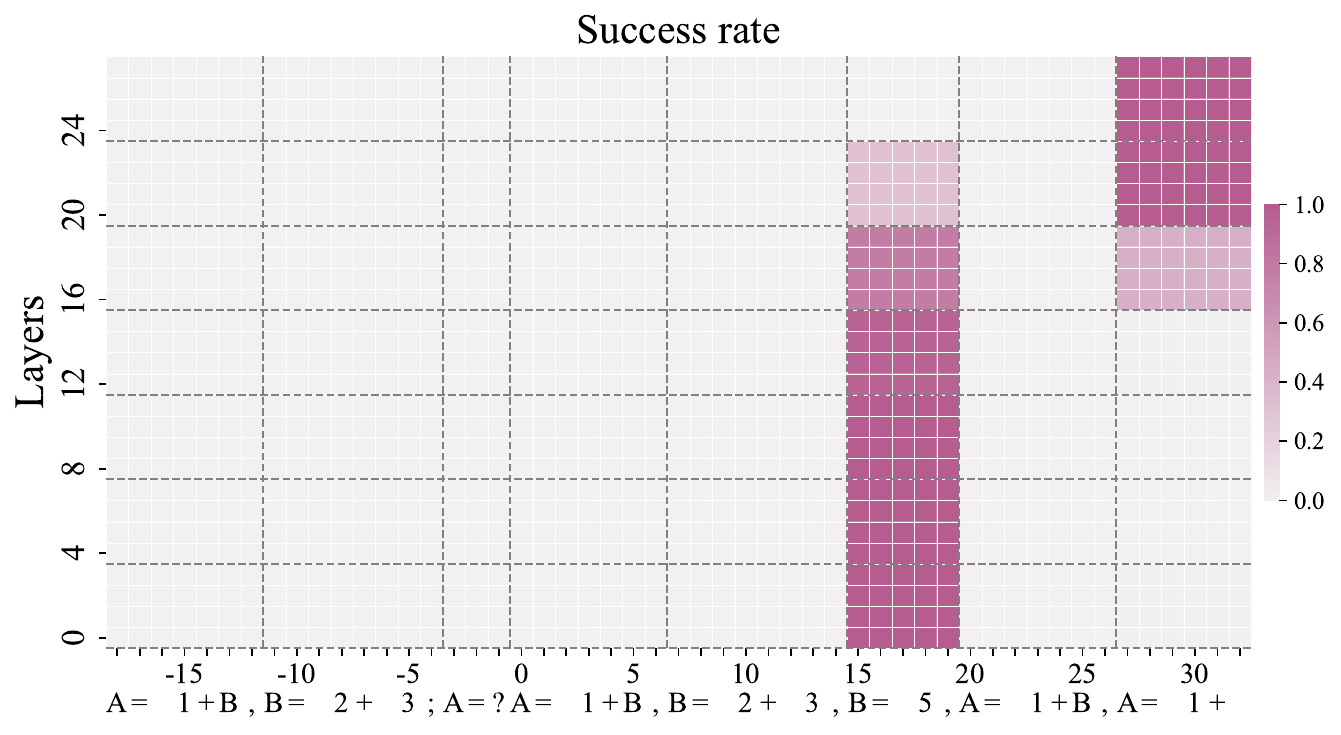}
  \caption{Results of the causal intervention on Llama-3.2-3B. Each grid cell shows the success rate when $z_{32}$ \(({\underline{\textcolor{gray}{\mathrm{A}=1+}5}}_{\hspace{0.05cm}\mathbf{4}})\) is the target token.}
  \label{fig:intervention_llama3.2_3B_last_5}
\end{figure}

\begin{figure}[t]
  \centering
  \includegraphics[width=0.97\linewidth]{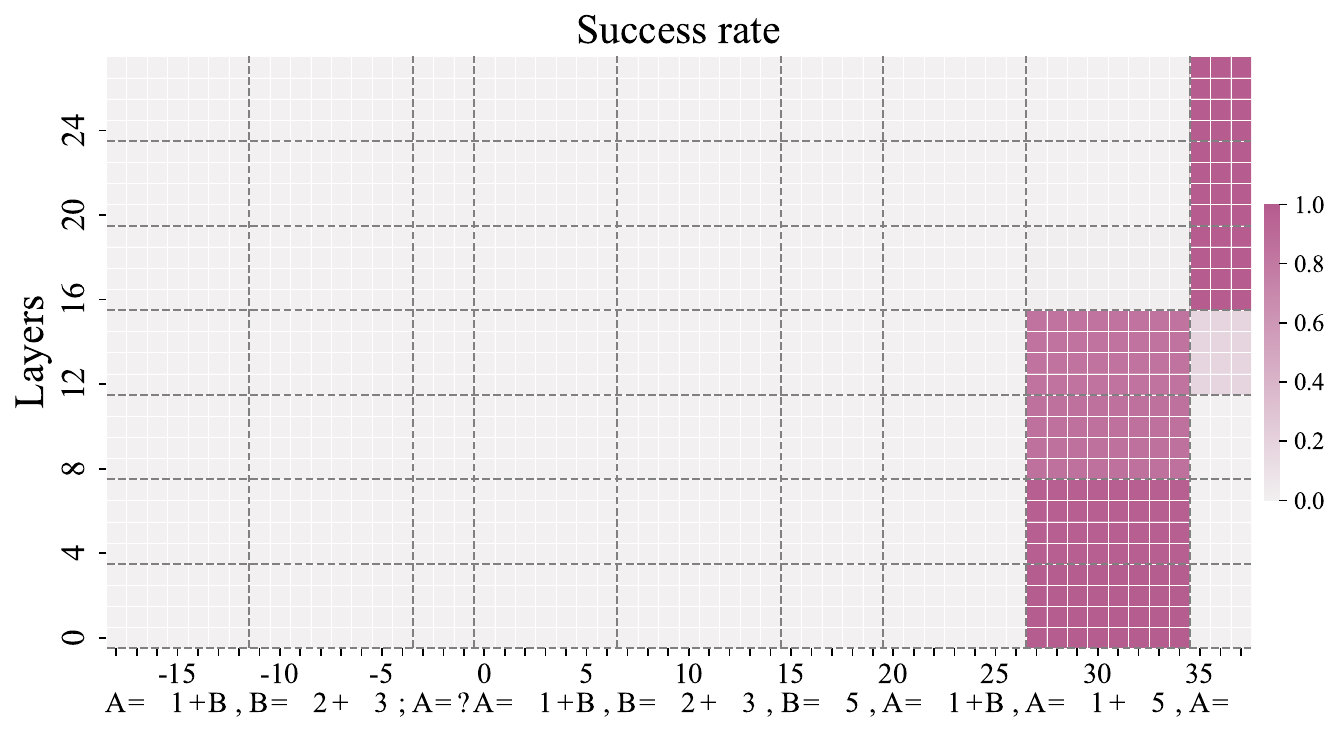}
  \caption{Results of the causal intervention on Llama-3.2-3B. Each grid cell shows the success rate when the final answer $y$ \(({\underline{\textcolor{gray}{\mathrm{A}=}\textbf{6}}}_{\hspace{0.05cm}\mathbf{5}})\) is the target token.}
  \label{fig:intervention_llama3.2_3B_last_answer}
\end{figure}

\begin{figure}[t]
  \centering
  \includegraphics[width=0.97\linewidth]{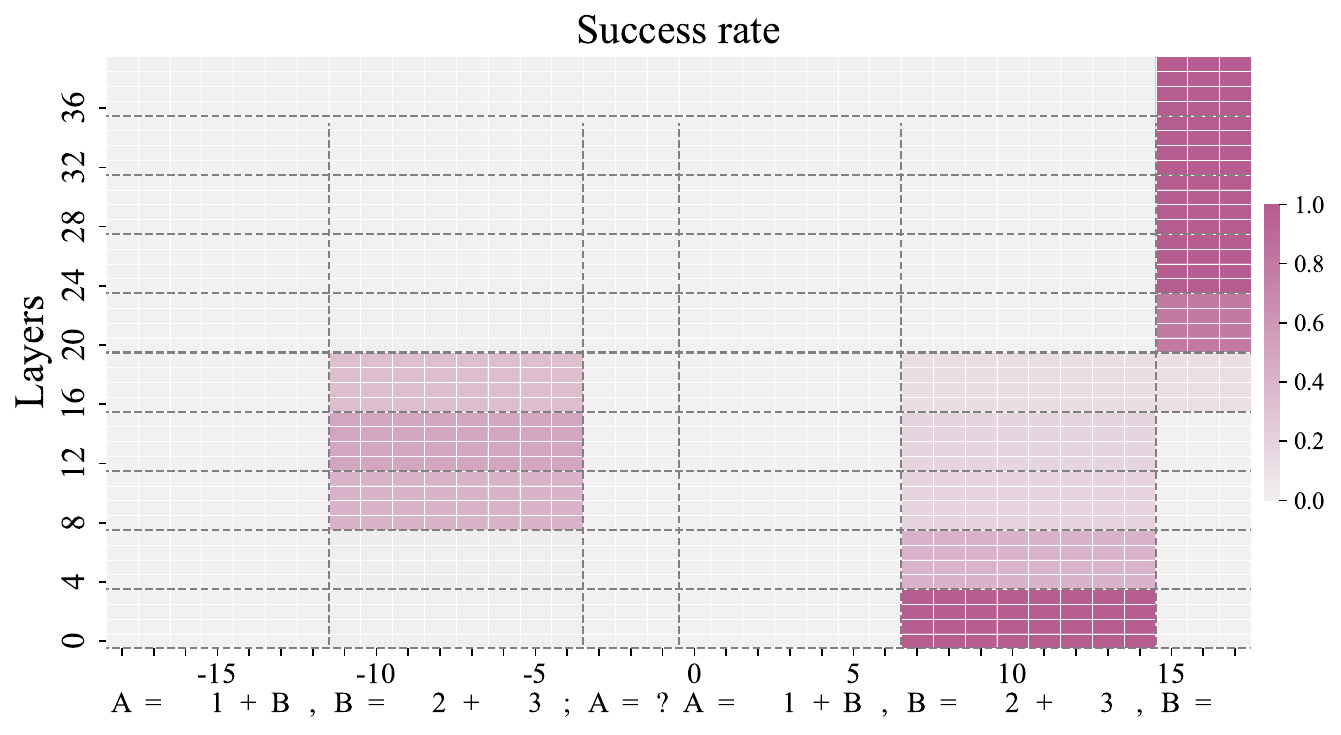}
  \caption{Results of the causal intervention on Mistral-Nemo-Base-2407. Each grid cell shows the success rate when the intermediate token $z_{17}$ \(({\underline{\textcolor{gray}{\mathrm{B}=}5}}_{\hspace{0.05cm}\mathbf{2}})\) is the target token.}
  \label{fig:intervention_mistral_nemo_base_2407_mid_5}
\end{figure}

\begin{figure}[t]
  \centering
  \includegraphics[width=0.97\linewidth]{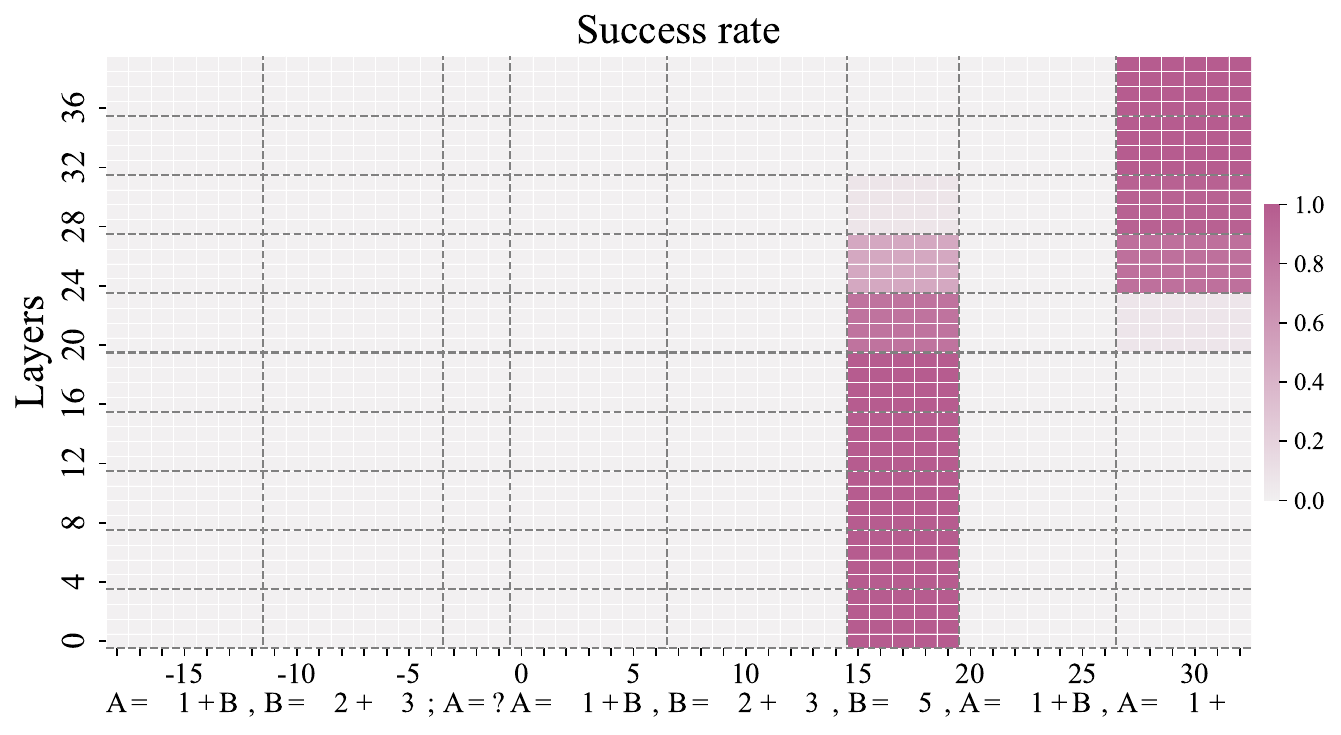}
  \caption{Results of the causal intervention on Mistral-Nemo-Base-2407. Each grid cell shows the success rate when $z_{32}$ \(({\underline{\textcolor{gray}{\mathrm{A}=1+}5}}_{\hspace{0.05cm}\mathbf{4}})\) is the target token.}
  \label{fig:intervention_mistral_nemo_base_2407_last_5}
\end{figure}

\begin{figure}[t]
  \centering
  \includegraphics[width=0.97\linewidth]{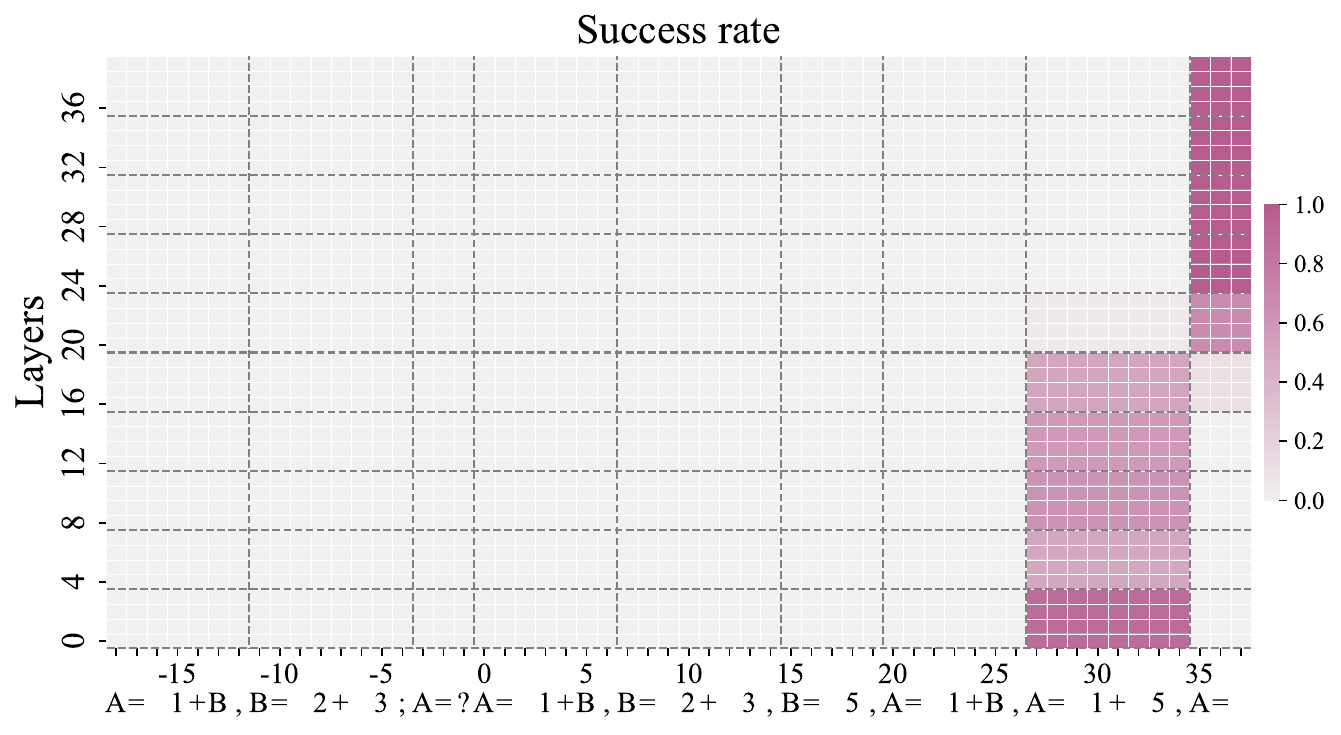}
  \caption{Results of the causal intervention on Mistral-Nemo-Base-2407. Each grid cell shows the success rate when the final answer $y$ \(({\underline{\textcolor{gray}{\mathrm{A}=}\textbf{6}}}_{\hspace{0.05cm}\mathbf{5}})\) is the target token.}
  \label{fig:intervention_mistral_nemo_base_2407_last_answer}
\end{figure}

\begin{figure}[t]
  \centering
  \includegraphics[width=0.97\linewidth]{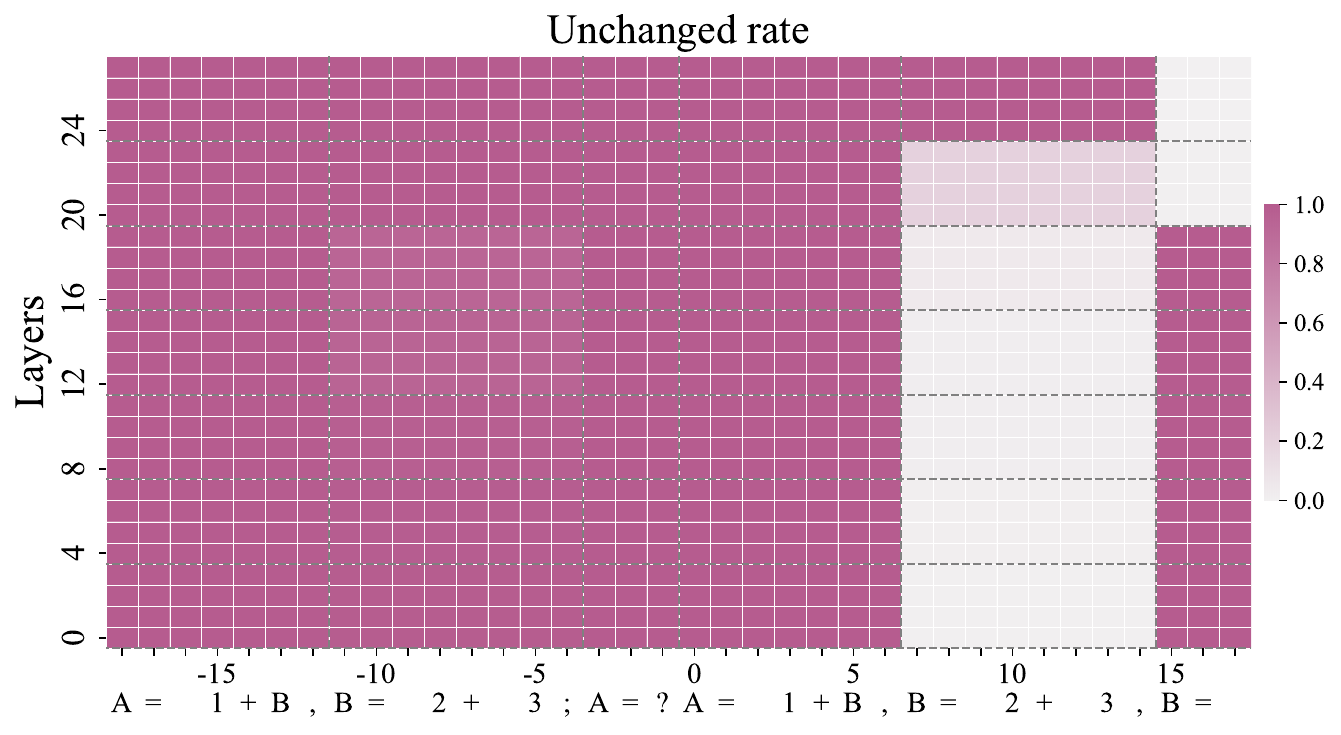}
  \caption{Results of the causal intervention on Qwen2.5-7B. Each grid cell shows the unchanged rate when the intermediate token $z_{17}$ \(({\underline{\textcolor{gray}{\mathrm{B}=}5}}_{\hspace{0.05cm}\mathbf{2}})\) is the target token.}
  \label{fig:intervention_qwen2.5_7B_mid_5_unchanged}
\end{figure}

\begin{figure}[t]
  \centering
  \includegraphics[width=0.97\linewidth]{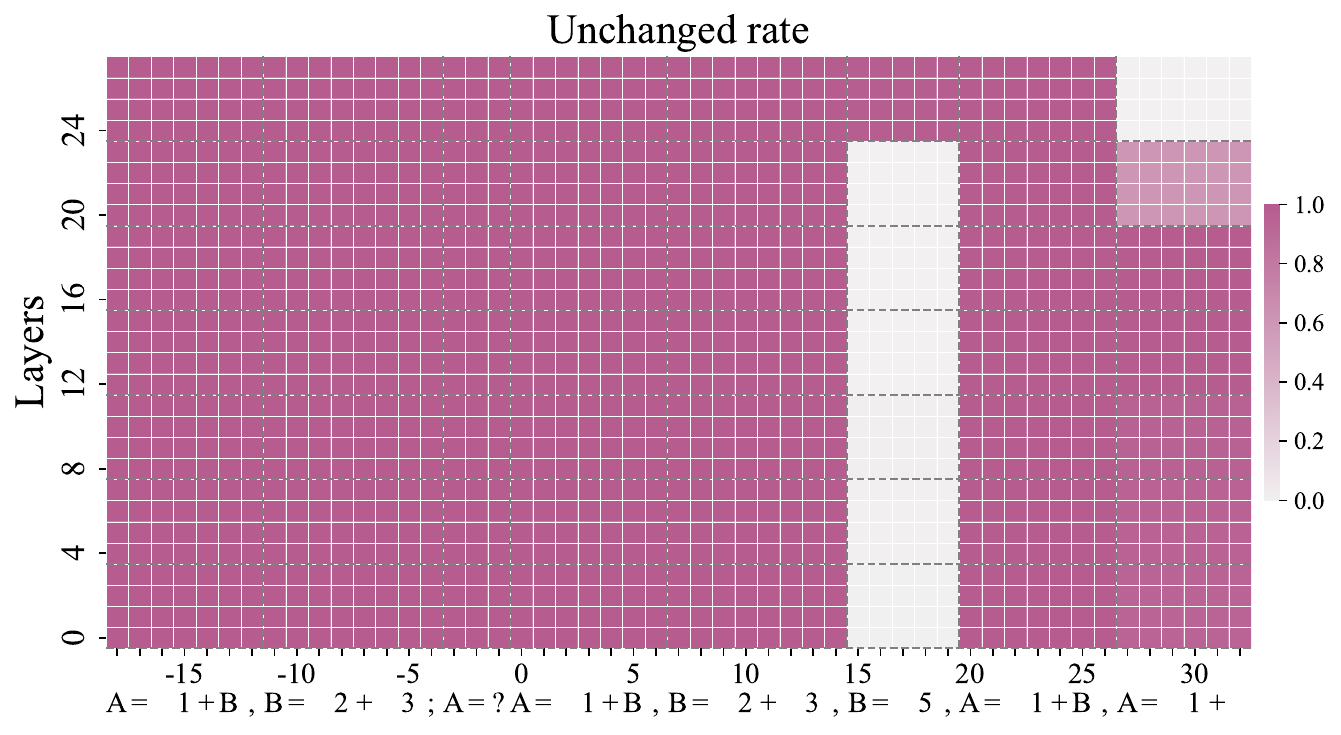}
  \caption{Results of the causal intervention on Qwen2.5-7B. Each grid cell shows the unchanged rate when $z_{32}$ \(({\underline{\textcolor{gray}{\mathrm{A}=1+}5}}_{\hspace{0.05cm}\mathbf{4}})\) is the target token.}
  \label{fig:intervention_qwen2.5_7B_last_5_unchanged}
\end{figure}

\begin{figure}[t]
  \centering
  \includegraphics[width=0.97\linewidth]{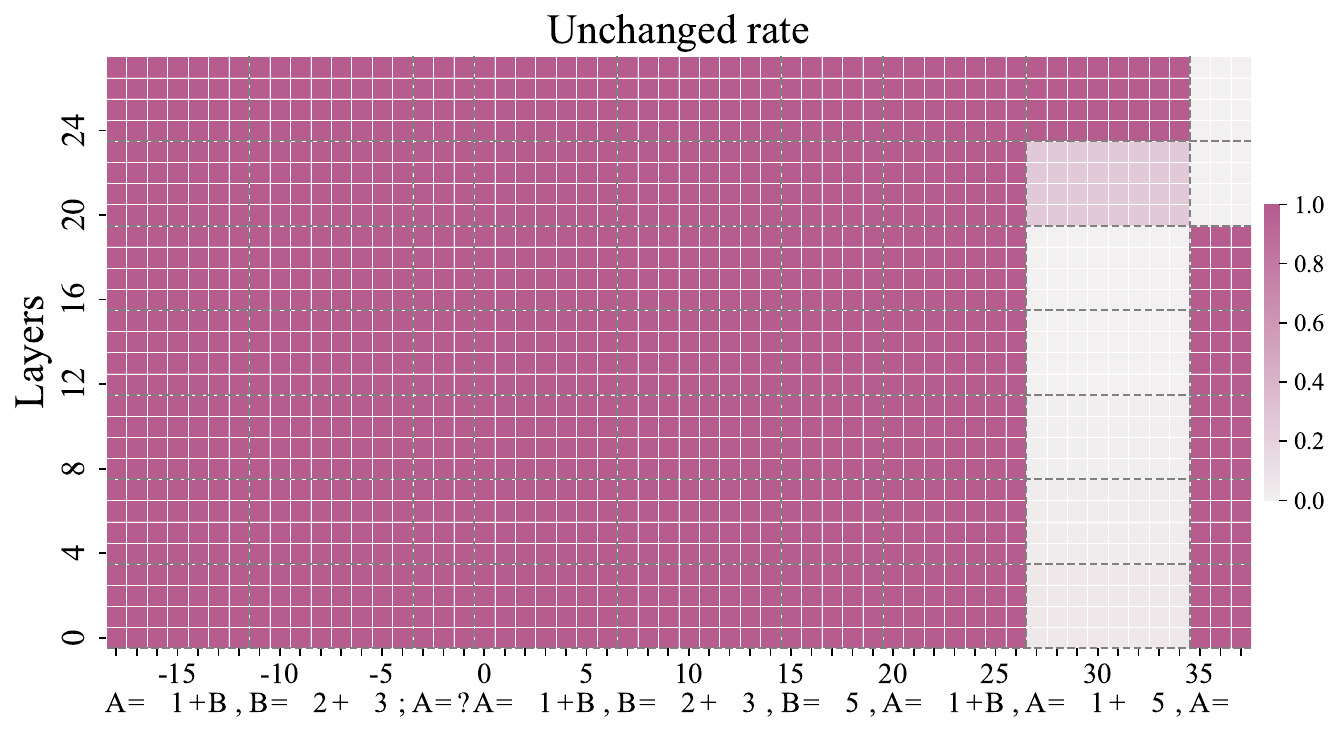}
  \caption{Results of the causal intervention on Qwen2.5-7B. Each grid cell shows the unchanged rate when the final answer $y$ \(({\underline{\textcolor{gray}{\mathrm{A}=}\textbf{6}}}_{\hspace{0.05cm}\mathbf{5}})\) is the target token.}
  \label{fig:intervention_qwen2.5_7B_last_answer_unchanged}
\end{figure}

\begin{figure}[t]
  \centering
  \includegraphics[width=0.97\linewidth]{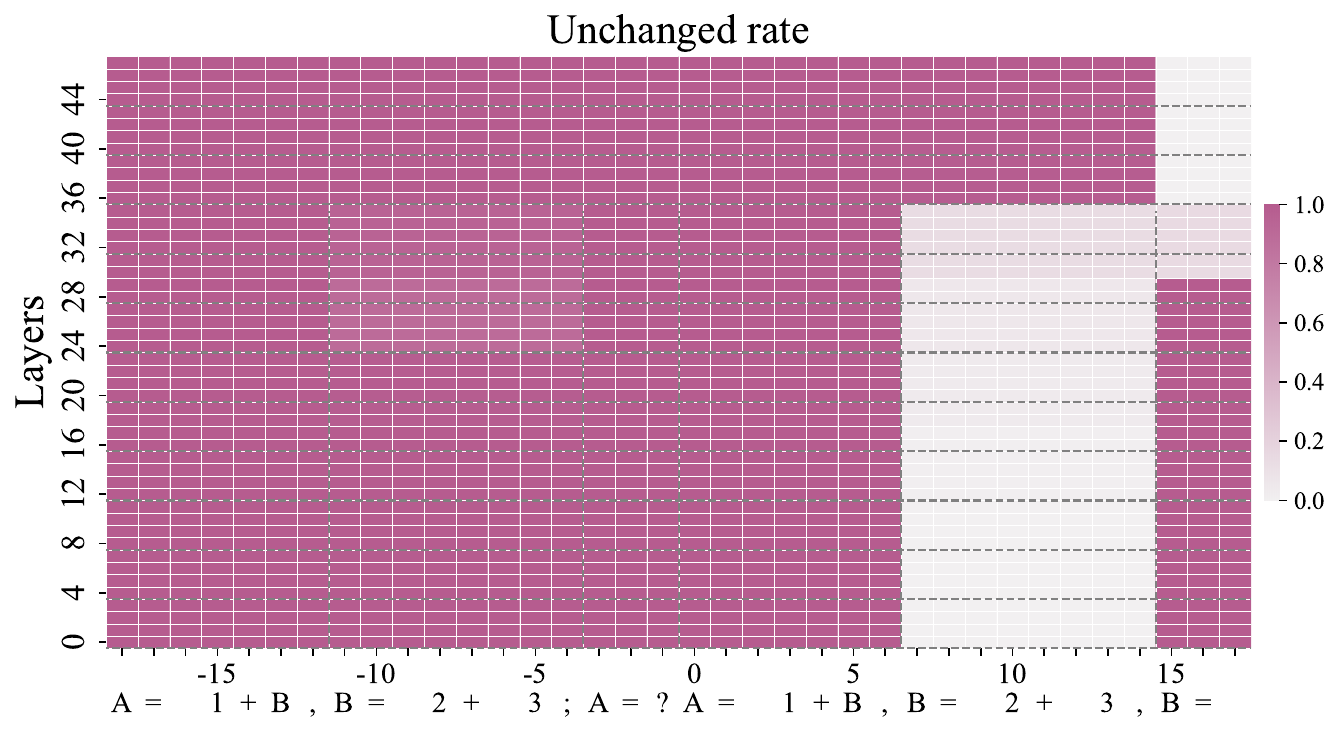}
  \caption{Results of the causal intervention on Qwen2.5-14B. Each grid cell shows the unchanged rate when the intermediate token $z_{17}$ \(({\underline{\textcolor{gray}{\mathrm{B}=}5}}_{\hspace{0.05cm}\mathbf{2}})\) is the target token.}
  \label{fig:intervention_qwen2.5_14B_mid_5_unchanged}
\end{figure}

\begin{figure}[t]
  \centering
  \includegraphics[width=0.97\linewidth]{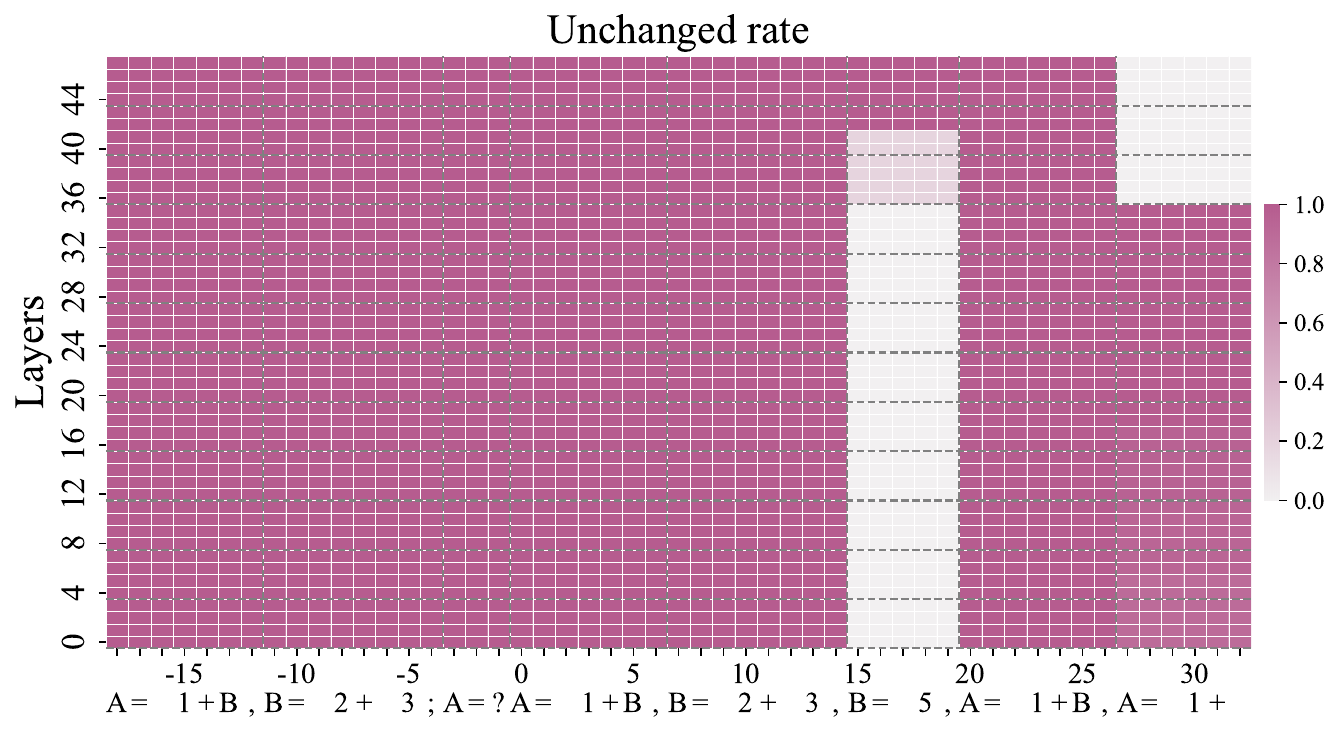}
  \caption{Results of the causal intervention on Qwen2.5-14B. Each grid cell shows the unchanged rate when $z_{32}$ \(({\underline{\textcolor{gray}{\mathrm{A}=1+}5}}_{\hspace{0.05cm}\mathbf{4}})\) is the target token.}
  \label{fig:intervention_qwen2.5_14B_last_5_unchanged}
\end{figure}

\begin{figure}[t]
  \centering
  \includegraphics[width=0.97\linewidth]{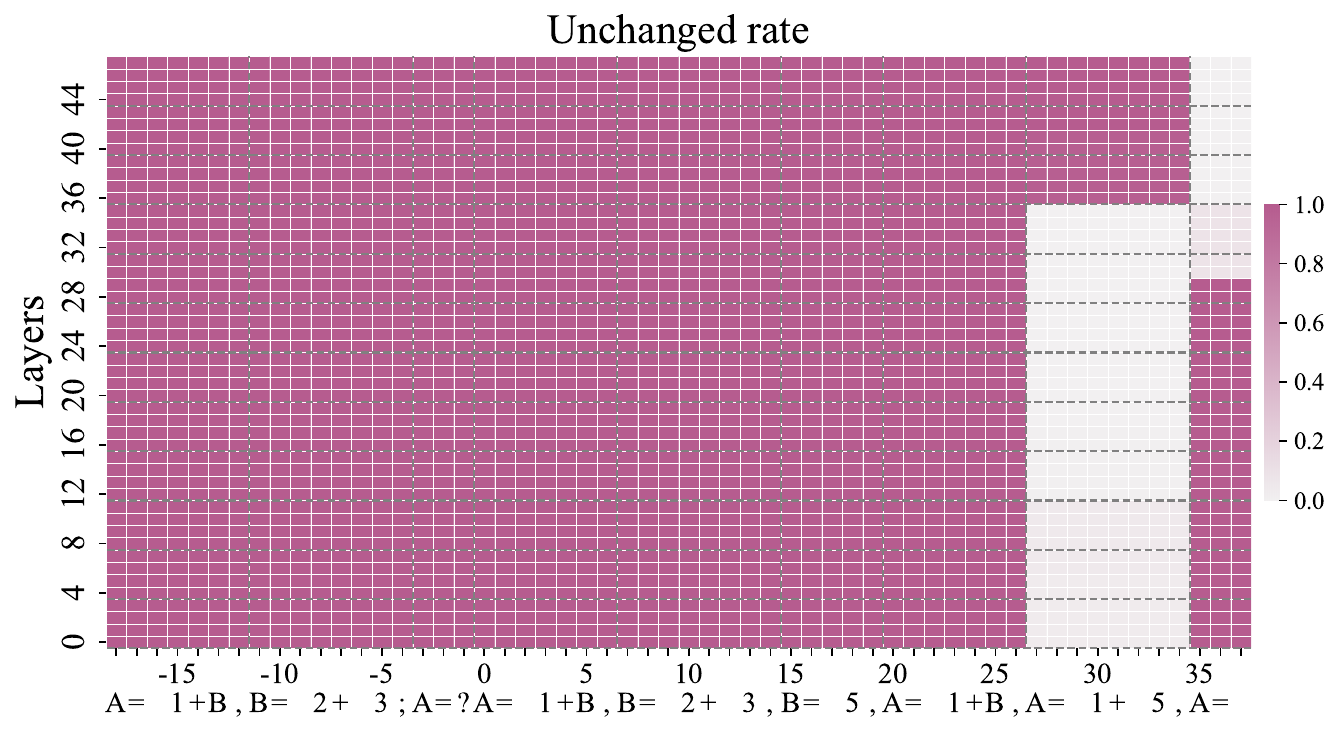}
  \caption{Results of the causal intervention on Qwen2.5-14B. Each grid cell shows the unchanged rate when the final answer $y$ \(({\underline{\textcolor{gray}{\mathrm{A}=}\textbf{6}}}_{\hspace{0.05cm}\mathbf{5}})\) is the target token.}
  \label{fig:intervention_qwen2.5_14B_last_answer_unchanged}
\end{figure}

\begin{figure}[t]
  \centering
  \includegraphics[width=0.97\linewidth]{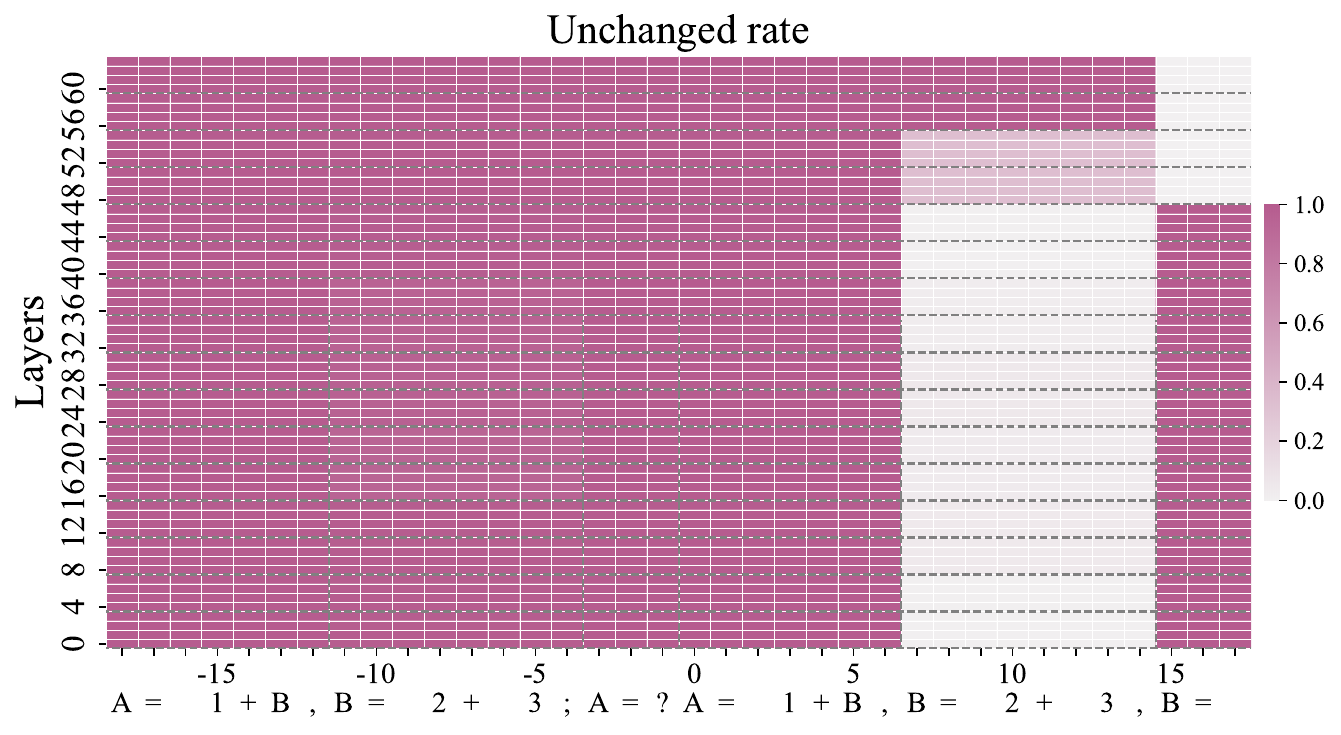}
  \caption{Results of the causal intervention on Qwen2.5-32B. Each grid cell shows the unchanged rate when the intermediate token $z_{17}$ \(({\underline{\textcolor{gray}{\mathrm{B}=}5}}_{\hspace{0.05cm}\mathbf{2}})\) is the target token.}
  \label{fig:intervention_qwen2.5_32B_mid_5_unchanged}
\end{figure}

\begin{figure}[t]
  \centering
  \includegraphics[width=0.97\linewidth]{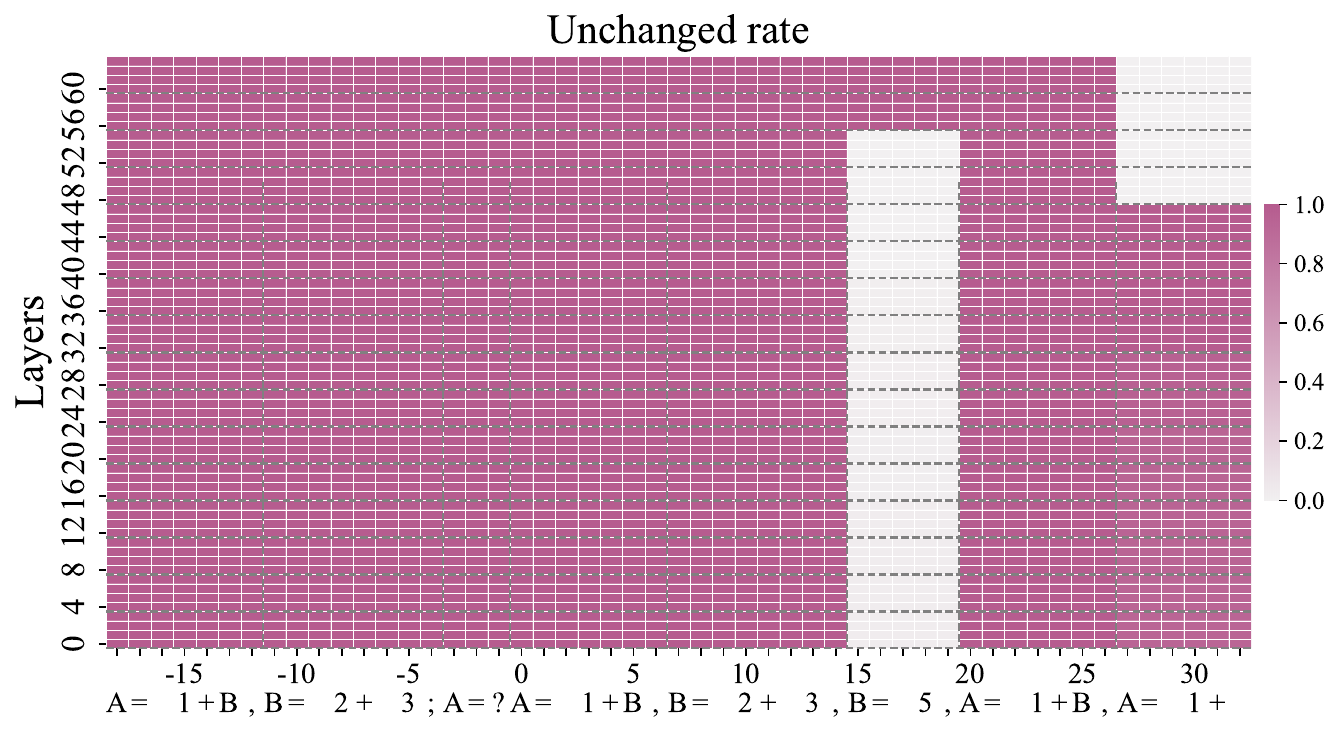}
  \caption{Results of the causal intervention on Qwen2.5-32B. Each grid cell shows the unchanged rate when $z_{32}$ \(({\underline{\textcolor{gray}{\mathrm{A}=1+}5}}_{\hspace{0.05cm}\mathbf{4}})\) is the target token.}
  \label{fig:intervention_qwen2.5_32B_last_5_unchanged}
\end{figure}

\begin{figure}[t]
  \centering
  \includegraphics[width=0.97\linewidth]{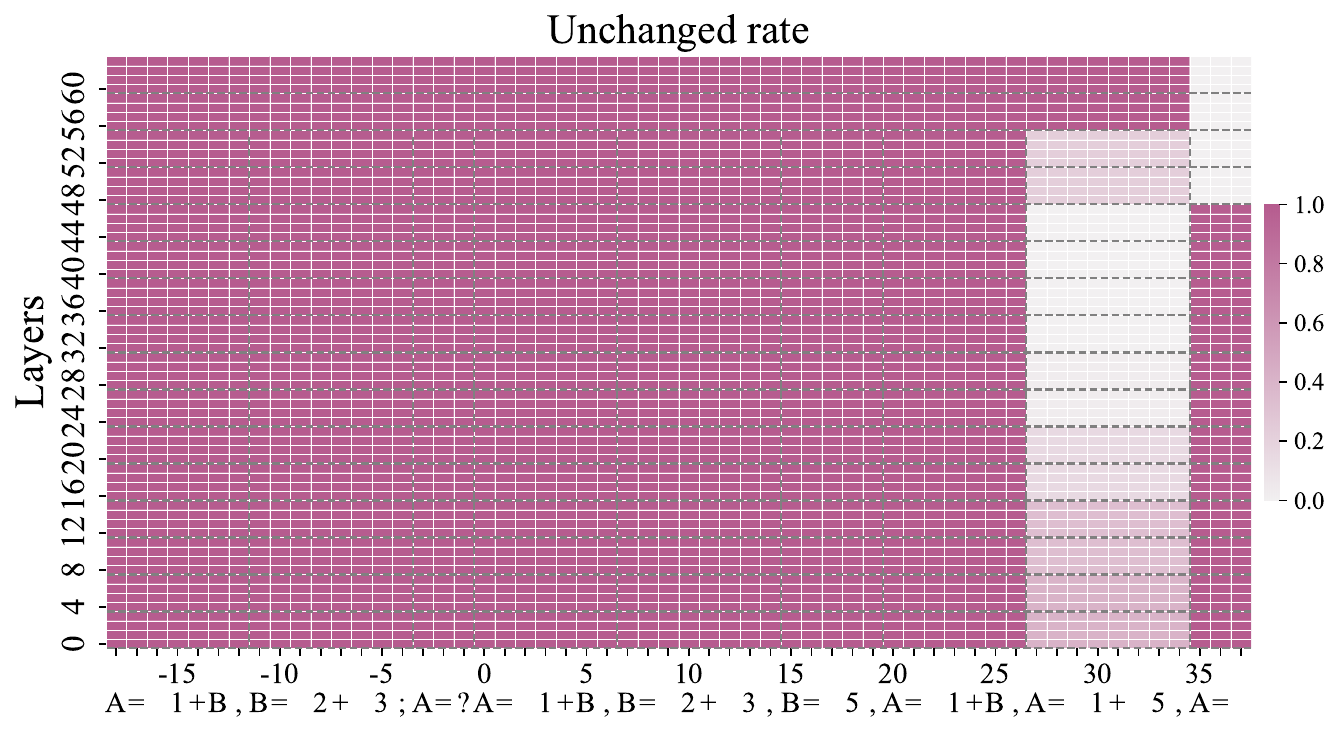}
  \caption{Results of the causal intervention on Qwen2.5-32B. Each grid cell shows the unchanged rate when the final answer $y$ \(({\underline{\textcolor{gray}{\mathrm{A}=}\textbf{6}}}_{\hspace{0.05cm}\mathbf{5}})\) is the target token.}
  \label{fig:intervention_qwen2.5_32B_last_answer_unchanged}
\end{figure}

\begin{figure}[t]
  \centering
  \includegraphics[width=0.97\linewidth]{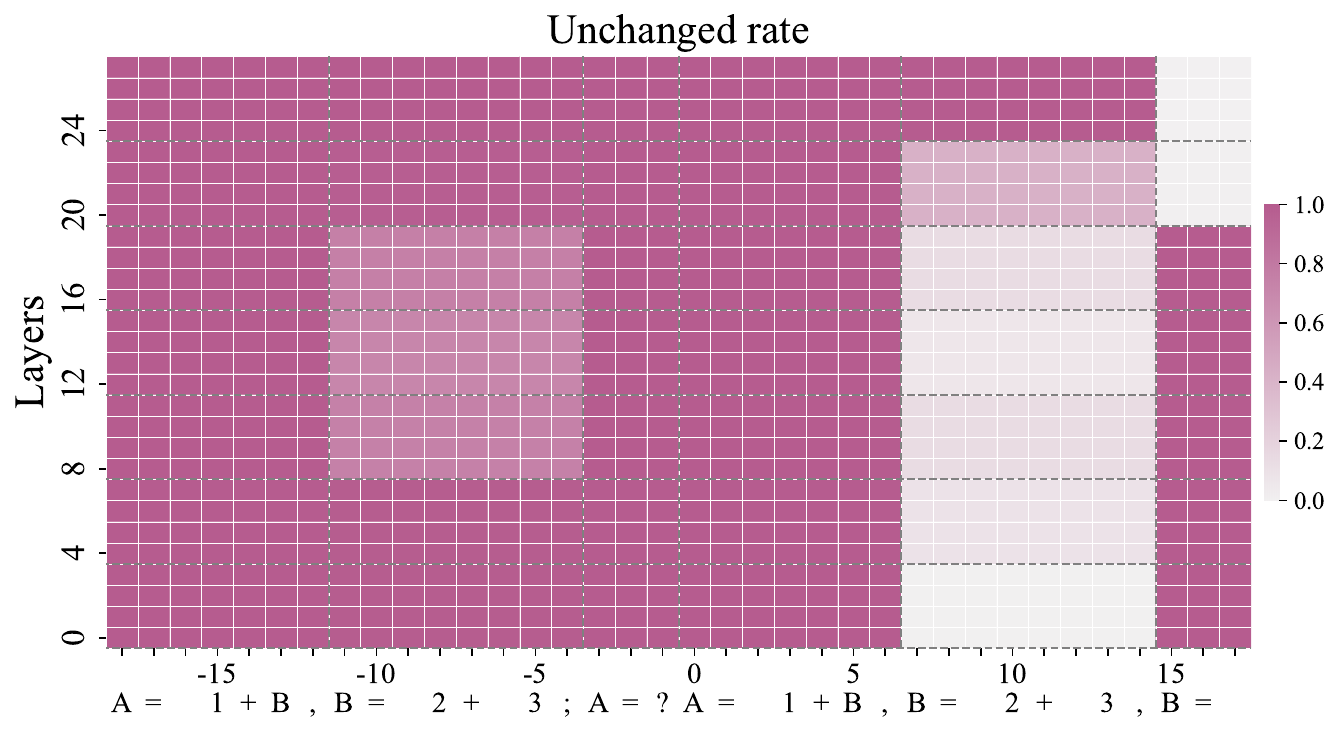}
  \caption{Results of the causal intervention on Qwen2.5-Math-7B. Each grid cell shows the unchanged rate when the intermediate token $z_{17}$ \(({\underline{\textcolor{gray}{\mathrm{B}=}5}}_{\hspace{0.05cm}\mathbf{2}})\) is the target token.}
  \label{fig:intervention_qwen2.5_math_7B_mid_5_unchanged}
\end{figure}

\begin{figure}[t]
  \centering
  \includegraphics[width=0.97\linewidth]{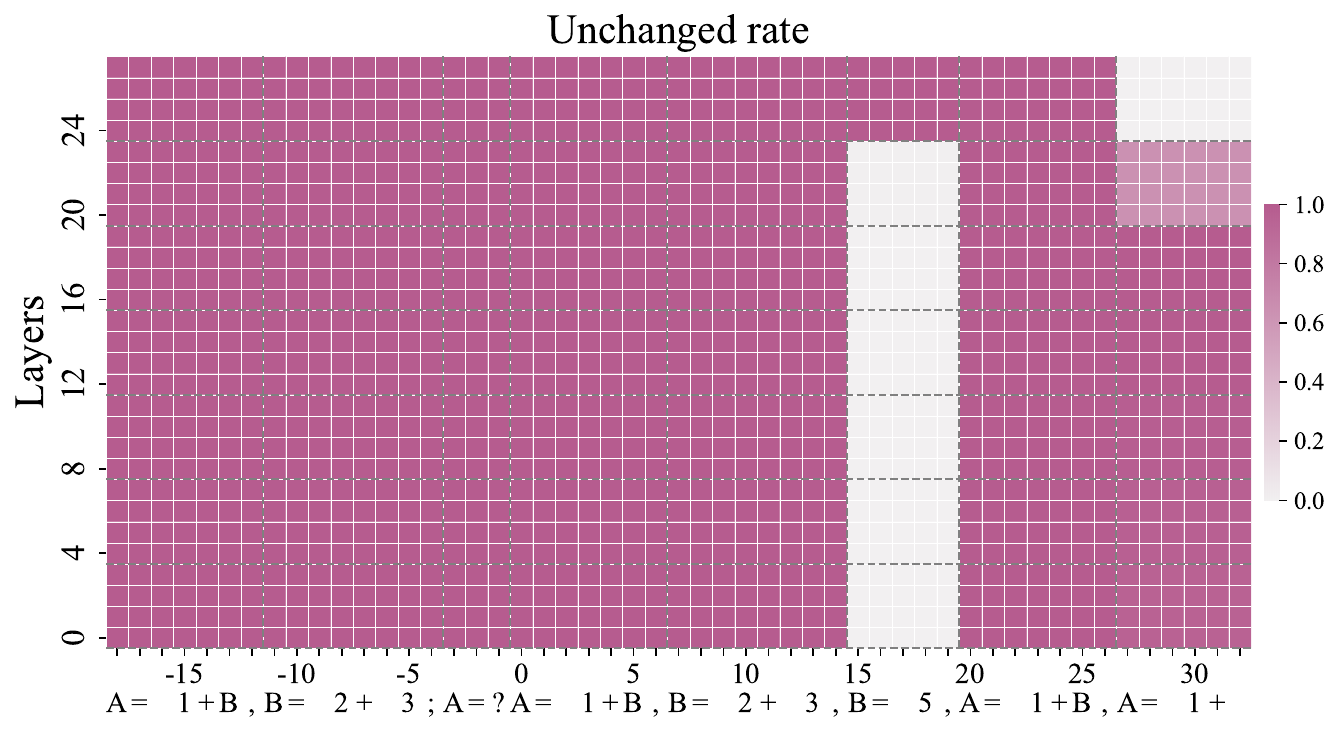}
  \caption{Results of the causal intervention on Qwen2.5-Math-7B. Each grid cell shows the unchanged rate when $z_{32}$ \(({\underline{\textcolor{gray}{\mathrm{A}=1+}5}}_{\hspace{0.05cm}\mathbf{4}})\) is the target token.}
  \label{fig:intervention_qwen2.5_math_7B_last_5_unchanged}
\end{figure}

\begin{figure}[t]
  \centering
  \includegraphics[width=0.97\linewidth]{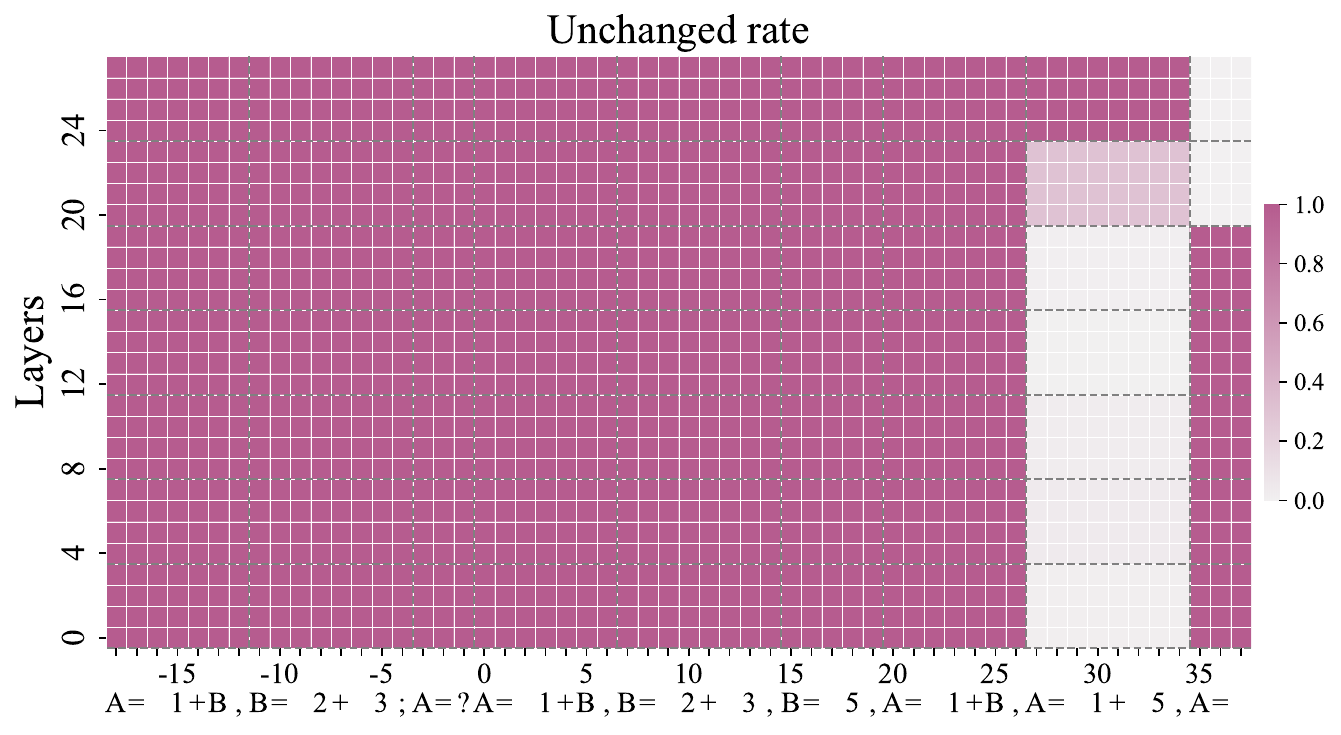}
  \caption{Results of the causal intervention on Qwen2.5-Math-7B. Each grid cell shows the unchanged rate when the final answer $y$ \(({\underline{\textcolor{gray}{\mathrm{A}=}\textbf{6}}}_{\hspace{0.05cm}\mathbf{5}})\) is the target token.}
  \label{fig:intervention_qwen2.5_math_7B_last_answer_unchanged}
\end{figure}

\begin{figure}[t]
  \centering
  \includegraphics[width=0.97\linewidth]{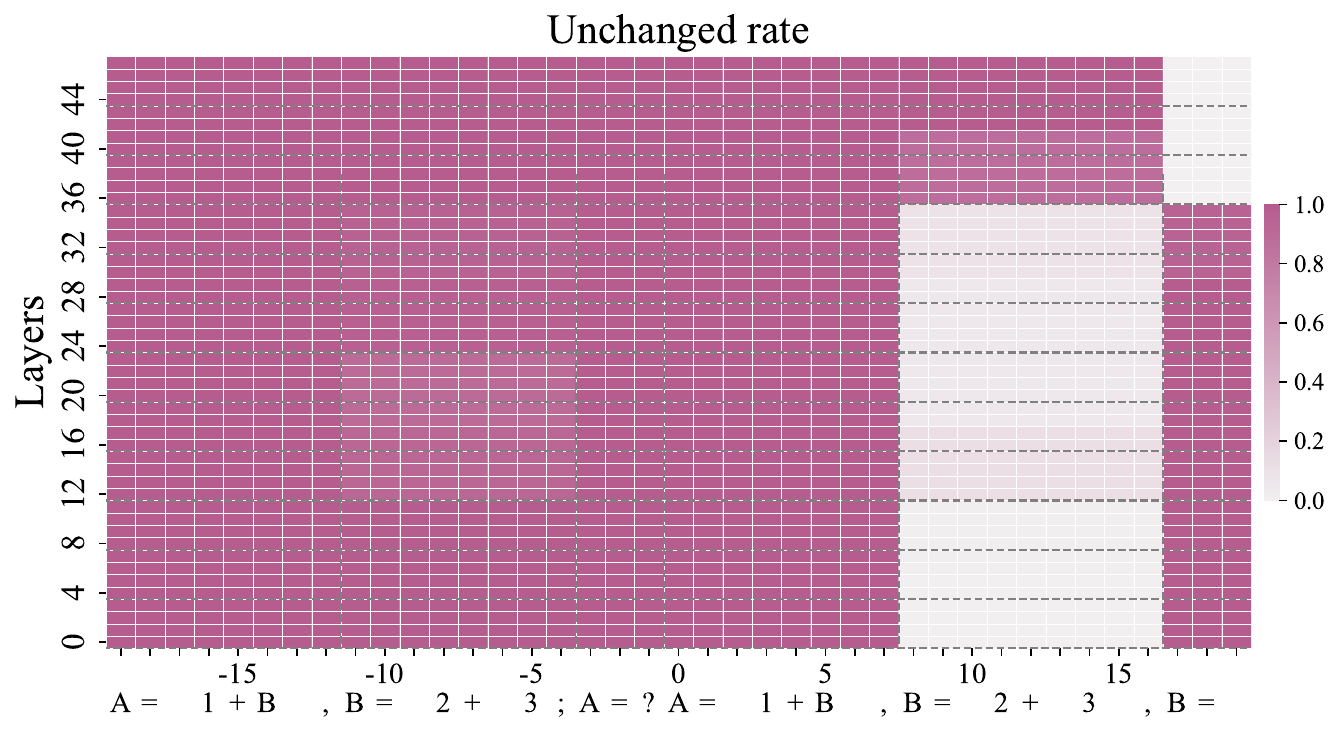}
  \caption{Results of the causal intervention on Yi-1.5-9B. Each grid cell shows the unchanged rate when the intermediate token $z_{17}$ \(({\underline{\textcolor{gray}{\mathrm{B}=}5}}_{\hspace{0.05cm}\mathbf{2}})\) is the target token.}
  \label{fig:intervention_yi_1.5_9B_mid_5_unchanged}
\end{figure}

\begin{figure}[t]
  \centering
  \includegraphics[width=0.97\linewidth]{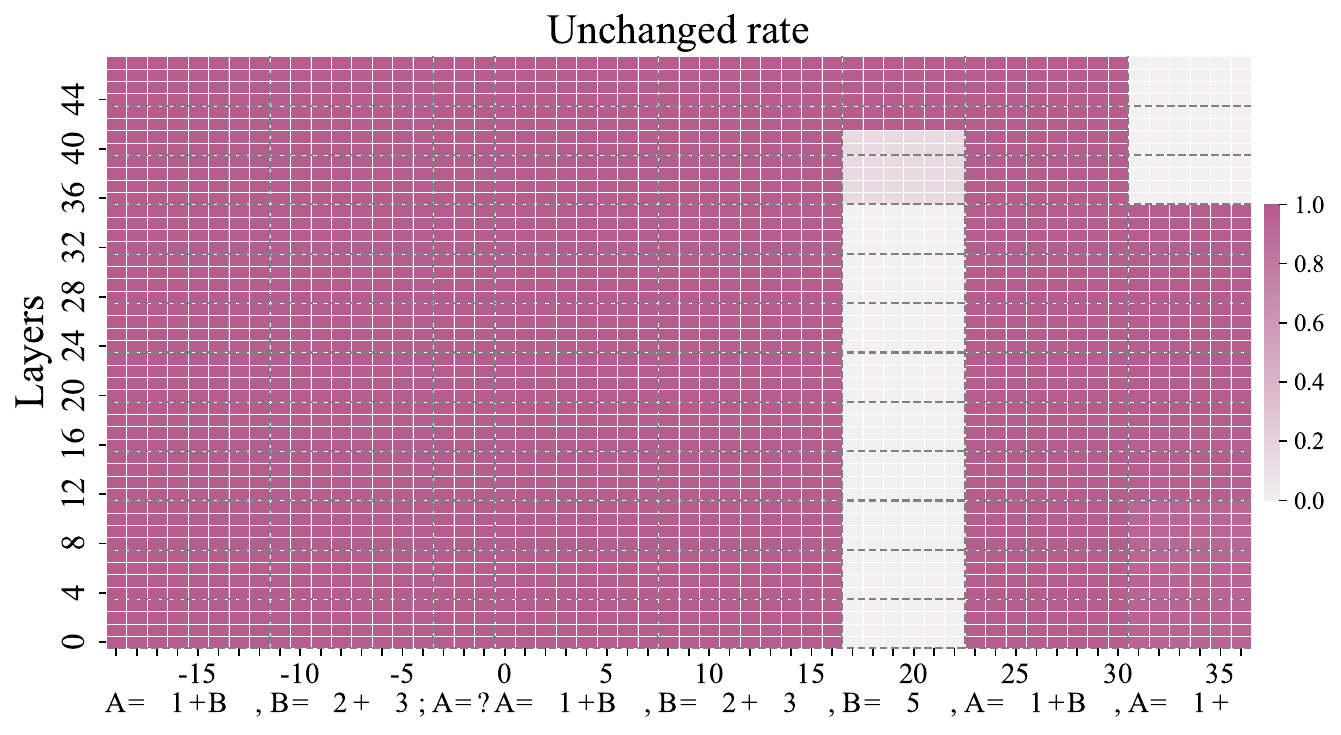}
  \caption{Results of the causal intervention on Yi-1.5-9B. Each grid cell shows the unchanged rate when $z_{32}$ \(({\underline{\textcolor{gray}{\mathrm{A}=1+}5}}_{\hspace{0.05cm}\mathbf{4}})\) is the target token.}
  \label{fig:intervention_yi_1.5_9B_last_5_unchanged}
\end{figure}

\begin{figure}[t]
  \centering
  \includegraphics[width=0.97\linewidth]{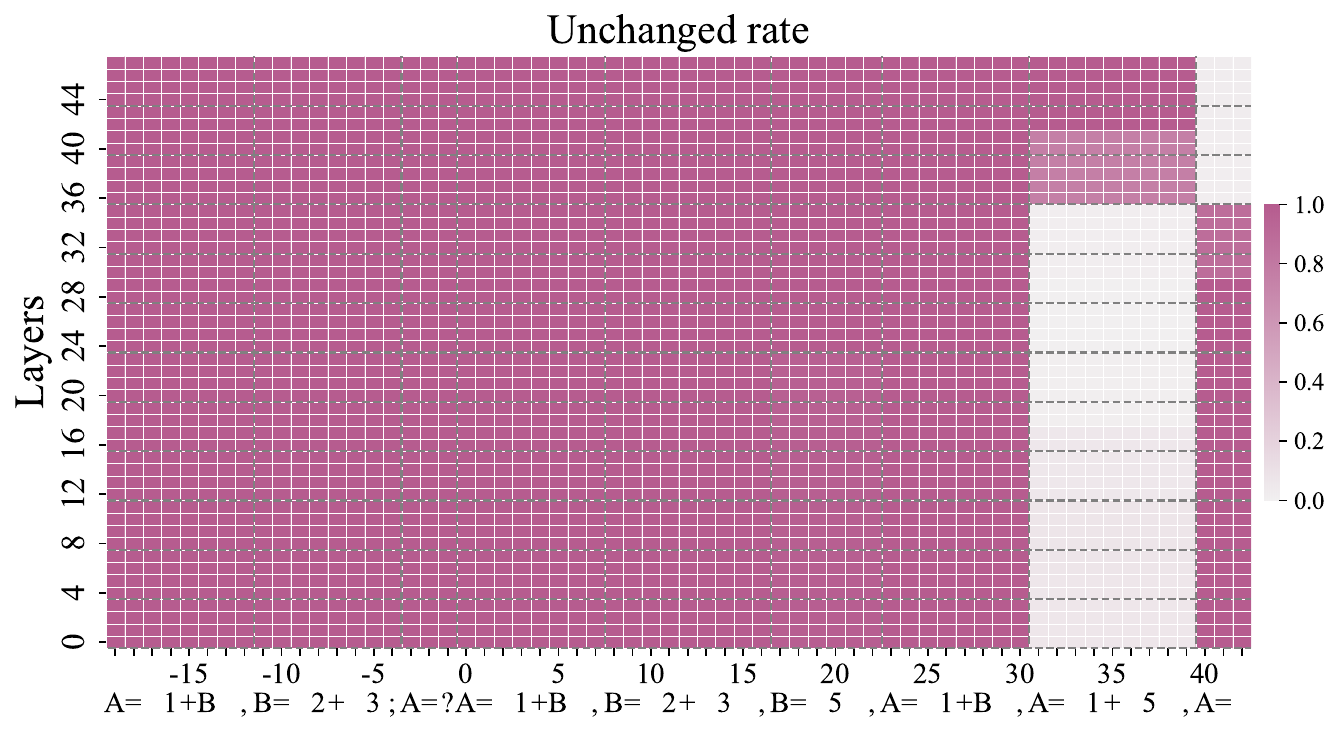}
  \caption{Results of the causal intervention on Yi-1.5-9B. Each grid cell shows the unchanged rate when the final answer $y$ \(({\underline{\textcolor{gray}{\mathrm{A}=}\textbf{6}}}_{\hspace{0.05cm}\mathbf{5}})\) is the target token.}
  \label{fig:intervention_yi_1.5_9B_last_answer_unchanged}
\end{figure}

\begin{figure}[t]
  \centering
  \includegraphics[width=0.97\linewidth]{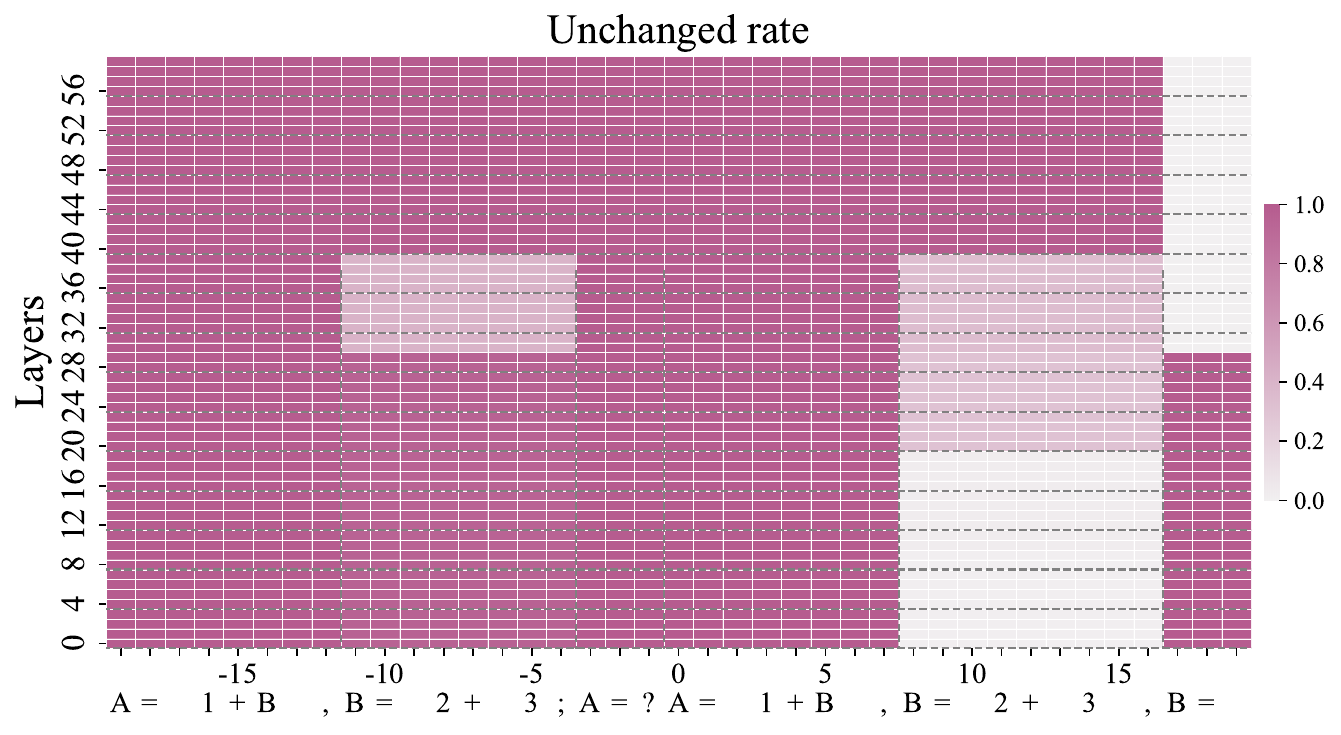}
  \caption{Results of the causal intervention on Yi-1.5-34B. Each grid cell shows the unchanged rate when the intermediate token $z_{17}$ \(({\underline{\textcolor{gray}{\mathrm{B}=}5}}_{\hspace{0.05cm}\mathbf{2}})\) is the target token.}
  \label{fig:intervention_yi_1.5_34B_mid_5_unchanged}
\end{figure}

\begin{figure}[t]
  \centering
  \includegraphics[width=0.97\linewidth]{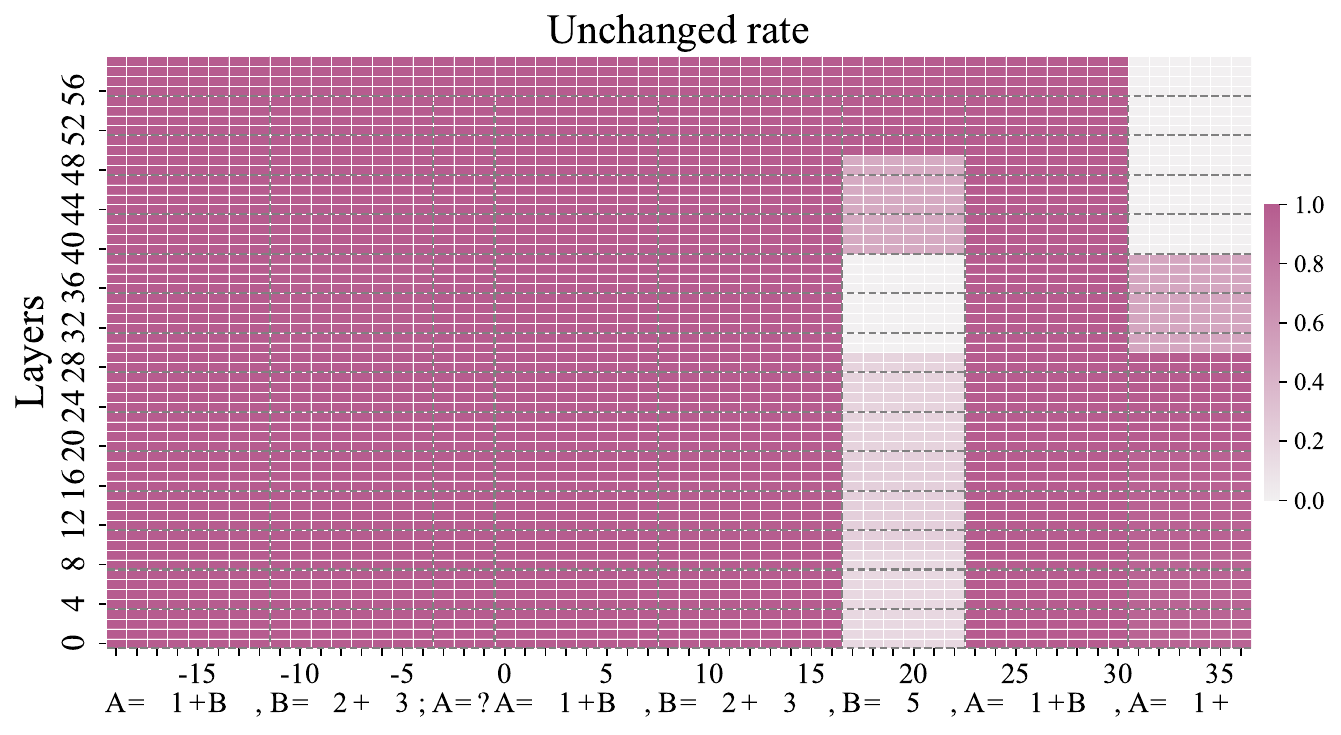}
  \caption{Results of the causal intervention on Yi-1.5-34B. Each grid cell shows the unchanged rate when $z_{32}$ \(({\underline{\textcolor{gray}{\mathrm{A}=1+}5}}_{\hspace{0.05cm}\mathbf{4}})\) is the target token.}
  \label{fig:intervention_yi_1.5_34B_last_5_unchanged}
\end{figure}

\begin{figure}[t]
  \centering
  \includegraphics[width=0.97\linewidth]{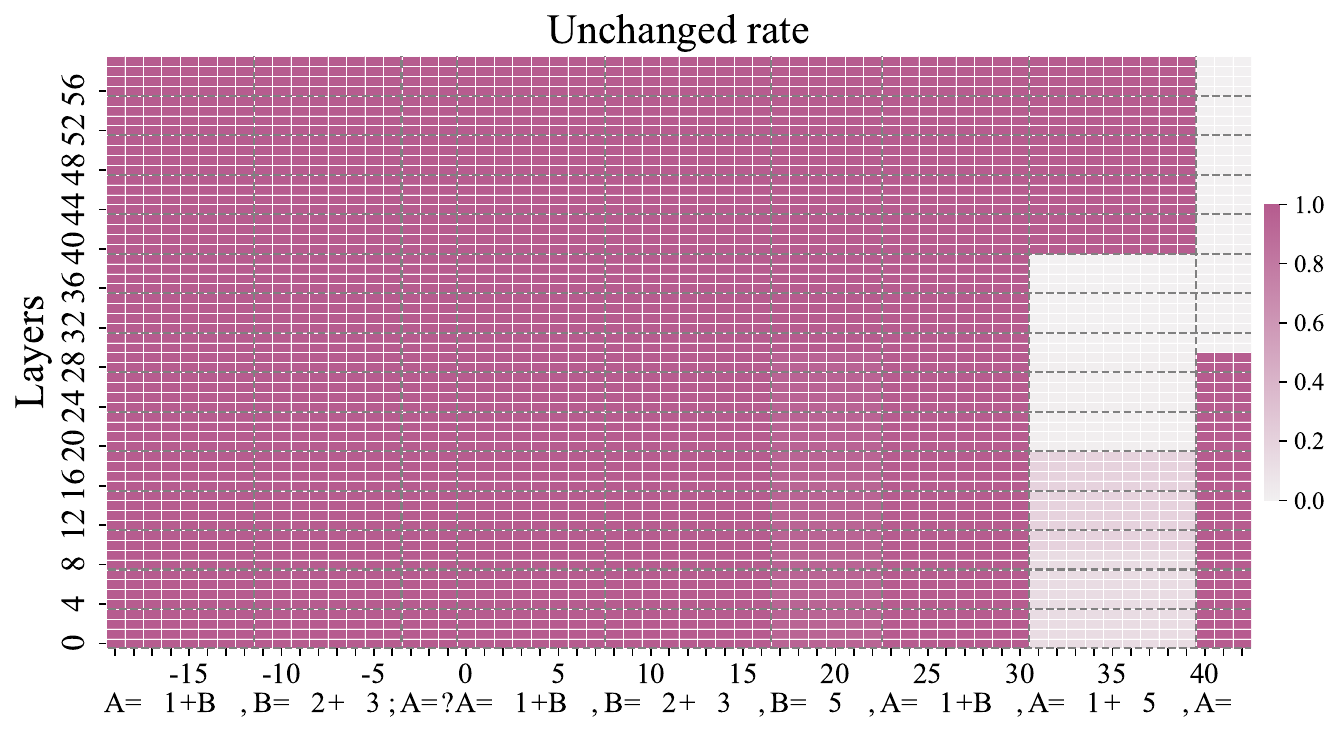}
  \caption{Results of the causal intervention on Yi-1.5-34B. Each grid cell shows the unchanged rate when the final answer $y$ \(({\underline{\textcolor{gray}{\mathrm{A}=}\textbf{6}}}_{\hspace{0.05cm}\mathbf{5}})\) is the target token.}
  \label{fig:intervention_yi_1.5_34B_last_answer_unchanged}
\end{figure}

\begin{figure}[t]
  \centering
  \includegraphics[width=0.97\linewidth]{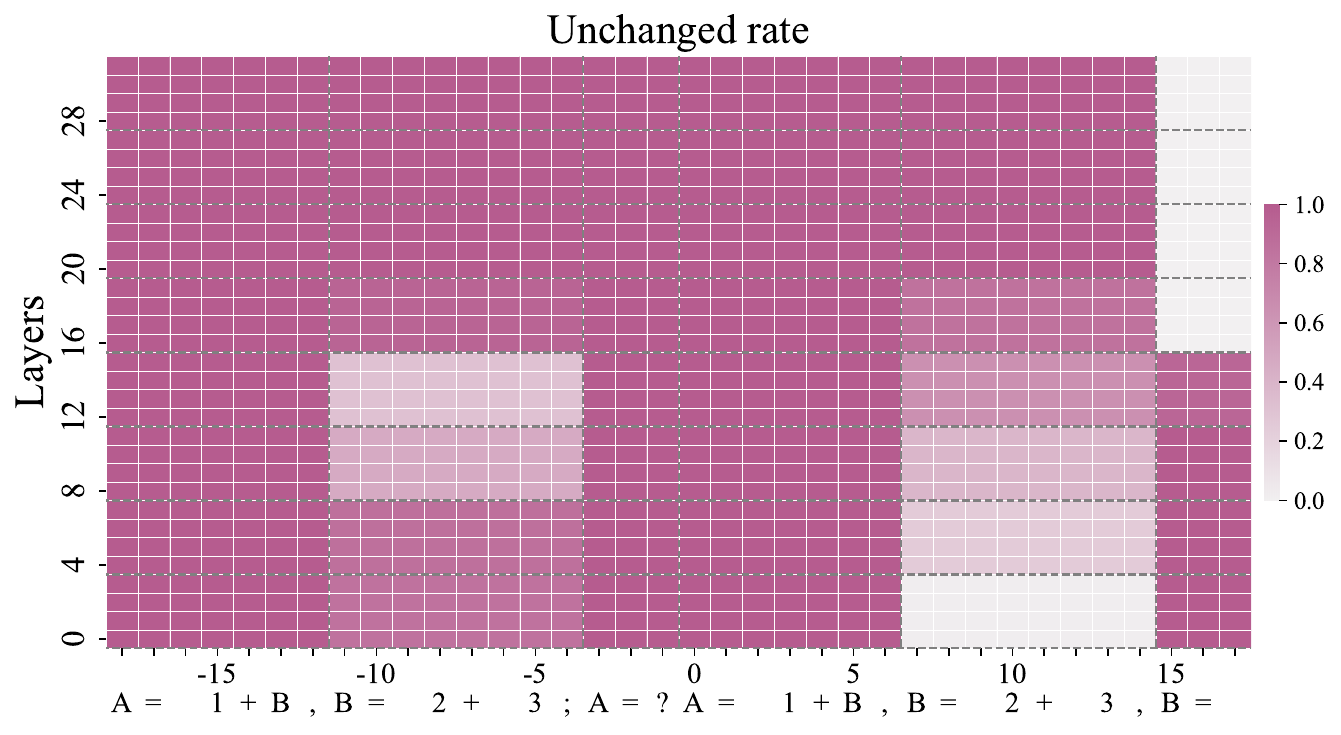}
  \caption{Results of the causal intervention on Llama-3.1-8B. Each grid cell shows the unchanged rate when the intermediate token $z_{17}$ \(({\underline{\textcolor{gray}{\mathrm{B}=}5}}_{\hspace{0.05cm}\mathbf{2}})\) is the target token.}
  \label{fig:intervention_llama3.1_8B_mid_5_unchanged}
\end{figure}

\begin{figure}[t]
  \centering
  \includegraphics[width=0.97\linewidth]{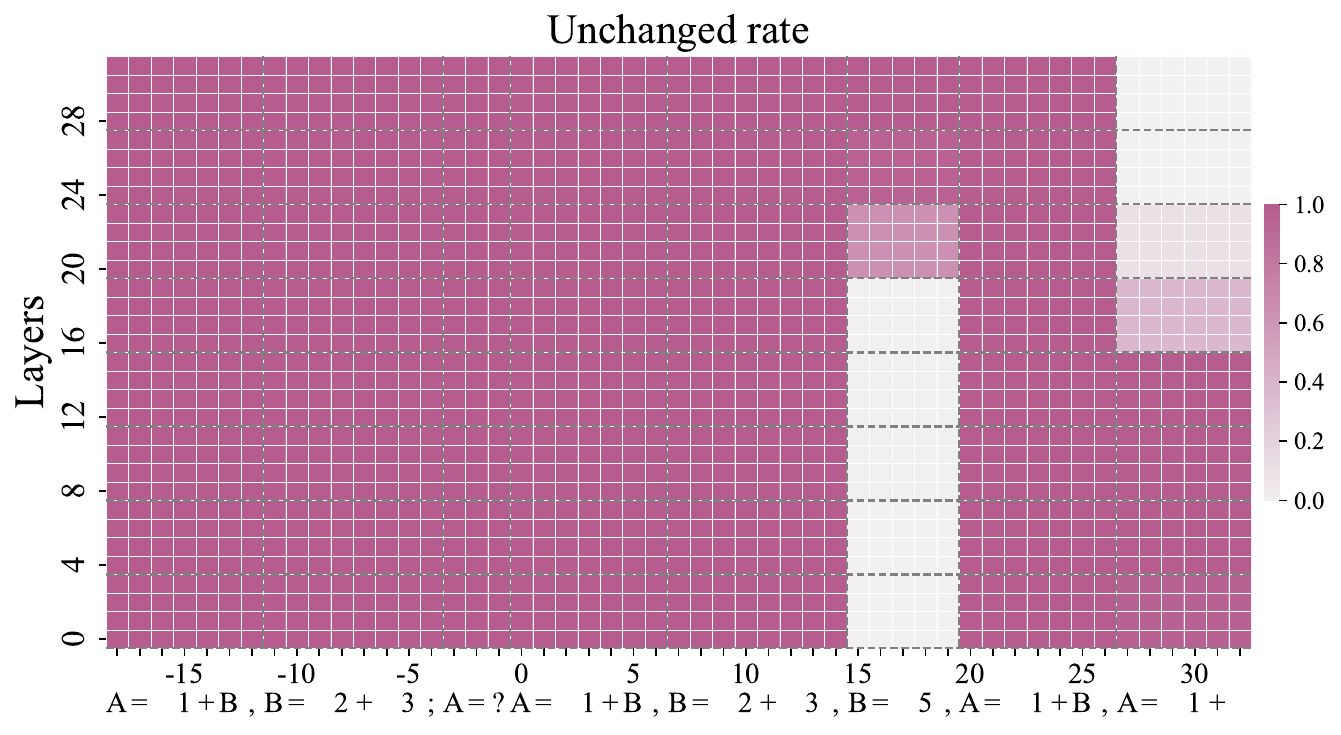}
  \caption{Results of the causal intervention on Llama-3.1-8B. Each grid cell shows the unchanged rate when $z_{32}$ \(({\underline{\textcolor{gray}{\mathrm{A}=1+}5}}_{\hspace{0.05cm}\mathbf{4}})\) is the target token.}
  \label{fig:intervention_llama3.1_8B_last_5_unchanged}
\end{figure}

\begin{figure}[t]
  \centering
  \includegraphics[width=0.97\linewidth]{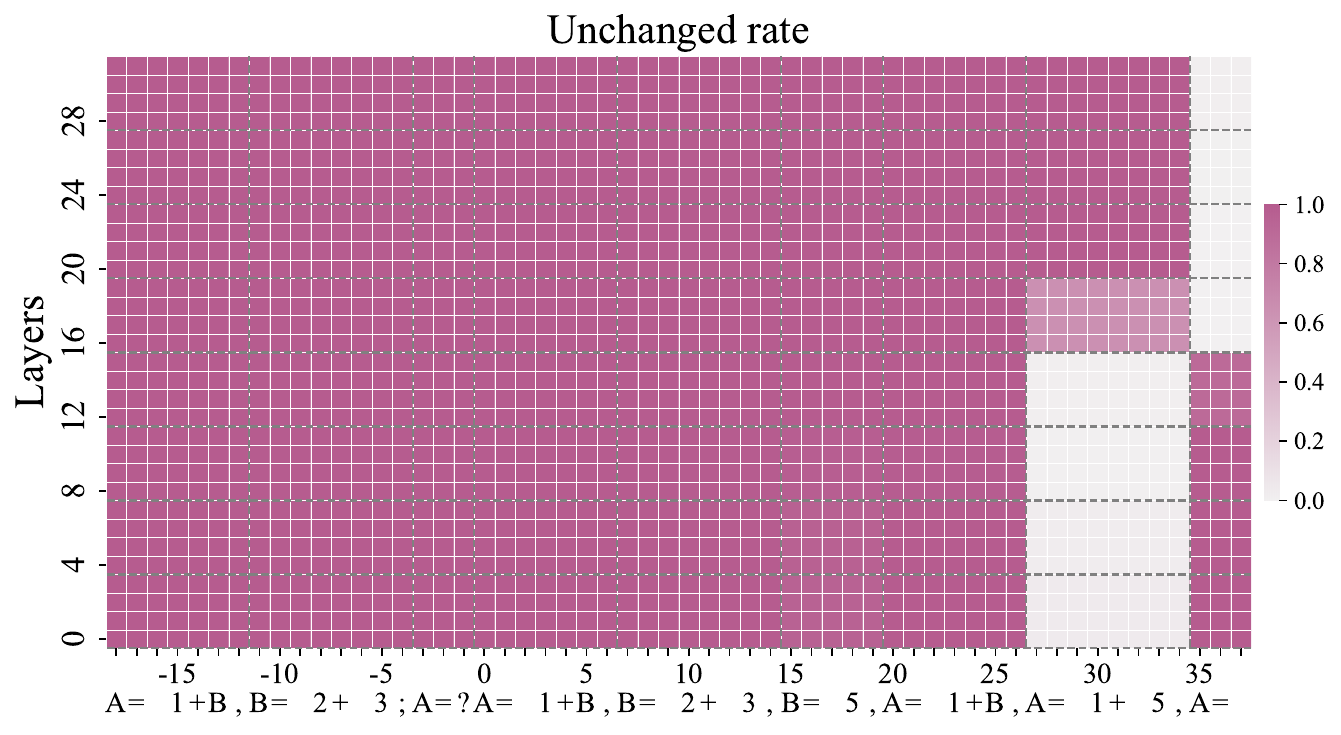}
  \caption{Results of the causal intervention on Llama-3.1-8B. Each grid cell shows the unchanged rate when the final answer $y$ \(({\underline{\textcolor{gray}{\mathrm{A}=}\textbf{6}}}_{\hspace{0.05cm}\mathbf{5}})\) is the target token.}
  \label{fig:intervention_llama3.1_8B_last_answer_unchanged}
\end{figure}

\begin{figure}[t]
  \centering
  \includegraphics[width=0.97\linewidth]{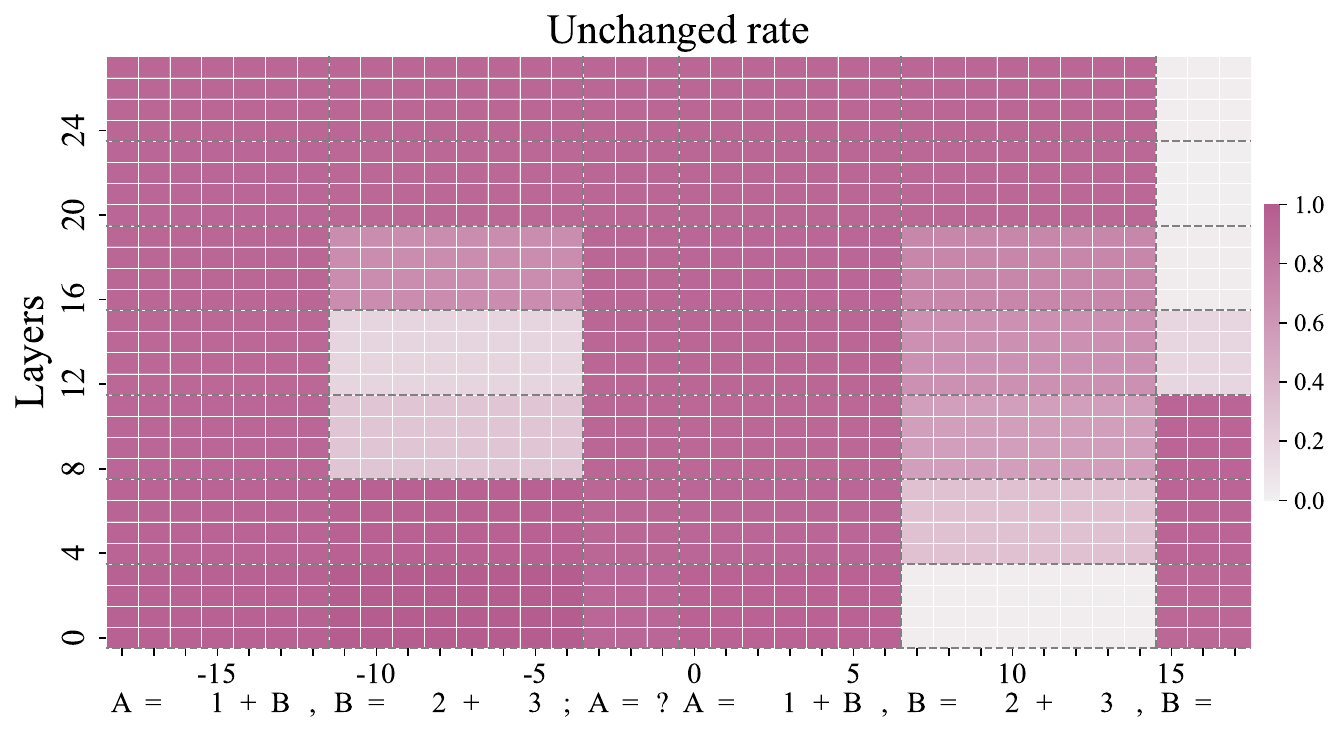}
  \caption{Results of the causal intervention on Llama-3.2-3B. Each grid cell shows the unchanged rate when the intermediate token $z_{17}$ \(({\underline{\textcolor{gray}{\mathrm{B}=}5}}_{\hspace{0.05cm}\mathbf{2}})\) is the target token.}
  \label{fig:intervention_llama3.2_3B_mid_5_unchanged}
\end{figure}

\begin{figure}[t]
  \centering
  \includegraphics[width=0.97\linewidth]{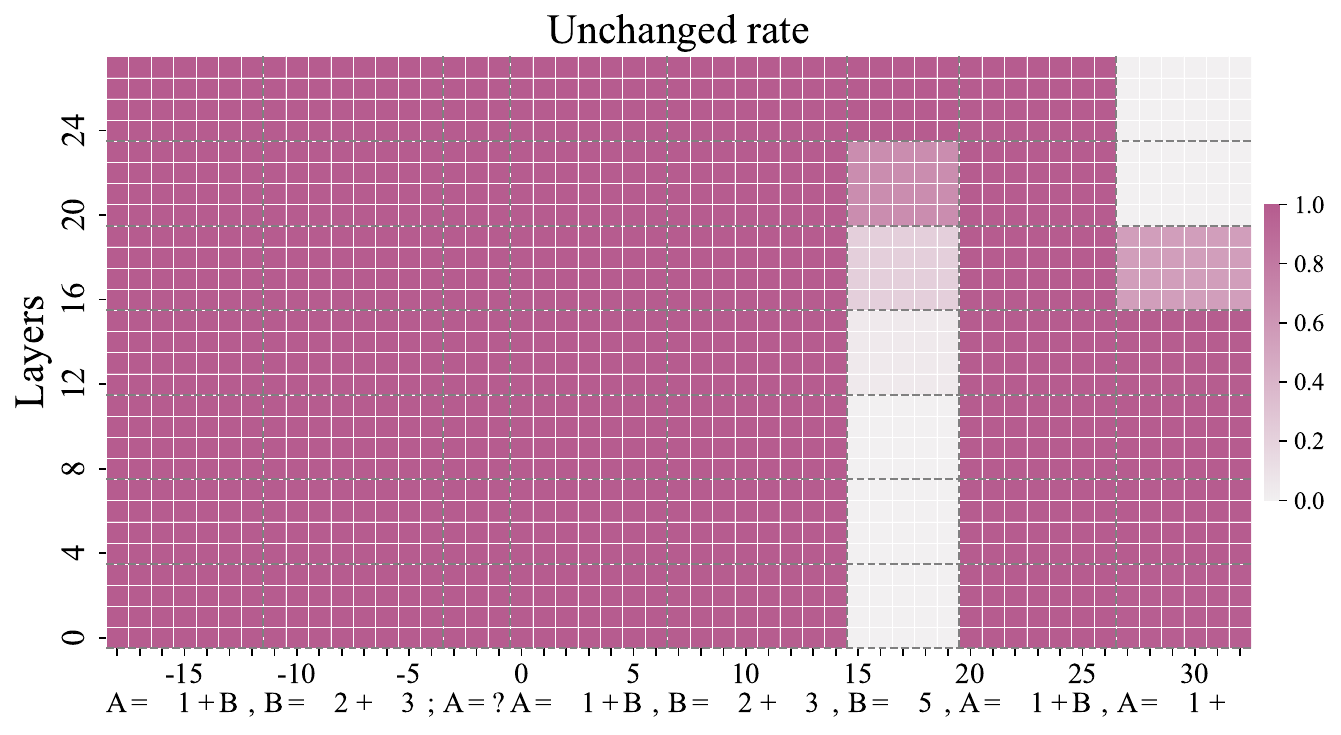}
  \caption{Results of the causal intervention on Llama-3.2-3B. Each grid cell shows the unchanged rate when $z_{32}$ \(({\underline{\textcolor{gray}{\mathrm{A}=1+}5}}_{\hspace{0.05cm}\mathbf{4}})\) is the target token.}
  \label{fig:intervention_llama3.2_3B_last_5_unchanged}
\end{figure}

\begin{figure}[t]
  \centering
  \includegraphics[width=0.97\linewidth]{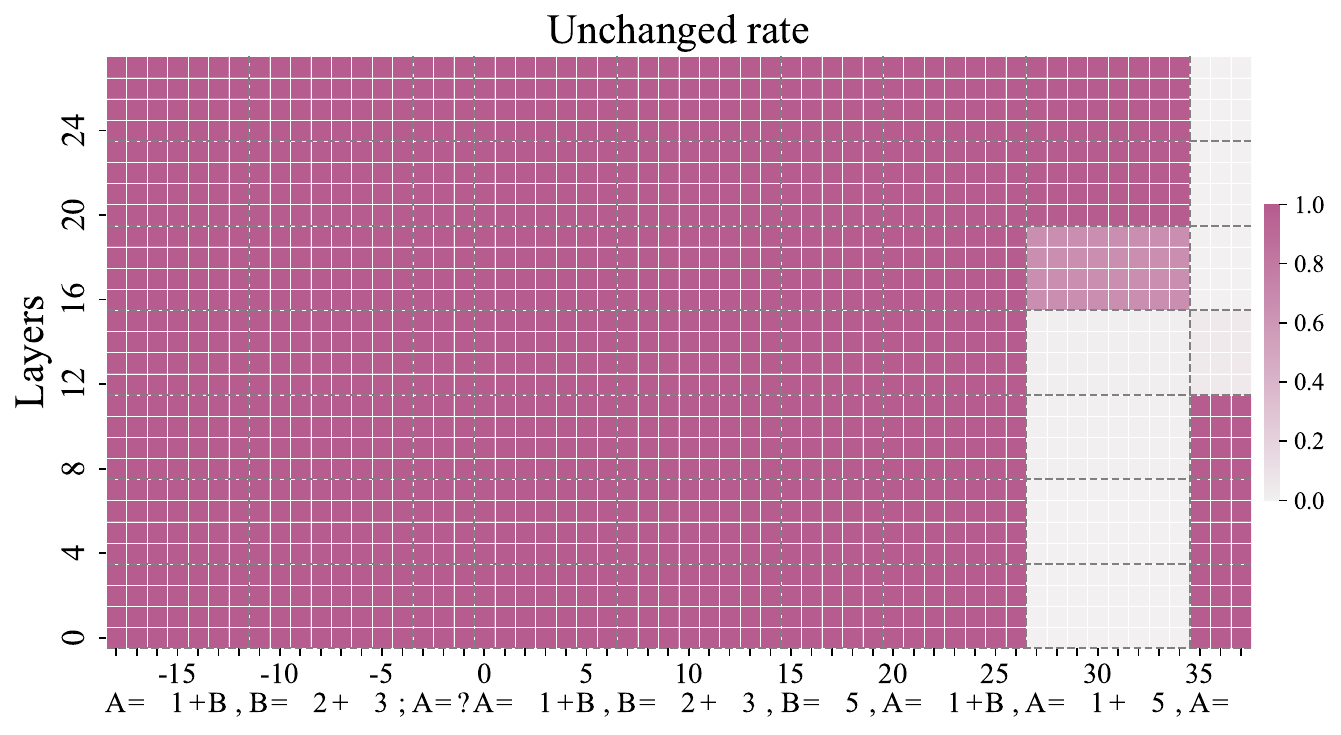}
  \caption{Results of the causal intervention on Llama-3.2-3B. Each grid cell shows the unchanged rate when the final answer $y$ \(({\underline{\textcolor{gray}{\mathrm{A}=}\textbf{6}}}_{\hspace{0.05cm}\mathbf{5}})\) is the target token.}
  \label{fig:intervention_llama3.2_3B_last_answer_unchanged}
\end{figure}

\begin{figure}[t]
  \centering
  \includegraphics[width=0.97\linewidth]{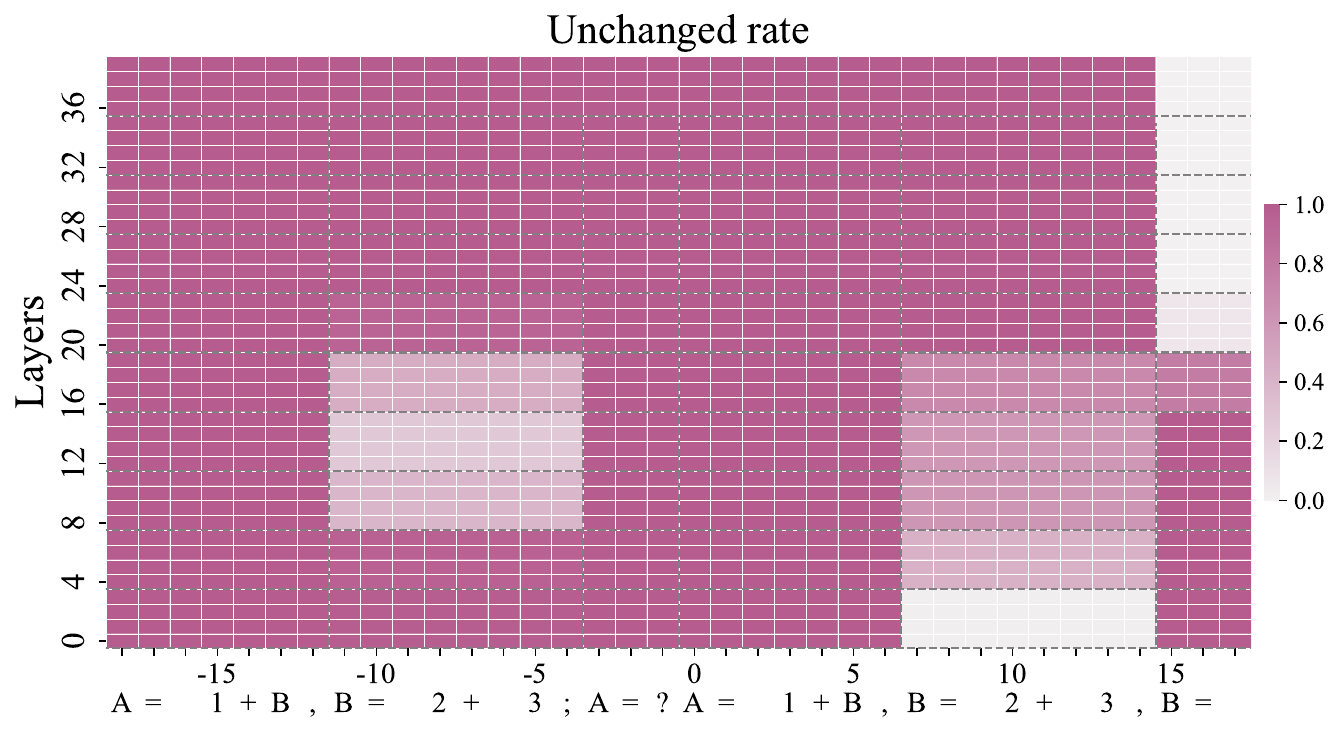}
  \caption{Results of the causal intervention on Mistral-Nemo-Base-2407. Each grid cell shows the unchanged rate when the intermediate token $z_{17}$ \(({\underline{\textcolor{gray}{\mathrm{B}=}5}}_{\hspace{0.05cm}\mathbf{2}})\) is the target token.}
  \label{fig:intervention_mistral_nemo_base_2407_mid_5_unchanged}
\end{figure}

\begin{figure}[t]
   \centering
   \includegraphics[width=0.97\linewidth]{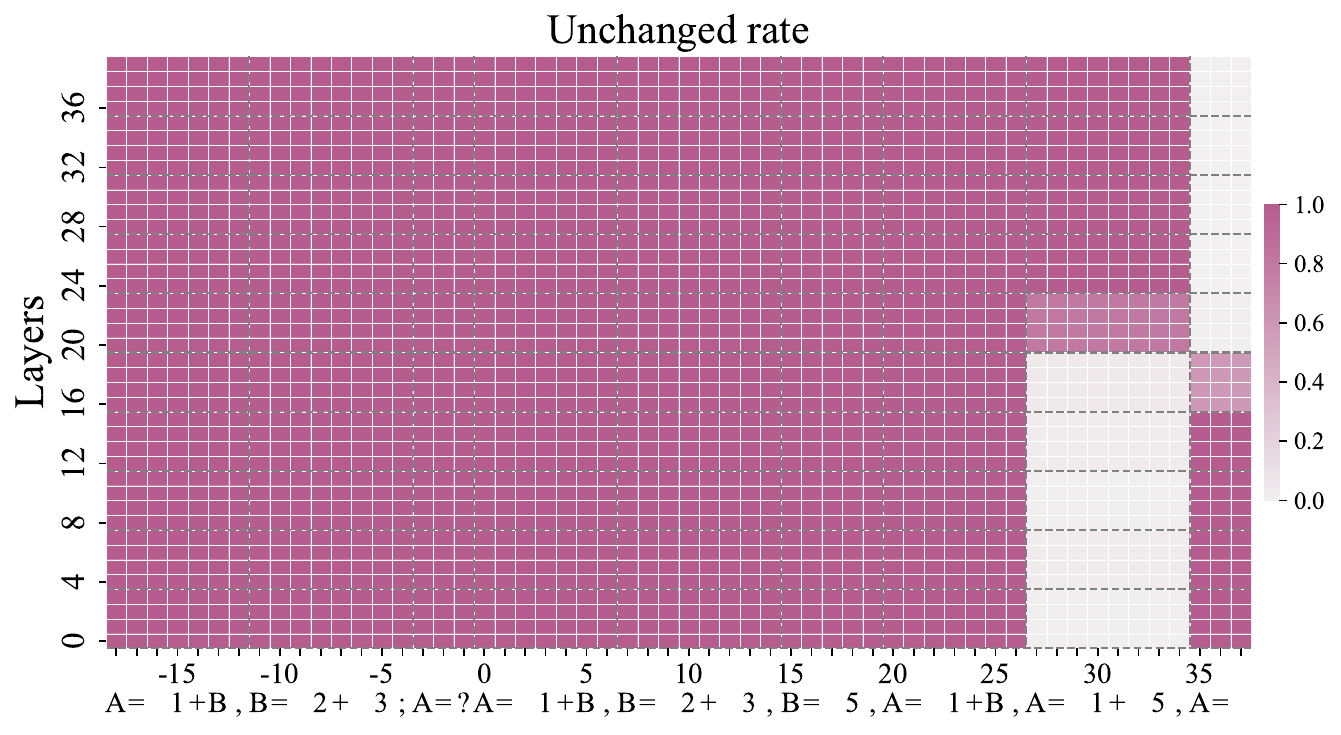}
   \caption{Results of the causal intervention on Mistral-Nemo-Base-2407. Each grid cell shows the unchanged rate when the final answer $y$ \(({\underline{\textcolor{gray}{\mathrm{A}=}\textbf{6}}}_{\hspace{0.05cm}\mathbf{5}})\) is the target token.}
   \label{fig:intervention_mistral_nemo_base_2407_last_answer_unchanged}
\end{figure}

Figures~\ref{fig:intervention_qwen2.5_7B_mid_5}-~\ref{fig:intervention_mistral_nemo_base_2407_last_answer_unchanged} show the causal intervention results for the target tokens $z_{17}$ and $z_{32}$, respectively.

Here, in addition to the Success rate, we also present the \textit{Unchanged rate} as a metric.
The Unchanged rate indicates how frequently (\%) the intervened output $\hat{y}^{\mathrm{patch}}$ remains the same as $y$.
If this value is small, it indicates that the patched hidden states do not affect the output.

\paragraph{Additional discussion: Memory vs. recomputation.}
Focusing on the case where the target token is
${\underline{\textcolor{gray}{\mathrm{A}=1+}\textbf{5}}}_{\hspace{0.05cm}\mathbf{4}}$ in Figure~\ref{fig:pooled_intervention_results}, if the model were recomputing when generating this $\textbf{5}$, it would be expected to show a causal relationship with the segment
${\underline{{\mathrm{B}=2+3}}}_{\hspace{0.05cm}\mathbf{1}}$, where the pre-computation equation information is expected to be explicitly represented.
However, in practice, a strong causal relationship was observed only with the hidden state of the immediately preceding segment
${\underline{\mathrm{B}=5}}_{\hspace{0.05cm}\mathbf{2}}$.
This pattern suggests that the model is not recomputing; instead, it likely relies on the immediately preceding intermediate result stored in the CoT text.

\section{Hyperparameters}
\label{appendix:sec:hyperparameters}
Table~\ref{table:probing_hyperparameter} shows the hyperparameters used for training the probes.

\section{Computational resources}
\label{appendix:sec:computational_resources}
We used NVIDIA A100 GPUs (40GB and 80GB memory) and NVIDIA H100 GPUs to conduct this study.

\section{Usage of AI assistants}
\label{appendix:sec:usage_of_ai_assistants}
For writing this paper and the source code for the experiments, we use AI assistants (e.g., ChatGPT, GitHub Copilot).
However, the use is limited to purposes such as code completion, translation, text editing, and table creation, and all content is solely based on the authors' ideas.

\clearpage

\begin{table}[p]
  \tabcolsep 0.3mm
  \footnotesize
  \centering
  \begin{tabular}{lcrrrrr}
\toprule
&&\multicolumn{2}{c}{When ($\downarrow$)} & \multicolumn{2}{c}{$\mathrm{Acc}$ ($\uparrow$)} \\
\cmidrule(lr){3-4} \cmidrule(lr){5-6}
Model & Variable & $t^*_\mathrm{eq}$ & $t^*$ & $\mathrm{\prec CoT}$ & $\mathrm{\succ CoT}$ \\ 
\midrule
Qwen2.5 (7B)  & $v_1$ ($\mathrm{A}$) & 4  & 27  & 35.8  & 100 \\  
\cite{qwen2.5}                    & $v_2$ ($\mathrm{B}$) & \textbf{$-$2} & \textbf{$-$5}  & 100  & 100 \\ 
\cmidrule(l){2-6}
Qwen2.5 (14B) & $v_1$ ($\mathrm{A}$) & 4  & 27  & 36.9  & 100 \\  
\cite{qwen2.5}                    & $v_2$ ($\mathrm{B}$) & \textbf{$-$2} & \textbf{$-$5}  & 100 & 100 \\  
\cmidrule(l){2-6}
Qwen2.5 (32B) & $v_1$ ($\mathrm{A}$) & 4  & 28  & 30.5  & 100 \\  
\cite{qwen2.5}                    & $v_2$ ($\mathrm{B}$) & \textbf{$-$2} & \textbf{$-$5}  & 100 & 100 \\ 
\cmidrule(l){2-6}
Qwen2.5\text{-}Math (7B) & $v_1$ ($\mathrm{A}$) & 4  & 27  & 41.8  & 100 \\  
\cite{yang2024qwen25mathtechnicalreportmathematical} & $v_2$ ($\mathrm{B}$) & \textbf{$-$2} & \textbf{$-$5}  & 100 & 100 \\  
\cmidrule(l){2-6}
Yi1.5 (9B) & $v_1$ ($\mathrm{A}$) & 4  & 32  & 28.1  & 100 \\  
\cite{DBLP:journals/corr/abs-2403-04652}      & $v_2$ ($\mathrm{B}$) & \textbf{$-$2} & \textbf{$-$5}  & 100  & 100 \\  
\cmidrule(l){2-6}
Yi1.5 (34B) & $v_1$ ($\mathrm{A}$) & 4  & 31  & 22.9  & 100 \\  
\cite{DBLP:journals/corr/abs-2403-04652}      & $v_2$ ($\mathrm{B}$) & \textbf{$-$2} & \textbf{$-$5}  & 100 & 100 \\ 
\cmidrule(l){2-6}
Llama3.1 (8B) & $v_1$ ($\mathrm{A}$) & 4  & 27  & 20.6  & 100 \\  
\cite{DBLP:journals/corr/abs-2407-21783}  & $v_2$ ($\mathrm{B}$) & \textbf{$-$2} & \textbf{$-$5}   & 100  & 100 \\ 
\cmidrule(l){2-6}
Llama3.2 (3B) & $v_1$ ($\mathrm{A}$) & 4  & 28  & 21.8  & 100.0 \\  
\cite{DBLP:journals/corr/abs-2407-21783}        & $v_2$ ($\mathrm{B}$) & \textbf{$-$2} & \textbf{$-$5}  & 100  & 100 \\ 
\cmidrule(l){2-6}
Mistral\text{-}Nemo (12B) & $v_1$ ($\mathrm{A}$) & 4  & 27  & 18.0  & 100 \\  
\cite{mistral-nemo}   & $v_2$ ($\mathrm{B}$) & \textbf{$-$2}  & \textbf{$-$5}  & 100  & 100 \\ 
\bottomrule
\end{tabular}

  \caption{
  Results for various models on the Level 1 task ($\tau=0.85$).
  }
  \label{table:rsp_model_dimmension_tau_0.85_level_1}
\end{table}

\begin{table}[p]
  \tabcolsep 0.3mm
  \footnotesize
  \centering
  \begin{tabular}{lcrrrrr}
\toprule
&&\multicolumn{2}{c}{When ($\downarrow$)} & \multicolumn{2}{c}{$\mathrm{Acc}$ ($\uparrow$)} \\
\cmidrule(lr){3-4} \cmidrule(lr){5-6}
Model & Variable & $t^*_\mathrm{eq}$ & $t^*$ & $\mathrm{\prec CoT}$ & $\mathrm{\succ CoT}$ \\ 
\midrule
Qwen2.5 (7B)  & $v_1$ ($\mathrm{A}$) & 2  & 16  & 49.2  & 100 \\  
\cite{qwen2.5}                    & $v_2$ ($\mathrm{B}$) & 5 & 35  & 21.2  & 100 \\ 
\cmidrule(l){2-6}
Qwen2.5 (14B)  & $v_1$ ($\mathrm{A}$) & 2  & 15  & 48.8  & 100 \\  
\cite{qwen2.5}                    & $v_2$ ($\mathrm{B}$) & 5 & 36  & 21.5  & 100 \\ 
\cmidrule(l){2-6}
Qwen2.5 (32B)  & $v_1$ ($\mathrm{A}$) & 2  & 15  & 66.4  & 100 \\  
\cite{qwen2.5}                    & $v_2$ ($\mathrm{B}$) & 5 & 36  & 21.3  & 100 \\ 
\cmidrule(l){2-6}
Qwen2.5\text{-}Math (7B)  & $v_1$ ($\mathrm{A}$) & 2  & 15  & 53.7  & 100 \\  
\cite{yang2024qwen25mathtechnicalreportmathematical} & $v_2$ ($\mathrm{B}$) & 5 & 35  & 22.1  & 100 \\  
\cmidrule(l){2-6}
Yi1.5 (9B)  & $v_1$ ($\mathrm{A}$) & 2  & 18  & 40.2  & 100 \\  
\cite{DBLP:journals/corr/abs-2403-04652}      & $v_2$ ($\mathrm{B}$) & 5 & 41  & 17.8  & 100 \\  
\cmidrule(l){2-6}
Yi1.5 (34B)  & $v_1$ ($\mathrm{A}$) & 2  & 18  & 35.6  & 100 \\  
\cite{DBLP:journals/corr/abs-2403-04652}      & $v_2$ ($\mathrm{B}$) & 5 & 41  & 18.3  & 100 \\ 
\cmidrule(l){2-6}
Llama3.1 (8B)  & $v_1$ ($\mathrm{A}$) & 2  & 15  & 31.9  & 100 \\  
\cite{DBLP:journals/corr/abs-2407-21783}  & $v_2$ ($\mathrm{B}$) & 5 & 35  & 17.8  & 100 \\ 
\cmidrule(l){2-6}
Llama3.2 (3B)  & $v_1$ ($\mathrm{A}$) & 2  & 16  & 36.2  & 99.9 \\  
\cite{DBLP:journals/corr/abs-2407-21783}        & $v_2$ ($\mathrm{B}$) & 5 & 36  & 17.8  & 99.9 \\ 
\cmidrule(l){2-6}
Mistral\text{-}Nemo (12B)  & $v_1$ ($\mathrm{A}$) & 2  & 16  & 30.8  & 100 \\  
\cite{mistral-nemo}   & $v_2$ ($\mathrm{B}$) & 5  & 36  & 17.8  & 100 \\ 
\bottomrule
\end{tabular}

  \caption{
  \tabcolsep 0.3mm
  Results for various models on the Level 2 task ($\tau=0.85$).
  }
  \label{table:rsp_model_dimmension_tau_0.85_level_2}
\end{table}

\begin{table}[p]
  \tabcolsep 0.3mm
  \footnotesize
  \centering
  \begin{tabular}{lcrrrrr}
\toprule
&&\multicolumn{2}{c}{When ($\downarrow$)} & \multicolumn{2}{c}{$\mathrm{Acc}$ ($\uparrow$)} \\
\cmidrule(lr){3-4} \cmidrule(lr){5-6}
Model & Variable & $t^*_\mathrm{eq}$ & $t^*$ & $\mathrm{\prec CoT}$ & $\mathrm{\succ CoT}$ \\ 
\midrule
Qwen2.5 (7B)  & $v_1$ ($\mathrm{A}$) & 5  & 35  & 17.9  & 100 \\  
\cite{qwen2.5}                    & $v_2$ ($\mathrm{B}$) & 2 & 16  & 50.5  & 100 \\ 
\cmidrule(l){2-6}
Qwen2.5 (14B)  & $v_1$ ($\mathrm{A}$) & 5  & 35  & 17.8  & 100 \\  
\cite{qwen2.5}                    & $v_2$ ($\mathrm{B}$) & 2 & 15  & 50.5  & 100 \\ 
\cmidrule(l){2-6}
Qwen2.5 (32B)  & $v_1$ ($\mathrm{A}$) & 5  & 36  & 17.8  & 100 \\  
\cite{qwen2.5}                    & $v_2$ ($\mathrm{B}$) & 2 & 15  & 67.4  & 100 \\ 
\cmidrule(l){2-6}
Qwen2.5\text{-}Math (7B)  & $v_1$ ($\mathrm{A}$) & 5  & 35  & 18.6  & 100 \\  
\cite{yang2024qwen25mathtechnicalreportmathematical} & $v_2$ ($\mathrm{B}$) & 2 & 15  & 56.1  & 100 \\  
\cmidrule(l){2-6}
Yi1.5 (9B)  & $v_1$ ($\mathrm{A}$) & 5  & 41  & 17.8  & 100 \\  
\cite{DBLP:journals/corr/abs-2403-04652}      & $v_2$ ($\mathrm{B}$) & 2 & 18  & 36.9  & 100 \\  
\cmidrule(l){2-6}
Yi1.5 (34B)  & $v_1$ ($\mathrm{A}$) & 5  & 41  & 22.4  & 100 \\  
\cite{DBLP:journals/corr/abs-2403-04652}      & $v_2$ ($\mathrm{B}$) & 2 & 18  & 37.4  & 100 \\ 
\cmidrule(l){2-6}
Llama3.1 (8B)  & $v_1$ ($\mathrm{A}$) & 5  & 35  & 26.0  & 100 \\  
\cite{DBLP:journals/corr/abs-2407-21783}  & $v_2$ ($\mathrm{B}$) & 2 & 16  & 29.6  & 100 \\ 
\cmidrule(l){2-6}
Llama3.2 (3B)  & $v_1$ ($\mathrm{A}$) & 5  & 36  & 17.8  & 93.2 \\  
\cite{DBLP:journals/corr/abs-2407-21783}        & $v_2$ ($\mathrm{B}$) & 2 & 16  & 33.2  & 95.4 \\ 
\cmidrule(l){2-6}
Mistral\text{-}Nemo (12B)  & $v_1$ ($\mathrm{A}$) & 5  & 36  & 17.8  & 100 \\  
\cite{mistral-nemo}   & $v_2$ ($\mathrm{B}$) & 2  & 16  & 28.9  & 100 \\ 
\bottomrule
\end{tabular}

  \caption{
  Results for various models on the Level 3 task ($\tau=0.85$).
  }
  \label{table:rsp_model_dimmension_tau_0.85_level_3}
\end{table}

\begin{table}[p]
  \tabcolsep 0.3mm
  \footnotesize
  \centering
  \begin{tabular}{lcrrrrr}
\toprule
&&\multicolumn{2}{c}{When ($\downarrow$)} & \multicolumn{2}{c}{$\mathrm{Acc}$ ($\uparrow$)} \\
\cmidrule(lr){3-4} \cmidrule(lr){5-6}
Model & Variable & $t^*_\mathrm{eq}$ & $t^*$ & $\mathrm{\prec CoT}$ & $\mathrm{\succ CoT}$ \\ 
\midrule
Qwen2.5 (7B)  & $v_1$ ($\mathrm{A}$) & 4  & 29  & 30.4  & 100 \\  
\cite{qwen2.5}                    & $v_2$ ($\mathrm{B}$) & 2 & 15  & 27.2  & 100 \\ 
                                   & $v_3$ ($\mathrm{C}$) & N/A & N/A & 18.5 & 17.6 \\
\cmidrule(l){2-6}
Qwen2.5 (14B)  & $v_1$ ($\mathrm{A}$) & 5  & 35  & 18.9  & 100 \\  
\cite{qwen2.5}                     & $v_2$ ($\mathrm{B}$) & 2 & 15  & 44.3  & 100 \\ 
                                    & $v_3$ ($\mathrm{C}$) & N/A & N/A & 40.4 & 26.8 \\
\cmidrule(l){2-6}
Qwen2.5 (32B)  & $v_1$ ($\mathrm{A}$) & 5  & 36  & 17.4  & 100 \\  
\cite{qwen2.5}                     & $v_2$ ($\mathrm{B}$) & 2 & 15  & 62.8  & 100 \\ 
                                    & $v_3$ ($\mathrm{C}$) & N/A & N/A & 64.4 & 32.6 \\
\cmidrule(l){2-6}
Qwen2.5\text{-}Math (7B)  & $v_1$ ($\mathrm{A}$) & 5  & 35  & 17.2  & 100 \\  
\cite{yang2024qwen25mathtechnicalreportmathematical} & $v_2$ ($\mathrm{B}$) & 2 & 15  & 55.6  & 100 \\ 
                                                            & $v_3$ ($\mathrm{C}$) & N/A & N/A & 47.8 & 29.4 \\
\cmidrule(l){2-6}
Yi1.5 (9B)  & $v_1$ ($\mathrm{A}$) & 5  & 41  & 17.8  & 100 \\  
\cite{DBLP:journals/corr/abs-2403-04652}      & $v_2$ ($\mathrm{B}$) & 2 & 18  & 43.5  & 100 \\ 
                                              & $v_3$ ($\mathrm{C}$) & N/A & N/A & 36.7 & 21.2 \\
\cmidrule(l){2-6}
Yi1.5 (34B)  & $v_1$ ($\mathrm{A}$) & 5  & 40  & 19.3  & 100 \\  
\cite{DBLP:journals/corr/abs-2403-04652}      & $v_2$ ($\mathrm{B}$) & 2 & 18  & 40.8  & 100 \\ 
                                              & $v_3$ ($\mathrm{C}$) & N/A & N/A & 27.9 & 26.2 \\
\cmidrule(l){2-6}
Llama3.1 (8B)  & $v_1$ ($\mathrm{A}$) & 4  & 29  & 30.4  & 100 \\  
\cite{DBLP:journals/corr/abs-2407-21783}      & $v_2$ ($\mathrm{B}$) & 2 & 15  & 27.2  & 100 \\ 
                                               & $v_3$ ($\mathrm{C}$) & N/A & N/A & 18.5 & 17.6 \\
\cmidrule(l){2-6}
Llama3.2 (3B)  & $v_1$ ($\mathrm{A}$) & 5  & 36  & 26.2  & 91.7 \\  
\cite{DBLP:journals/corr/abs-2407-21783}       & $v_2$ ($\mathrm{B}$) & 2 & 16  & 29.1  & 98.7 \\ 
                                                & $v_3$ ($\mathrm{C}$) & N/A & N/A & 18.3 & 17.3 \\
\cmidrule(l){2-6}
Mistral\text{-}Nemo (12B)  & $v_1$ ($\mathrm{A}$) & 5  & 36  & 17.2  & 100 \\  
\cite{mistral-nemo}                     & $v_2$ ($\mathrm{B}$) & 2 & 16  & 29.9  & 100 \\ 
                                        & $v_3$ ($\mathrm{C}$) & N/A & N/A & 22.0 & 19.8 \\
\bottomrule
\end{tabular}

  \caption{
  Results for various models on the Level 4 task ($\tau=0.85$).
  }
  \label{table:rsp_model_dimmension_tau_0.85_level_4}
\end{table}

\begin{table}[p]
  \tabcolsep 0.3mm
  \footnotesize
  \centering
  \begin{tabular}{lcrrrrr}
\toprule
&&\multicolumn{2}{c}{When ($\downarrow$)} & \multicolumn{2}{c}{$\mathrm{Acc}$ ($\uparrow$)} \\
\cmidrule(lr){3-4} \cmidrule(lr){5-6}
Model & Variable & $t^*_\mathrm{eq}$ & $t^*$ & $\mathrm{\prec CoT}$ & $\mathrm{\succ CoT}$ \\ 
\midrule
Qwen2.5 (7B)  & $v_1$ ($\mathrm{A}$) & 9  & 62  & 18.1  & 100 \\  
\cite{qwen2.5}                    & $v_2$ ($\mathrm{B}$) & 6  & 42  & 22.6  & 100 \\ 
                                   & $v_3$ ($\mathrm{C}$) & 3  & 23  & 50.6  & 100 \\
\cmidrule(l){2-6}
Qwen2.5 (14B) & $v_1$ ($\mathrm{A}$) & 9  & 63  & 18.1  & 98.8 \\  
\cite{qwen2.5}                    & $v_2$ ($\mathrm{B}$) & 6  & 42  & 18.7  & 98.9 \\  
                                   & $v_3$ ($\mathrm{C}$) & 3  & 23  & 42.2  & 100 \\
\cmidrule(l){2-6}
Qwen2.5 (32B) & $v_1$ ($\mathrm{A}$) & 9  & 63  & 18.7  & 100 \\  
\cite{qwen2.5}                    & $v_2$ ($\mathrm{B}$) & 6  & 43  & 22.6  & 100 \\ 
                                   & $v_3$ ($\mathrm{C}$) & 3  & 23  & 62.4  & 100 \\
\cmidrule(l){2-6}
Qwen2.5\text{-}Math (7B) & $v_1$ ($\mathrm{A}$) & 9  & 62  & 18.1  & 100 \\  
\cite{yang2024qwen25mathtechnicalreportmathematical} & $v_2$ ($\mathrm{B}$) & 6  & 42  & 22.6  & 100 \\  
                                   & $v_3$ ($\mathrm{C}$) & 3  & 22  & 54.5  & 100 \\
\cmidrule(l){2-6}
Yi1.5 (34B) & $v_1$ ($\mathrm{A}$) & 9  & 71  & 18.1  & 100 \\  
\cite{DBLP:journals/corr/abs-2403-04652}      & $v_2$ ($\mathrm{B}$) & 6  & 49  & 22.6  & 100 \\  
                                   & $v_3$ ($\mathrm{C}$) & 3  & 26  & 41.2  & 100 \\
\cmidrule(l){2-6}
Llama3.1 (8B) & $v_1$ ($\mathrm{A}$) & 9  & 62  & 16.0  & 99.5 \\  
\cite{DBLP:journals/corr/abs-2407-21783}  & $v_2$ ($\mathrm{B}$) & 6  & 43  & 20.0  & 99.5 \\ 
                                   & $v_3$ ($\mathrm{C}$) & 3  & 23  & 30.6  & 99.8 \\
\cmidrule(l){2-6}
Llama3.2 (3B) & $v_1$ ($\mathrm{A}$) & N/A  & N/A  & 14.2  & 43.7 \\  
\cite{DBLP:journals/corr/abs-2407-21783}        & $v_2$ ($\mathrm{B}$) & N/A  & N/A  & 26.3  & 47.4 \\ 
                                   & $v_3$ ($\mathrm{C}$) & N/A  & N/A  & 37.7  & 71.7 \\
\cmidrule(l){2-6}
Mistral\text{-}Nemo (12B) & $v_1$ ($\mathrm{A}$) & 9  & 63  & 18.1  & 99.9 \\  
\cite{mistral-nemo}   & $v_2$ ($\mathrm{B}$) & 6  & 43  & 16.3  & 99.9 \\ 
                                   & $v_3$ ($\mathrm{C}$) & 3  & 23  & 32.0  & 99.9 \\
\bottomrule
\end{tabular}

  \caption{
  Results for various models on the Level 5 task ($\tau=0.85$).
  }
  \label{table:rsp_model_dimmension_tau_0.85_level_5}
\end{table}

\begin{table}[p]
  \tabcolsep 0.3mm
  \footnotesize
  \centering
  \begin{tabular}{lcrrrrr}
\toprule
&&\multicolumn{2}{c}{When ($\downarrow$)} & \multicolumn{2}{c}{$\mathrm{Acc}$ ($\uparrow$)} \\
\cmidrule(lr){3-4} \cmidrule(lr){5-6}
Model & Variable & $t^*_\mathrm{eq}$ & $t^*$ & $\mathrm{\prec CoT}$ & $\mathrm{\succ CoT}$ \\ 
\midrule
Qwen2.5 (7B)  & $v_1$ ($\mathrm{A}$) & 4  & 27  & 35.8  & 100 \\  
\cite{qwen2.5}                    & $v_2$ ($\mathrm{B}$) & \textbf{$-$2} & \textbf{$-$5}  & 100  & 100 \\ 
\cmidrule(l){2-6}
Qwen2.5 (14B) & $v_1$ ($\mathrm{A}$) & 4  & 27  & 36.9  & 100 \\  
\cite{qwen2.5}                    & $v_2$ ($\mathrm{B}$) & \textbf{$-$2} & \textbf{$-$5}  & 100 & 100 \\  
\cmidrule(l){2-6}
Qwen2.5 (32B) & $v_1$ ($\mathrm{A}$) & 4  & 28  & 30.5  & 100 \\  
\cite{qwen2.5}                    & $v_2$ ($\mathrm{B}$) & \textbf{$-$2} & \textbf{$-$5}  & 100 & 100 \\ 
\cmidrule(l){2-6}
Qwen2.5-Math (7B) & $v_1$ ($\mathrm{A}$) & 4  & 27  & 41.8  & 100 \\  
\cite{yang2024qwen25mathtechnicalreportmathematical} & $v_2$ ($\mathrm{B}$) & \textbf{$-$2} & \textbf{$-$5}  & 100 & 100 \\  
\cmidrule(l){2-6}
Yi1.5 (9B) & $v_1$ ($\mathrm{A}$) & 4  & 32  & 28.1  & 100 \\  
\cite{DBLP:journals/corr/abs-2403-04652}      & $v_2$ ($\mathrm{B}$) & \textbf{$-$2} & \textbf{$-$5}  & 100  & 100 \\  
\cmidrule(l){2-6}
Yi1.5 (34B) & $v_1$ ($\mathrm{A}$) & 4  & 32  & 22.9  & 100 \\  
\cite{DBLP:journals/corr/abs-2403-04652}      & $v_2$ ($\mathrm{B}$) & \textbf{$-$2} & \textbf{$-$5}  & 100 & 100 \\ 
\cmidrule(l){2-6}
Llama3.1 (8B) & $v_1$ ($\mathrm{A}$) & 4  & 27  & 20.6  & 100 \\  
\cite{DBLP:journals/corr/abs-2407-21783}  & $v_2$ ($\mathrm{B}$) & \textbf{$-$2} & \textbf{$-$5}   & 100  & 100 \\ 
\cmidrule(l){2-6}
Llama3.2 (3B) & $v_1$ ($\mathrm{A}$) & 4  & 28  & 21.8  & 100.0 \\  
\cite{DBLP:journals/corr/abs-2407-21783}        & $v_2$ ($\mathrm{B}$) & \textbf{$-$2} & \textbf{$-$5}  & 100  & 100 \\ 
\cmidrule(l){2-6}
Mistral-Nemo (12B) & $v_1$ ($\mathrm{A}$) & 4  & 28  & 18.0  & 100 \\  
\cite{mistral-nemo}   & $v_2$ ($\mathrm{B}$) & \textbf{$-$2}  & \textbf{$-$5}  & 100  & 100 \\ 
\bottomrule
\end{tabular}

  \caption{
  Results for various models on the Level 1 task ($\tau=0.90$).
  }
  \label{table:rsp_model_dimmension_tau_0.9_level_1}
\end{table}

\begin{table}[p]
  \tabcolsep 0.3mm
  \footnotesize
  \centering
  \begin{tabular}{lcrrrrr}
\toprule
&&\multicolumn{2}{c}{When ($\downarrow$)} & \multicolumn{2}{c}{$\mathrm{Acc}$ ($\uparrow$)} \\
\cmidrule(lr){3-4} \cmidrule(lr){5-6}
Model & Variable & $t^*_\mathrm{eq}$ & $t^*$ & $\mathrm{\prec CoT}$ & $\mathrm{\succ CoT}$ \\ 
\midrule
Qwen2.5 (7B)  & $v_1$ ($\mathrm{A}$) & 2  & 16  & 49.2  & 100 \\  
\cite{qwen2.5}                    & $v_2$ ($\mathrm{B}$) & 5 & 35  & 21.2  & 100 \\ 
\cmidrule(l){2-6}
Qwen2.5 (14B)  & $v_1$ ($\mathrm{A}$) & 2  & 16  & 48.8  & 100 \\  
\cite{qwen2.5}                    & $v_2$ ($\mathrm{B}$) & 5 & 36  & 21.5  & 100 \\ 
\cmidrule(l){2-6}
Qwen2.5 (32B)  & $v_1$ ($\mathrm{A}$) & 2  & 16  & 66.4  & 100 \\  
\cite{qwen2.5}                    & $v_2$ ($\mathrm{B}$) & 5 & 36  & 21.3  & 100 \\ 
\cmidrule(l){2-6}
Qwen2.5\text{-}Math (7B)  & $v_1$ ($\mathrm{A}$) & 2  & 15  & 53.7  & 100 \\  
\cite{yang2024qwen25mathtechnicalreportmathematical} & $v_2$ ($\mathrm{B}$) & 5 & 35  & 22.1  & 100 \\  
\cmidrule(l){2-6}
Yi1.5 (9B)  & $v_1$ ($\mathrm{A}$) & 2  & 18  & 40.2  & 100 \\  
\cite{DBLP:journals/corr/abs-2403-04652}      & $v_2$ ($\mathrm{B}$) & 5 & 41  & 17.8  & 100 \\  
\cmidrule(l){2-6}
Yi1.5 (34B)  & $v_1$ ($\mathrm{A}$) & 2  & 18  & 35.6  & 100 \\  
\cite{DBLP:journals/corr/abs-2403-04652}      & $v_2$ ($\mathrm{B}$) & 5 & 41  & 18.3  & 100 \\ 
\cmidrule(l){2-6}
Llama3.1 (8B)  & $v_1$ ($\mathrm{A}$) & 2  & 15  & 31.9  & 100 \\  
\cite{DBLP:journals/corr/abs-2407-21783}  & $v_2$ ($\mathrm{B}$) & 5 & 35  & 17.8  & 100 \\ 
\cmidrule(l){2-6}
Llama3.2 (3B)  & $v_1$ ($\mathrm{A}$) & 2  & 16  & 36.2  & 99.9 \\  
\cite{DBLP:journals/corr/abs-2407-21783}        & $v_2$ ($\mathrm{B}$) & 5 & 36  & 17.8  & 99.9 \\ 
\cmidrule(l){2-6}
Mistral-Nemo (12B)  & $v_1$ ($\mathrm{A}$) & 2  & 16  & 30.8  & 100 \\  
\cite{mistral-nemo}   & $v_2$ ($\mathrm{B}$) & 5  & 36  & 17.8  & 100 \\ 
\bottomrule
\end{tabular}

  \caption{
  Results for various models on the Level 2 task ($\tau=0.90$).
  }
  \label{table:rsp_model_dimmension_tau_0.9_level_2}
\end{table}

\begin{table}[p]
  \tabcolsep 0.3mm
  \footnotesize
  \centering
  \begin{tabular}{lcrrrrr}
\toprule
&&\multicolumn{2}{c}{When ($\downarrow$)} & \multicolumn{2}{c}{$\mathrm{Acc}$ ($\uparrow$)} \\
\cmidrule(lr){3-4} \cmidrule(lr){5-6}
 & Variable & $t^*_\mathrm{eq}$ & $t^*$ & $\mathrm{\prec CoT}$ & $\mathrm{\succ CoT}$ \\ 
\midrule
Qwen2.5 (7B)  & $v_1$ ($\mathrm{A}$) & 5  & 36  & 17.9  & 100 \\  
\cite{qwen2.5}                    & $v_2$ ($\mathrm{B}$) & 2 & 16  & 50.5  & 100 \\ 
\cmidrule(l){2-6}
Qwen2.5 (14B)  & $v_1$ ($\mathrm{A}$) & 5  & 35  & 17.8  & 100 \\  
\cite{qwen2.5}                    & $v_2$ ($\mathrm{B}$) & 2 & 16  & 50.5  & 100 \\ 
\cmidrule(l){2-6}
Qwen2.5 (32B)  & $v_1$ ($\mathrm{A}$) & 5  & 36  & 17.8  & 100 \\  
\cite{qwen2.5}                    & $v_2$ ($\mathrm{B}$) & 2 & 15  & 67.4  & 100 \\ 
\cmidrule(l){2-6}
Qwen2.5-Math (7B)  & $v_1$ ($\mathrm{A}$) & 5  & 35  & 18.6  & 100 \\  
\cite{yang2024qwen25mathtechnicalreportmathematical} & $v_2$ ($\mathrm{B}$) & 2 & 15  & 56.1  & 100 \\  
\cmidrule(l){2-6}
Yi1.5 (9B)  & $v_1$ ($\mathrm{A}$) & 5  & 41  & 17.8  & 100 \\  
\cite{DBLP:journals/corr/abs-2403-04652}      & $v_2$ ($\mathrm{B}$) & 2 & 18  & 36.9  & 100 \\  
\cmidrule(l){2-6}
Yi1.5 (34B)  & $v_1$ ($\mathrm{A}$) & 5  & 41  & 22.4  & 100 \\  
\cite{DBLP:journals/corr/abs-2403-04652}      & $v_2$ ($\mathrm{B}$) & 2 & 18  & 37.4  & 100 \\ 
\cmidrule(l){2-6}
Llama3.1 (8B)  & $v_1$ ($\mathrm{A}$) & 5  & 35  & 26.0  & 100 \\  
\cite{DBLP:journals/corr/abs-2407-21783}  & $v_2$ ($\mathrm{B}$) & 2 & 16  & 29.6  & 100 \\ 
\cmidrule(l){2-6}
Llama3.2 (3B)  & $v_1$ ($\mathrm{A}$) & 5  & 36  & 17.8  & 93.2 \\  
\cite{DBLP:journals/corr/abs-2407-21783}        & $v_2$ ($\mathrm{B}$) & 2 & 17  & 33.2  & 95.4 \\ 
\cmidrule(l){2-6}
Mistral-Nemo (12B)  & $v_1$ ($\mathrm{A}$) & 5  & 36  & 17.8  & 100 \\  
\cite{mistral-nemo}   & $v_2$ ($\mathrm{B}$) & 2  & 16  & 28.9  & 100 \\ 
\bottomrule
\end{tabular}

  \caption{
  Results for various models on the Level 3 task ($\tau=0.90$).
  }
  \label{table:rsp_model_dimmension_tau_0.9_level_3}
\end{table}

\begin{table}[p]
  \tabcolsep 0.3mm
  \footnotesize
  \centering
  \begin{tabular}{lcrrrrr}
\toprule
&&\multicolumn{2}{c}{When ($\downarrow$)} & \multicolumn{2}{c}{$\mathrm{Acc}$ ($\uparrow$)} \\
\cmidrule(lr){3-4} \cmidrule(lr){5-6}
Model & Variable & $t^*_\mathrm{eq}$ & $t^*$ & $\mathrm{\prec CoT}$ & $\mathrm{\succ CoT}$ \\ 
\midrule
Qwen2.5 (7B)  & $v_1$ ($\mathrm{A}$) & 5  & 35  & 17.2  & 100 \\  
\cite{qwen2.5}                    & $v_2$ ($\mathrm{B}$) & 2 & 16  & 47.7  & 100 \\ 
                                   & $v_3$ ($\mathrm{C}$) & N/A & N/A & 43.7 & 23.7 \\
\cmidrule(l){2-6}
Qwen2.5 (14B)  & $v_1$ ($\mathrm{A}$) & 5  & 36  & 18.9  & 100 \\  
\cite{qwen2.5}                     & $v_2$ ($\mathrm{B}$) & 2 & 15  & 44.3  & 100 \\ 
                                    & $v_3$ ($\mathrm{C}$) & N/A & N/A & 40.4 & 26.8 \\
\cmidrule(l){2-6}
Qwen2.5 (32B)  & $v_1$ ($\mathrm{A}$) & 5  & 36  & 17.4  & 100 \\  
\cite{qwen2.5}                     & $v_2$ ($\mathrm{B}$) & 2 & 15  & 62.8  & 100 \\ 
                                    & $v_3$ ($\mathrm{C}$) & N/A & N/A & 64.4 & 32.6 \\
\cmidrule(l){2-6}
Qwen2.5\text{-}Math (7B)  & $v_1$ ($\mathrm{A}$) & 5  & 35  & 17.2  & 100 \\  
\cite{yang2024qwen25mathtechnicalreportmathematical} & $v_2$ ($\mathrm{B}$) & 2 & 15  & 55.6  & 100 \\ 
                                                            & $v_3$ ($\mathrm{C}$) & N/A & N/A & 47.8 & 29.4 \\
\cmidrule(l){2-6}
Yi1.5 (9B)  & $v_1$ ($\mathrm{A}$) & 5  & 41  & 17.8  & 100 \\  
\cite{DBLP:journals/corr/abs-2403-04652}      & $v_2$ ($\mathrm{B}$) & 2 & 18  & 43.5  & 100 \\ 
                                              & $v_3$ ($\mathrm{C}$) & N/A & N/A & 36.7 & 21.2 \\
\cmidrule(l){2-6}
Yi1.5 (34B)  & $v_1$ ($\mathrm{A}$) & 5  & 40  & 19.3  & 100 \\  
\cite{DBLP:journals/corr/abs-2403-04652}      & $v_2$ ($\mathrm{B}$) & 2 & 18  & 40.8  & 100 \\ 
                                              & $v_3$ ($\mathrm{C}$) & N/A & N/A & 27.9 & 26.2 \\
\cmidrule(l){2-6}
Llama3.1 (8B)  & $v_1$ ($\mathrm{A}$) & 5  & 35  & 30.4  & 100 \\  
\cite{DBLP:journals/corr/abs-2407-21783}      & $v_2$ ($\mathrm{B}$) & 2 & 16  & 27.2  & 100 \\ 
                                               & $v_3$ ($\mathrm{C}$) & N/A & N/A & 18.5 & 17.6 \\
\cmidrule(l){2-6}
Llama3.2 (3B)  & $v_1$ ($\mathrm{A}$) & 5  & 37  & 26.2  & 91.7 \\  
\cite{DBLP:journals/corr/abs-2407-21783}       & $v_2$ ($\mathrm{B}$) & 2 & 17  & 29.1  & 98.7 \\ 
                                                & $v_3$ ($\mathrm{C}$) & N/A & N/A & 18.3 & 17.3 \\
\cmidrule(l){2-6}
Mistral-Nemo (12B)  & $v_1$ ($\mathrm{A}$) & 5  & 36  & 17.2  & 100 \\  
\cite{mistral-nemo}                     & $v_2$ ($\mathrm{B}$) & 2 & 16  & 29.9  & 100 \\ 
                                        & $v_3$ ($\mathrm{C}$) & N/A & N/A & 22.0 & 19.8 \\
\bottomrule
\end{tabular}

  \caption{
  Results for various models on the Level 4 task ($\tau=0.90$).
  }
  \label{table:rsp_model_dimmension_tau_0.9_level_4}
\end{table}

\begin{table}[p]
  \tabcolsep 0.3mm
  \footnotesize
  \centering
  \begin{tabular}{lcrrrrr}
\toprule
&&\multicolumn{2}{c}{When ($\downarrow$)} & \multicolumn{2}{c}{$\mathrm{Acc}$ ($\uparrow$)} \\
\cmidrule(lr){3-4} \cmidrule(lr){5-6}
Model & Variable & $t^*_\mathrm{eq}$ & $t^*$ & $\mathrm{\prec CoT}$ & $\mathrm{\succ CoT}$ \\ 
\midrule
Qwen2.5 (7B)  & $v_1$ ($\mathrm{A}$) & 9  & 63  & 18.1  & 100 \\  
\cite{qwen2.5}                    & $v_2$ ($\mathrm{B}$) & 6  & 42  & 22.6  & 100 \\ 
& $v_3$ ($\mathrm{C}$) & 3  & 23  & 50.6  & 100 \\
\cmidrule(l){2-6}
Qwen2.5 (14B) & $v_1$ ($\mathrm{A}$) & 9  & 63  & 18.1  & 98.8 \\  
\cite{qwen2.5}                    & $v_2$ ($\mathrm{B}$) & 6  & 42  & 18.7  & 98.9 \\  
& $v_3$ ($\mathrm{C}$) & 3  & 23  & 42.2  & 100 \\
\cmidrule(l){2-6}
Qwen2.5 (32B) & $v_1$ ($\mathrm{A}$) & 9  & 63  & 18.7  & 100 \\  
\cite{qwen2.5}                    & $v_2$ ($\mathrm{B}$) & 6  & 43  & 22.6  & 100 \\ 
& $v_3$ ($\mathrm{C}$) & 3  & 23  & 62.4  & 100 \\
\cmidrule(l){2-6}
Qwen2.5\text{-}Math (7B) & $v_1$ ($\mathrm{A}$) & 9  & 62  & 18.1  & 100 \\  
\cite{yang2024qwen25mathtechnicalreportmathematical} & $v_2$ ($\mathrm{B}$) & 6  & 42  & 22.6  & 100 \\  
& $v_3$ ($\mathrm{C}$) & 3  & 22  & 54.5  & 100 \\
\cmidrule(l){2-6}
Yi1.5 (34B) & $v_1$ ($\mathrm{A}$) & 9  & 72  & 18.1  & 100 \\  
\cite{DBLP:journals/corr/abs-2403-04652}      & $v_2$ ($\mathrm{B}$) & 6  & 49  & 22.6  & 100 \\  
& $v_3$ ($\mathrm{C}$) & 3  & 26  & 41.2  & 100 \\
\cmidrule(l){2-6}
Llama3.1 (8B) & $v_1$ ($\mathrm{A}$) & 9  & 62  & 16.0  & 99.5 \\  
\cite{DBLP:journals/corr/abs-2407-21783}  & $v_2$ ($\mathrm{B}$) & 6  & 43  & 20.0  & 99.5 \\ 
& $v_3$ ($\mathrm{C}$) & 3  & 23  & 30.6  & 99.8 \\
\cmidrule(l){2-6}
Llama3.2 (3B) & $v_1$ ($\mathrm{A}$) & N/A  & N/A  & 14.2  & 43.7 \\  
\cite{DBLP:journals/corr/abs-2407-21783}        & $v_2$ ($\mathrm{B}$) & N/A  & N/A  & 26.3  & 47.4 \\ 
& $v_3$ ($\mathrm{C}$) & N/A  & N/A  & 37.7  & 71.7 \\
\cmidrule(l){2-6}
Mistral-Nemo (12B) & $v_1$ ($\mathrm{A}$) & 9  & 63  & 18.1  & 99.9 \\  
\cite{mistral-nemo}   & $v_2$ ($\mathrm{B}$) & 6  & 43  & 16.3  & 99.9 \\ 
& $v_3$ ($\mathrm{C}$) & 3  & 23  & 32.0  & 99.9 \\
\bottomrule
\end{tabular}

  \caption{
  Results for various models on the Level 5 task ($\tau=0.90$).
  }
  \label{table:rsp_model_dimmension_tau_0.9_level_5}
\end{table}

\begin{table}[p]
  \tabcolsep 0.3mm
  \footnotesize
  \centering
  \begin{tabular}{lcrrrrr}
\toprule
&&\multicolumn{2}{c}{When ($\downarrow$)} & \multicolumn{2}{c}{$\mathrm{Acc}$ ($\uparrow$)} \\
\cmidrule(lr){3-4} \cmidrule(lr){5-6}
Model & Variable & $t^*_\mathrm{eq}$ & $t^*$ & $\mathrm{\prec CoT}$ & $\mathrm{\succ CoT}$ \\ 
\midrule
Qwen2.5 (7B)  & $v_1$ ($\mathrm{A}$) & 4  & 27  & 35.8  & 100 \\  
\cite{qwen2.5}                    & $v_2$ ($\mathrm{B}$) & \textbf{$-$2} & \textbf{$-$5}  & 100  & 100 \\ 
\cmidrule(l){2-6}
Qwen2.5 (14B) & $v_1$ ($\mathrm{A}$) & 4  & 28  & 36.9  & 100 \\  
\cite{qwen2.5}                    & $v_2$ ($\mathrm{B}$) & \textbf{$-$2} & \textbf{$-$5}  & 100 & 100 \\  
\cmidrule(l){2-6}
Qwen2.5 (32B) & $v_1$ ($\mathrm{A}$) & 4  & 28  & 30.5  & 100 \\  
\cite{qwen2.5}                    & $v_2$ ($\mathrm{B}$) & \textbf{$-$2} & \textbf{$-$5}  & 100 & 100 \\ 
\cmidrule(l){2-6}
Qwen2.5\text{-}Math (7B) & $v_1$ ($\mathrm{A}$) & 4  & 27  & 41.8  & 100 \\  
\cite{yang2024qwen25mathtechnicalreportmathematical} & $v_2$ ($\mathrm{B}$) & \textbf{$-$2} & \textbf{$-$5}  & 100 & 100 \\  
\cmidrule(l){2-6}
Yi1.5 (9B) & $v_1$ ($\mathrm{A}$) & 4  & 32  & 28.1  & 100 \\  
\cite{DBLP:journals/corr/abs-2403-04652}      & $v_2$ ($\mathrm{B}$) & \textbf{$-$2} & \textbf{$-$5}  & 100  & 100 \\  
\cmidrule(l){2-6}
Yi1.5 (34B) & $v_1$ ($\mathrm{A}$) & 4  & 32  & 22.9  & 100 \\  
\cite{DBLP:journals/corr/abs-2403-04652}      & $v_2$ ($\mathrm{B}$) & \textbf{$-$2} & \textbf{$-$5}  & 100 & 100 \\ 
\cmidrule(l){2-6}
Llama3.1 (8B) & $v_1$ ($\mathrm{A}$) & 4  & 27  & 20.6  & 100 \\  
\cite{DBLP:journals/corr/abs-2407-21783}  & $v_2$ ($\mathrm{B}$) & \textbf{$-$2} & \textbf{$-$5}   & 100  & 100 \\ 
\cmidrule(l){2-6}
Llama3.2 (3B) & $v_1$ ($\mathrm{A}$) & 4  & 28  & 21.8  & 100.0 \\  
\cite{DBLP:journals/corr/abs-2407-21783}        & $v_2$ ($\mathrm{B}$) & \textbf{$-$2} & \textbf{$-$5}  & 100  & 100 \\ 
\cmidrule(l){2-6}
Mistral\text{-}Nemo (12B) & $v_1$ ($\mathrm{A}$) & 4  & 28  & 17.9  & 100 \\  
\cite{mistral-nemo}   & $v_2$ ($\mathrm{B}$) & \textbf{$-$2}  & \textbf{$-$5}  & 100  & 100 \\ 
\bottomrule
\end{tabular}

  \caption{
  Results for various models on the Level 1 task ($\tau=0.95$).
  }
  \label{table:rsp_model_dimmension_tau_0.95_level_1}
\end{table}

\begin{table}[p]
  \tabcolsep 0.3mm
  \footnotesize
  \centering
  \begin{tabular}{lcrrrrr}
\toprule
&&\multicolumn{2}{c}{When ($\downarrow$)} & \multicolumn{2}{c}{$\mathrm{Acc}$ ($\uparrow$)} \\
\cmidrule(lr){3-4} \cmidrule(lr){5-6}
Model & Variable & $t^*_\mathrm{eq}$ & $t^*$ & $\mathrm{\prec CoT}$ & $\mathrm{\succ CoT}$ \\ 
\midrule
Qwen2.5 (7B)  & $v_1$ ($\mathrm{A}$) & 2  & 16  & 49.2  & 100 \\  
\cite{qwen2.5}                    & $v_2$ ($\mathrm{B}$) & 5 & 36  & 21.2  & 100 \\ 
\cmidrule(l){2-6}
Qwen2.5 (14B)  & $v_1$ ($\mathrm{A}$) & 2  & 16  & 48.8  & 100 \\  
\cite{qwen2.5}                    & $v_2$ ($\mathrm{B}$) & 5 & 36  & 21.5  & 100 \\ 
\cmidrule(l){2-6}
Qwen2.5 (32B)  & $v_1$ ($\mathrm{A}$) & 2  & 16  & 66.4  & 100 \\  
\cite{qwen2.5}                    & $v_2$ ($\mathrm{B}$) & 5 & 36  & 21.3  & 100 \\ 
\cmidrule(l){2-6}
Qwen2.5\text{-}Math (7B)  & $v_1$ ($\mathrm{A}$) & 2  & 15  & 53.7  & 100 \\  
\cite{yang2024qwen25mathtechnicalreportmathematical} & $v_2$ ($\mathrm{B}$) & 5 & 36  & 22.1  & 100 \\  
\cmidrule(l){2-6}
Yi1.5 (9B)  & $v_1$ ($\mathrm{A}$) & 2  & 18  & 40.2  & 100 \\  
\cite{DBLP:journals/corr/abs-2403-04652}      & $v_2$ ($\mathrm{B}$) & 5 & 41  & 17.8  & 100 \\  
\cmidrule(l){2-6}
Yi1.5 (34B)  & $v_1$ ($\mathrm{A}$) & 2  & 18  & 35.6  & 100 \\  
\cite{DBLP:journals/corr/abs-2403-04652}      & $v_2$ ($\mathrm{B}$) & 5 & 41  & 18.3  & 100 \\ 
\cmidrule(l){2-6}
Llama3.1 (8B)  & $v_1$ ($\mathrm{A}$) & 2  & 15  & 31.9  & 100 \\  
\cite{DBLP:journals/corr/abs-2407-21783}  & $v_2$ ($\mathrm{B}$) & 5 & 35  & 17.8  & 100 \\ 
\cmidrule(l){2-6}
Llama3.2 (3B)  & $v_1$ ($\mathrm{A}$) & 2  & 16  & 36.2  & 99.9 \\  
\cite{DBLP:journals/corr/abs-2407-21783}        & $v_2$ ($\mathrm{B}$) & 5 & 36  & 17.8  & 99.9 \\ 
\cmidrule(l){2-6}
Mistral\text{-}Nemo (12B)  & $v_1$ ($\mathrm{A}$) & 2  & 16  & 30.8  & 100 \\  
\cite{mistral-nemo}   & $v_2$ ($\mathrm{B}$) & 5  & 36  & 17.8  & 100 \\ 
\bottomrule
\end{tabular}

  \caption{
  Results for various models on the Level 2 task ($\tau=0.95$).
  }
  \label{table:rsp_model_dimmension_tau_0.95_level_2}
\end{table}

\begin{table}[p]
  \tabcolsep 0.3mm
  \footnotesize
  \centering
  \begin{tabular}{lcrrrrr}
\toprule
&&\multicolumn{2}{c}{When ($\downarrow$)} & \multicolumn{2}{c}{$\mathrm{Acc}$ ($\uparrow$)} \\
\cmidrule(lr){3-4} \cmidrule(lr){5-6}
Model & Variable & $t^*_\mathrm{eq}$ & $t^*$ & $\mathrm{\prec CoT}$ & $\mathrm{\succ CoT}$ \\ 
\midrule
Qwen2.5 (7B)  & $v_1$ ($\mathrm{A}$) & 5  & 36  & 17.9  & 100 \\  
\cite{qwen2.5}                    & $v_2$ ($\mathrm{B}$) & 2  & 16  & 50.5  & 100 \\ 
\cmidrule(l){2-6}
Qwen2.5 (14B) & $v_1$ ($\mathrm{A}$) & 5  & 36  & 17.8  & 100 \\  
\cite{qwen2.5}                    & $v_2$ ($\mathrm{B}$) & 2  & 16  & 50.5  & 100 \\  
\cmidrule(l){2-6}
Qwen2.5 (32B) & $v_1$ ($\mathrm{A}$) & 5  & 36  & 17.8  & 100 \\  
\cite{qwen2.5}                    & $v_2$ ($\mathrm{B}$) & 2  & 15  & 67.4  & 100 \\ 
\cmidrule(l){2-6}
Qwen2.5\text{-}Math (7B) & $v_1$ ($\mathrm{A}$) & 5  & 35  & 18.6  & 100 \\  
\cite{yang2024qwen25mathtechnicalreportmathematical} & $v_2$ ($\mathrm{B}$) & 2  & 15  & 56.1  & 100 \\  
\cmidrule(l){2-6}
Yi1.5 (9B) & $v_1$ ($\mathrm{A}$) & 5  & 41  & 17.8  & 100 \\  
\cite{DBLP:journals/corr/abs-2403-04652}      & $v_2$ ($\mathrm{B}$) & 2  & 18  & 36.9  & 100 \\  
\cmidrule(l){2-6}
Yi1.5 (34B) & $v_1$ ($\mathrm{A}$) & 5  & 41  & 22.4  & 100 \\  
\cite{DBLP:journals/corr/abs-2403-04652}      & $v_2$ ($\mathrm{B}$) & 2  & 18  & 37.4  & 100 \\ 
\cmidrule(l){2-6}
Llama3.1 (8B) & $v_1$ ($\mathrm{A}$) & 5  & 35  & 26.0  & 100 \\  
\cite{DBLP:journals/corr/abs-2407-21783}  & $v_2$ ($\mathrm{B}$) & 2  & 16  & 29.6  & 100 \\ 
\cmidrule(l){2-6}
Llama3.2 (3B) & $v_1$ ($\mathrm{A}$) & N/A  & N/A  & 17.8  & 93.2 \\  
\cite{DBLP:journals/corr/abs-2407-21783}        & $v_2$ ($\mathrm{B}$) & 2  & 17  & 33.2  & 95.4 \\ 
\cmidrule(l){2-6}
Mistral\text{-}Nemo (12B) & $v_1$ ($\mathrm{A}$) & 5  & 36  & 17.8  & 100 \\  
\cite{mistral-nemo}   & $v_2$ ($\mathrm{B}$) & 2  & 16  & 28.9  & 100 \\ 
\bottomrule
\end{tabular}

  \caption{
  Results for various models on the Level 3 task ($\tau=0.95$).
  }
  \label{table:rsp_model_dimmension_tau_0.95_level_3}
\end{table}

\begin{table}[p]
  \tabcolsep 0.3mm
  \footnotesize
  \centering
  \begin{tabular}{lcrrrrr}
\toprule
&&\multicolumn{2}{c}{When ($\downarrow$)} & \multicolumn{2}{c}{$\mathrm{Acc}$ ($\uparrow$)} \\
\cmidrule(lr){3-4} \cmidrule(lr){5-6}
Model & Variable & $t^*_\mathrm{eq}$ & $t^*$ & $\mathrm{\prec CoT}$ & $\mathrm{\succ CoT}$ \\ 
\midrule
Qwen2.5 (7B)  & $v_1$ ($\mathrm{A}$) & 5  & 36  & 17.2  & 100 \\  
\cite{qwen2.5}                    & $v_2$ ($\mathrm{B}$) & 2 & 16  & 47.7  & 100 \\ 
                                   & $v_3$ ($\mathrm{C}$) & N/A & N/A & 43.7 & 23.7 \\
\cmidrule(l){2-6}
Qwen2.5 (14B) & $v_1$ ($\mathrm{A}$) & 5  & 36  & 18.9  & 100 \\  
\cite{qwen2.5}                    & $v_2$ ($\mathrm{B}$) & 2 & 15  & 44.3  & 100 \\  
                                   & $v_3$ ($\mathrm{C}$) & N/A & N/A & 40.4 & 26.8 \\
\cmidrule(l){2-6}
Qwen2.5 (32B) & $v_1$ ($\mathrm{A}$) & 5  & 36  & 17.4  & 100 \\  
\cite{qwen2.5}                    & $v_2$ ($\mathrm{B}$) & 2 & 16  & 62.8  & 100 \\ 
                                   & $v_3$ ($\mathrm{C}$) & N/A & N/A & 64.4 & 32.6 \\
\cmidrule(l){2-6}
Qwen2.5\text{-}Math (7B) & $v_1$ ($\mathrm{A}$) & 5  & 35  & 17.2  & 100 \\  
\cite{yang2024qwen25mathtechnicalreportmathematical} & $v_2$ ($\mathrm{B}$) & 2 & 15  & 55.6  & 100 \\  
                                   & $v_3$ ($\mathrm{C}$) & N/A & N/A & 47.8 & 29.4 \\
\cmidrule(l){2-6}
Yi1.5 (9B)    & $v_1$ ($\mathrm{A}$) & 5  & 41  & 17.8  & 100 \\  
\cite{DBLP:journals/corr/abs-2403-04652}      & $v_2$ ($\mathrm{B}$) & 2 & 18  & 43.5  & 100 \\  
                                   & $v_3$ ($\mathrm{C}$) & N/A & N/A & 36.7 & 21.2 \\
\cmidrule(l){2-6}
Yi1.5 (34B)   & $v_1$ ($\mathrm{A}$) & 5  & 41  & 19.3  & 100 \\  
\cite{DBLP:journals/corr/abs-2403-04652}      & $v_2$ ($\mathrm{B}$) & 2 & 18  & 40.8  & 100 \\ 
                                   & $v_3$ ($\mathrm{C}$) & N/A & N/A & 27.9 & 26.2 \\
\cmidrule(l){2-6}
Llama3.1 (8B) & $v_1$ ($\mathrm{A}$) & 5  & 35  & 30.4  & 100 \\  
\cite{DBLP:journals/corr/abs-2407-21783}  & $v_2$ ($\mathrm{B}$) & 2 & 16  & 27.2  & 100 \\ 
                                   & $v_3$ ($\mathrm{C}$) & N/A & N/A & 18.5 & 17.6 \\
\cmidrule(l){2-6}
Llama3.2 (3B) & $v_1$ ($\mathrm{A}$) & N/A & N/A & 26.2 & 91.7 \\  
\cite{DBLP:journals/corr/abs-2407-21783}        & $v_2$ ($\mathrm{B}$) & 2 & 17  & 29.1  & 98.7 \\ 
                                   & $v_3$ ($\mathrm{C}$) & N/A & N/A & 18.3 & 17.3 \\
\cmidrule(l){2-6}
Mistral\text{-}Nemo (12B) & $v_1$ ($\mathrm{A}$) & 5  & 36  & 17.2  & 100 \\  
\cite{mistral-nemo}   & $v_2$ ($\mathrm{B}$) & 2  & 16  & 29.9  & 100 \\ 
                                   & $v_3$ ($\mathrm{C}$) & N/A & N/A & 22.0 & 19.8 \\
\bottomrule
\end{tabular}

  \caption{
  Results for various models on the Level 4 task ($\tau=0.95$).
  }
  \label{table:rsp_model_dimmension_tau_0.95_level_4}
\end{table}

\begin{table}[p]
  \tabcolsep 0.3mm
  \footnotesize
  \centering
  \begin{tabular}{lcrrrrr}
\toprule
&&\multicolumn{2}{c}{When ($\downarrow$)} & \multicolumn{2}{c}{$\mathrm{Acc}$ ($\uparrow$)} \\
\cmidrule(lr){3-4} \cmidrule(lr){5-6}
Model & Variable & $t^*_\mathrm{eq}$ & $t^*$ & $\mathrm{\prec CoT}$ & $\mathrm{\succ CoT}$ \\ 
\midrule
Qwen2.5 (7B)  & $v_1$ ($\mathrm{A}$) & 9  & 63  & 18.1  & 100 \\  
\cite{qwen2.5}                    & $v_2$ ($\mathrm{B}$) & 6  & 43  & 22.6  & 100 \\ 
                                   & $v_3$ ($\mathrm{C}$) & 3  & 23  & 50.6  & 100 \\
\cmidrule(l){2-6}
Qwen2.5 (14B) & $v_1$ ($\mathrm{A}$) & 9  & 63  & 18.1  & 98.8 \\  
\cite{qwen2.5}                    & $v_2$ ($\mathrm{B}$) & 6  & 43  & 18.7  & 98.9 \\  
                                   & $v_3$ ($\mathrm{C}$) & 3  & 23  & 42.2  & 100 \\
\cmidrule(l){2-6}
Qwen2.5 (32B) & $v_1$ ($\mathrm{A}$) & 9  & 63  & 18.7  & 100 \\  
\cite{qwen2.5}                    & $v_2$ ($\mathrm{B}$) & 6  & 43  & 22.6  & 100 \\ 
                                   & $v_3$ ($\mathrm{C}$) & 3  & 23  & 62.4  & 100 \\
\cmidrule(l){2-6}
Qwen2.5\text{-}Math (7B) & $v_1$ ($\mathrm{A}$) & 9  & 63  & 18.1  & 100 \\  
\cite{yang2024qwen25mathtechnicalreportmathematical} & $v_2$ ($\mathrm{B}$) & 6  & 42  & 22.6  & 100 \\  
                                   & $v_3$ ($\mathrm{C}$) & 3  & 22  & 54.5  & 100 \\
\cmidrule(l){2-6}
Yi1.5 (34B) & $v_1$ ($\mathrm{A}$) & 9  & 72  & 18.1  & 100 \\  
\cite{DBLP:journals/corr/abs-2403-04652}      & $v_2$ ($\mathrm{B}$) & 6  & 49  & 22.6  & 100 \\  
                                   & $v_3$ ($\mathrm{C}$) & 3  & 26  & 41.2  & 100 \\
\cmidrule(l){2-6}
Llama3.1 (8B) & $v_1$ ($\mathrm{A}$) & 9  & 62  & 16.0  & 99.5 \\  
\cite{DBLP:journals/corr/abs-2407-21783}  & $v_2$ ($\mathrm{B}$) & 6  & 43  & 20.0  & 99.5 \\ 
                                   & $v_3$ ($\mathrm{C}$) & 3  & 23  & 30.6  & 99.8 \\
\cmidrule(l){2-6}
Llama3.2 (3B) & $v_1$ ($\mathrm{A}$) & N/A  & N/A  & 14.1  & 43.7 \\  
\cite{DBLP:journals/corr/abs-2407-21783}        & $v_2$ ($\mathrm{B}$) & N/A  & N/A  & 26.3  & 47.4 \\ 
                                   & $v_3$ ($\mathrm{C}$) & N/A  & N/A  & 37.7  & 71.7 \\
\cmidrule(l){2-6}
Mistral\text{-}Nemo (12B) & $v_1$ ($\mathrm{A}$) & 9  & 63  & 18.1  & 99.9 \\  
\cite{mistral-nemo}   & $v_2$ ($\mathrm{B}$) & 6  & 43  & 16.3  & 99.9 \\ 
                                   & $v_3$ ($\mathrm{C}$) & 3  & 23  & 32.0  & 99.9 \\
\bottomrule
\end{tabular}

  \caption{
  Results for various models on the Level 5 task ($\tau=0.95$).
  }
  \label{table:rsp_model_dimmension_tau_0.95_level_5}
\end{table}

\clearpage

\clearpage

\begin{figure}[t]
  \centering
  \includegraphics[width=0.97\linewidth]{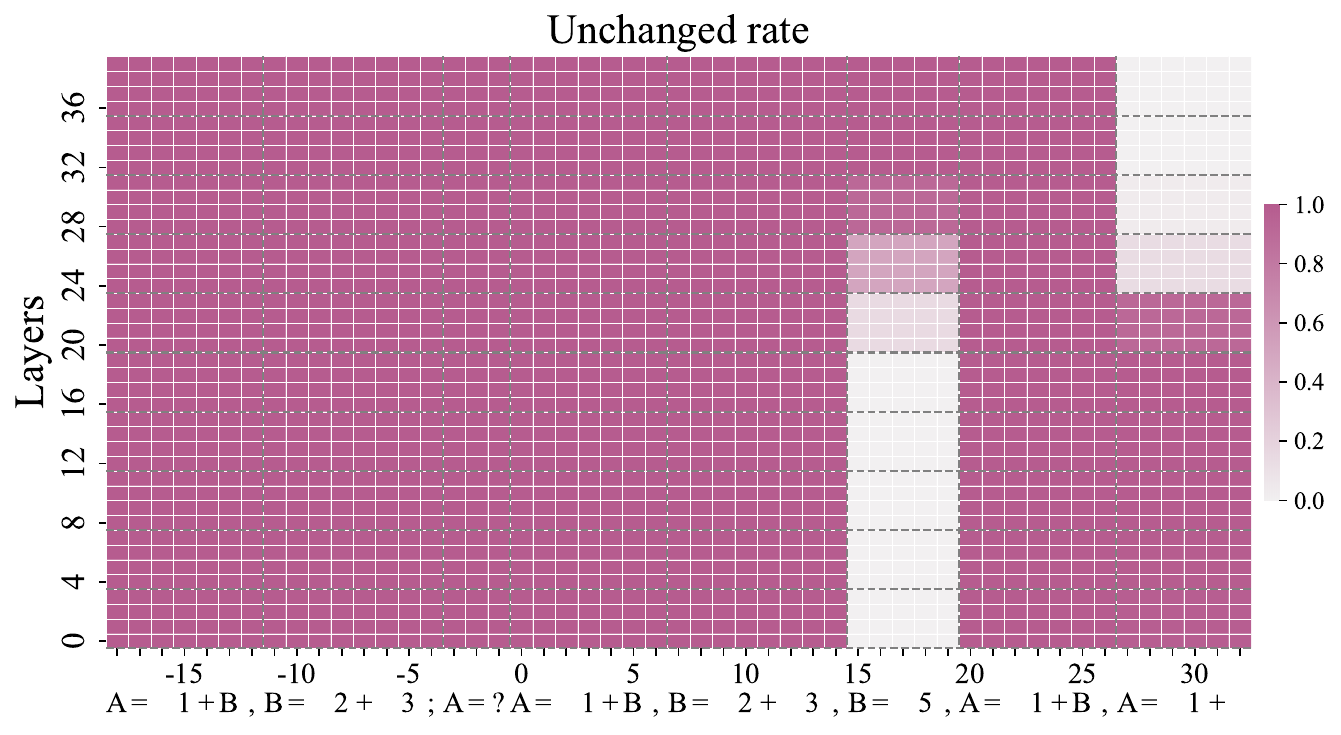}
  \caption{Results of the causal intervention on Mistral-Nemo-Base-2407. Each grid cell shows the unchanged rate when $z_{32}$ \(({\underline{\textcolor{gray}{\mathrm{A}=1+}5}}_{\hspace{0.05cm}\mathbf{4}})\) is the target token.}
  \label{fig:intervention_mistral_nemo_base_2407_last_5_unchanged}
\end{figure}

\end{document}